\documentclass[10pt]{article} % For LaTeX2e
\usepackage[dvipsnames]{xcolor}
\usepackage[accepted]{tmlr}
\usepackage{subcaption}
\usepackage{float}
\usepackage{booktabs}
\usepackage{graphicx}
\usepackage{bbm}
\DeclareUnicodeCharacter{202F}{\,}      % map to a thin space
\usepackage{amssymb}
\usepackage{pifont}   % for \ding
\newcommand{\xmark}{\ding{55}}
\usepackage{array}
\usepackage{pdflscape}
\usepackage{booktabs}
\usepackage{multirow} 
\usepackage{rotating}
\usepackage{comment}
\usepackage{amsmath,amsfonts,bm}

\def\eqref#1{equation~\ref{#1}}
\def\1{\bm{1}}

\DeclareMathAlphabet{\mathsfit}{\encodingdefault}{\sfdefault}{m}{sl}
\SetMathAlphabet{\mathsfit}{bold}{\encodingdefault}{\sfdefault}{bx}{n}

\usepackage{hyperref}
\usepackage{url}

\title{SafeOR-Gym: A Benchmark Suite for Safe Reinforcement Learning Algorithms on Practical Operations Research Problems}

\author{Asha Ramanujam\textsuperscript{1}, 
Adam Elyoumi\textsuperscript{1}, 
Hao Chen\textsuperscript{1}, 
Sai Madhukiran Kompalli\textsuperscript{1},\\
\textbf{Akshdeep Singh Ahluwalia\textsuperscript{1}, 
Shraman Pal\textsuperscript{1}, 
Dimitri J. Papageorgiou\textsuperscript{2}, 
Can Li\textsuperscript{1} }\\
\textsuperscript{1}Davidson School of Chemical Engineering, Purdue University, West Lafayette, IN \\
\textsuperscript{2}Energy Sciences, ExxonMobil Technology and Engineering Company, Annadale, NJ \\
\texttt{\{aramanuj, aelyoumi, chen4433, skompall, ahluwala, pal52, canli\}@purdue.edu} \\
\texttt{dimitri.j.papageorgiou@exxonmobil.com}}

\def\month{May}
\def\year{2026}
\def\openreview{\url{https://openreview.net/forum?id=3EREfePUvi}}

\begin{document}

\maketitle

\begin{abstract}
Most existing safe reinforcement learning (RL) benchmarks focus on robotics and control tasks, offering limited relevance to high-stakes domains that involve structured constraints, mixed-integer decisions, and industrial complexity. This gap hinders the advancement and deployment of safe RL in critical areas such as energy systems, manufacturing, and supply chains. To address this limitation, we present SafeOR-Gym, a benchmark suite of nine operations research (OR) environments tailored for safe RL under complex constraints. Each environment captures a realistic planning, scheduling, or control problems characterized by cost-based constraint violations, planning horizons, and hybrid discrete-continuous action spaces. The suite integrates seamlessly with the Constrained Markov Decision Process (CMDP) interface provided by OmniSafe. We evaluate several state-of-the-art safe RL algorithms across these environments, revealing a wide range of performance: while some tasks are tractable, others expose fundamental limitations in current approaches. SafeOR-Gym provides a challenging and practical testbed that aims to catalyze future research in safe RL for real-world decision-making problems. 
\end{abstract}
\section{Introduction}
Real-world reinforcement learning (RL) applications often demand that agents respect strict safety requirements at all times. \textit{Safe reinforcement learning} \citep{garcia2015comprehensive,gu2022review} addresses this need by maximizing long-term rewards while satisfying safety constraints. In contrast to standard RL, which might treat safety violations as mere negative rewards, safe RL explicitly enforces constraints throughout training and deployment. This capability is vital in safety-critical domains such as autonomous driving, robotics, and power systems, where an agent’s actions can lead to irreversible damage or hazards if constraints such as speed limits, stability margins, or resource capacities are violated. The formalism of constrained Markov decision processes (CMDPs) \citep{altman2021constrained} provides a natural framework for such problems, requiring the learned policy to remain within a set of safe outcomes at all times. In practice, safe RL algorithms incorporate cost signals or penalties for unsafe behavior and aim to ensure constraint satisfaction during both learning and execution. This paradigm has gained traction as a cornerstone for deploying RL in high-stakes environments.

Progress in safe RL has been accelerated by the development of specialized libraries and benchmark suites. A prominent example is \textit{OmniSafe} \citep{ji2024omnisafe}, an open-source infrastructure designed specifically for safe RL research. OmniSafe offers a unified, modular framework with built-in support for constraint handling and a comprehensive collection of constrained RL algorithms. Notably, it extends the standard OpenAI Gym interface by supporting constrained Markov decision processes (CMDPs), enabling explicit modeling of safety constraints through cost signals. These capabilities make OmniSafe and similar frameworks highly valuable for evaluating safe RL methods in a standardized and extensible setting. However, despite advances in algorithms and software infrastructure, the diversity and realism of benchmark environments tailored for safe RL remain limited.

Most existing RL benchmarks were not designed with safety constraints in mind. Classic control tasks and popular continuous control domains such as CartPole, Pendulum, and MuJoCo locomotion simulations \citep{todorov2012mujoco,brockman2016gym} offer simple dynamics with no explicit safety constraints. Crucially, these environments rarely involve the mixed discrete–continuous decision-making or constraints that characterize real operational problems. Even specialized safe RL environment suites focus primarily on robotic control and do not capture the structured complexity of industrial decision problems. This gap in benchmarks makes it challenging to rigorously evaluate how well safe RL algorithms would perform on realistic, safety-critical tasks that involve complex constraints and decision structures.

Operations research (OR) problems \citep{rardin1998optimization} present an appealing opportunity to fill this gap. OR encompasses a broad class of decision-making tasks with rich combinatorial structure, explicit constraints, and often long planning horizons. These problems inherently require balancing long-term objectives with immediate feasibility and safety. For instance, planning the operation of an energy storage system demands making multi-year investments and dispatch decisions that remain robust to short-term operational constraints and rare events\citep{ramanujam2025tutorial,li2022mixed}. In general, OR formulations like scheduling, resource allocation, and supply chain management force agents to respect resource capacities, timing deadlines, and logical constraints at every step, exactly the kind of requirements safe RL is meant to handle. Moreover, OR problems frequently arise in safety-critical domains such as energy systems, transportation, supply chains, and chemical process operations. In these settings, violating a constraint such as overloading a network, missing a maintenance schedule, running a process outside safe limits, etc., can lead to severe real-world consequences. The structured nature and practical relevance of OR tasks make them well-suited as benchmark environments to stress-test safe RL algorithms on realistic problems that go beyond the toy examples commonly used.

We introduce SafeOR-Gym, a benchmark suite for safe reinforcement learning that features a diverse set of operations research environments spanning industrial planning and real-time control. The suite includes nine environments with varying structures, time horizons, and decision complexities. 

% RTN Scheduling is a batch production task modeled as a resource-task network, while STN Scheduling focuses on timing and precedence constraints in a state-task network. ASU Scheduling addresses production planning in an energy-intensive air separation unit. Maintenance Scheduling involves optimizing the timing of maintenance actions to ensure system reliability. The Blending environment models the safe mixing of raw materials to meet product specifications. Capacity Expansion is a long-term infrastructure planning problem under budget and reliability constraints. Unit Commitment schedules power generation to meet hourly electricity demand safely. The Supply Chain environment manages multi-echelon inventory and distribution across multiple stages. Finally, the Energy Storage environment involves the seasonal control of long-duration energy storage systems. Together, these environments provide a rich and realistic testbed for evaluating safe reinforcement learning algorithms.

In summary, this work makes the following contributions:
\begin{enumerate}
    \item We develop a suite of nine OR-based benchmark environments tailored for safe RL, addressing the need for more realistic and structured evaluation tasks.
    \item We release Gym-compatible implementations of these environments with native integration into the OmniSafe framework, enabling out-of-the-box use of constraint-handling algorithms.
    \item We evaluate the environments using several on-policy algorithms in OminiSafe and discuss the current limitations of existing safe RL algorithms in solving highly constrained OR problems.
\end{enumerate}
The primary goal of this paper is to stress-test safe reinforcement learning algorithms across environments of increasing structural complexity. Rather than proposing new hybrid algorithms, we aim to diagnose structural failure modes and identify directions for future OR–RL integration. All the code used is available at \url{https://github.com/li-group/SafeOR-Gym}.
\section{Related Work}

Standard RL benchmarks, such as the OpenAI Gym toolkit \citep{brockman2016gym} with classic control and MuJoCo-based continuous control tasks, have been foundational for evaluating reinforcement learning algorithms. However, they typically lack built-in safety constraints or the structured decision-making found in operations research (OR) problems. Most environments are designed as simplified tasks with minimal realism or constraints.

To address the need for safety-aware evaluation, specialized benchmark libraries have emerged. OpenAI’s \textit{Safety Gym} \citep{ray2019benchmarking} introduced a set of environments that simulate continuous control tasks with hazards, cost signals, and explicit safety constraints. More recently, the \textit{Safety Gymnasium} \citep{ji2023safetygymnasium} suite has extended this idea to include both single- and multi-agent safety-critical environments. The \textit{OmniSafe} \citep{ji2024omnisafe} framework further consolidates these contributions by integrating a variety of safe RL environments and algorithms into a unified and modular platform. It supports constraint-handling algorithms, constrained policy optimization methods, and provides compatibility with Gym-style APIs.

In parallel, a growing body of work has focused on bringing OR and classical control problems into Gym-compatible environments. \textit{OR-Gym} \citep{hubbs2020orgym} provides a library of OR formulations, such as knapsack, bin packing, and supply chain management, as RL tasks, enabling comparison between RL policies and traditional optimization techniques. \textit{PC-Gym} \citep{bloor2024pcgym} and \citep{park2025reinforcement} introduce environments that model realistic chemical process dynamics, control disturbances, and enforce operational safety. \textit{SustainGym} \citep{yeh2023sustaingym} contributes several sustainability-focused environments such as electric vehicle charging and data center scheduling, which feature complex operational constraints, distribution shifts, and hybrid discrete-continuous action spaces.

Despite these efforts, existing benchmarks for operations research problems exhibit important limitations. Existing works are only compatible with standard Gymnasium environments, rather than the CMDP interface required by frameworks like OmniSafe. They typically handle constraint violations by penalizing the reward function, rather than modeling them as explicit cost signals. Furthermore, many environments simplify or abstract away industrial complexity, limiting their utility for evaluating algorithms in realistic, safety-critical scenarios. \textit{OR-Gym}, for example, primarily consists of toy problems such as knapsack and bin packing, which lack the structural and constraint richness of real-world applications. \textit{SustainGym} centers on sustainability-oriented problems, but does not incorporate the nonlinear nonconvex constraints in many OR applications, such as those found in the blending problem. These gaps motivate the development of a benchmark suite featuring rich, constrained, and practically relevant OR problems that can serve as a rigorous testbed for safe RL algorithms.  These comparisons are summarized in Table~\ref{tab:comp}. A more detailed comparison of the environments is show in Appendix \ref{sec:appendix:envcomp}.

\begin{table}[htbp]
\centering
\caption{Comparison of Gym environments with operations research applications}\label{tab:comp}
\label{tab:env_comparison}
\setlength{\tabcolsep}{4pt}
\renewcommand{\arraystretch}{1.3}

% ---------- Subtable (a) ----------
\begin{subtable}[t]{\textwidth}
\centering
\caption{Environment class, constraints, and SafeRL compatibility}
\label{tab:env_comparison_a}
\begin{center}
\begin{tabular}{llll}
%\toprule
\textbf{Work} & \textbf{Env. Class} & \textbf{Constraint Handling} & \textbf{SafeRL} \\
%\midrule
\hline \\
OR-Gym     & Gymnasium & truncation, reward penalties & $\times$ \\
SustainGym & Gymnasium & truncation, reward penalties & $\times$ \\
SafeOR-Gym & Gymnasium + CMDP & truncation, reward penalties, explicit costs & $\checkmark$ \\
%\bottomrule
\end{tabular}
\end{center}
\end{subtable}

\vspace{0.8em}

% ---------- Subtable (b) ----------
\begin{subtable}[t]{\textwidth}
\centering
\caption{problem size, constraint complexity, and application domains}
\label{tab:env_comparison_b}
\begin{center}
\begin{tabular}{llll}
%\toprule
\textbf{Work} & \textbf{Obs / Action (mean, max)} & \textbf{Nonconvex} & \textbf{Domain} \\
%\midrule
\hline \\
OR-Gym     & (242, 2501) / (57, 200) & $\times$     & classical OR \\
SustainGym & (79, 150) / (33, 72)    & $\times$     & sustainable energy \\
SafeOR-Gym & (86, 4280) / (32, 272)  & $\checkmark$ & planning, power, chemical, scheduling \\
%\bottomrule
\end{tabular}
\end{center}
\end{subtable}

\end{table}

\section{Environments} \label{sec:environments}
This section provides an overview of the environments in \textit{SafeOR-Gym}. The detailed descriptions of all the environments including the mathematical details, explanations, and illustrative figures can be found in Appendix \ref{sec:appnedix:env}. 
\subsection{Resource and State Task Network Environments (\texttt{RTNEnv}, \texttt{STNEnv})}

\textbf{Problem Description} 
The Resource Task Network (RTN) \citep{pantelides1994unified} is a foundational model for scheduling in process industries such as chemical manufacturing and oil and gas refining. It represents the system as a network of materials (reactants, intermediates, products, and equipment) and processing steps (\textit{tasks}) that transform inputs into outputs. Each task is executed in discrete quantities (\textit{batches}) and requires specific equipment units. In \texttt{RTNEnv}, the agent controls task-level batch sizes at each timestep over a finite horizon to meet demand while minimizing cost under constraints. The environment captures inventory dynamics, delayed production, and equipment reuse governed by task durations.

The State Task Network (STN) extends RTN by explicitly modeling task–equipment assignments and material \textit{states}. Each task must be executed on a compatible unit with unit-specific durations and capacity bounds. In \texttt{STNEnv}, the agent selects task–equipment pairings and batch sizes at each timestep. This introduces explicit equipment occupancy, pair-specific processing durations, and compatibility constraints, yielding a more structured and constrained scheduling problem.

\textbf{State Space} 
In both environments, the state includes: (i) inventory levels of all materials (RTN) or states (STN), including equipment; (ii) a buffer of pending outputs representing delayed deliveries from in-progress tasks; and (iii) future product demand from time $t$ to $T$, right-padded with zeros.

\textbf{Action Space} 
In \texttt{RTNEnv}, the action is a continuous vector with one entry per task, representing batch sizes. In \texttt{STNEnv}, the action is a continuous matrix over task–equipment pairs, representing batch sizes executed on specific units, subject to compatibility and capacity constraints.

\textbf{Transition Dynamics} 
Both environments apply task stoichiometry to compute material inflows and outflows, update inventories, and maintain a buffer of delayed outputs that are realized after task-specific processing times. Actions are sanitized to enforce feasibility by clipping batch sizes based on available inventory and constraints. In \texttt{RTNEnv}, transitions capture inventory consumption, delayed production, and equipment reuse constraints. In \texttt{STNEnv}, transitions additionally enforce unit compatibility and track equipment occupancy explicitly, ensuring that tasks cannot be assigned to unavailable units.

\textbf{Cost} 
In both environments, the cost consists of four components: (i) penalties for inventory bound violations (upper and lower), (ii) penalties for infeasible equipment usage, (iii) penalties for constraint violations arising from task execution, and (iv) a sanitization cost that penalizes deviations between proposed and feasible actions. In \texttt{STNEnv}, these penalties additionally reflect unit-level feasibility and assignment constraints.

\textbf{Reward} 
In both environments, the reward is the net economic objective: revenue from fulfilling product demand minus operational and holding costs. This includes utility costs for task execution, penalties for unmet demand, and costs associated with procuring reactants when inventories fall below zero. The formulations are structurally identical, with \texttt{STNEnv} reflecting tighter coupling between tasks and equipment.

 \subsection{Unit Commitment (\texttt{UCEnv})}

 \textbf{Problem Description} The Unit Commitment problem is a cornerstone optimization task in power systems. It determines when to turn power generators on or off and how much electricity each should produce, aiming to minimize operational cost while ensuring supply-demand balance and meeting system reliability constraints \citep{knueven2020}. In \texttt{UCEnv}, the agent controls binary on/off decisions and continuous dispatch levels for each generator over a finite planning horizon. These decisions must satisfy safety constraints such as ramping limits, minimum up/down times, and transmission capacity bounds. The environment captures realistic power grid dynamics under uncertainty, reflecting the challenges of operating with variable demand and fluctuating renewable generation.

 \textbf{State Space} The state includes generator on/off sequences; power output of each generator; voltage angles at each bus; and forecast electricity demand for future time steps.

 \textbf{Action Space} The action includes on/off status and power dispatch level for each generator, along with voltage angle assignments at each bus.

 \textbf{Transition Dynamics} Applies feasibility checks and automatic repairs to enforce minimum up/down times and ramp rate limits; update load unfulfillment and power reserve; clips power output and voltage angles to stay within generator and grid limits; updates generator status sequences and forecasts.

 \textbf{Cost} The environment penalizes violations of minimum up/down time constraints; ramp rate limit violations; tranmision capacity violations.

 \textbf{Reward} The reward is the negative of total operating cost, including: quadratic generation costs; startup and shutdown costs; load shedding penalties for unmet demand; and reserve shortfall penalties for failing to maintain spinning reserves.

\subsection{Generation and Transmission Expansion Planning (\texttt{GTEPEnv})}

\textbf{Problem Description} 
Generation and transmission expansion planning is a long-term power systems problem involving sequential decisions on where and when to install generators and transmission lines to meet evolving electricity demand \citep{li2022mixed}. In \texttt{GTEPEnv}, the agent determines generator deployments and inter-regional power flows over a multi-period horizon, subject to installation bounds, flow limits, and demand constraints. Safety-critical violations, such as exceeding generator limits or under-supplying electricity, incur penalties. This environment captures planning under uncertainty, where irreversible investments must anticipate changing demand.

\textbf{State Space} The state includes current generator installations per region; transmission line installations; forecasted regional electricity demand over a lookahead window; and the current time step.

\textbf{Action Space} The action specifies: the number of generators of each type to install in each region; and the power flow in each candidate transmission line.

\textbf{Transition Dynamics} Generator counts and line installation flags are updated based on actions. The available power in each region is computed based on local generation and net imports via transmission lines. Future demand forecasts are rolled forward. State variable bounds are clipped if violation happens.

\textbf{Cost} The environment penalizes: exceeding the maximum allowed generators per region (violations are penalized and corrected to remain feasible);  and failing to meet regional electricity demand. 

\textbf{Reward} The reward is the negative of total economic costs, including: installation costs for new generators and transmission lines.

\subsection{Multiperiod Blending Problem (\texttt{BlendingEnv})}

\textbf{Problem Description} The multiperiod blending problem arises in industries such as refining and chemical processing, where raw materials with different properties must be blended over time to produce saleable products meeting strict quality constraints \citep{Chen2020}. In \texttt{BlendingEnv}, the agent decides how much of each source stream to purchase, how to route flows through a network of blenders, and how much product to sell at each time step. Safety constraints include property bounds on final products and storage limits on inventories. The environment reflects operational complexities such as nonlinear blending effects, capacity limits, and quality enforcement.

\textbf{State Space} The state includes current inventory levels of sources, blenders, and demand nodes; material properties of blenders; future availability of sources and product demand over a lookahead window, and the current time step.

\textbf{Action Space} The action specifies: source stream purchase quantities; product sale quantities; flow rates from source inventories to blenders, between blenders, and from blenders to product inventories.

\textbf{Transition Dynamics} Updates inventories using material balances; clips or adjusts values to satisfy inventory bounds when violations occur (depending on the constraint-handling strategy); updates material properties in blenders based on flow composition; rolls forward source availability and demand forecasts. The dynamics of mixing material properties involve nonlinear nonconvex constraints.

\textbf{Cost} The environment penalizes: violating inventory bounds; violating the no simultaneous ``in-out'' flow rule for blenders; and producing out-of-spec blends.

\textbf{Reward} The reward includes: revenue from product sales; minus cost of purchasing source streams; minus variable and fixed costs for flow operations.

 \subsection{Multi-Echelon Inventory Management (\texttt{InvMgmtEnv})}

 \textbf{Problem Description} The multi-echelon inventory management problem models the operation of complex supply chains involving multiple interconnected tiers such as suppliers, producers, distributors, retailers, and markets \citep{perez2021algorithmic}. In \texttt{InvMgmtEnv}, the agent decides ordering quantities across various routes to satisfy uncertain market demand while minimizing operational costs and penalties. The environment features fixed lead times, capacity constraints, and stochastic demand, simulating real-world supply chain dynamics and risks. Safe operations require respecting bounds on order volumes, inventory capacities, and maintaining service levels under varying demand.

 \textbf{State Space} The state includes current inventory levels at all nodes; quantities in pipeline inventory by route and lead time; historical and forecasted market demand; backlog of unmet demand; and sales at the current period.

 \textbf{Action Space} The action specifies continuous reorder quantities along each supply chain route.

 \textbf{Transition Dynamics} Orders enter pipelines and propagate forward according to lead times; arrivals update on-hand inventories; update demand forecast; sales and backlog are computed based on available inventory.

 \textbf{Cost} Penalties are incurred for orders exceeding quantity or capacity limits, for any state (inventory or pipeline) falling outside its valid bounds, for negative or infeasible inventory levels, and for backlogs beyond the allowed threshold.

 \textbf{Reward} The reward includes: revenue from market sales; minus procurement costs, inventory holding costs (both on-hand and in-transit), operating expenses, and backlog penalties.

 \subsection{Grid-Integrated Energy Storage (\texttt{GridStorageEnv})}

 \textbf{Problem Description}
 \texttt Grid-Integrated Energy Storage problem simulates hourly dispatch on a transmission grid where batteries share buses with thermal generators and a prescheduled subset of lines may be switched off. During each hour the agent selects generator set‑points, battery charge and discharge rates, deliberate load‑shedding, and controllable voltage angles at all non‑reference buses; the DC power‑flow equations then determine line flows, and any de‑energised line is forced to zero. Batteries evolve through a simple state‑of‑charge (SOC) recursion that applies round‑trip efficiency and an inter‑period carry‑over factor, and the episode spans a fixed horizon without early termination.

 \textbf{State Space}
 State space contains normalised battery SOC, voltage‑angle differences, line‑loading flows, slack generation, a rolling demand‑forecast window, and a normalised time. Each component is clipped to its admissible interval, with the clipping distance recorded for later penalties.

 \textbf{Action Space}
 The policy emits a normalised vector that is affinely mapped to physical generator outputs, battery charge and discharge rates, load‑shedding, and bus angles. Values that exceed their limits are clipped, and every excess contributes to an action‑bound penalty.

 \textbf{Transition Dynamics}
 After decoding and clipping the action, the environment updates battery SOC, caps load‑shedding, solves the DC power flow with zero flow on de‑energised lines, introduces slack generation to balance each bus, reconstructs the next observation, and logs penalties for any bound, flow, SOC, angle, or power‑balance violation encountered along the way

 \textbf{Cost}
 At each step the environment charges three kinds of penalties: the action‑bound penalty for any control component that is clipped, the aggregated observation‑bound penalty for state variables that drift outside their safe intervals, and a network‑balance penalty that scales with the absolute mismatch between nodal injections and the DC power‑flow equations.

 \textbf{Reward}
 The reward delivered to the agent is the negative sum of generator fuel cost, slack‑generation cost, and load‑shedding cost, so maximising reward is equivalent to minimising these economic charges. This formulation steers the policy toward meeting demand with the lowest feasible operating expense while respecting all system constraints.

\subsection{Integrated Scheduling and Maintenance (\texttt{SchedMaintEnv})}

\textbf{Problem Description} The Integrated Scheduling and Maintenance environment models the daily operation of compressors in an Air Separation Unit (ASU), where gaseous product demand must be met without inventory buffers \citep{Xenos2016Operational}. The agent must coordinate production and maintenance actions for each compressor while optionally procuring external supply to satisfy demand. At each time step, the agent decides the fraction of each compressor’s capacity to operate, whether to initiate maintenance, and how much external product to purchase. Maintenance policies are constrained by compressor-specific conditions, such as mean time to failure (MTTF), maintenance duration, and cooldown periods. Note that the machine failures can also be uncertain. Safety arises from the need to avoid compressor breakdowns due to delayed maintenance, premature maintenance interventions, or ramping during repair periods. 

\textbf{State Space} The state includes: forecasted product demand and electricity prices over a fixed horizon; compressor-level indicators for time since last maintenance, time remaining to complete maintenance, and eligibility to enter maintenance.

\textbf{Action Space} The action at each time step consists of: a binary vector for maintenance decisions (schedule or not); a continuous vector for compressor production rates; and a continuous external purchase action representing the fraction of a predefined maximum external capacity.

\textbf{Transition Dynamics} Compressor state evolves based on maintenance initiation, progress, and cooldown periods. Upate electricity prices and product demand follow forecasts. Maintenance eligibility resets once a cooldown period has elapsed. The environment enforces repair continuity and halts production while repair.

\textbf{Cost} The environment penalizes: early maintenance actions; failure to perform maintenance before MTTF; ramping while under maintenance; and disruption of ongoing maintenance. Additionally, unmet or overmet demand incurs a penalty proportional to the deviation from forecast.

\textbf{Reward} The agent receives a negative reward equal to the total cost incurred: production cost (based on power price and compressor load), external purchase cost.

 \subsection{Production Scheduling in Air Separation Unit (\texttt{ASUEnv})}

 \textbf{Problem Description}  
 The production scheduling problem in an Air Separation Unit (ASU) involves determining hourly production levels for multiple liquid products—LIN (liquid nitrogen), LOX (liquid oxygen), and LAR (liquid argon)—to minimize energy costs while satisfying daily demand and inventory constraints. In \texttt{ASUEnv}, the agent selects convex combinations of historical production modes that obey plant feasibility constraints. The environment simulates inventory accumulation, daily shipping to meet forecasted demand, and electricity price fluctuations. It penalizes inventory overflows and demand shortfalls while enabling flexible production responses through convex scheduling. This environment captures the temporal coupling of inventory and demand, along with realistic production feasibility via data-driven convex sets.

 \textbf{State Space}  
 The state includes: hourly electricity prices and demand forecasts for each product; and current inventory levels for all products. 

 \textbf{Action Space}  
 The action consists of convex weights over a fixed set of feasible production profiles, derived from historical data, to specify hourly production rates for each product.

 \textbf{Transition Dynamics}  
 Electricity price and demand forecasts are updated via a rolling mechanism. Inventories evolve via accumulation of production and daily shipping, with demand fulfillment capped by availability. A sanitization step clips inventory levels to avoid infeasible overflows.

 \textbf{Cost}  
 Costs include: inventory overflow penalties if product levels exceed maximum capacity; and daily demand shortfall penalties computed at the end of each day. Overflow costs are applied instantly; demand shortfall is assessed based on shipping at the daily time boundary.

\textbf{Reward}  
 The reward is the negative of the total production cost, which includes a fixed cost (if any production occurs) and a variable cost based on the electricity price and production volume. 
\section{Implementation, Compatibility, and Extensibility}
All environments are implemented on top of the Gymnasium API \citep{brockman2016gym}, with an additional Constrained Markov Decision Process (CMDP) wrapper \citep{ji2024omnisafe}. This wrapper consists of fewer than 50 lines of code but provides compatibility with Safe RL algorithms that explicitly handle constraints. While OmniSafe is used as the primary reference implementation due to its breadth and active development, SafeOR-Gym can be adapted to other Safe RL libraries with only minimal modifications, since most Safe RL algorithms are CMDP-based and rely on Gymnasium as the base environment class.

Each environment in SafeOR-Gym is initialized with a small illustrative instance, but the underlying data structures are independent of any specific problem setup. Instance-specific parameters such as network topologies, renewable generation profiles, or equipment characteristics are stored in external JSON files. This separation of code and data makes it straightforward to create new problem instances or modify existing ones by editing the JSON inputs, without altering the environment code. As a result, stochastic elements such as randomized demand realizations or sampled renewable generation profiles can be introduced naturally by providing alternative data inputs. 

Although the present work focuses on deterministic environments, SafeOR-Gym is designed to accommodate stochasticity and non-stationarity. This is particularly important for real-world OR problems, which often involve renewable generation variability, equipment failures, or evolving market conditions. As a demonstration, we extended the maintenance scheduling environment to incorporate random machine failures, producing \texttt{SchedMaintEnv-v1}, a stochastic variant of the original deterministic environment (\texttt{SchedMaintEnv-v0}). Importantly, this extension was achieved by inheriting the original environment rather than reimplementing it from scratch, underscoring the ease of extending SafeOR-Gym to uncertainty-aware benchmarks.  In this work, we focus primarily on deterministic environments because most existing safe RL algorithms already face substantial challenges in solving even deterministic OR problems with complex constraints, which will be shown in section \ref{sec:experiments}. Demonstrating these limitations in a controlled deterministic setting helps establish a clear baseline before introducing additional sources of uncertainty.
\section{Experiments}\label{sec:experiments}
We evaluate a suite of safe reinforcement learning (RL) algorithms across multiple environments subject to safety constraints. Each experiment involves one or more deterministic case studies, designed to test algorithmic robustness and generalization. In addition, we also include \texttt{SchedMaintEnv-v1}, a stochastic variant of the original deterministic environment (\texttt{SchedMaintEnv-v0}). We use the safe RL algorithms implemented in OmniSafe, leveraging its \texttt{ExperimentGrid} feature for parallelized training and evaluation with configurable model and training settings. The safe RL algorithms tested include Constrained Policy Optimization (CPO) \citep{pmlr-v70-achiam17a}, the Lagrangian version of trust regions policy optimization (TRPOLag) \citep{ray2019benchmarking}, Penalized Proximal Policy Optimization (P3O) \citep{ijcai2022p520}, Constraint Rectified Policy Optimization (OnCRPO) \citep{pmlr-v139-xu21a}, the Lagrangian version of Deep Deterministic Policy Gradient (DDPGLag) \citep{lillicrap2019continuouscontroldeepreinforcement}, First Order Constrained Optimization in Policy Space (FOCOPS) \citep{zhang2020first}, PID version of SACLag (SACPID) \citep{stooke2020responsive}, and the Lagrangian version of Soft Actor-Critic (SAC) algorithm (SACLag) \citep{pmlr-v80-haarnoja18b}. All the experiments were conducted on AWS servers using g4dn.xlarge instances equipped with NVIDIA T4 GPUs, providing sufficient computational resources for training and evaluation of all the algorithms. The training times for the various algorithms across different experiments are provided in the supplementary material.

\subsection{Results and Discussion}
\begin{figure}[htbp]
    \centering
    %ROw 1
    % Panel 1
    \begin{subfigure}[t]{0.48\textwidth}
        \centering
        \includegraphics[width=0.49\textwidth]{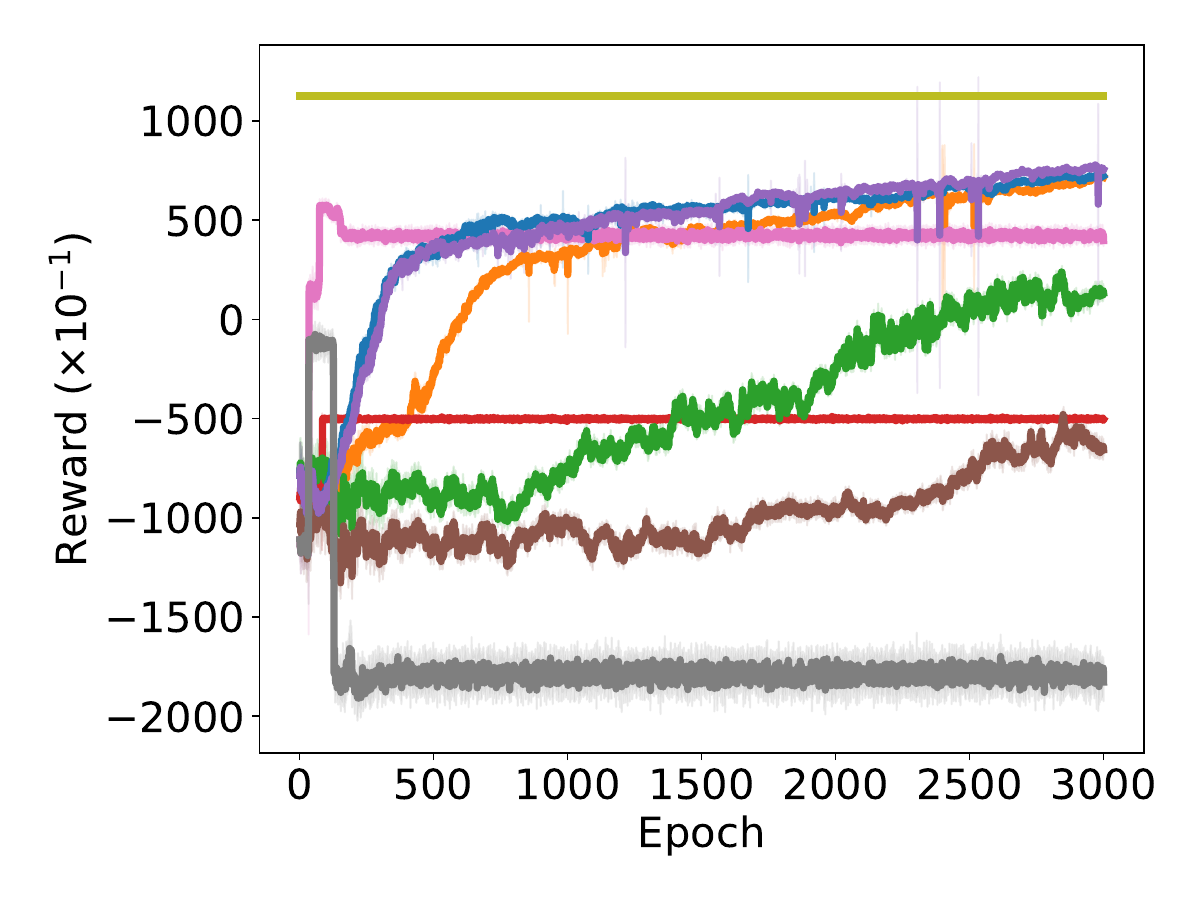}
        \includegraphics[width=0.49\textwidth]{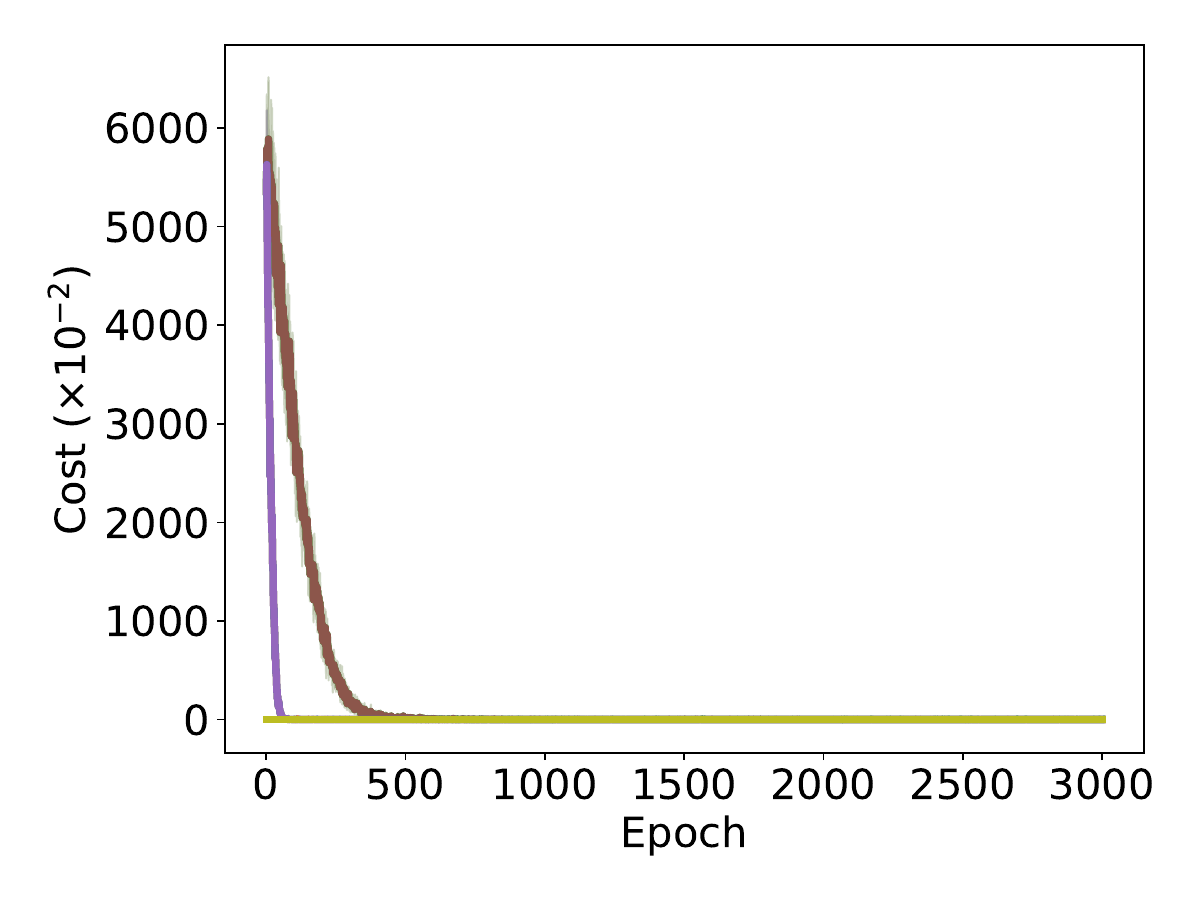}
        \caption{InvMgmtEnv}
        \label{fig:InvMgmtEnv}
    \end{subfigure}
    \hfill
    % Panel 2
    \begin{subfigure}[t]{0.48\textwidth}
        \centering
        \includegraphics[width=0.49\textwidth]{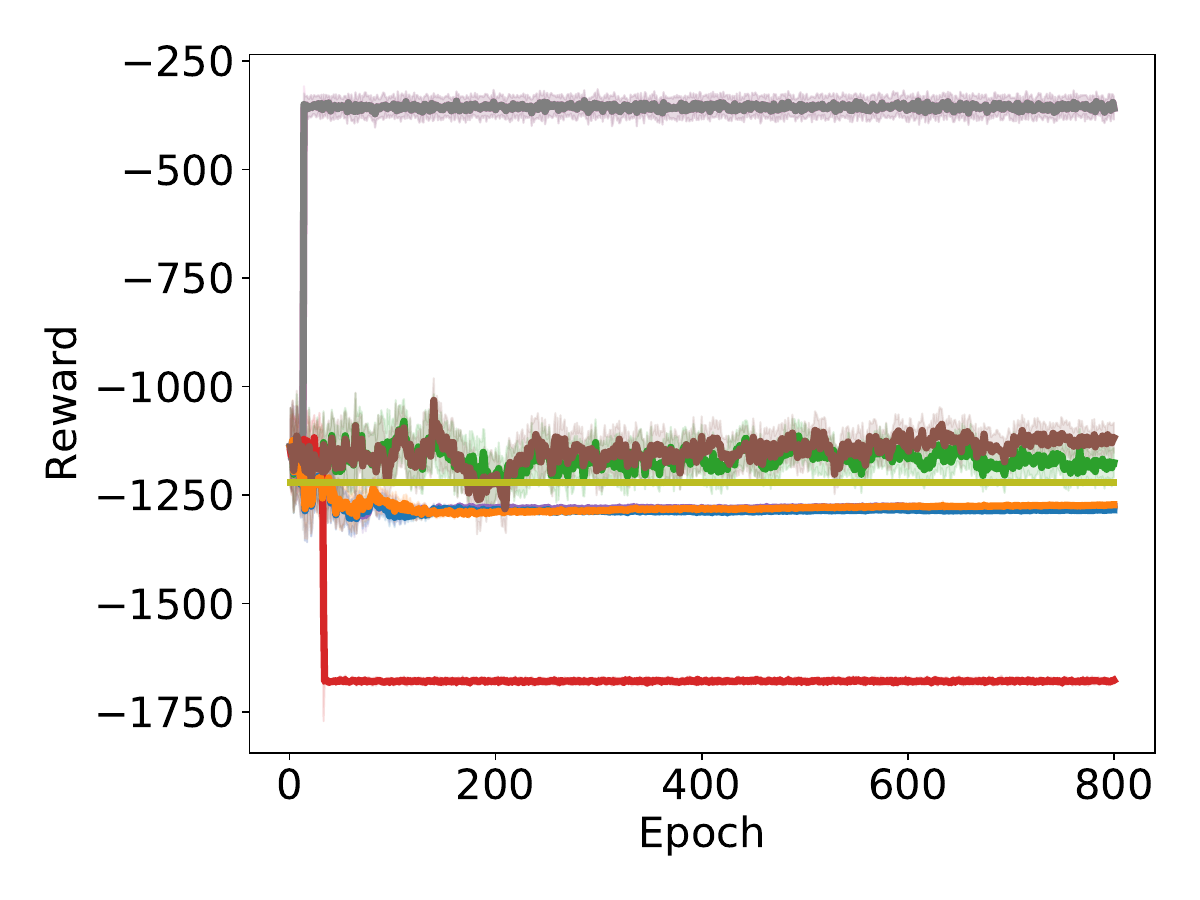}
        \includegraphics[width=0.49\textwidth]{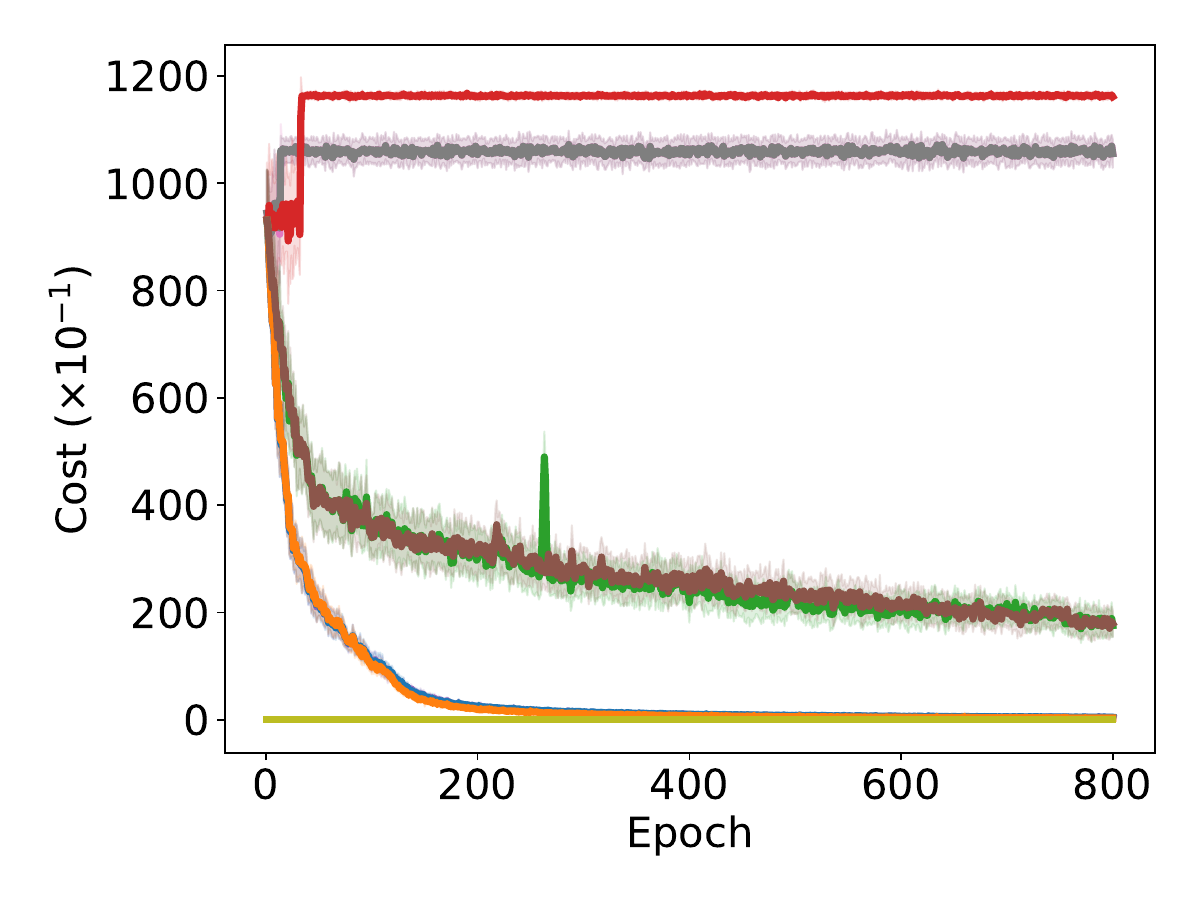}
        \caption{SchedMaintEnv-v0}
        \label{fig:SchedMaintEnv-v0}
    \end{subfigure}

    \vspace{1em}

    % ------------------ Row 2 ------------------
    % Panel 3
    \begin{subfigure}[t]{0.48\textwidth}
        \centering
        \includegraphics[width=0.49\textwidth]{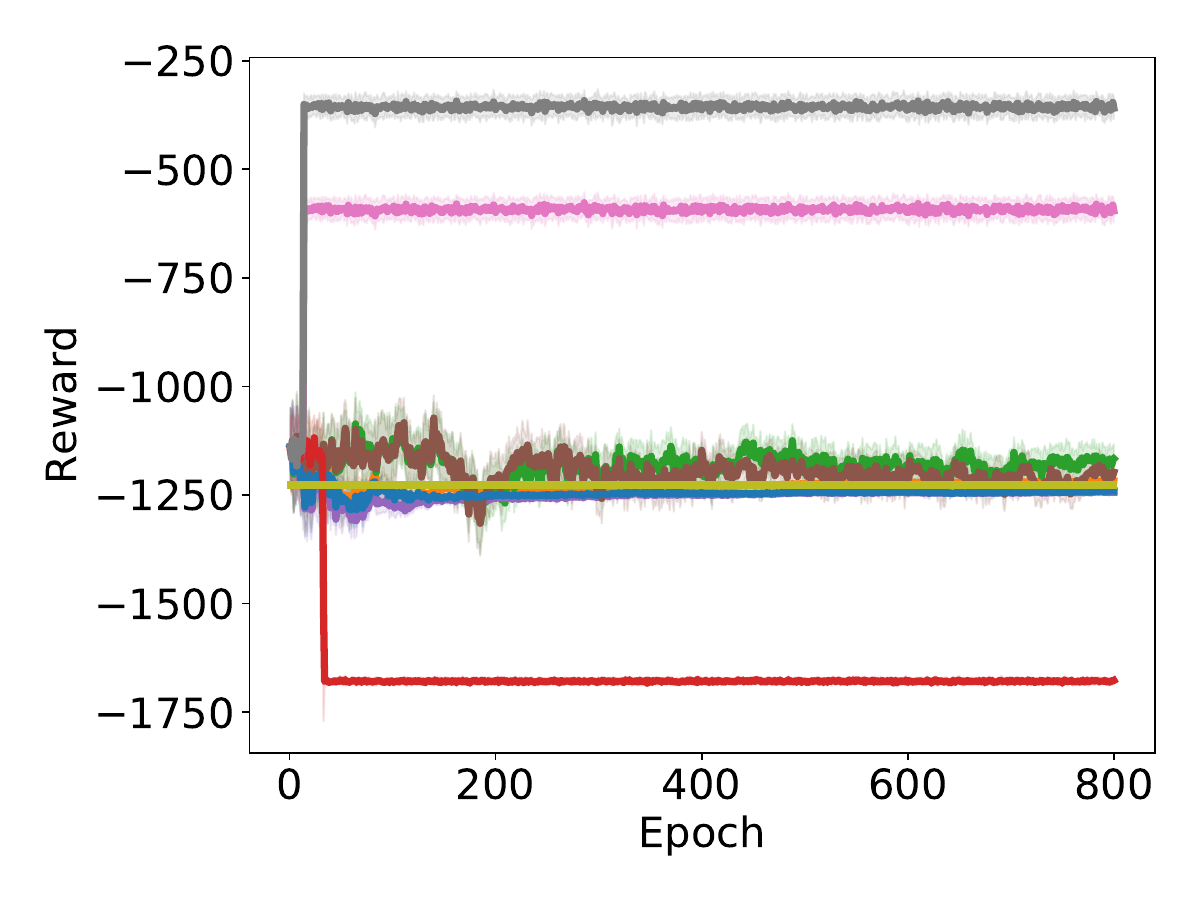}
        \includegraphics[width=0.49\textwidth]{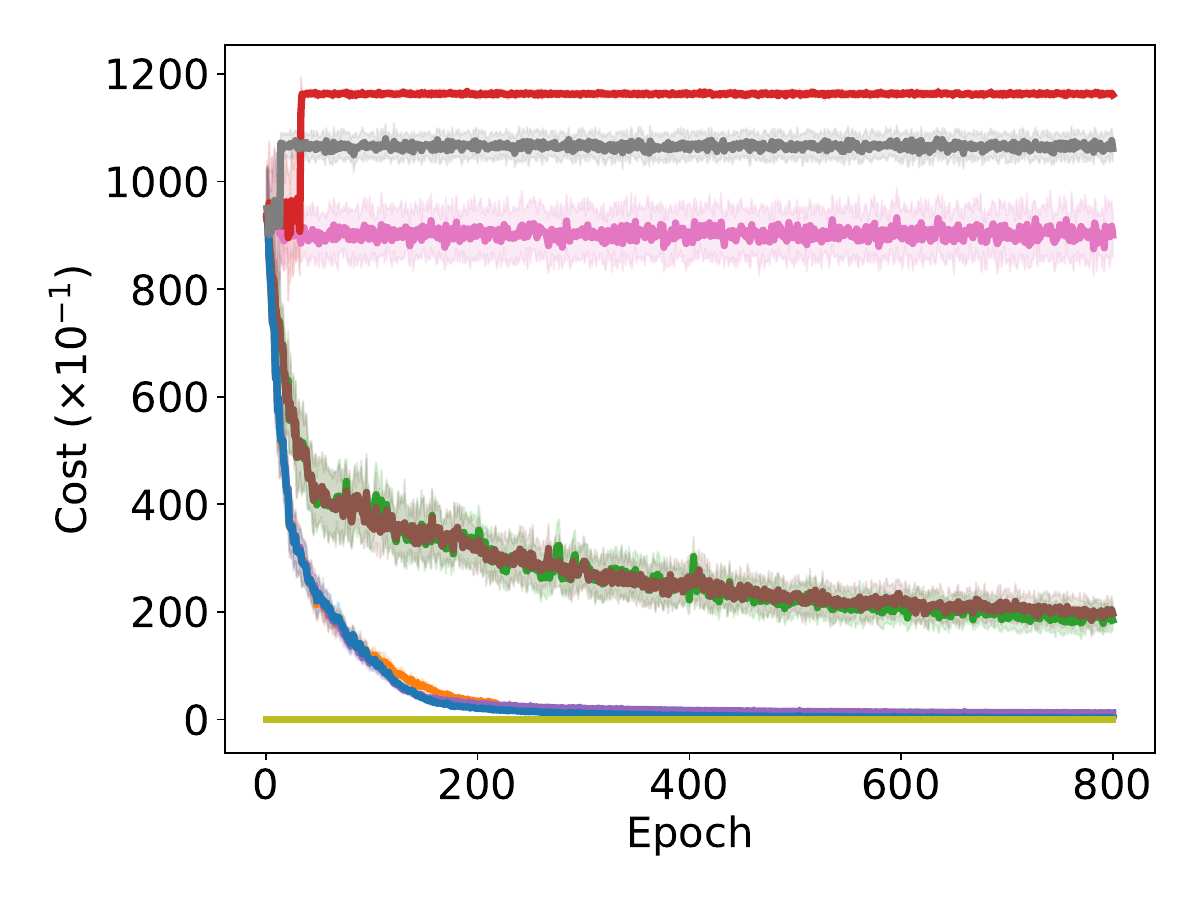}
        \caption{SchedMaintEnv-v1}
        \label{fig:SchedMaintEnv-v1}
    \end{subfigure}
    \hfill
    % Panel 4
    \begin{subfigure}[t]{0.48\textwidth}
        \centering
        \includegraphics[width=0.49\textwidth]{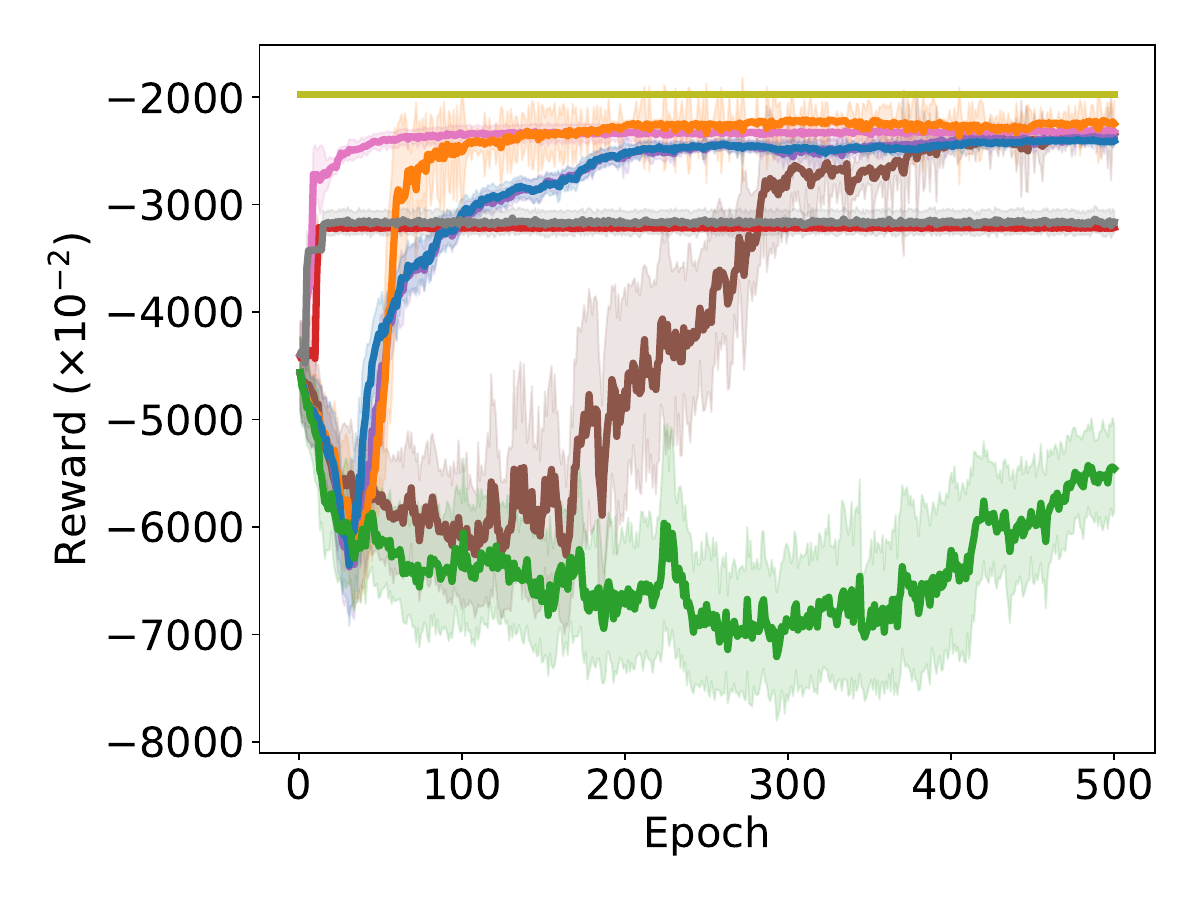}
        \includegraphics[width=0.49\textwidth]{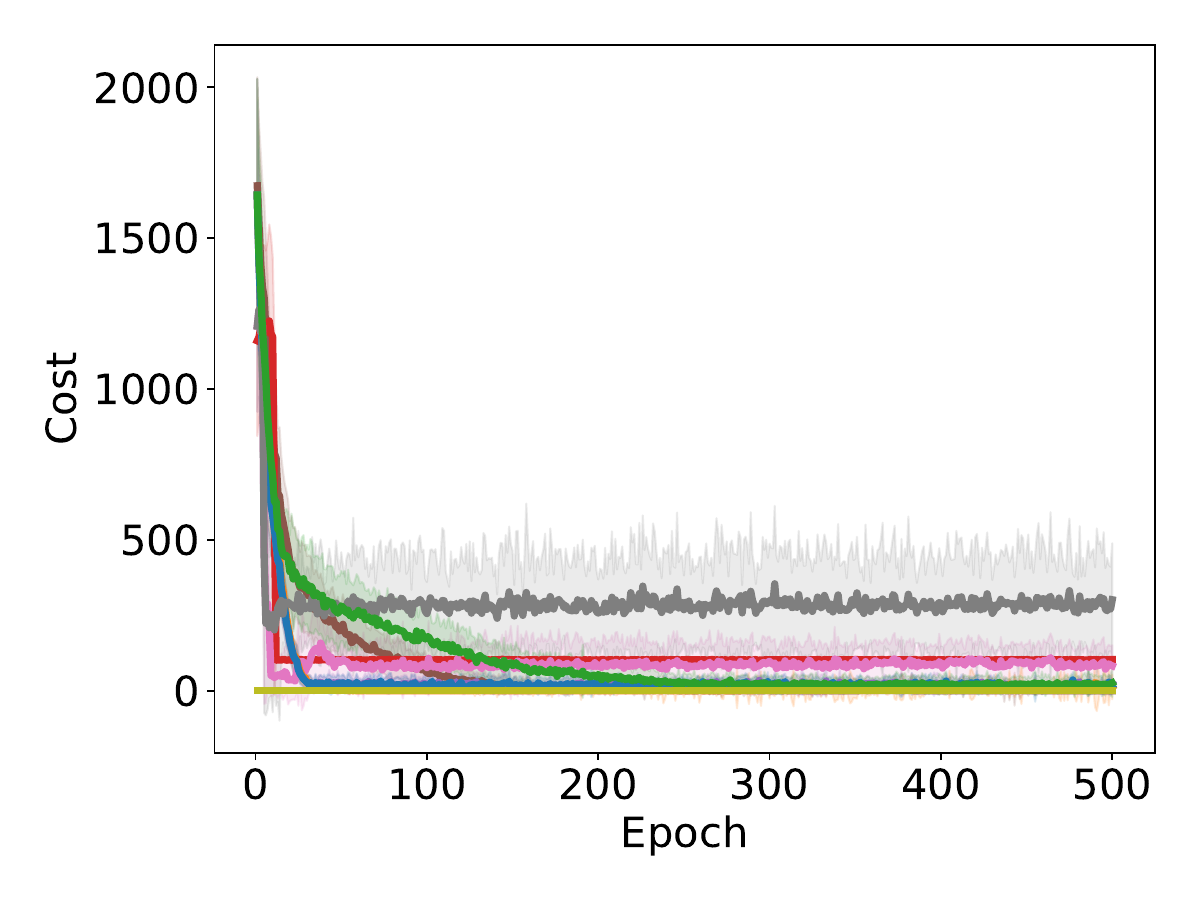}
        \caption{UCEnv}
        \label{fig:UCEnv}
    \end{subfigure}

%Row 3
 \begin{subfigure}[t]{0.48\textwidth}
        \centering
        \includegraphics[width=0.49\textwidth]{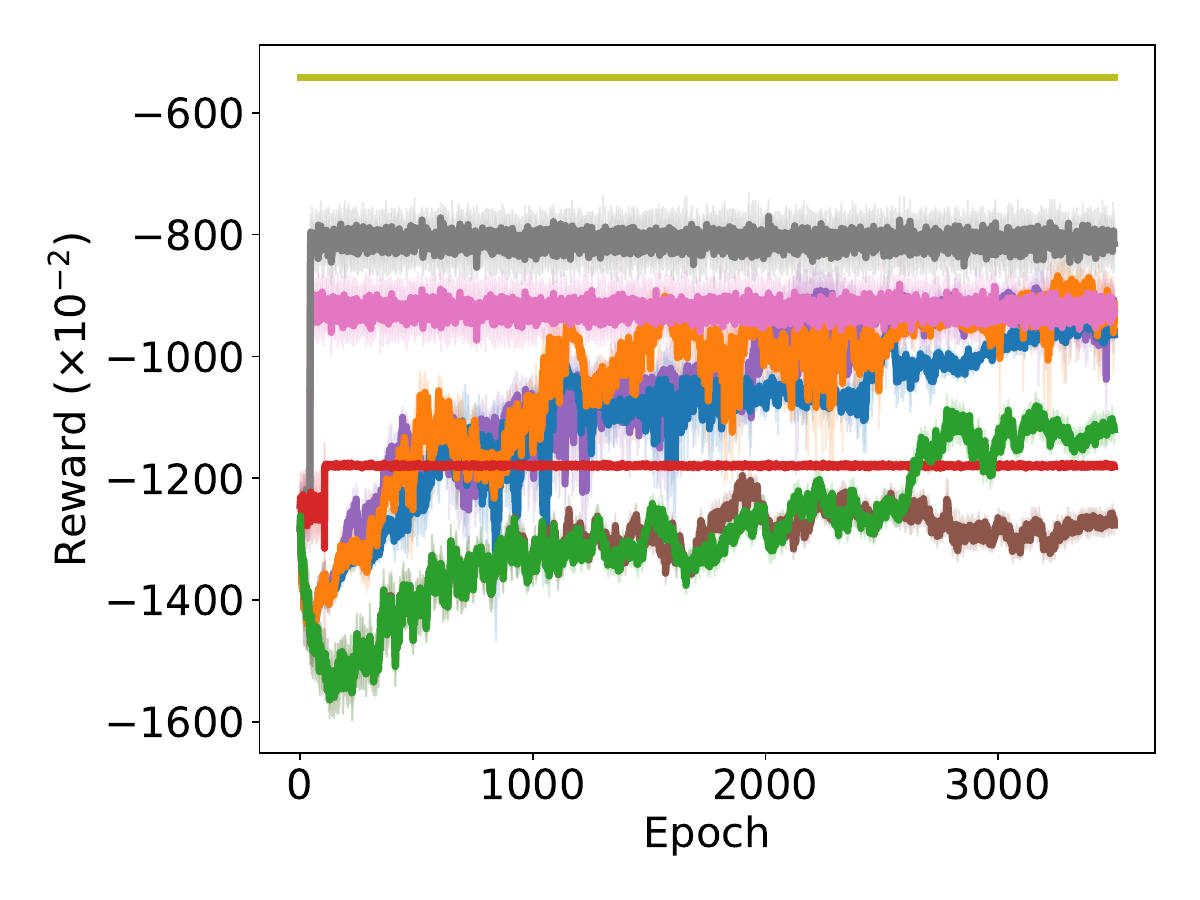}
        \includegraphics[width=0.49\textwidth]{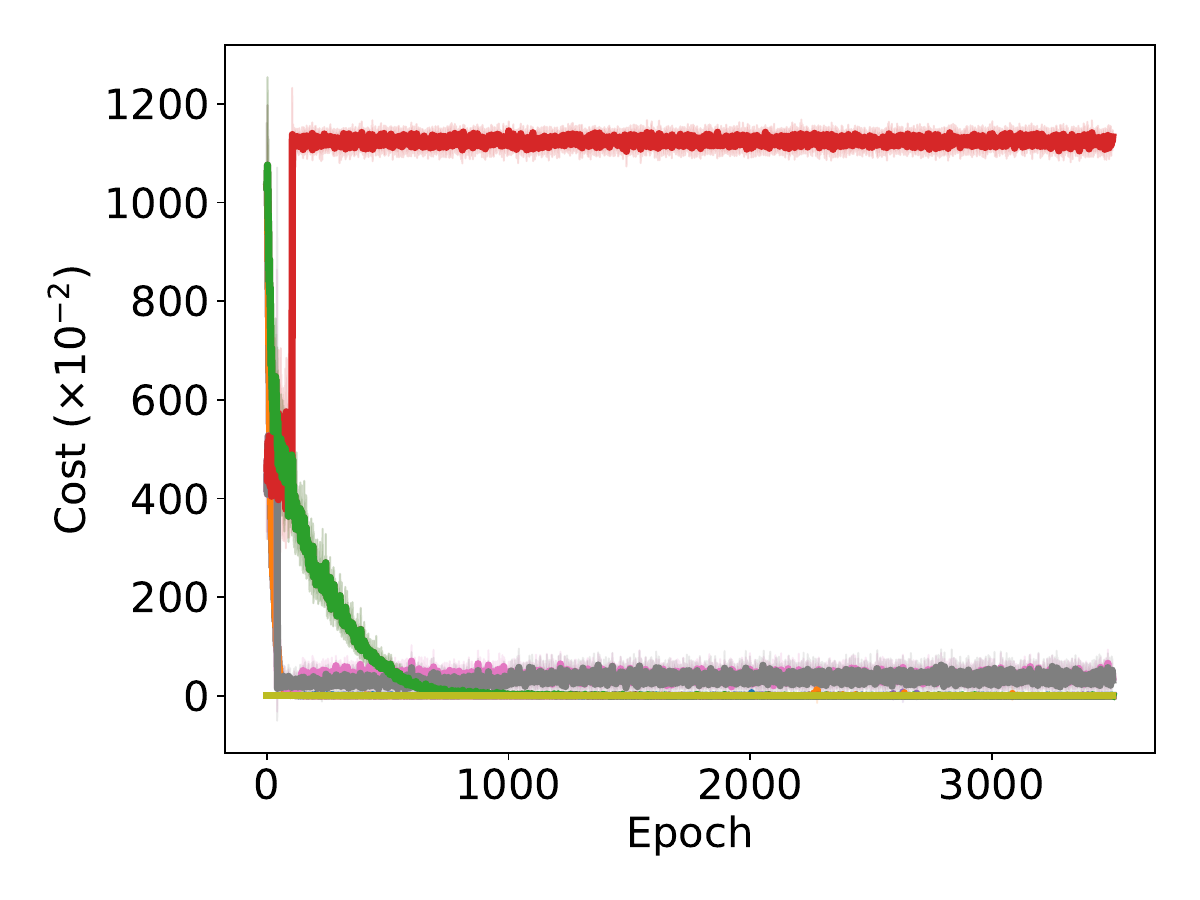}
        \caption{GridStorageEnv}
        \label{fig:Grid}
    \end{subfigure}
    \hfill
\begin{subfigure}[t]{0.48\textwidth}
        \centering
        \includegraphics[width=0.49\textwidth]{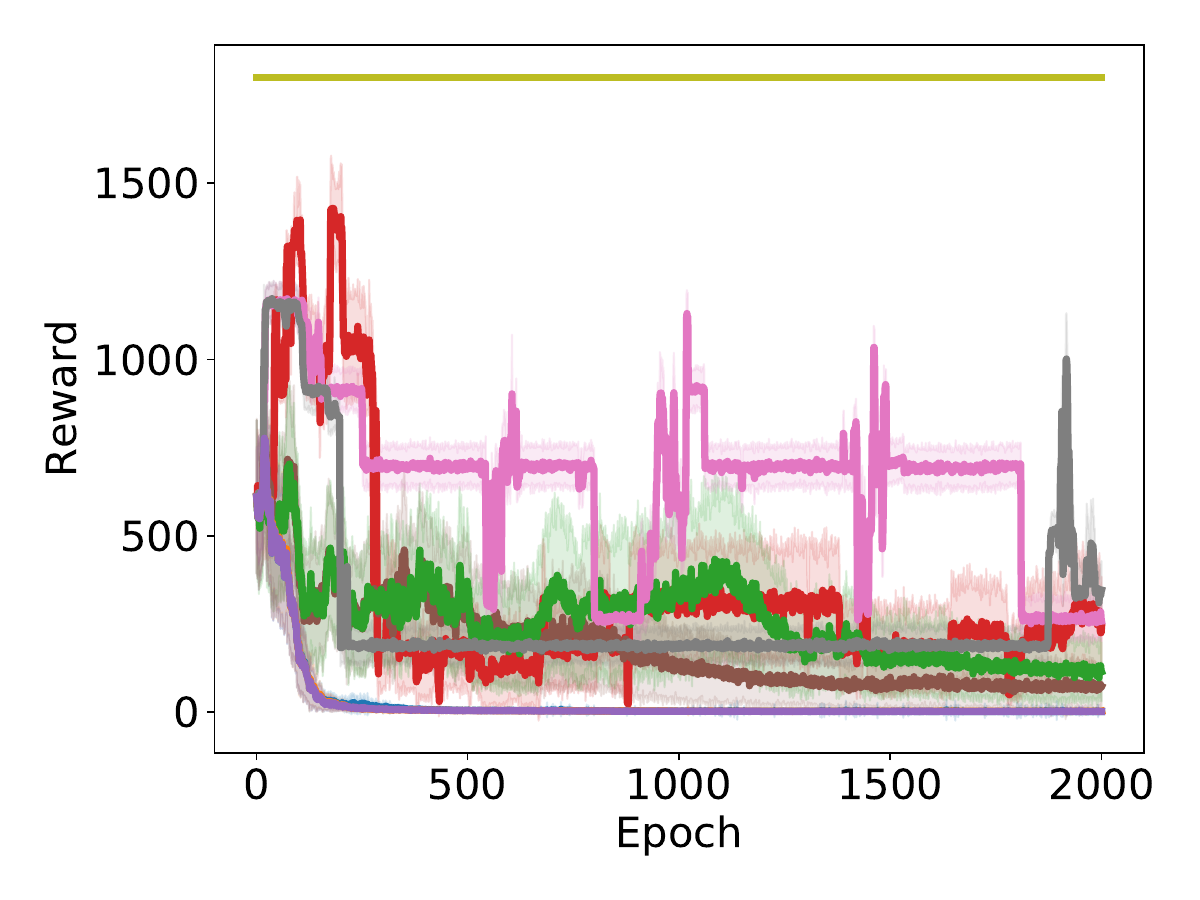}
        \includegraphics[width=0.49\textwidth]{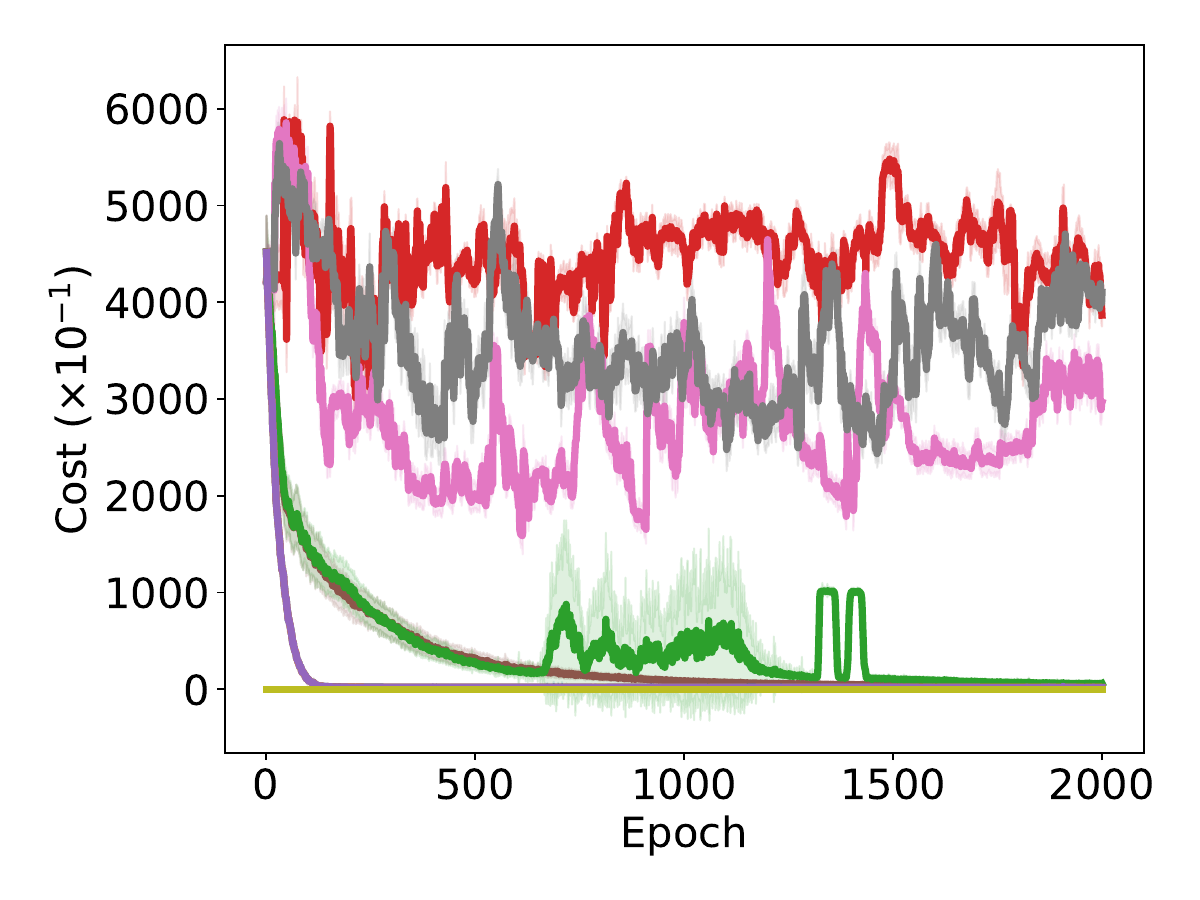}
        \caption{BlendingEnv}
        \label{fig:Blend}
    \end{subfigure}
    % Panel 4

%Row 4
\begin{subfigure}[t]{0.48\textwidth}
        \centering
        \includegraphics[width=0.49\textwidth]{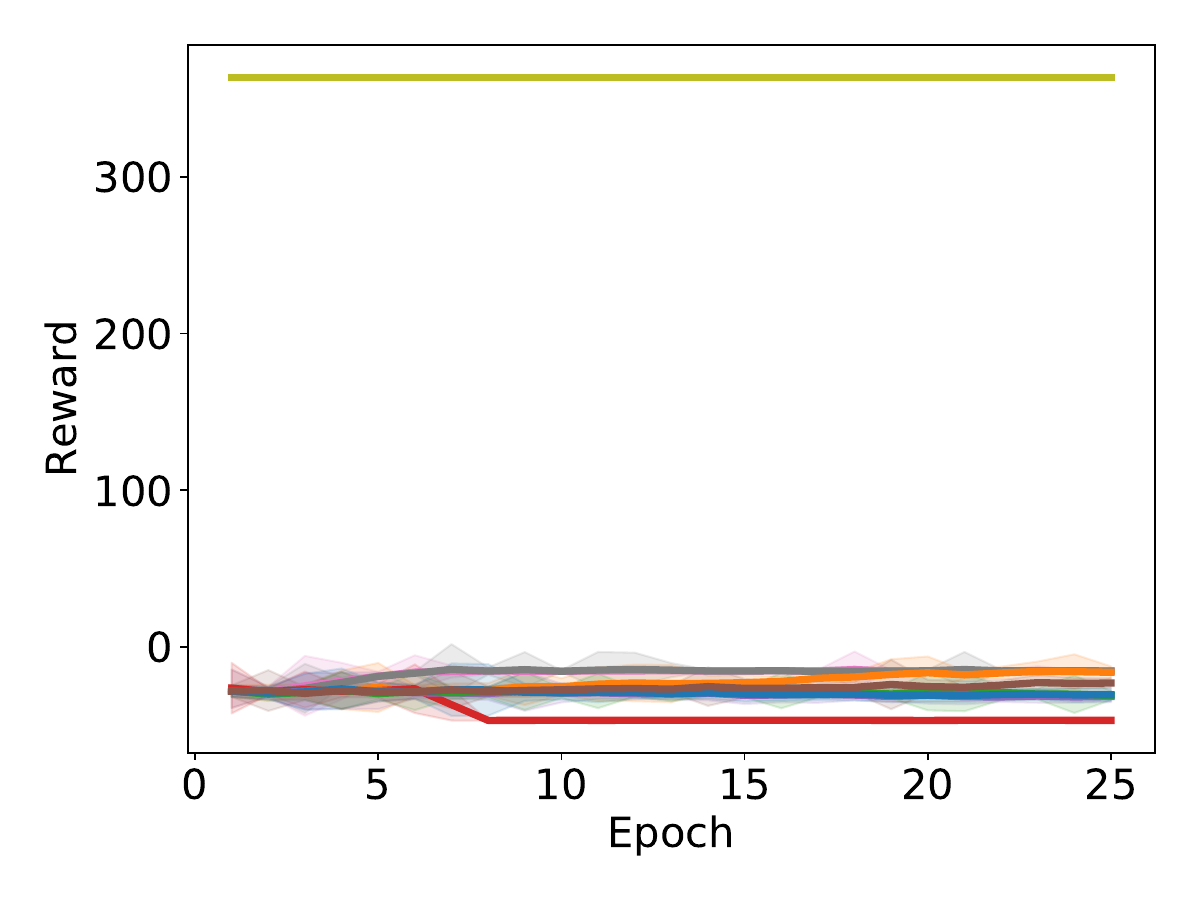}
        \includegraphics[width=0.49\textwidth]{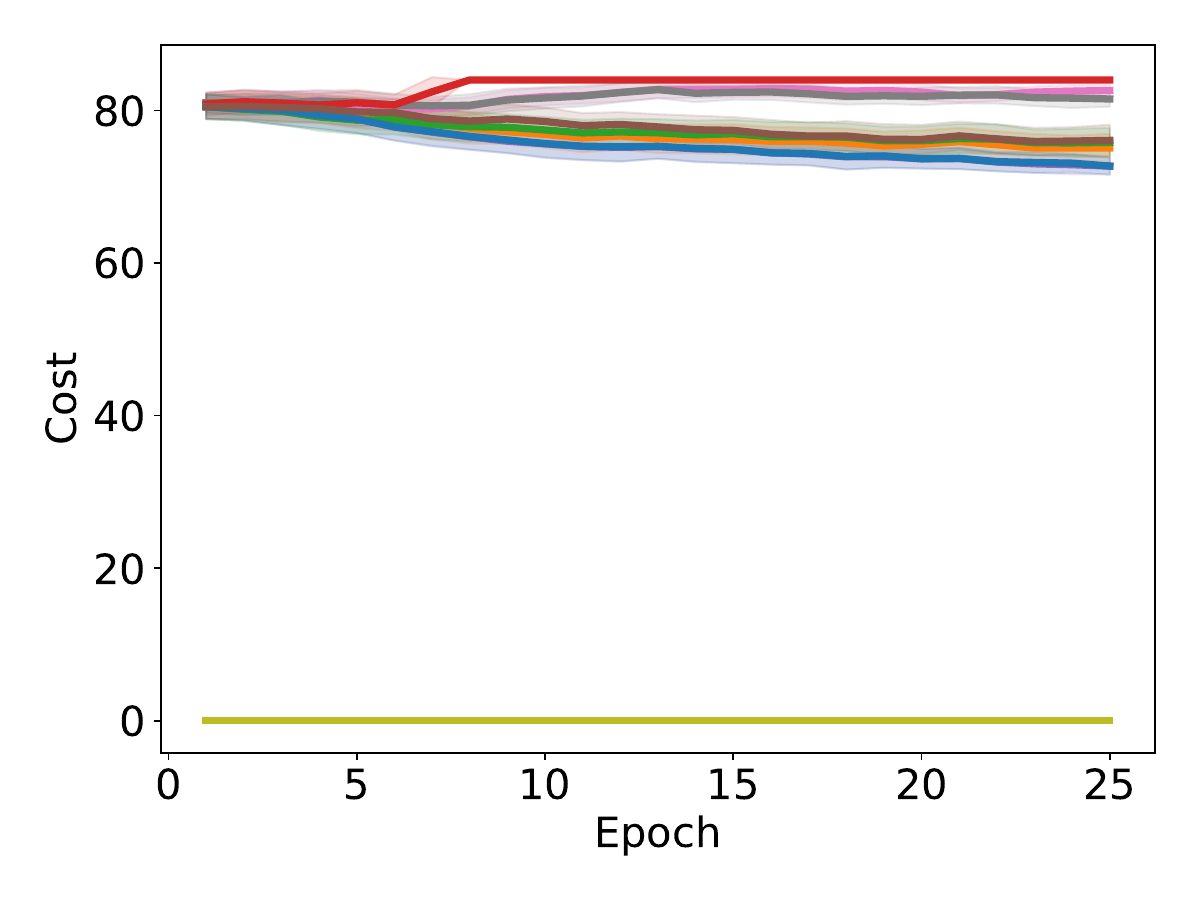}
        \caption{RTNEnv}
        \label{fig:RTN}
    \end{subfigure}
    \hfill
    % Panel 4
    \begin{subfigure}[t]{0.48\textwidth}
        \centering
        \includegraphics[width=0.49\textwidth]{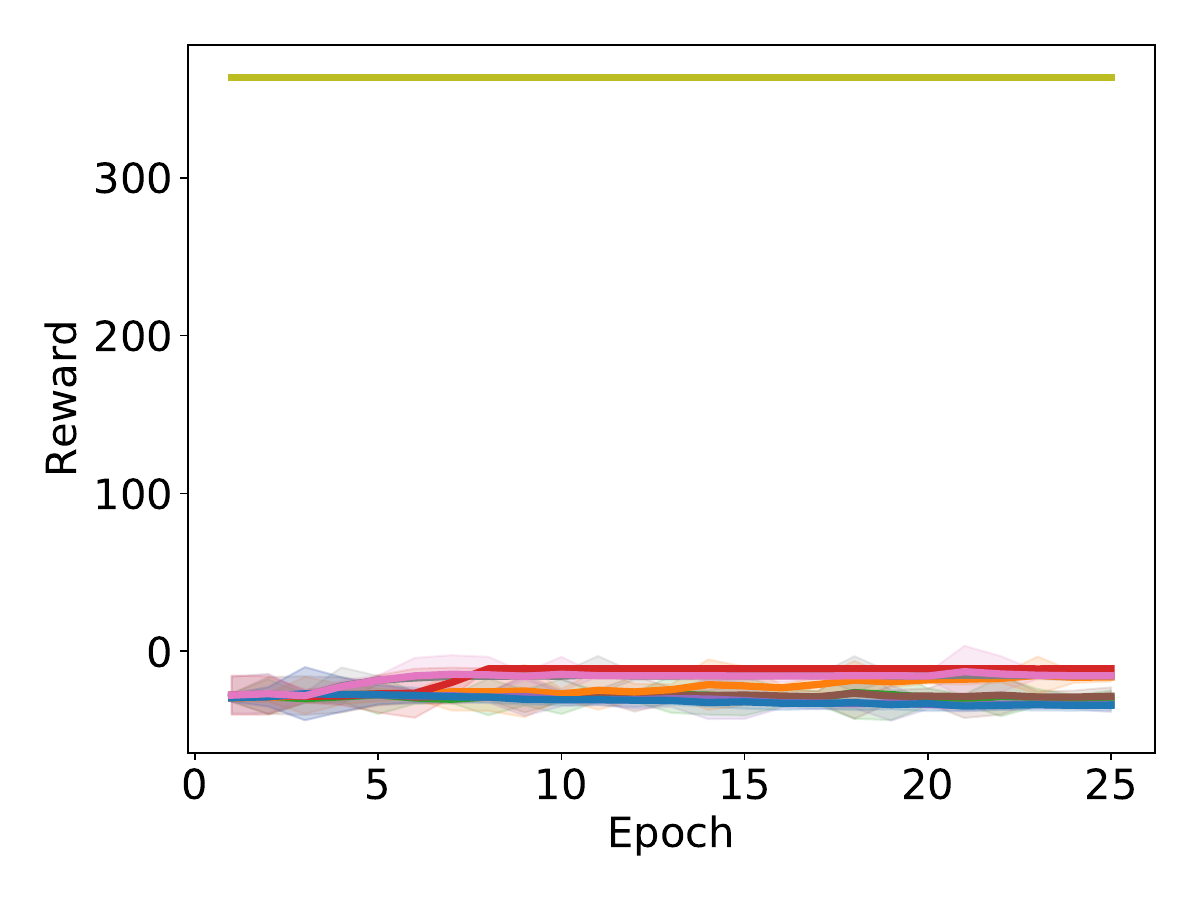}
        \includegraphics[width=0.49\textwidth]{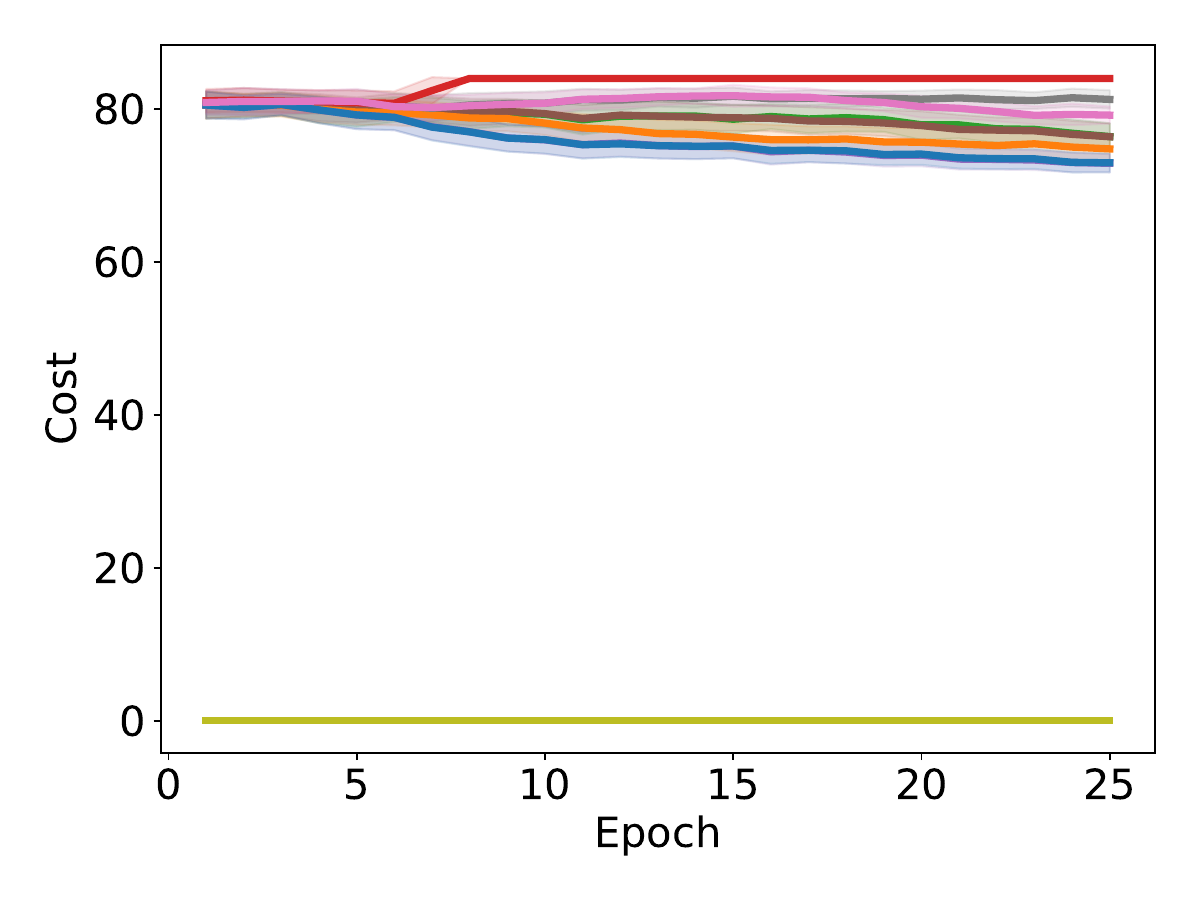}
        \caption{STNEnv}
        \label{fig:STN}
    \end{subfigure}
    \vspace{0.5em}
    \centering
    \includegraphics[width=1\textwidth]{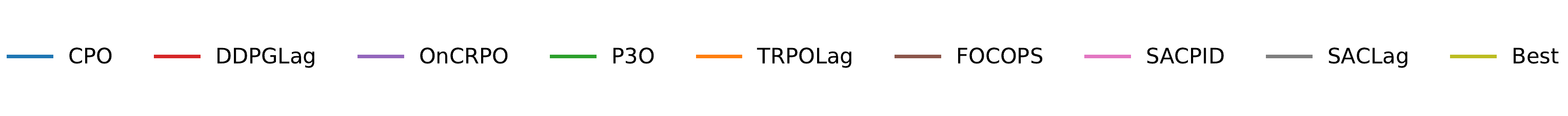}
    \caption{Training curves showing the average reward and cost per episode over training epochs across all case studies.}
    \label{fig:all_panels}
\end{figure}

We benchmark each algorithm's performance across environments during both training and evaluation. Figure~\ref{fig:all_panels} presents the average reward and cost per training epoch, with shaded regions indicating one standard deviation around the mean. 

\begin{figure}[htbp]
    \centering
    %ROw 1
    % Panel 1
    \begin{subfigure}[t]{0.48\textwidth}
        \centering
        \includegraphics[width=0.49\textwidth]{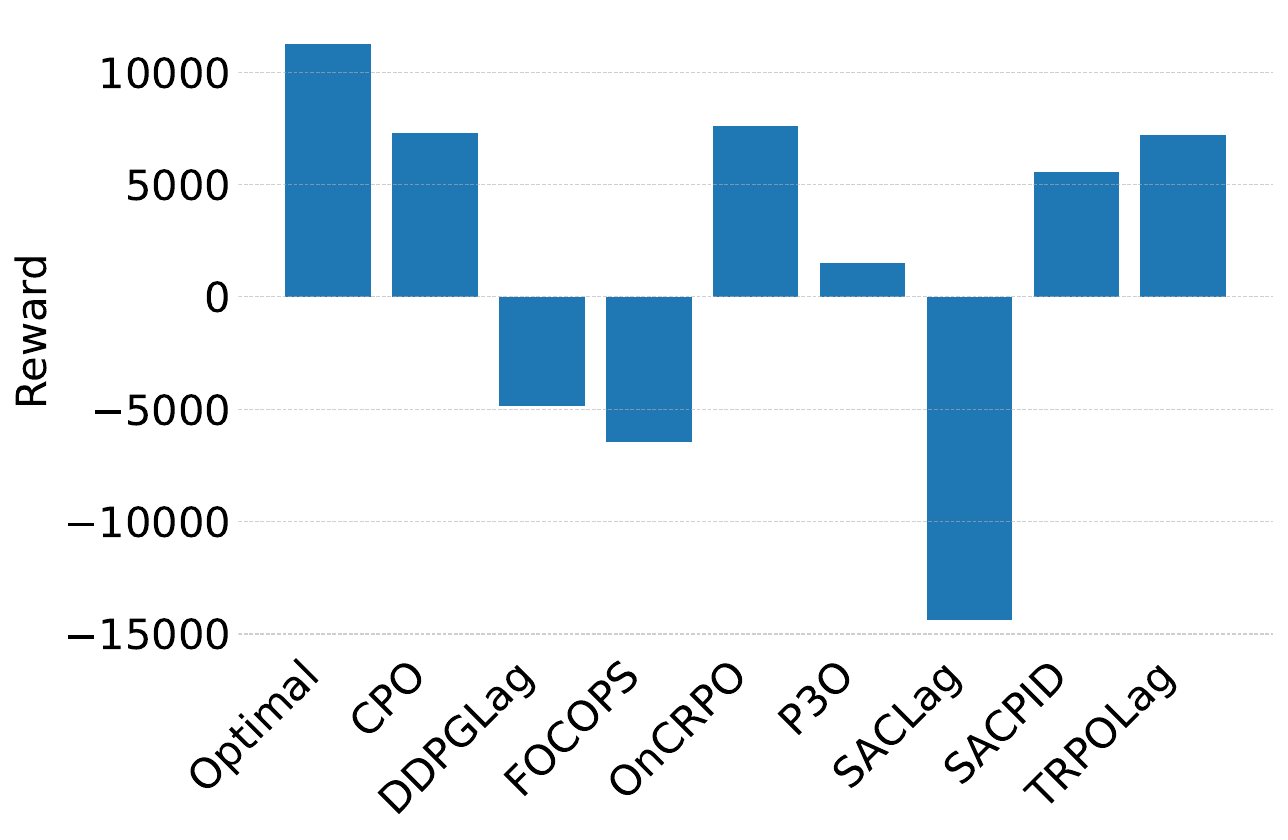}
        \includegraphics[width=0.49\textwidth]{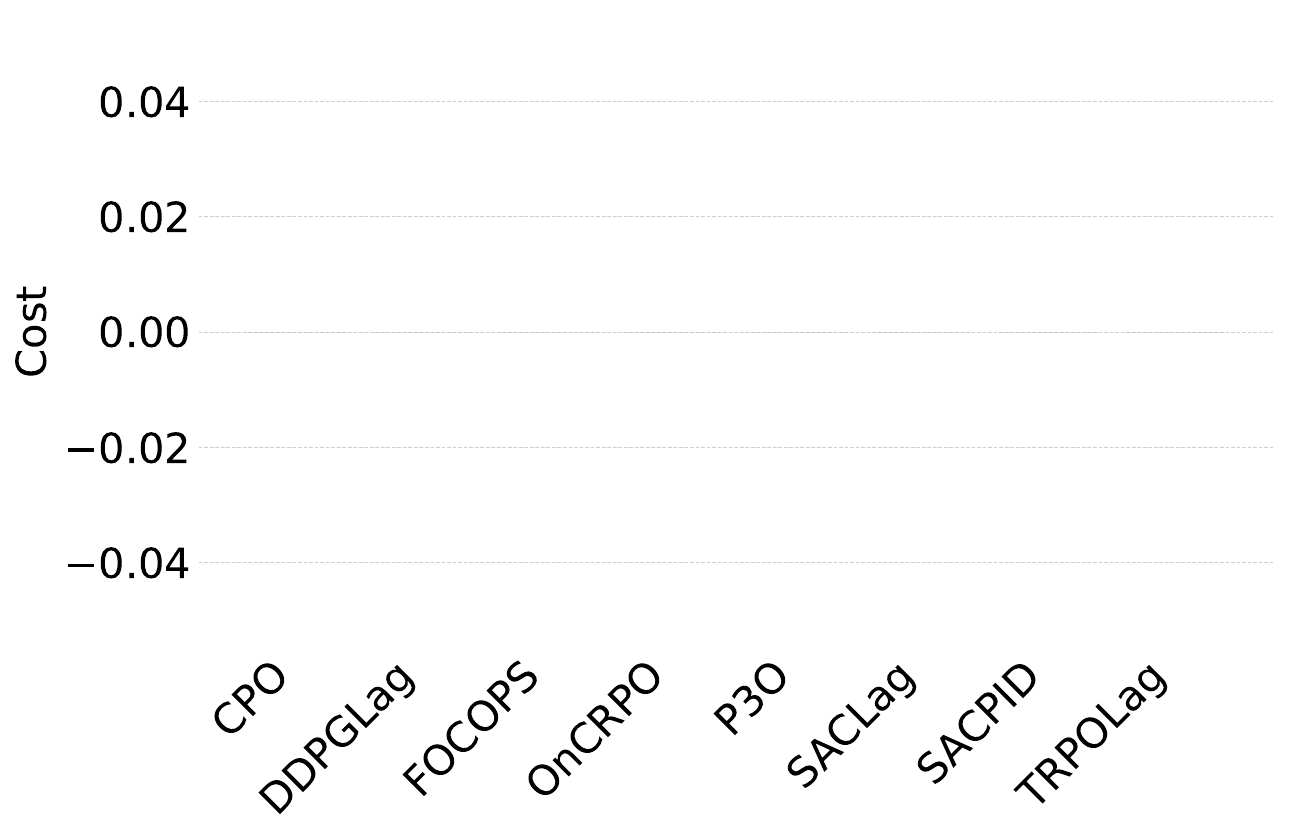}
        \caption{InvMgmtEnv}
        \label{fig:InvMgmtEnv_1}
    \end{subfigure}
    \hfill
    % Panel 2
    \begin{subfigure}[t]{0.48\textwidth}
        \centering
        \includegraphics[width=0.49\textwidth]{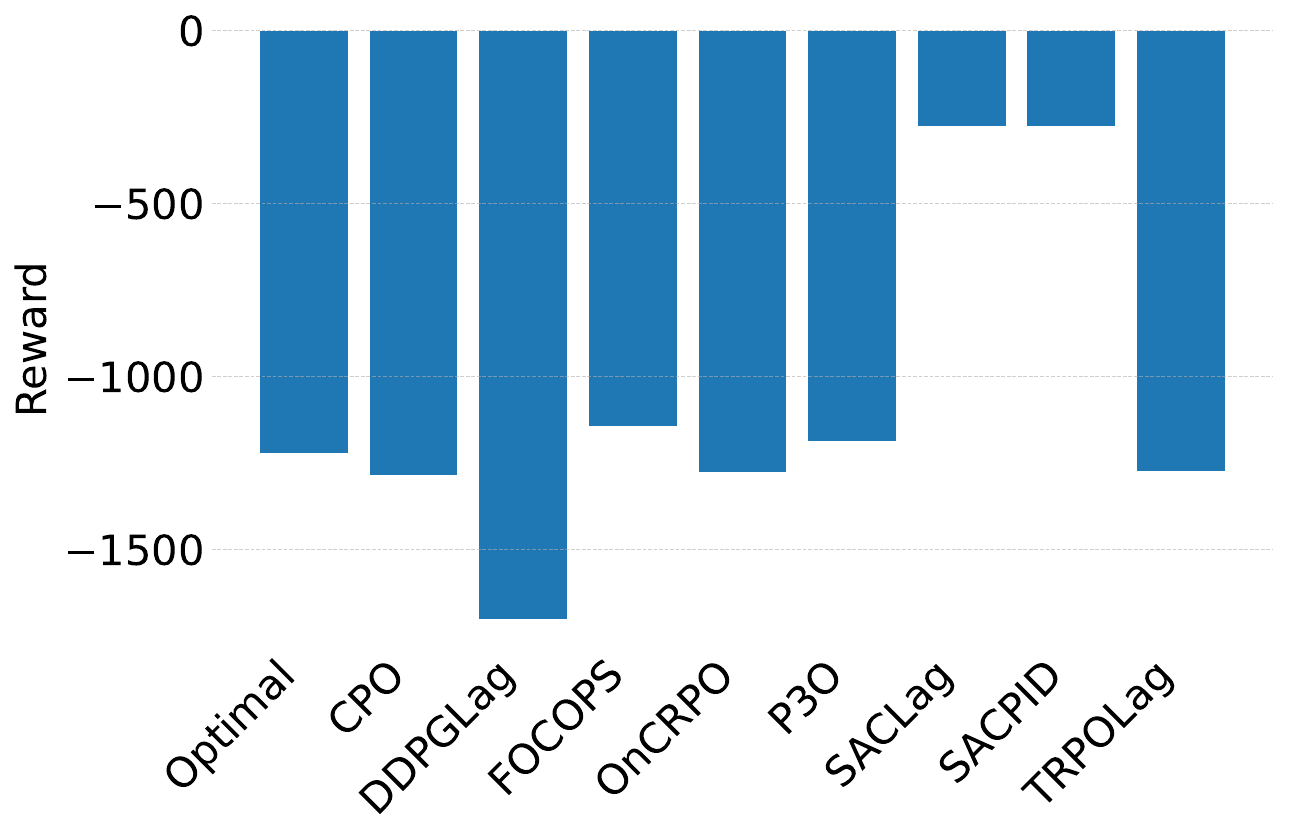}
        \includegraphics[width=0.49\textwidth]{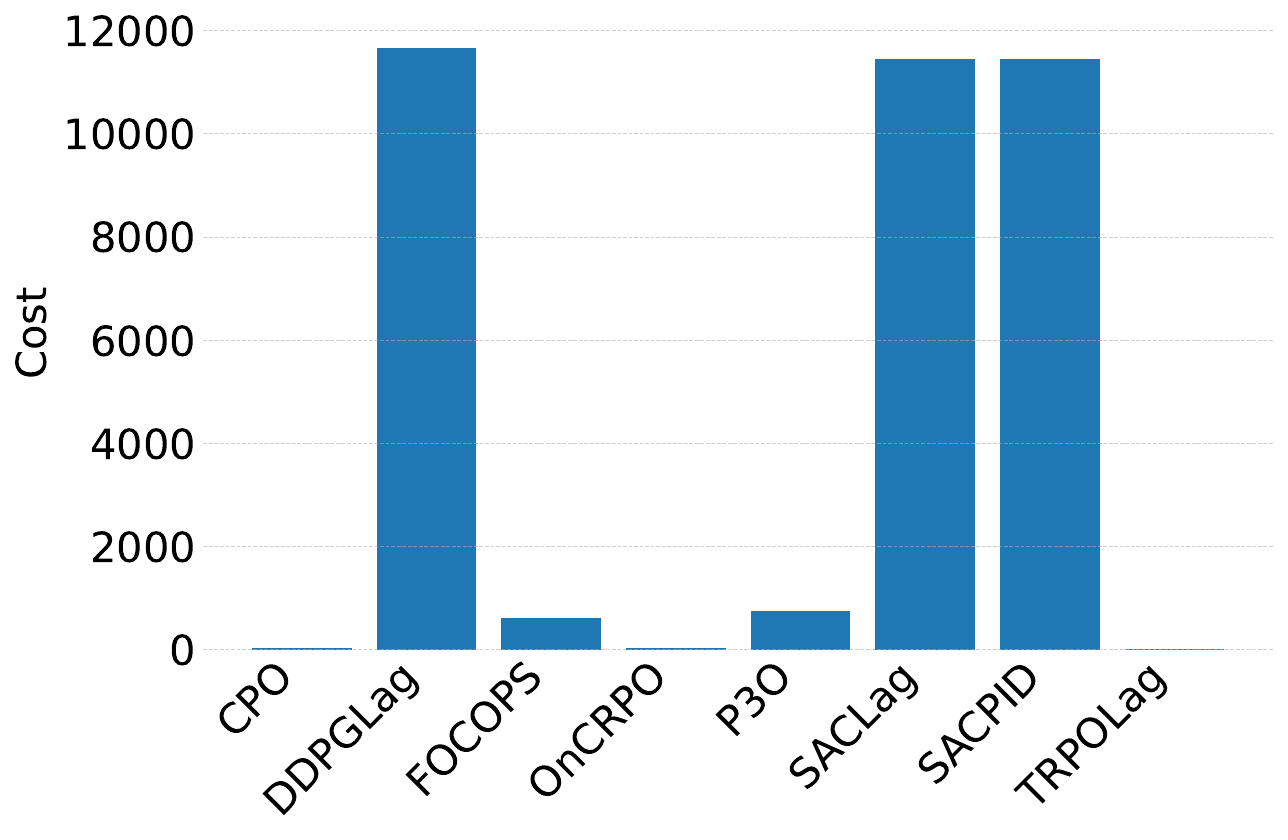}
        \caption{SchedMaintEnv-v0}
        \label{fig:SchedMaintEnv-v0_1}
    \end{subfigure}

    \vspace{1em}

    % ------------------ Row 2 ------------------
    % Panel 3
    \begin{subfigure}[t]{0.48\textwidth}
        \centering
        \includegraphics[width=0.49\textwidth]{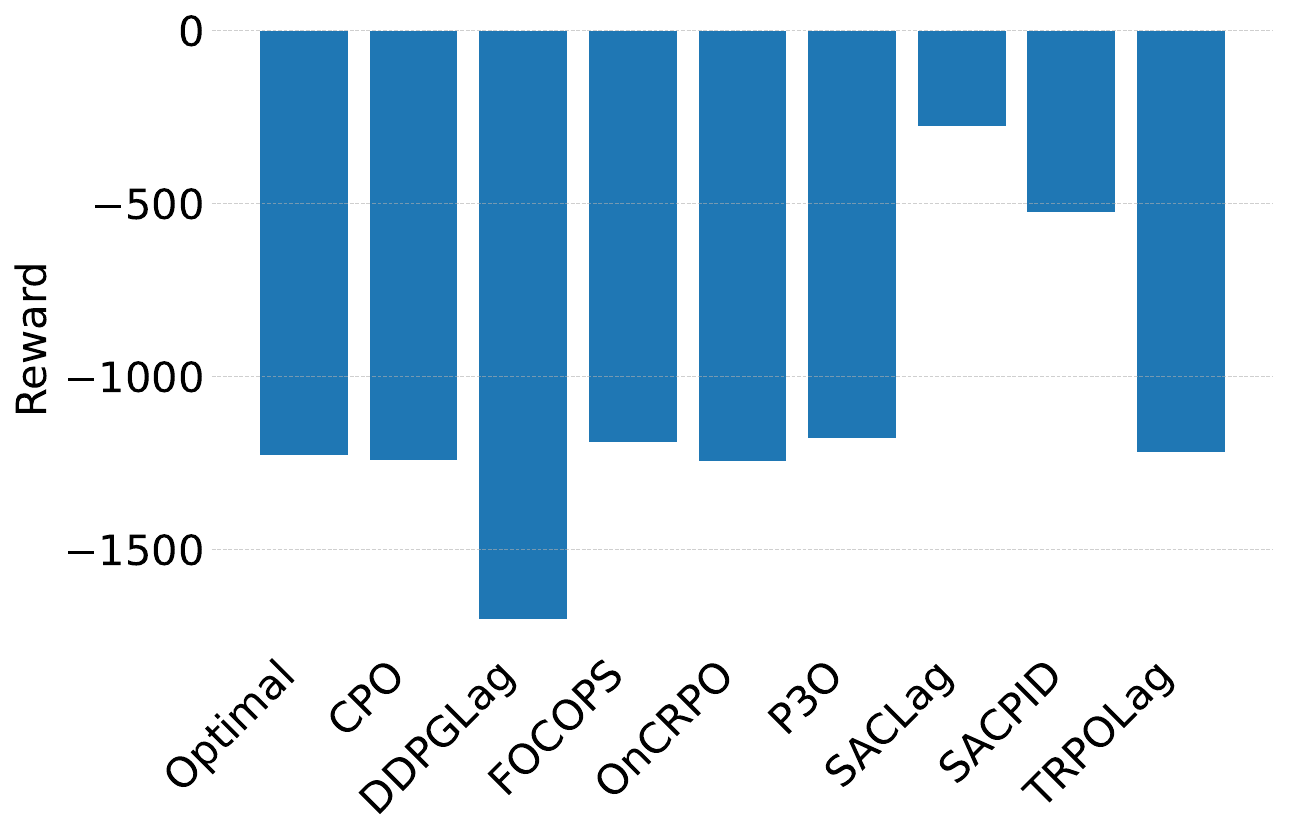}
        \includegraphics[width=0.49\textwidth]{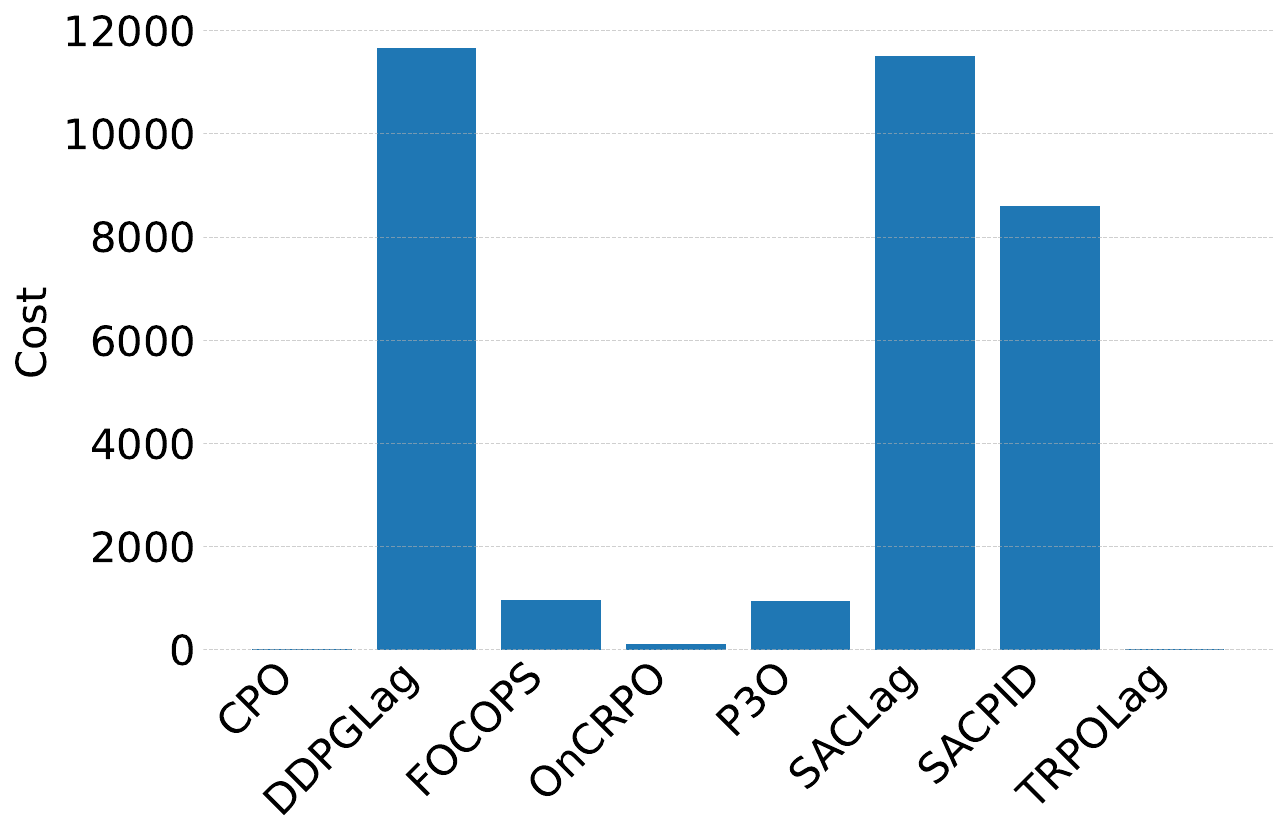}
        \caption{SchedMaintEnv-v1}
        \label{fig:SchedMaintEnv-v1_1}
    \end{subfigure}
    \hfill
    % Panel 4
    \begin{subfigure}[t]{0.48\textwidth}
        \centering
        \includegraphics[width=0.49\textwidth]{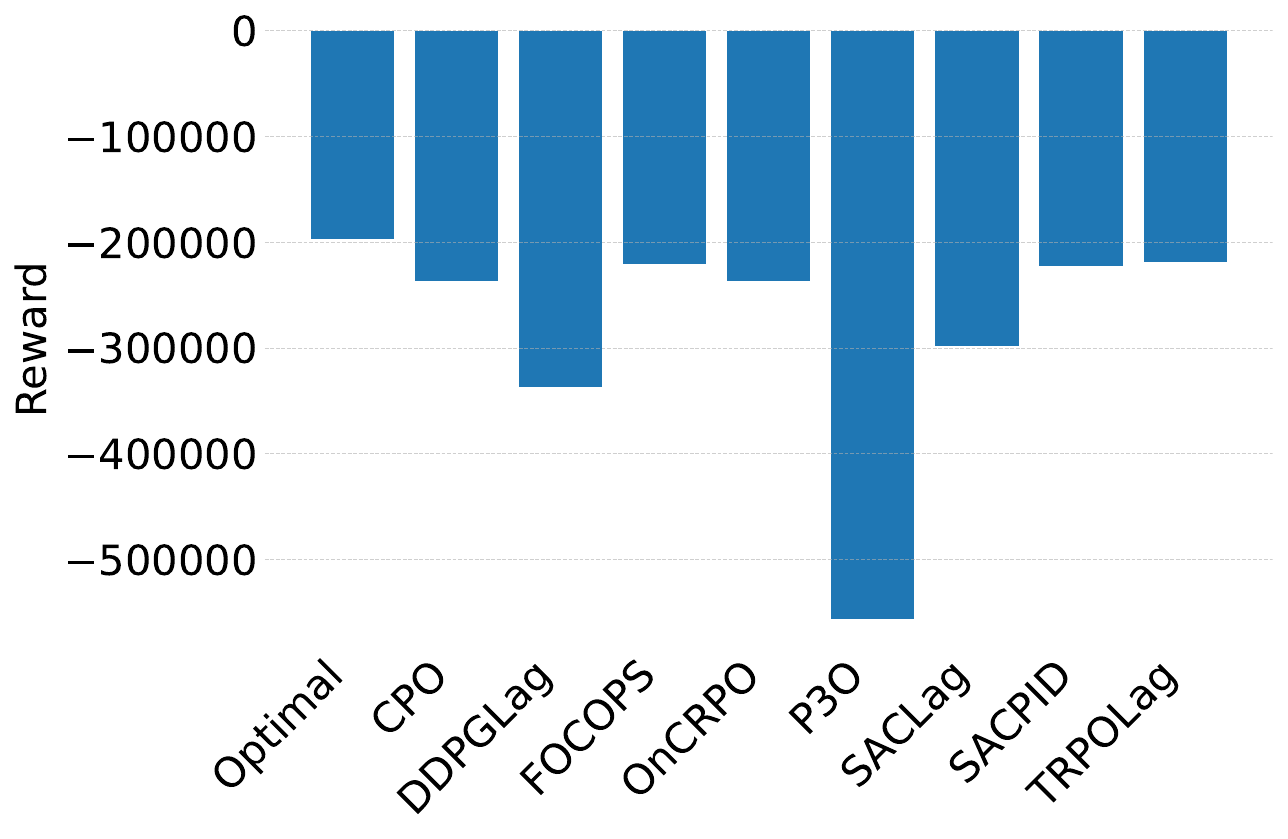}
        \includegraphics[width=0.49\textwidth]{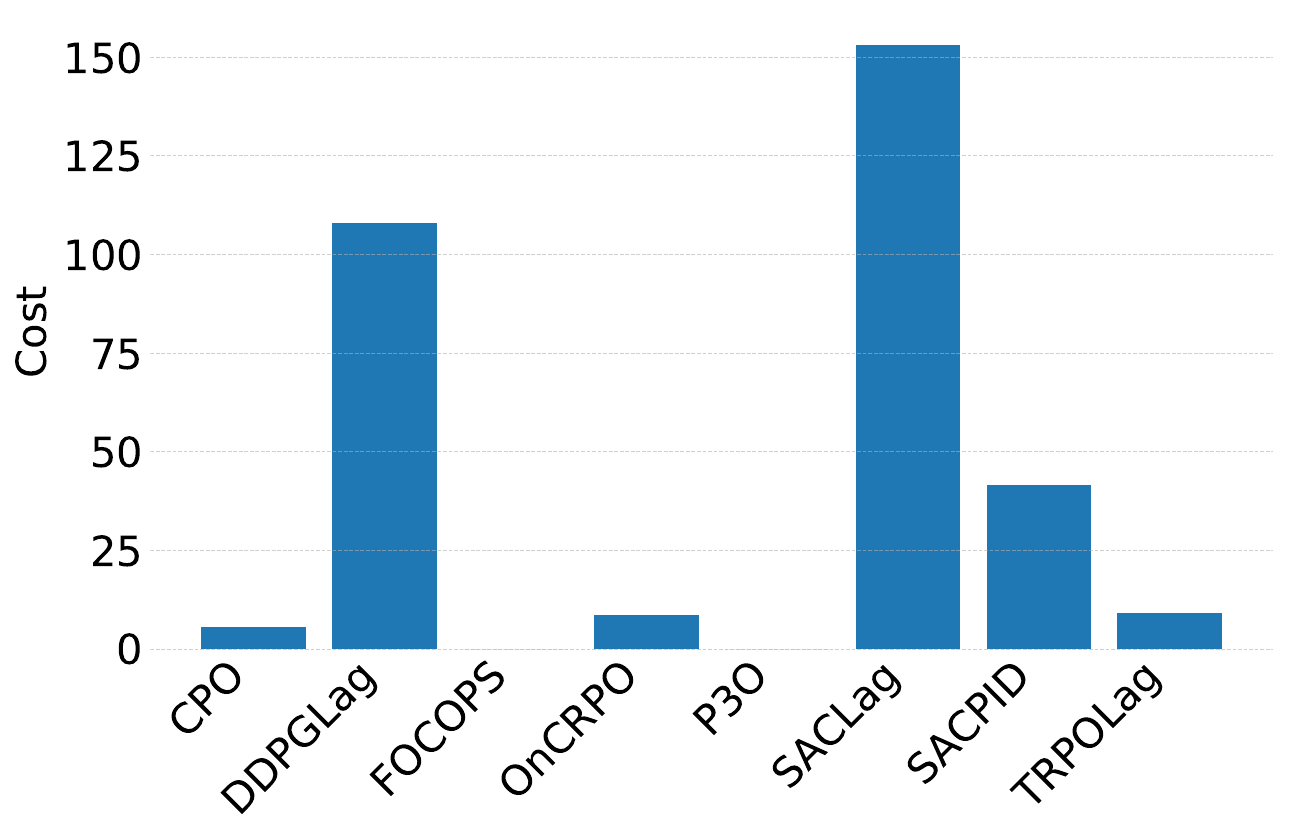}
        \caption{UCEnv}
        \label{fig:UCEnv_1}
    \end{subfigure}

%Row 3
 \begin{subfigure}[t]{0.48\textwidth}
        \centering
        \includegraphics[width=0.49\textwidth]{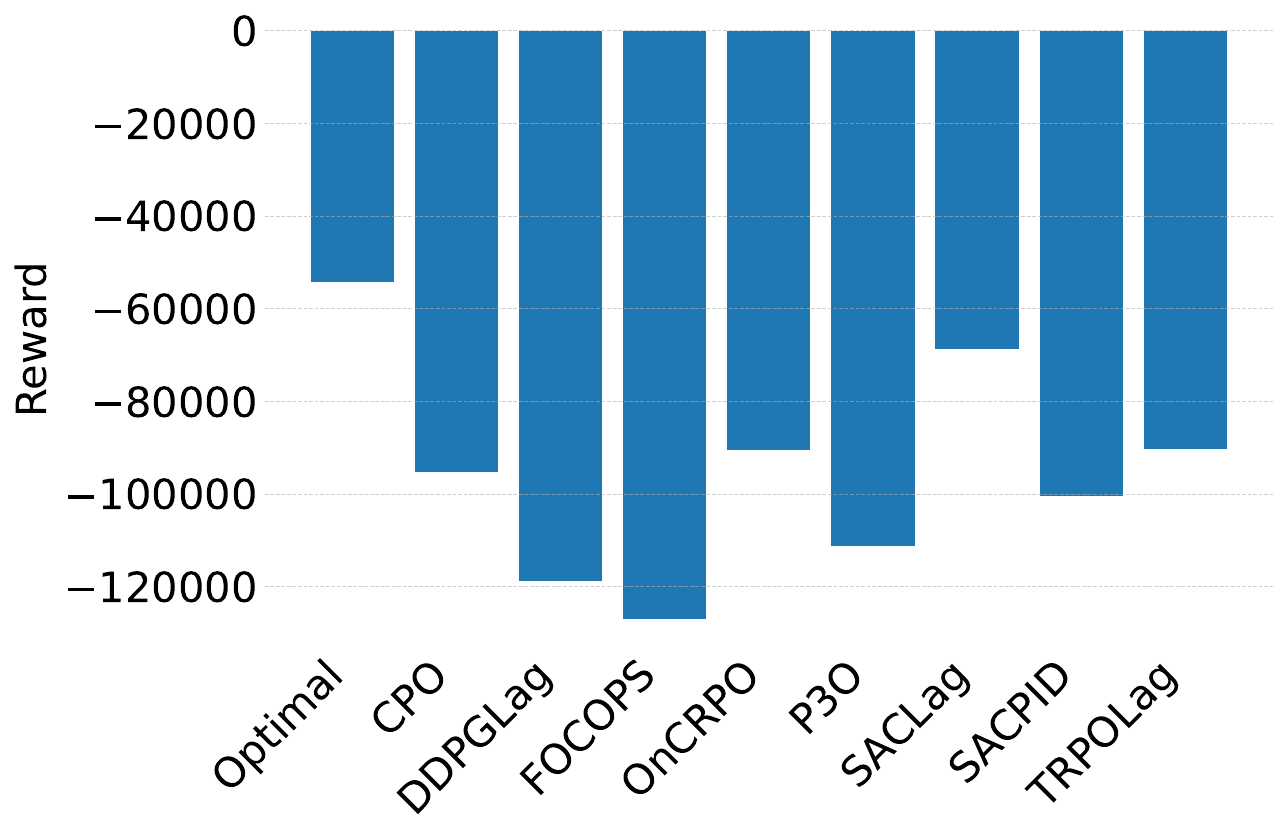}
        \includegraphics[width=0.49\textwidth]{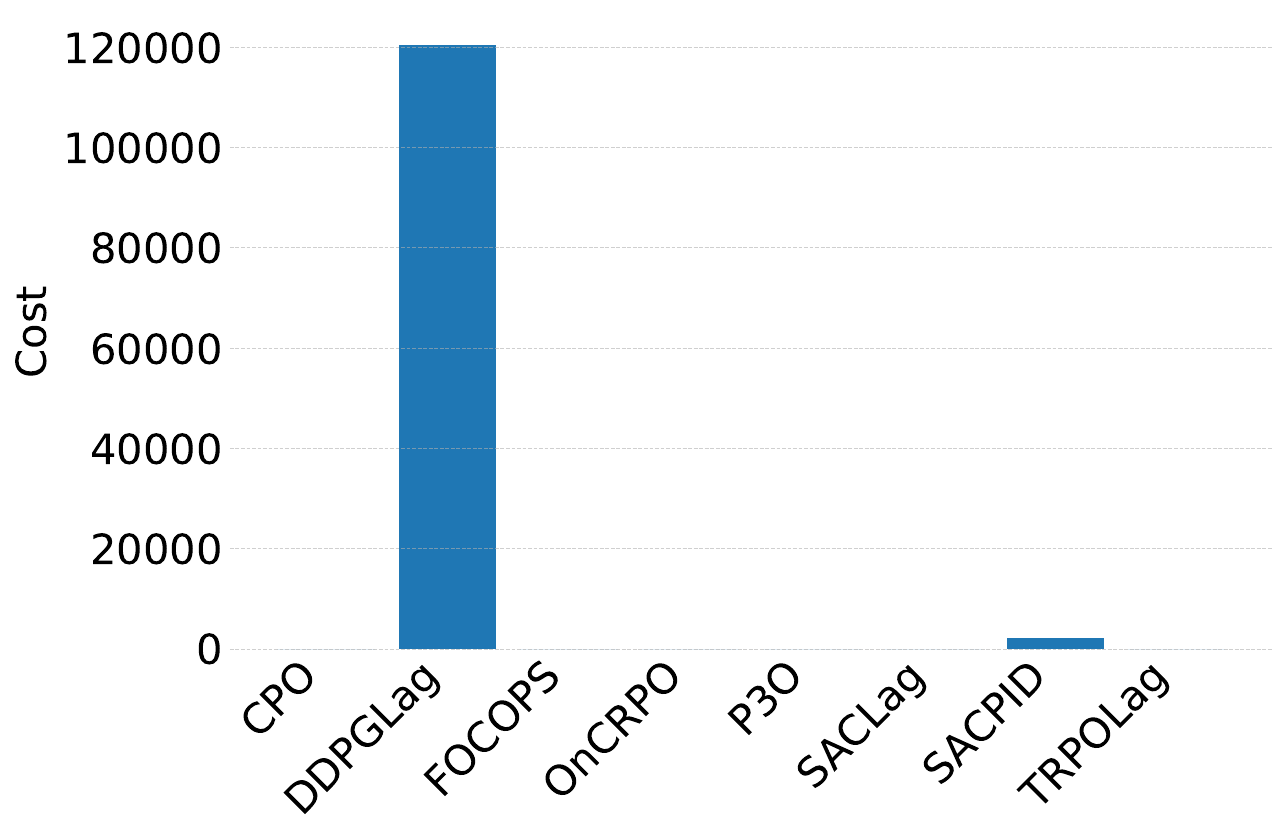}
        \caption{GridStorageEnv}
        \label{fig:Grid_1}
    \end{subfigure}
    \hfill
\begin{subfigure}[t]{0.48\textwidth}
        \centering
        \includegraphics[width=0.49\textwidth]{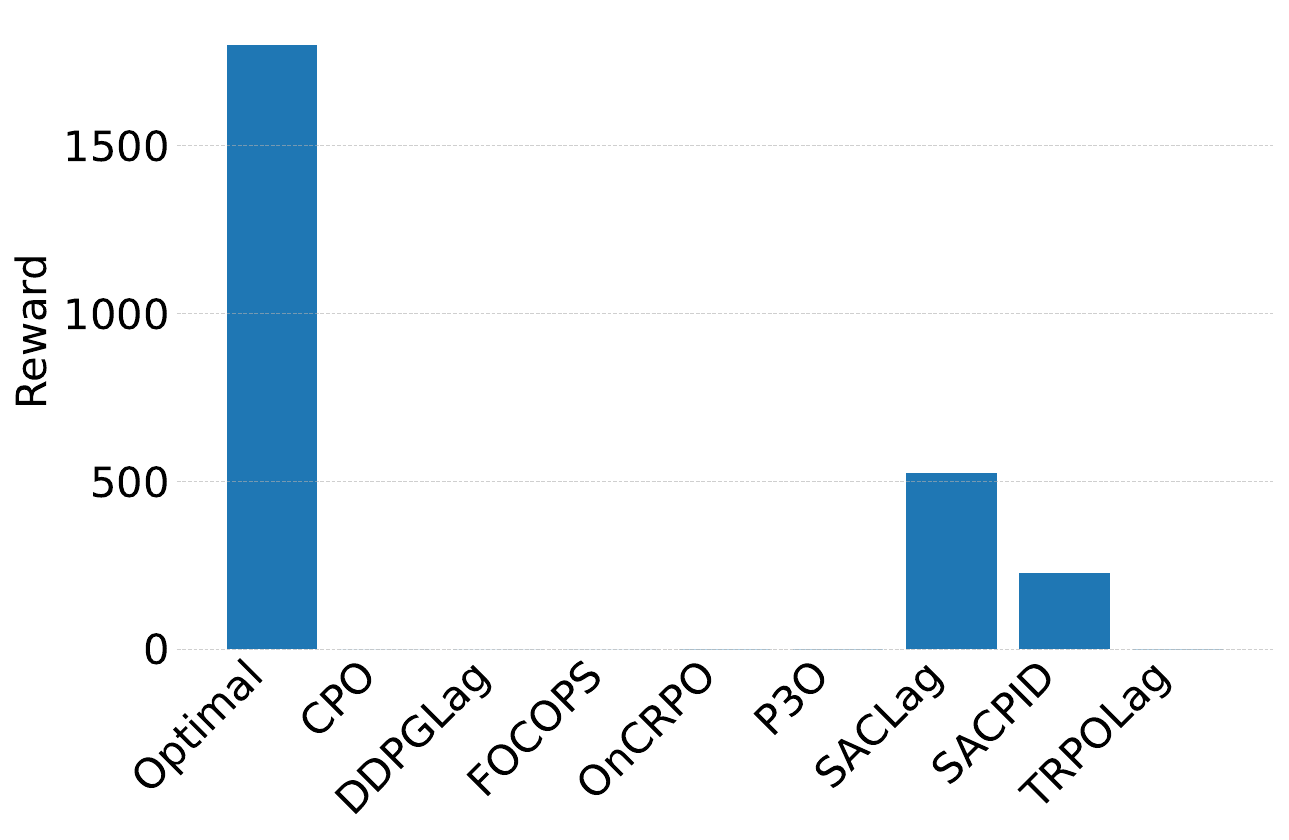}
        \includegraphics[width=0.49\textwidth]{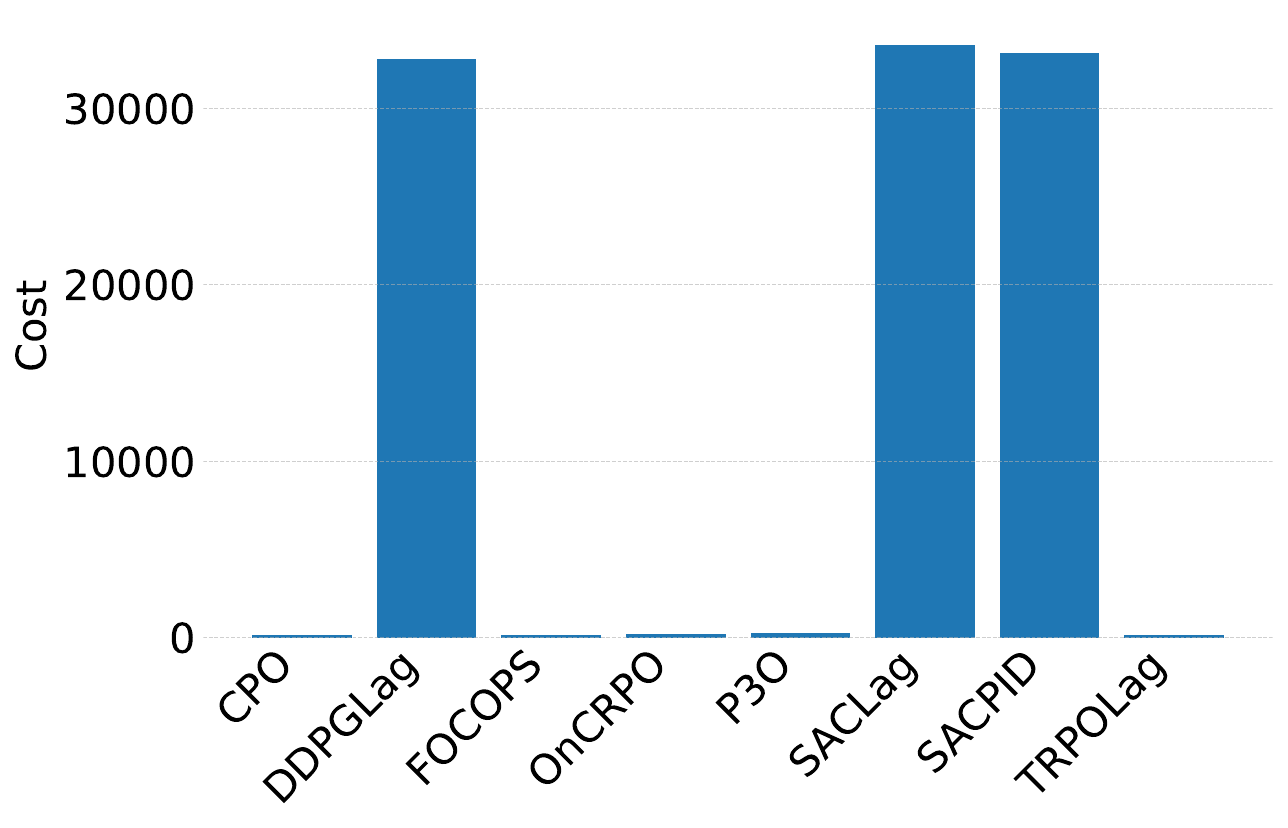}
        \caption{BlendingEnv}
        \label{fig:Blend_1}
    \end{subfigure}
    % Panel 4

%Row 4
\begin{subfigure}[t]{0.48\textwidth}
        \centering
        \includegraphics[width=0.49\textwidth]{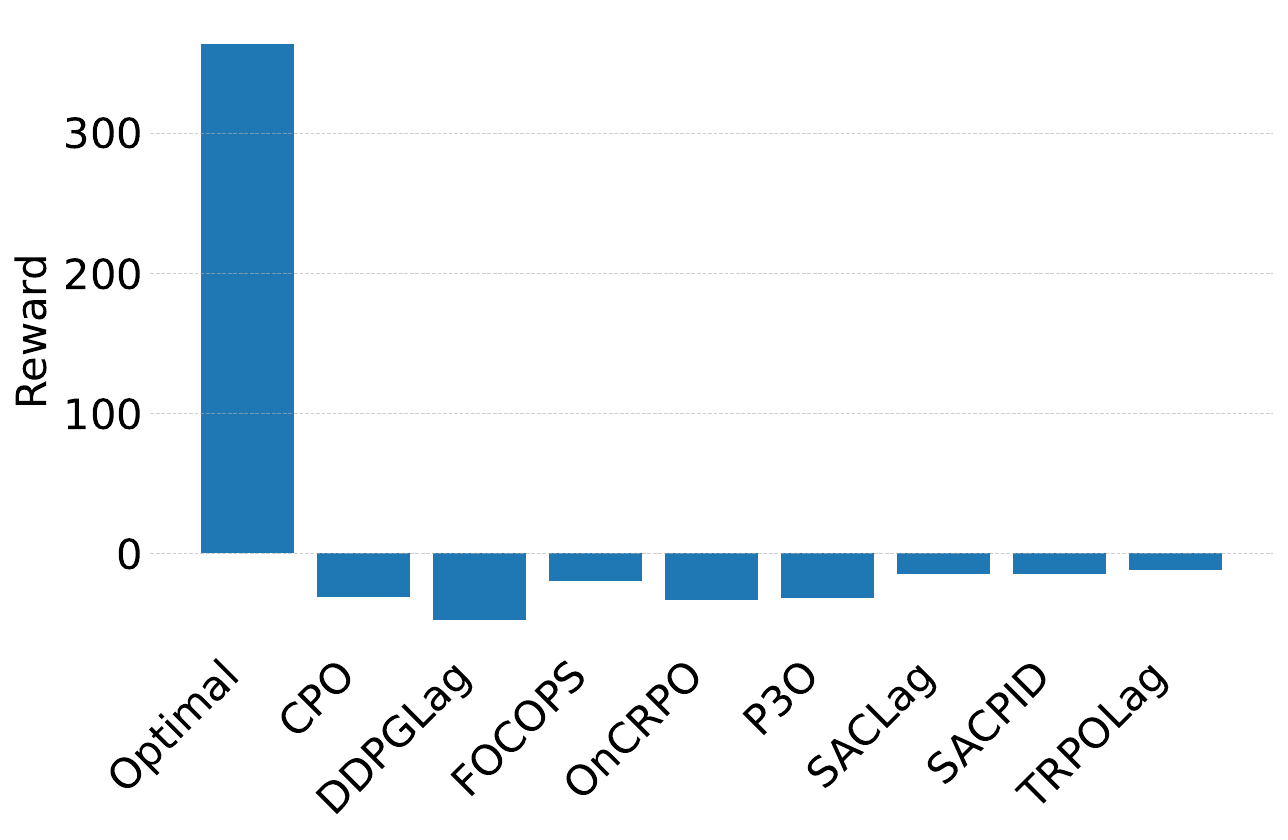}
        \includegraphics[width=0.49\textwidth]{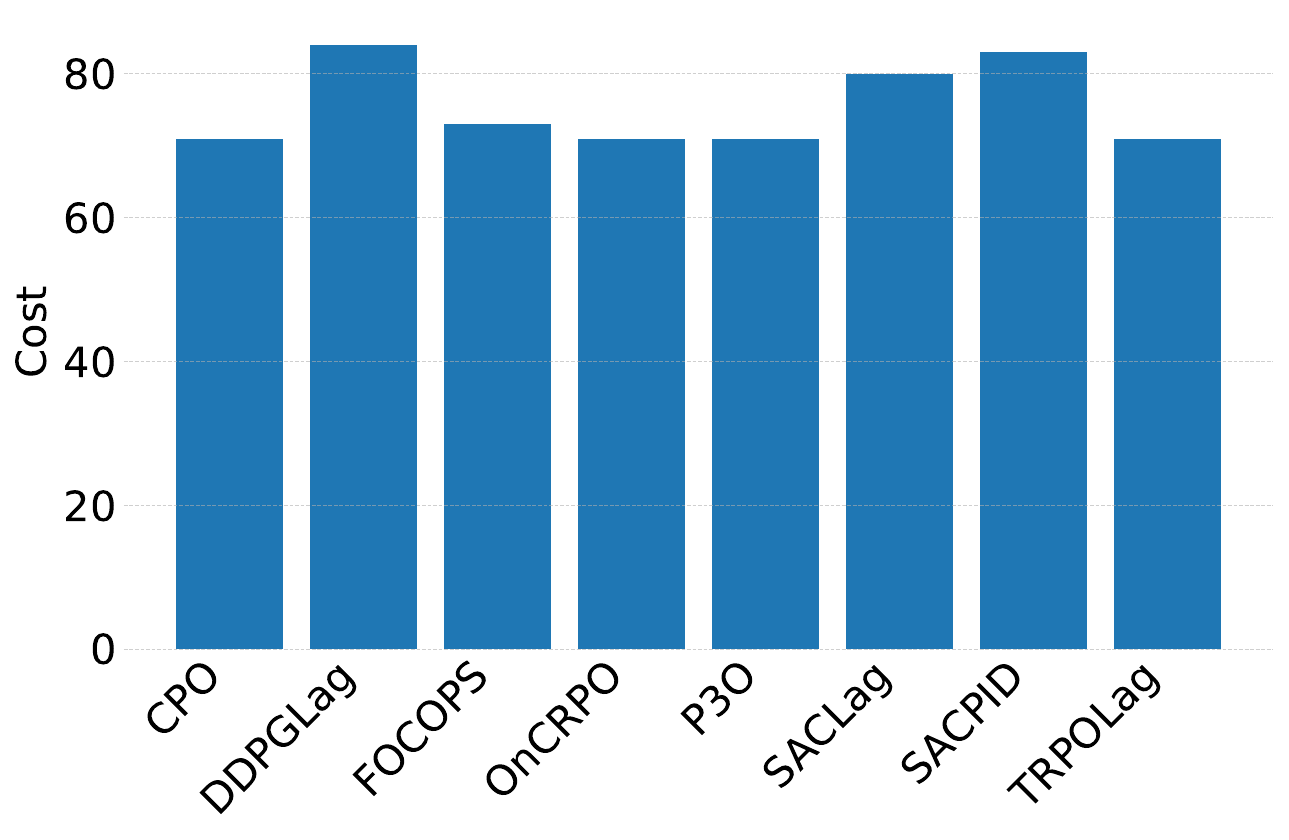}
        \caption{RTNEnv}
        \label{fig:RTN_1}
    \end{subfigure}
    \hfill
    % Panel 4
    \begin{subfigure}[t]{0.48\textwidth}
        \centering
        \includegraphics[width=0.49\textwidth]{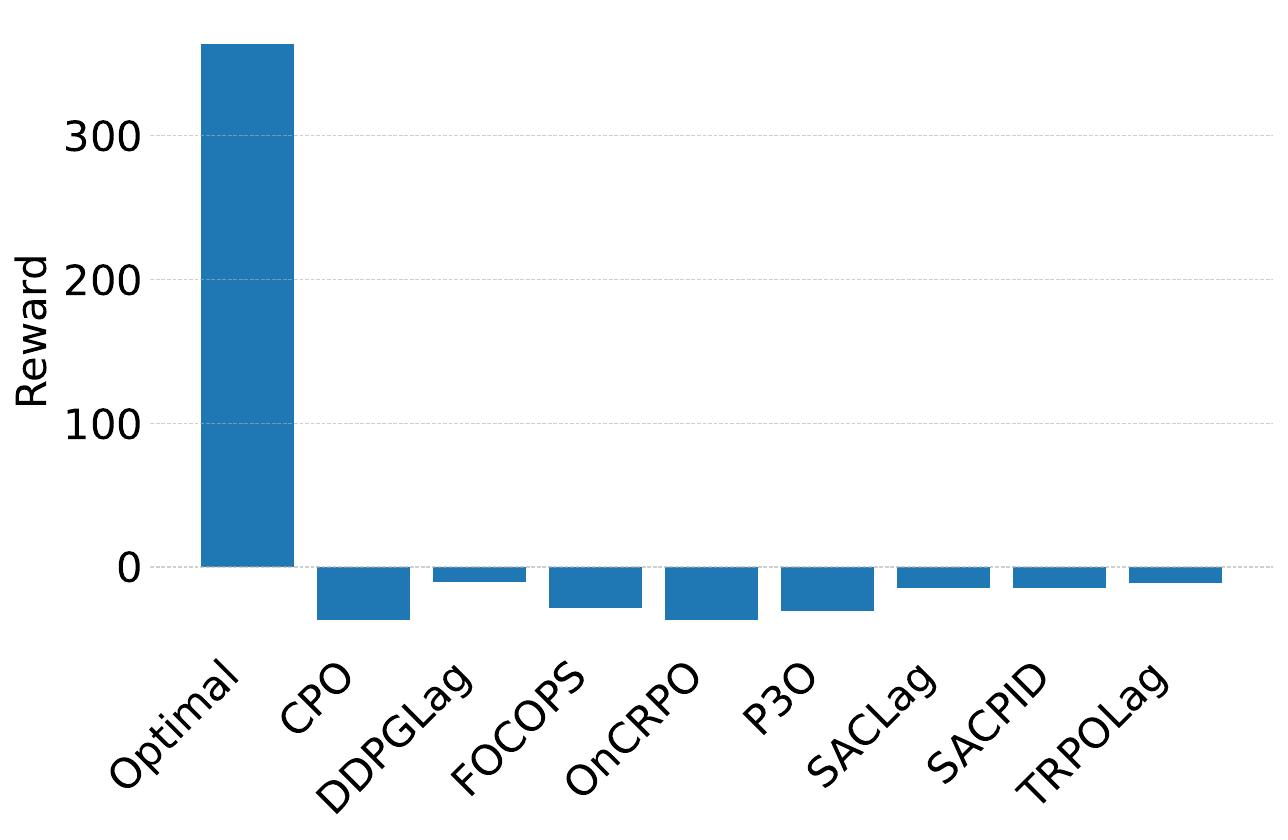}
        \includegraphics[width=0.49\textwidth]{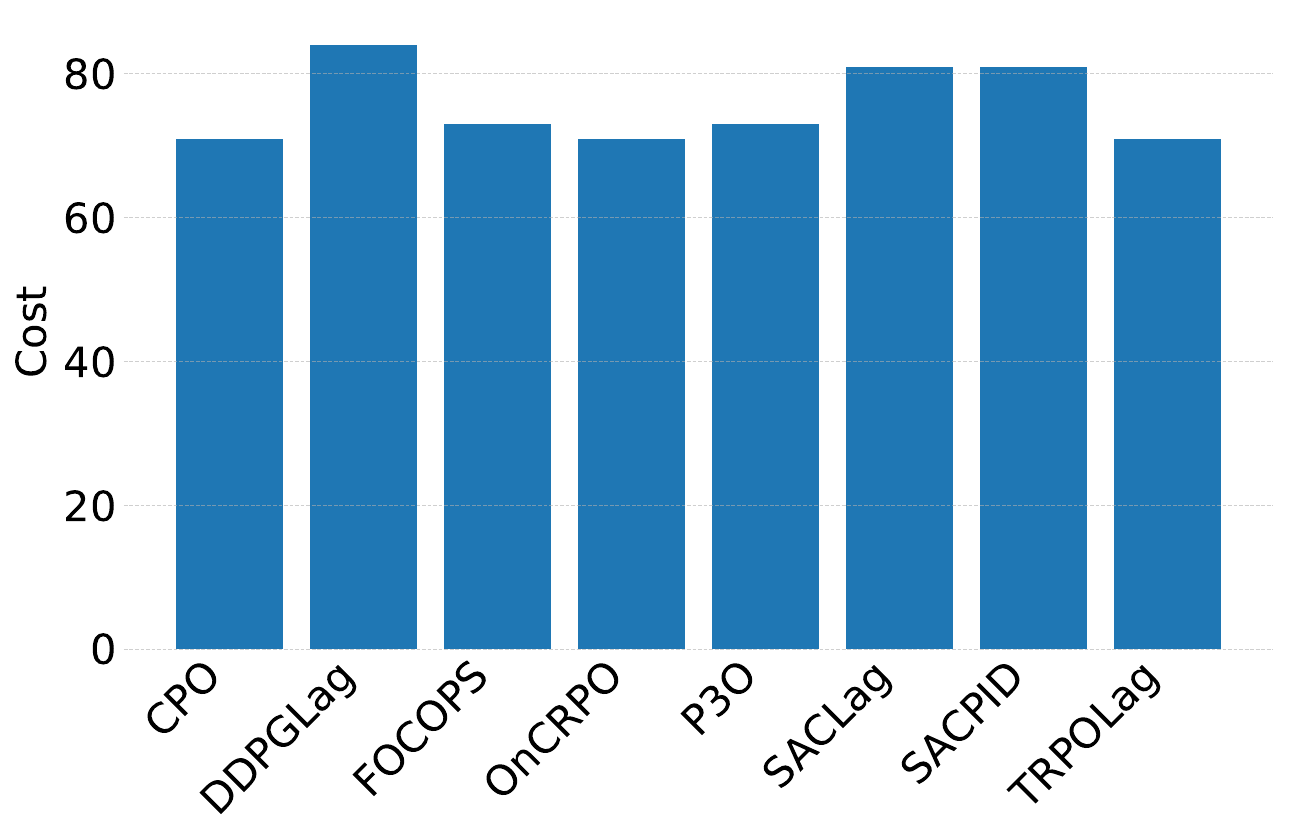}
        \caption{STNEnv}
        \label{fig:STN_1}
    \end{subfigure}
    
    \caption{Average evaluation rewards and costs}
    \label{fig:all_panels_evaluation}
\end{figure}

\clearpage
Figure~\ref{fig:all_panels_evaluation} presents the average evaluation reward and cost for each algorithm across a representative set of environments, computed over 10 evaluation episodes. For clarity of visualization, the figure reports only the mean values and does not display standard deviations. A complete tabular version of the results, including standard deviations (which are primarily non-negligible in the stochastic environments), is provided in Table~\ref{table:evaluation} in the appendix. The standard deviation of the evaluation rewards and costs is generally small in the deterministic environments, indicating consistent performance in the deterministic environments.

\textbf{Optimal reward of the environments:} One of the advantages of benchmarking safe RL algorithms using environments based on deterministic opeartions research problems is that state-of-the-art optimization solvers such as Gurobi \citep{gurobi} can be used to solve the nonconvex problems to global optimality while strictly enforcing all the constraints. The optimal reward from the optimization solvers can be seen as the ``ground truth'' of the environments, shown in the first column of Table~\ref{table:evaluation} as well as the first bar of the represetation of rewards in Figure ~\ref{fig:all_panels_evaluation}. The optimal reward of \texttt{SchedMaintEnv-v1} is obtained by assuming a ``perfect information'' lookahead using the Gurobi solver, which provide a theoretical upper bound of the optimal expected reward.

\textbf{Evaluation Criteria:} \label{section:eval_criteria}

We evaluate safe RL algorithms under two competing objectives: (i) reward maximization and (ii) constraint violation minimization. Since these objectives are inherently conflicting, aggregating them into a single scalar metric can obscure meaningful trade-offs. We therefore adopt a multi-objective evaluation framework.

For each environment, our evaluation proceeds in three steps:
\begin{enumerate}
    \item \textbf{Pareto identification:} We identify Pareto-efficient algorithms in the reward–violation space. An algorithm is Pareto-efficient if no other method achieves both higher reward and lower violation.
    
    \item \textbf{Scalarization of trade-offs:} To compare Pareto-efficient methods, we use a convex scalarization
    \[
    S_\alpha = \alpha \tilde{R} - (1-\alpha)\tilde{C}, \quad \alpha \in [0,1],
    \]
    where $\tilde{R}$ and $\tilde{C}$ denote normalized reward and violation, respectively. The parameter $\alpha$ captures the relative preference between performance and safety. We normalize reward and violation to $[0,1]$ within each environment and evaluate $S_\alpha$ over a uniform grid of 101 values of $\alpha$ in $[0,1]$, including 0, 0.5, and 1.
    
    \item \textbf{Selection of best-performing algorithm:} 
For each $\alpha$, we identify the Pareto-efficient algorithm that maximizes $S_\alpha$. 
Algorithms that achieve the highest scalarized score over the largest portion of the $\alpha$-range are classified as the representative best-performing methods, reflecting robustness across a wide range of trade-off preferences. 
The best-performing algorithms for each environment discussed in the main paper are summarized in Table~\ref{table:best_algo}. Note that normalization may reduce large differences in constraint violations, occasionally favoring high-reward methods over a broader range of $\alpha$.
\end{enumerate}

\begin{table}[htbp]
\centering
    \caption{Best performing algorithm}
    \label{table:best_algo}
\begin{tabular}{lll}
Environment & Pareto efficient algorithms & Best performing algorithm \\
\hline 
InvMgmtEnv & OnCRPO & OnCRPO \\
SchedMaintEnv-v0 & FOCOPS, SACLag, SACPID, TRPOLag & SACLag,SACPID \\
SchedMaintEnv-v1 & CPO, P3O, SACLag, SACPID, TRPOLag & TRPOLag \\
UCEnv & FOCOPS, TRPOLag & FOCOPS \\
GridStorageEnv & SACLag, TRPOLag & SACLag \\
BlendingEnv & FOCOPS, P3O, SACLag, SACPID, TRPOLag & FOCOPS \\
RTNEnv & TRPOLag & TRPOLag \\
STNEnv & DDPGLag, TRPOLag & TRPOLag
\end{tabular}
\end{table}

\begin{figure}[htbp]
\centering

\begin{subfigure}[t]{0.24\textwidth}
    \centering
    \includegraphics[width=\linewidth]{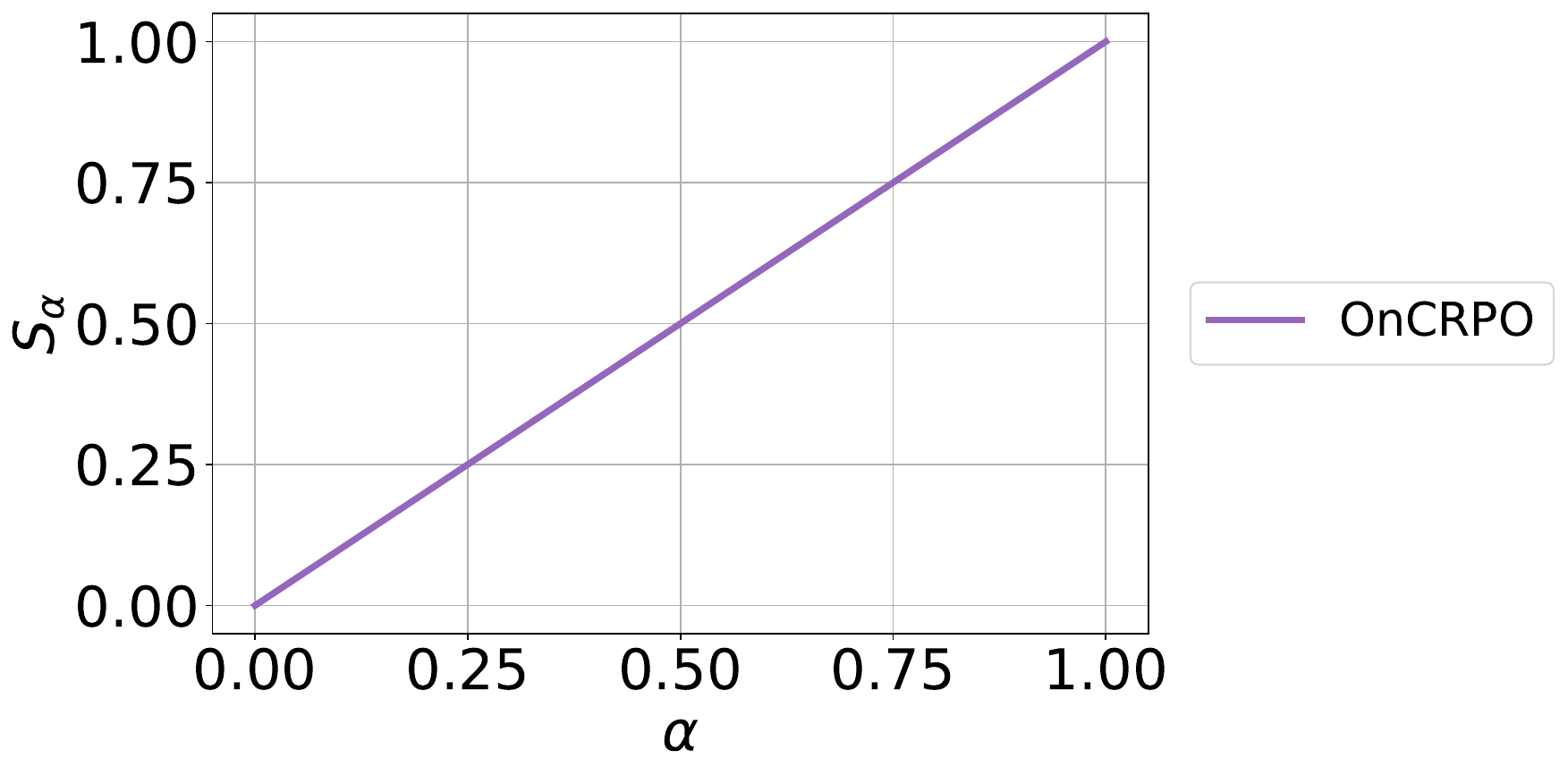}
    \caption{InvMgmtEnv}
\end{subfigure}
\hfill
\begin{subfigure}[t]{0.24\textwidth}
    \centering
    \includegraphics[width=\linewidth]{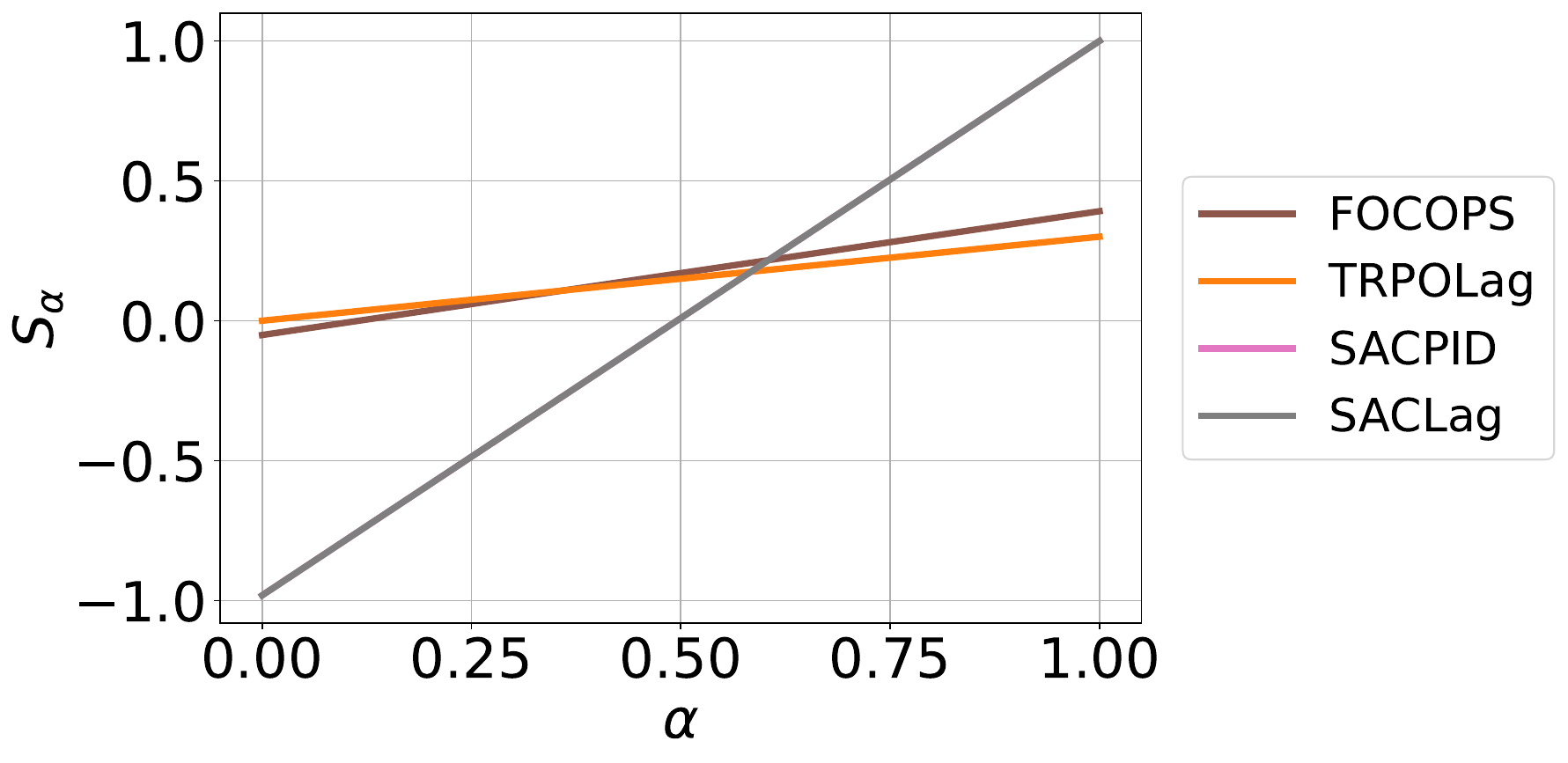}
    \caption{SchedMaintEnv-v0}
\end{subfigure}
\hfill
\begin{subfigure}[t]{0.24\textwidth}
    \centering
    \includegraphics[width=\linewidth]{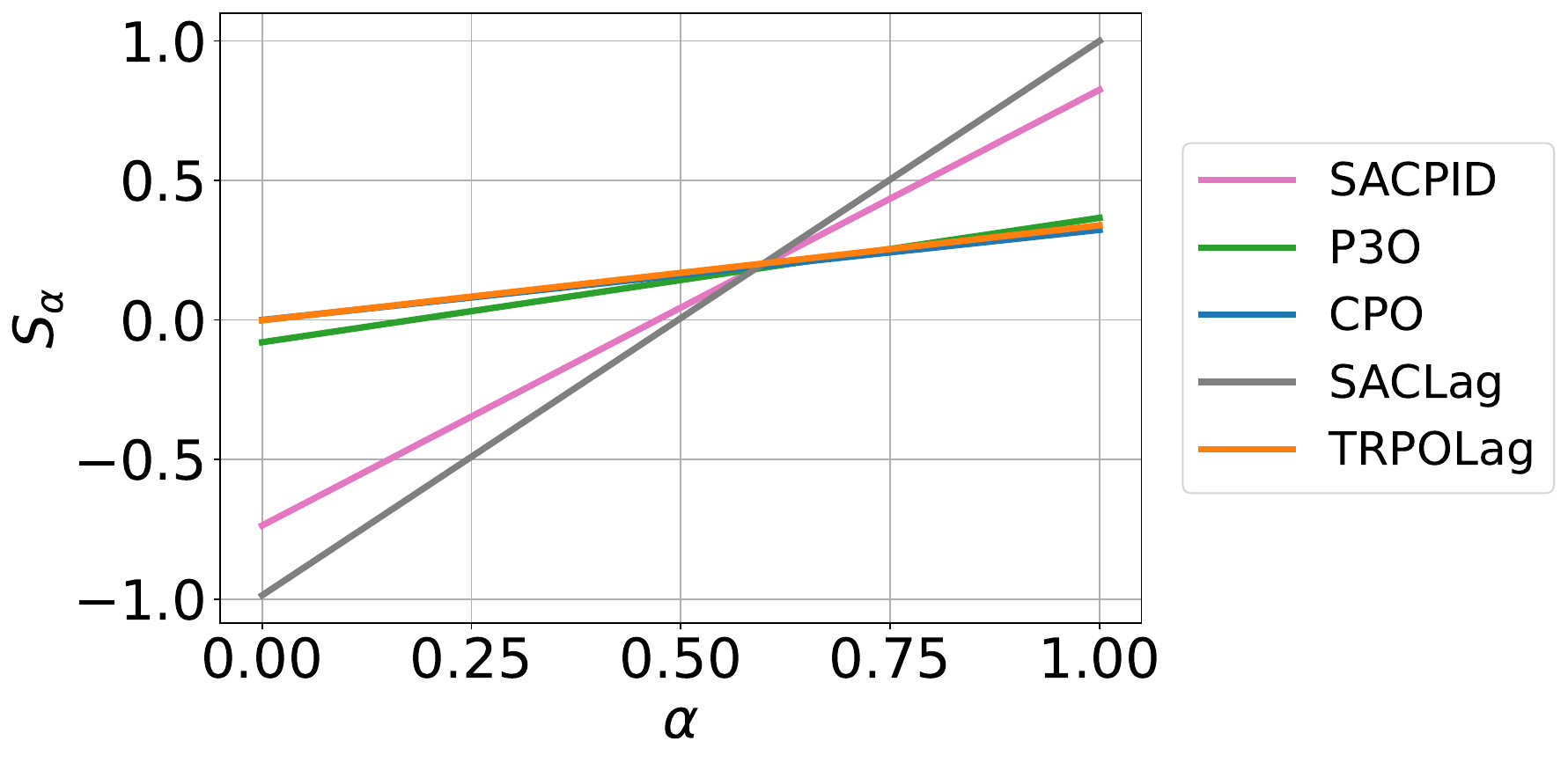}
    \caption{SchedMaintEnv-v1}
\end{subfigure}
\hfill
\begin{subfigure}[t]{0.24\textwidth}
    \centering
    \includegraphics[width=\linewidth]{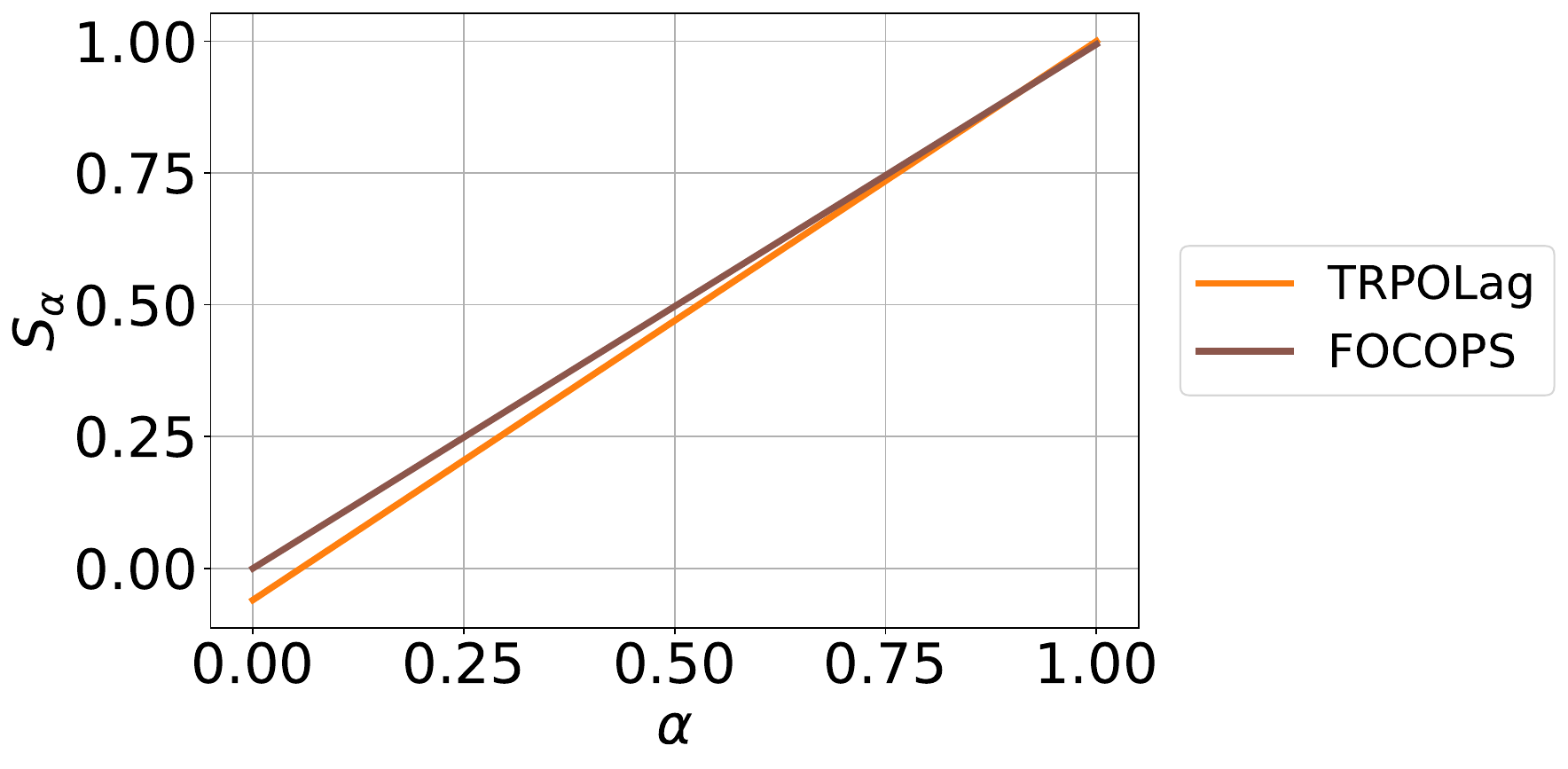}
    \caption{UCEnv}
\end{subfigure}
\vspace{1em}
\begin{subfigure}[t]{0.24\textwidth}
    \centering
    \includegraphics[width=\linewidth]{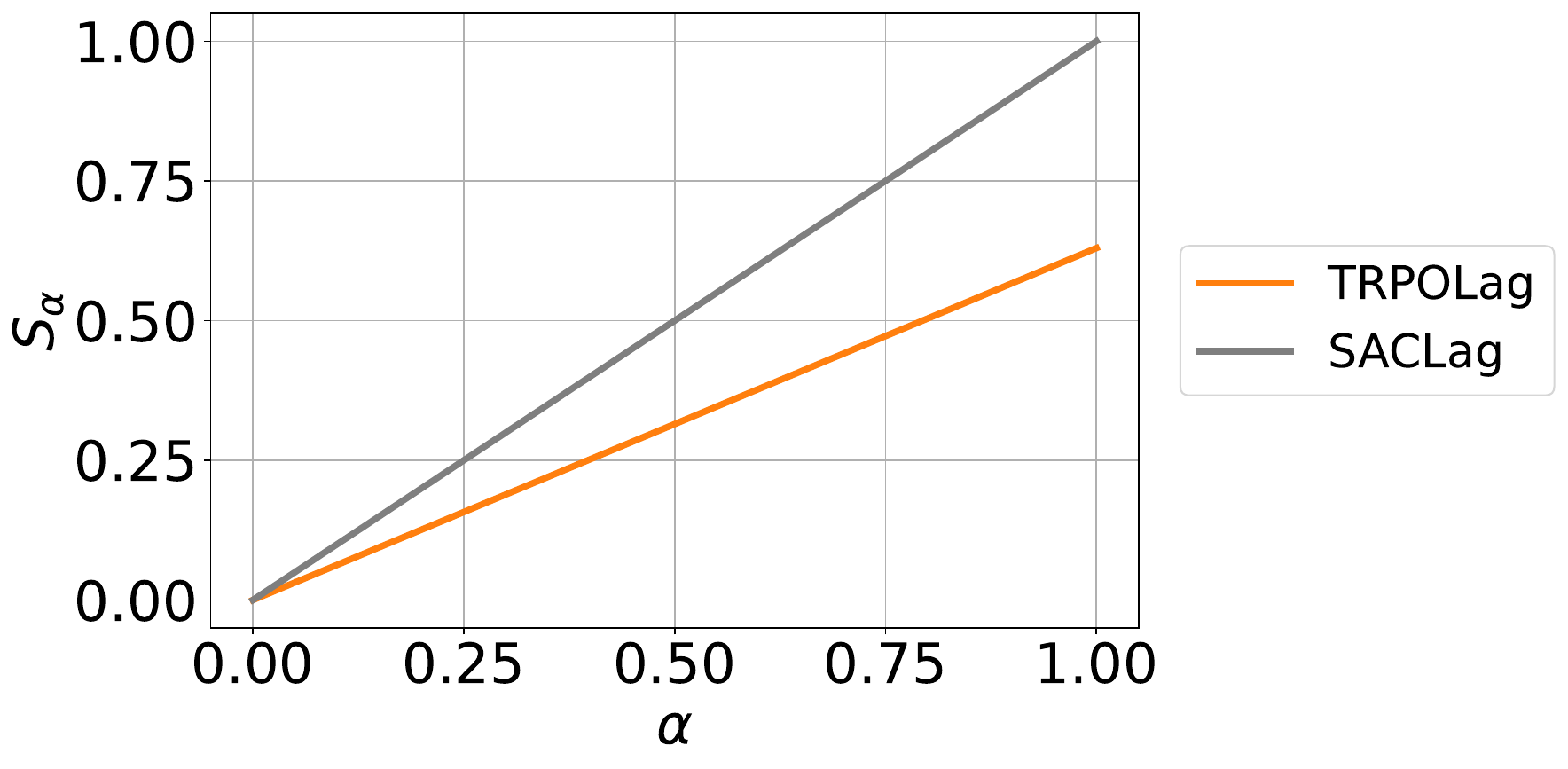}
    \caption{GridStorageEnv}
\end{subfigure}
\hfill
\begin{subfigure}[t]{0.24\textwidth}
    \centering
    \includegraphics[width=\linewidth]{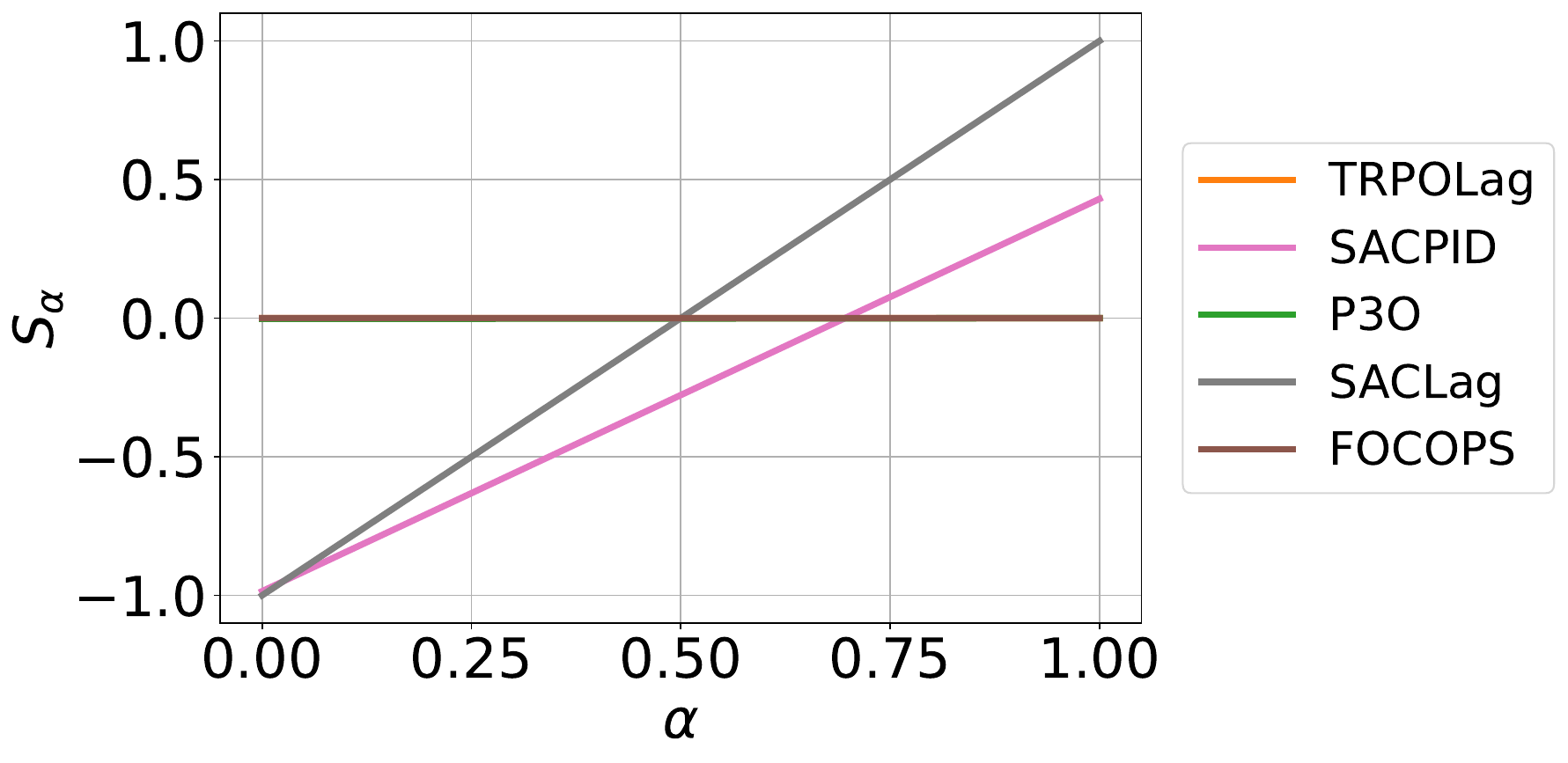}
    \caption{BlendingEnv}
\end{subfigure}
\hfill
\begin{subfigure}[t]{0.24\textwidth}
    \centering
    \includegraphics[width=\linewidth]{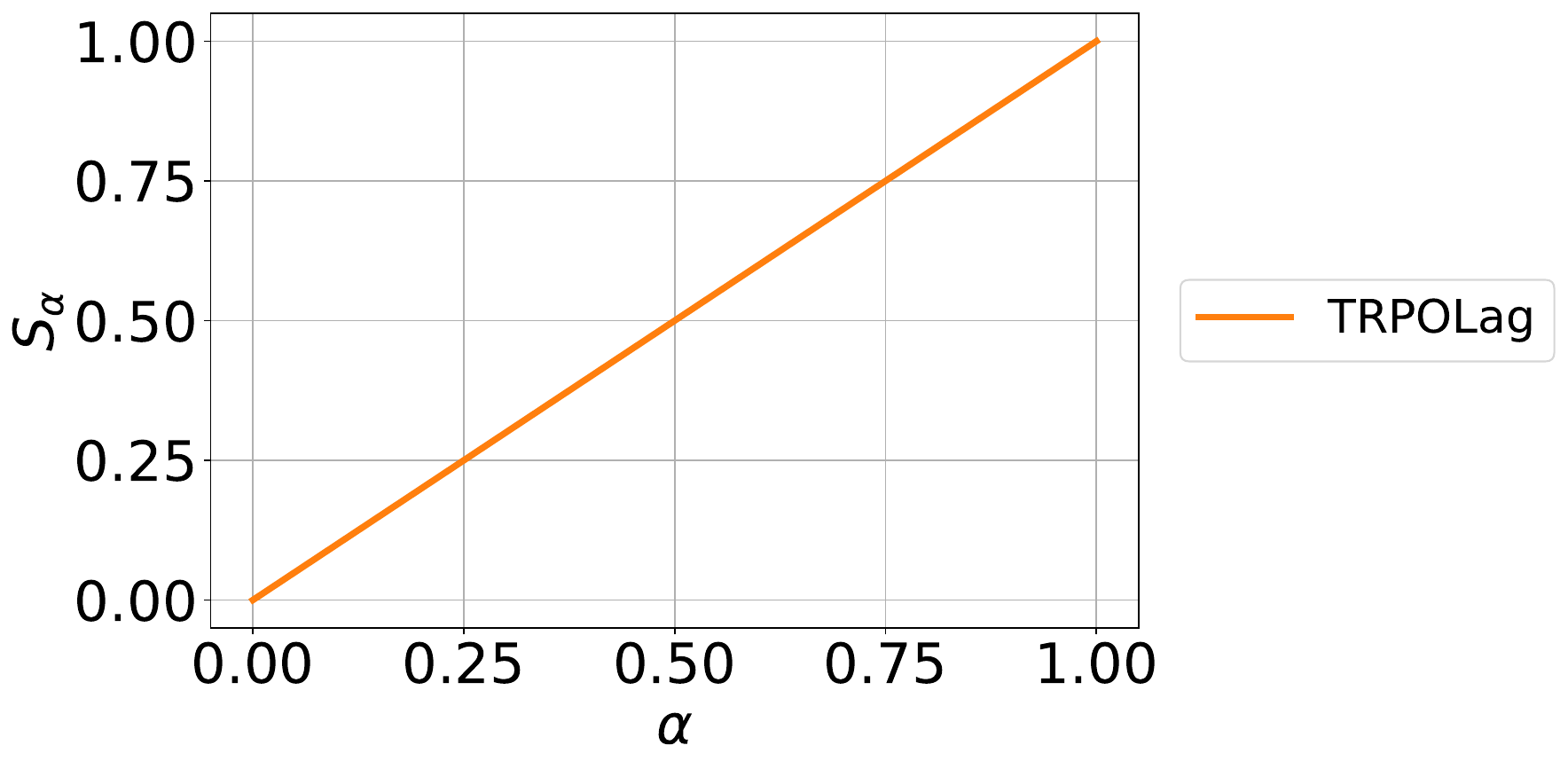}
    \caption{RTNEnv}
\end{subfigure}
\hfill
\begin{subfigure}[t]{0.24\textwidth}
    \centering
    \includegraphics[width=\linewidth]{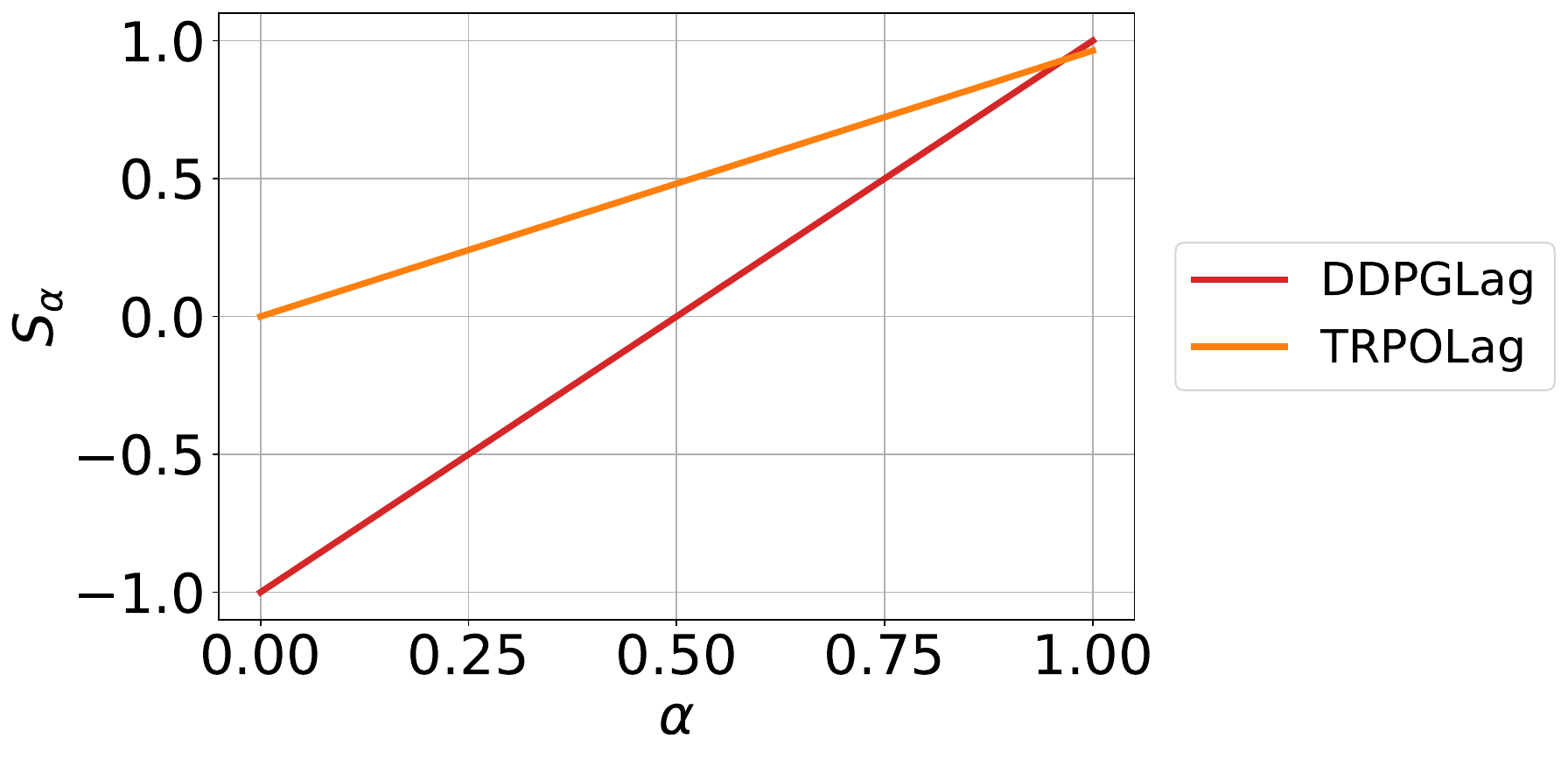}
    \caption{STNEnv}
\end{subfigure}
\caption{Scalarized Pareto score across environments.}
\label{fig:alpha_row1}

\end{figure}

For visualization, we plot $S_\alpha$ as a function of $\alpha$ for all Pareto-efficient algorithms in each environment (Figure~\ref{fig:alpha_row1}), illustrating how rankings change under different safety–performance preferences for each environment discussed in the main paper.

\clearpage
\medskip
\noindent \textbf{Additional evaluation diagnostics:}
In addition to the multi-objective analysis, we consider two complementary diagnostics.

First, we identify significant discrepancies between training and evaluation. The training result refers to the average episode reward from the final training epoch. A gap is considered significant if the absolute difference exceeds 100 and the relative difference exceeds 30\%. When the magnitude of the training result is below 0.1, only the absolute difference is used.

Second, we classify an environment as being solved to \emph{reasonable optimality} if an algorithm achieves (i) a reward within 35\% of the optimal reward and (ii) a constraint violation below a threshold of 25. All metrics are computed using average evaluation reward and cost.
\subsubsection{Environments trained to reasonable optimality}
The following environments illustrate scenarios in which at least one algorithm successfully solves an environment to reasonable optimality:

\begin{itemize}
   % \item \textbf{GTEPEnv:} This case study involves a five-region power system with two generators and possible transmission lines between all region pairs over a 10-period planning horizon. We train with 100 episodes per epoch. P3O shows a significant gap between training and evaluation in both reward and cost. Notably, DDPGLag learns a policy that achieves high rewards at the expense of significantly increased costs. To preserve clarity in visual comparisons, we omit the training curve for DDPGLag in Figure~\ref{fig:GTEPEnv}.

    \item \textbf{InvMgmtEnv:} This case study involves a multi-echelon inventory network comprising one market, one retailer, two distributors, three producers, and two raw-material suppliers, connected via eleven reorder routes and governed by one-period lead times. Each episode spans 30 days, and training is performed with 10 episodes per epoch. P3O exhibits a significant gap between training and evaluation costs and SACPID exhibits a significant gap between training and evaluation rewards.

    \item \textbf{SchedMaintEnv-v0:} We consider an air separation unit with three compressors and optional external product purchases to compensate for maintenance downtime. The planning horizon is 30 days, and each episode is 31 days long. Training is conducted using 25 episodes per epoch. A significant gap between training and evaluation costs is observed for P3O and FOCOPS.
    \item \textbf{SchedMaintEnv-v1:} We consider a similar setup to the \textbf{SchedMaintEnv-v0}. We include uncertainty in the environment by incorporating random machine failures. Training is conducted using 25 episodes per epoch. We observe a significant gap between training and evaluation costs for P3O and FOCOPS. We also find that the performance of most algorithms in this stochastic environment closely matches their behavior in the deterministic counterpart, likely due to the underlying physics of the environment. However, a subset of algorithms shows elevated evaluation standard deviations, reflecting the inherent stochasticity of the environment.

    \item \textbf{UCEnv:} This example considers a single-bus unit commitment power system with five generators operating over a 24-hour horizon, with hourly updates to demand forecasts and generator states. Training is done with 100 episodes per epoch.  A significant gap between training and evaluation costs is observed for SACLag. 

    \item \textbf{GridStorageEnv:} This case study includes a three-bus network with transmission limits of 80MW, 120MW (de-energized under wildfire risk), and 90MW; one generator per bus rated at 100MW, 90MW, and 80MW; and batteries sized 1.0, 1.2, and 0.8 p.u.\ respectively. The horizon spans 24 hours, with perfect charging, discharging, and carry-over efficiency. Experiments were run with 10 episodes per epoch. We observe a significant gap between training and evaluation costs for SACPID and SACLag.

\end{itemize}

\subsubsection{Environments not trained to reasonable optimality}

The following environments presented significant challenges during training, preventing the agents from learning reasonably optimal policies.

\begin{itemize}
    
    \item \textbf{BlendingEnv:} The case study includes 2 input streams, 2 output streams, 2 quality properties, 4 blenders, and 6 time periods per episode with the prop strategy used to handle infeasible actions. No flow is allowed between 2 of the blenders. Experiments were run with 100 episodes per epoch. A significant gap is observed between training and evaluation for costs in P3O and FOCOPS, and for rewards in P3O, DDPGLag and SACLag. As shown in Figure~\ref{fig:Blend}, some algorithms exhibit considerable oscillations in both reward and cost over training epochs. This instability may be attributed to the highly non-convex nature of the underlying optimization problem and the resulting non-smooth feasible action space, which makes consistent policy improvement more difficult.

    \item \textbf{RTNEnv:} This case study includes 3 raw materials, 2 intermediates, 2 products, 3 tasks, 3 pieces of equipment, and 2 utilities. The planning horizon spans 30 time periods with fixed product demands. Training used 128 episodes per epoch. As shown in Figure~\ref{fig:RTN}, both reward and cost exhibit slow learning across all algorithms. This sluggish progress may be attributed to the combination of non-convexities introduced by integer-based constraints such as task-equipment assignments and equipment availability and the indirect, time-coupled linear constraints that the action space must satisfy. The presence of temporal dependencies, material balances, and combinatorial task scheduling further compounds the difficulty of policy learning in these environments.

    \item \textbf{STNEnv:} This case study builds on the same network as RTNEnv but includes extended task-to-equipment mappings and product-specific processing times. To enable comparison with RTNEnv, mappings were kept unique. The planning horizon and demand profiles remain unchanged. Experiments were run with 128 episodes per epoch. TRPOLag performs the best, while OnCRPO performs the worst. Similar to RTNEnv, both reward and cost exhibit slow learning across all the algorithms in STNEnv, likely due to a similar combination of non-convexities introduced by integer-based constraints and the indirect linear constraints imposed on the action space.
\end{itemize}

Taken together, these environments illustrate a broader structural challenge for safe reinforcement learning in nonconvex and mixed-integer settings. In contrast to standard continuous-control benchmarks, the feasible action space in these problems is often piecewise-defined and shaped by discrete scheduling or assignment constraints. Small perturbations in actions can trigger abrupt changes in feasibility, such as activating a different task sequence or violating time-coupled resource balances. As a result, the reward and constraint landscapes become non-smooth and locally discontinuous. For gradient-based safe RL methods, this leads to unstable policy updates, noisy Lagrangian multiplier dynamics, and oscillatory training behavior. Moreover, delayed constraint feedback across time periods further complicates credit assignment, making it difficult for policies to learn consistent reward–violation trade-offs. These structural properties help explain the slow convergence and instability observed across algorithms in these environments.

\subsubsection{Discussion: Algorithm based performance analysis}
The summary table reports, for each algorithm, the number of environments in which it (i) achieves feasibility under the prescribed cost budget (25 in our experiments), (ii) achieves reasonable optimality, (iii) is Pareto-efficient, and (iv) is classified as best-performing under the scalarization-based trade-off analysis described in Section~\ref{section:eval_criteria}.
This characterization avoids heuristic aggregation of reward and cost into a single scalar metric and instead highlights structural trade-offs between performance and constraint satisfaction. These results are summarized in Table~\ref{table:evaluation_algo}.

\begin{table}[htbp]
\centering
    \caption{Summary of performance of algorithms}
    \label{table:evaluation_algo}
        \centering
        \begin{tabular}{lllll}
           % \toprule
            & Frequency of & Frequency of & Frequency of & Frequency of \\
            Algorithm & acheiving  feasibility & acheiving reasonable optimality &  being Pareto-efficient & being best performing\\
            \hline \\
            CPO & 4 & 2 & 1 & 0 \\
DDPGLag & 1 & 0 &  1 & 0 \\
OnCRPO & 3 & 2 &  1 & 1 \\
P3O & 3 & 0 & 2 & 0\\
TRPOLag & 5 & 3 &  7 & 3 \\
FOCOPS & 3 & 1 &  3 & 2\\
SACPID & 1 & 0 &  3  & 1\\
SACLag & 2 & 1  & 4 & 2\\
            %\bottomrule
        \end{tabular}
    \end{table}

%Across most environments, TRPOLag perform best, while DDPGLag consistently underperforms. Interestingly, the P3O, FOCOPS, and SACLag algorithms showed a significant gap between training and evaluation performance in several settings. While some environments allowed agents to learn policies that achieved near-optimal rewards with minimal constraint violations, other environments posed significant challenges. These more difficult environments highlight the limitations of current methods and underscore the need for more sophisticated approaches to safe reinforcement learning.

%\paragraph{Algorithm-type–driven performance analysis:}

To interpret the results in Table~\ref{table:evaluation_algo}, we categorize the evaluated algorithms along two design dimensions that strongly influence performance in constrained reinforcement learning: 
\emph{(i) the data regime and update style (on-policy trust-region versus off-policy replay-based actor--critic), and (ii) the mechanism used to enforce safety constraints (primal constrained updates, Lagrangian dual variables, or adaptive penalties/controllers).}
This structured grouping clarifies the observed feasibility, Pareto efficiency, and best-performing frequencies across environments.

\paragraph{On-policy trust-region methods:}
Algorithms relying on on-policy data and conservative trust-region updates \textbf{CPO}, \textbf{TRPOLag}, and \textbf{FOCOPS} exhibit the most stable multi-objective behavior overall. 
All three methods appear repeatedly on the Pareto frontier, indicating competitive reward–violation trade-offs without being strictly dominated by alternative approaches.

Among them, \textbf{TRPOLag} emerges as the strongest overall performer. It achieves the highest Pareto-efficiency frequency (7 environments) and the highest frequency of being best-performing (3 environments). Its hybrid design—combining trust-region stability with adaptive Lagrangian penalties enables strong reward maximization while maintaining controlled constraint violations. 
\textbf{FOCOPS} also appears frequently on the Pareto frontier (3 environments) and is best-performing in 2 environments, though with lower overall feasibility compared to TRPOLag.
\textbf{CPO} maintains moderate feasibility but appears less frequently as Pareto-efficient and never emerges as best-performing, reflecting its more conservative update structure.

\paragraph{On-policy switching and penalty-based PPO variants:}
\textbf{OnCRPO} and \textbf{P3O} adopt PPO-style updates with distinct constraint-handling mechanisms. 
\textbf{OnCRPO} appears Pareto-efficient in one environment and best-performing once, indicating that objective switching can occasionally achieve strong trade-offs. 
\textbf{P3O} appears Pareto-efficient in two environments but does not emerge as best-performing, suggesting that penalty-based formulations can produce competitive trade-offs yet may struggle to dominate consistently across environments.
Both methods exhibit moderate feasibility but higher variability relative to trust-region approaches, reflecting sensitivity to penalty adaptation and objective switching dynamics.

\paragraph{Off-policy Lagrangian actor--critic methods:}
Replay-based methods \textbf{DDPGLag} and \textbf{SACLag} show less consistent multi-objective performance. 
Although \textbf{SACLag} appears Pareto-efficient in four environments and best-performing in two, its feasibility frequency remains relatively low, indicating instability in maintaining constraint satisfaction. 
\textbf{DDPGLag} appears Pareto-efficient only once and never emerges as best-performing, consistent with weaker constraint regulation. 
These results suggest that replay-based cost estimation may introduce distribution mismatch, weakening reliable reward–violation balancing.

\paragraph{Controller-based constraint enforcement:}
\textbf{SACPID}, which regulates violations via PID-style feedback rather than explicit constrained optimization, appears Pareto-efficient in three environments and best-performing once. However, its overall feasibility frequency remains low. This indicates that while controller-based regulation can occasionally produce competitive trade-offs, it lacks the systematic stability of trust-region or Lagrangian-based constrained updates.

\paragraph{Overall multi-objective trends:}
Several consistent patterns emerge:
\begin{itemize}
    \item Trust-region–based methods, particularly \textbf{TRPOLag}, appear prominently on the Pareto frontier, indicating strong reward–violation trade-offs.
    \item The hybrid Lagrangian–trust-region design of \textbf{TRPOLag} yields the strongest overall multi-objective performance, as reflected by its highest Pareto-efficiency and best-performing frequencies.
    \item Replay-based and controller-based approaches exhibit greater variability in feasibility and reward trade-offs, whereas trust-region methods demonstrate comparatively more stable evaluation-time behavior across environments.
\end{itemize}

Importantly, the Pareto-efficient framing reveals that no algorithm universally optimizes both reward and constraint satisfaction. Instead, different methods occupy distinct regions of the trade-off space. This multi-objective characterization provides a principled evaluation of safe RL behavior, avoiding heuristic scalarization and highlighting structural performance differences across environments.
\paragraph{Training–evaluation gaps:}
While training–evaluation gaps are observed for several algorithms in specific environments, the nature of these gaps differs across algorithm classes. For off-policy Lagrangian methods such as \textbf{DDPGLag} and \textbf{SACLag}, replay-based updates can introduce distribution mismatch between policy and constraint estimates, leading to discrepancies in both reward and feasibility at evaluation time. Penalty-based methods such as \textbf{P3O} may exhibit gaps due to delayed or unstable penalty adaptation, particularly in environments with sharp feasibility boundaries. In contrast, on-policy primal constrained methods such as \textbf{FOCOPS} and \textbf{CPO} comparatively stronger feasibility relative to replay-based methods, and any observed training–evaluation differences primarily reflect conservative policy updates that limit reward improvement rather than failures in constraint satisfaction. As a result, these gaps do not undermine the algorithm-type trends reported here, which are driven by evaluation-time feasibility and reward trade-offs.

% To improve performance in these environments, several strategies could be explored. First, further tuning of cost parameters may help the agent better balance rewards and penalties. Second, rescaling and restructuring the action space could improve alignment with the underlying constraint geometry, reducing the frequency of infeasible actions. Finally, incorporating a differentiable projection layer into the policy network could help project raw actions onto the feasible set, promoting better constraint adherence during both training and evaluation.

\section{Conclusions and Future Directions}

We presented SafeOR-Gym, a suite of nine operations research environments tailored to benchmark safe RL algorithms in complex, realistic settings. This work is intended as a benchmarking and diagnostic study rather than a methodological contribution, aimed at systematically evaluating the capabilities and limitations of existing safe RL algorithms. These environments introduce structured constraints, mixed-integer decisions, and discrete–continuous actions, going beyond the scope of conventional safe RL benchmarks. While existing algorithms can solve some tasks, they perform poorly on problems involving mixed-integer variables or nonlinear, nonconvex constraints. These limitations point to broader challenges in applying safe RL to industrial domains.

To bridge this gap, future research can pursue several promising directions.  One avenue is to broaden the benchmark environments to cover uncertainty and multiagent interactions, which are central to many operations research applications. Another direction is to extend the benchmark beyond traditional optimization-based solutions. For example, we have shown how the classical base-stock reorder policy can serve as a heuristic baseline for the inventory management environment in Appendix~\ref{sec:appendix:ssrq}. Developing heuristics that can respect nontrivial constraints in operations research problems is an ongoing research direction. In this spirit, SafeOR-Gym could be enriched with algorithms drawn from a unified framework of sequential decision-making proposed by \citet{powell2022reinforcement}. In particular, hybrid OR–RL approaches that combine optimization-based reasoning with learning-based policies offer a promising direction for addressing the structural challenges identified in this work.

Finally, \textit{SafeOR-Gym} highlights fundamental limitations of current safe RL approaches that require sustained research. Safe RL methods often fail when problem structures involve nonconvex or mixed-integer constraints, raising questions about their robustness and reliability in safety-critical domains. Algorithms like CPO require sensitive hyperparameter tuning to balance performance and feasibility, suggesting that automated approaches to constraint-aware parameter selection could reduce manual overhead and improve reproducibility. Moreover, current methods often rely on penalty-based formulations or post-hoc projection, which cannot guarantee feasibility. Recent work on action-constrained RL \citep{hung2025efficientactionconstrainedreinforcementlearning} points toward approaches that can enforce hard safety constraints directly during action selection. Another promising direction lies in encoding safety constraints within neural network architectures themselves, for instance through differentiable constraint satisfaction layers \citep{CHEN2024108764} that enforce feasibility by design.

    \section{Acknowledgments and Disclosure of Funding}
    We would like to acknowledge the financial support from ExxonMobil Corporation, NSF awards CBET-2441184, DMS-2424004, ONR award N000142412641.

\bibliography{references}
\bibliographystyle{tmlr}

\appendix

\section{Problem Environment Description}\label{sec:appnedix:env}

\subsection{Resource Task Network Environment (\texttt{RTNEnv})}

\subsubsection{Overview}
\begin{figure}[htbp]
    \centering
    \includegraphics[width=0.85\linewidth]{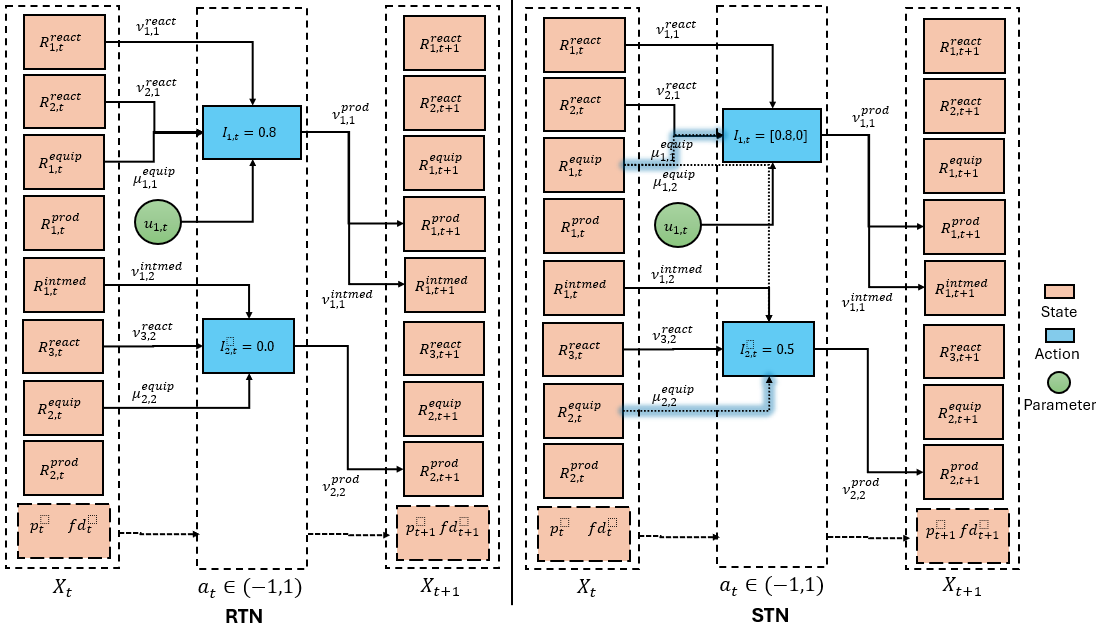}
    \caption{Schematic of RTN and STN: Solid lines in both RTN and STN represent the flow of material through the network. Other inventory levels are included in the state vector. The dashed lines in STN from equipments to tasks represent the choices available where the highlighted lines show the choice taken. The actions are scaled as described in the transition dynamics subsection.}
    \label{fig:rtn_stn_schematic}
\end{figure}

The Resource Task Network (RTN) \citep{pantelides1994unified} is a mathematical modeling framework used for plant scheduling problems. It optimally schedules a set of interdependent production \textit{tasks} executed on equipment, which transform reactants into products via intermediates. The full set of reactants, intermediates, products, and equipment are collectively referred to as \textit{resources}. Products are produced to fulfill time-varying demand while minimizing operational costs. The \texttt{RTNEnv} simulates this system in discrete time, where the agent chooses task batch sizes at each timestep and observes the resulting state transitions, constraint violations, and rewards. The schematic of RTN has been illustrated in \ref{fig:rtn_stn_schematic}

\subsubsection{Problem Setup}
The RTN is defined over a finite time horizon \(T\). Reactants, intermediates, products, equipment, tasks, and utilities are defined as follows:

\begin{itemize}
    \item \textbf{Reactants}: Consumable raw materials that can be ordered (at a cost).
    \item \textbf{Intermediates}: Internally produced and consumed materials. Cannot be ordered or sold.
    \item \textbf{Products}: Final deliverables with external demand and associated revenue and penalty structures.
    \item \textbf{Equipments}: Units required for task execution. Each is consumed when a task starts and returned after the task completes.
    \item \textbf{Tasks}: Tasks produce a subset of resources from another distinct subset of resources using a subset of equipments after a specified time period.
    \item \textbf{Utilities}: Time-varying operational costs incurred per unit utility consumed by tasks.
    \item \textbf{Demand}: Defined only for products, specifying \(d_{p,t}\) at each timestep \(t\).
\end{itemize}

\paragraph{Sets}
\begin{itemize}
    \item \( \mathcal{R} \): All resources (reactants, intermediates, products, equipment)
    \item \( \mathcal{R}^{\text{react}} \): Reactants
    \item \( \mathcal{R}^{\text{int}} \): Intermediates
    \item \( \mathcal{R}^{\text{prod}} \): Products
    \item \( \mathcal{R}^{\text{equip}} \): Equipments
    \item \( \mathcal{R}^{\text{pend}} \subset \mathcal{R}^{\text{int}} \cup \mathcal{R}^{\text{prod}} \cup \mathcal{R}^{\text{equip}} \): Resources which are produced/replenished.
    \item \( \mathcal{I} \): Tasks
    \item \( \mathcal{K}_i \subset \mathcal{R}^{\text{equip}} \): Equipment used by task \(i\)
    \item \( \mathcal{U} \): Utilities
    \item \( \mathcal{U}_i \subset \mathcal{U} \): Utilities consumed by task \(i\)
\end{itemize}

\paragraph{Parameters}
\begin{itemize}
    \item \(T\): Time horizon
    \item \(\tau_i\): Processing time of task \(i\)
    \item \(\tau_{\text{max}} = \max \limits_i \tau_i \) : Maximum processing time.
    \item \(V_i^{\min}, V_i^{\max}\): Batch size bounds for task \(i\)
    \item \(X_r^0, X_r^{\min}, X_r^{\max}\): Initial and bounded inventories for resource \(r\)
    \item \(\nu_{i,r}\): Stoichiometric coefficient of resource \(r\) in task \(i\)
    \item \(\text{cost}_r\): Unit cost of ordering reactant \(r\)
    \item \(\text{price}_p\): Unit price of product \(p\)
    \item \(d_{p,t}\): Demand for product \(p\) at time \(t\)
    \item \(u_{u,t}\): Cost of utility \(u\) at time \(t\)
    \item \(\lambda_{\text{sanit}}\): Sanitization penalty coefficient
    \item \(\mathbb{I}[\cdot]\): Indicator function
    \item \(\oplus\) : vector concatenation
\end{itemize}

\subsubsection{State Space}
At timestep \(t\), the agent observes:
\begin{itemize}
    \item Inventory vector \(X_t \in \mathbb{R}^{|\mathcal{R}|}\) : Inventory of all resources.
    \item Pending outputs \(p_t \in \mathbb{R}^{\tau_{\max} \times |\mathcal{R}^{\text{pend}}|}\) : Intermediates and products which will be delivered in upcoming timesteps.
    \item Future demand \(fd_t \in \mathbb{R}^{T \times |\mathcal{R}^{\text{prod}}|}\) : Demand of all products from timestep \(t\) to \(T\), post-padded with \(0\)s to maintain consistent shapes.
\end{itemize}

\subsubsection{Action Space}
The action \(a_t \in [-1,1]^{|\mathcal{I}|}\) is scaled to batch sizes via:
\[
a^{\text{scaled}}_{i,t} = \frac{a_{i,t} + 1}{2} \cdot (V_i^{\max} - V_i^{\min}) + V_i^{\min}
\]
For numerical stability, any \(|a_{i,t}| \leq 10^{-3} \) is set to \(0\).

\subsubsection{Transition Dynamics}
At each step:
\begin{enumerate}
    \item \textbf{Sanitize Action}: Prevent resource violations and enforce equipment availability by calculating the maximum inventory available for a resource, maximum batch size that can be processed based on inventory levels, and clipping between the batch size bounds accordingly.
    \begin{equation*}
        \begin{split}
        b_{i,r,t} \;&=\; \frac{\max(0, X_{r,t} - X_r^{\min})}{|\nu_{i,r}|} \\
        b_{i, t} \;&=\; \min \limits_r b_{i,r,t} \\
        a_i^{\text{clip}} \;&=\;
    %\begin{cases}
        0 \;\; \text{if } b_{i,t} < V_i^{\min} \\
        %\min(a_i^{\text{scaled}}, V_i^{\max}), & \text{otherwise}
    %\end{cases} \\
    a_{i,t}^{\text{final}} \;&=\;
    \begin{cases}
        0, & \text{if } \exists e \in \mathcal{K}_i \text{ with } X_{e,t} = 0 \\
        a_i^{\text{clip}}, & \text{otherwise}
    \end{cases}
    \end{split}
    \end{equation*}
    \item \textbf{Inventory Update}: Consumes the inputs to a task immediately.
    \[
    X_{r,t+1} = X_{r,t} + \sum_{i} \min(\nu_{i,r}, 0) \cdot a_{i,t}^{\text{final}}
    \]

    \item \textbf{Pending Outputs}: Add outputs of a task to the pending output buffer and update inventory of resources that are being delivered in the next timestep.
    \begin{equation}
        \begin{split}
            X_{r,t+1} \;&=\; X_{r,t} + p_{t,r,0} \\
            p_t \;&=\;  p_{t-1, r, 1 : \tau_{\text{max}}} \oplus \sum \limits_i \max \{\nu_{i,r}, 0\}\cdot a_{i,t}^{\text{final}} \\
        \end{split}
    \end{equation}

    %\item \textbf{Equipment Update}:
    %\[
    %X_{e,t+1} = X_{e,t} - \sum_{i : e \in \mathcal{K}_i} \mathbb{I}[a_{i,t} > 0] + \sum_{i : e \in \mathcal{K}_i} \mathbb{I}[a_{i,t - \tau_i} > 0]
    %\]

    \item \textbf{Inventory Enforcement}: Ensures inventory bounds are not violated. A part of the cost is calculated based on this. Refer to \ref{rtn cost func}.
    \[
    X_{r,t+1} = \min(X_r^{\max}, \max(X_r^{\min}, X_{r,t+1}))
    \]
\end{enumerate}

\subsubsection{Cost Function}
\label{rtn cost func}

The total cost at each timestep \(t\) is given by:
\[
C_t = C_t^{\text{lb}} + C_t^{\text{ub}} + C_t^{\text{eq}} + \lambda_{\text{sanit}} C_t^{\text{sanit}}
\]

\begin{itemize}
    \item \textbf{Bound Violation}: Penalizes the resource inventory going out of bounds. (Reactants going below lower bounds are assumed to imply ordering of reactants).
    \[
    C_t^{\text{lb}} = \sum_{r \in \mathcal{R}^{\text{int}} \cup \mathcal{R}^{\text{prod}}} \mathbb{I}[X_{r,t} < X_r^{\min}]
    \]
    \[
    C_t^{\text{ub}} = \sum_{r \in \mathcal{R}} \mathbb{I}[X_{r,t} > X_r^{\max}]
    \]
    \item \textbf{Equipment Feasibility Cost:} Penalizes execution of tasks when any required equipment is unavailable.
    \[
    C_t^{\text{eq}} = \sum_{i} \sum_{e \in \mathcal{K}_i} \mathbb{I}[a_{i,t}^{\text{final}} > 0] \cdot \mathbb{I}[X_{e,t} < 1]
    \]

    \item \textbf{Sanitization Cost:} Penalizes deviations between the raw and final actions due to constraint handling.
    \[
    C_t^{\text{sanit}} = \sum_{i} \left|a_{i,t}^{\text{final}} - a_{i,t}^{\text{scaled}}\right|
    \]
\end{itemize}

\subsubsection{Reward Function}
The total reward at each timestep \(t\) is:
\[
\Pi_t = \Pi_t^{\text{rev}} - \Pi_t^{\text{util}} - \Pi_t^{\text{unmet}} - \Pi_t^{\text{order}}
\]

\begin{itemize}
    \item \textbf{Revenue:} Earned by fulfilling product demand, constrained by available inventory.
    \[
    \Pi_t^{\text{rev}} = \sum_{p \in \mathcal{R}^{\text{prod}}} \min(X_{p,t} - X_p^{\min}, d_{p,t}) \cdot \text{price}_p
    \]

    \item \textbf{Unmet Demand Penalty:} Penalizes unmet product demand at 1.5× product price.
    \[
    \Pi_t^{\text{unmet}} = 1.5 \cdot \sum_{p \in \mathcal{R}^{\text{prod}}} \max(d_{p,t} - (X_{p,t} - X_p^{\min}), 0) \cdot \text{price}_p
    \]

    \item \textbf{Reactant Ordering Cost:} Charged when reactant inventory drops below zero (interpreted as external procurement)
    \[
    \Pi_t^{\text{order}} = \sum_{r \in \mathcal{R}^{\text{react}}} \max(-X_{r,t}, 0) \cdot \text{cost}_r
    \]

    \item \textbf{Utility Cost:} Incurred from executing tasks using utilities.
    \[
    \Pi_t^{\text{util}} = \sum_{i} \sum_{j \in \mathcal{U}_i } a_{i,t}^{\text{final}} \cdot u_{j,t}
    \]
\end{itemize}

\subsubsection{Episode Termination}
The episode ends when \(t = T\). Truncation does not occur even under constraint violations.

\subsection{State Task Network (\texttt{STNEnv})}

\subsubsection{Overview}
The State Task Network (STN) \citep{pantelides1994unified} is a mathematical modeling framework for short-term production scheduling. It focuses on scheduling a set of production \textit{tasks} on processing units, where each task transforms material \textit{states} over time. Unlike the RTN, which models resources more abstractly, the STN emphasizes the evolution of \textit{material states} through task execution and transfer between units. The objective is to satisfy time-varying demand for final products while minimizing operational costs. The \texttt{STNEnv} simulates this system in discrete time, where an agent selects task batch sizes on specific units at each timestep and observes the resulting state transitions, constraint violations, and rewards. The schematic of STN has been illustrated in \ref{fig:rtn_stn_schematic}

\subsubsection{Problem Setup}
The STN is defined over a finite time horizon \(T\). Material states, processing units, tasks, and utilities are defined as follows:

\begin{itemize}
    \item \textbf{Reactants}: Consumable raw materials that can be externally procured at a cost. They serve as initial inputs to tasks.
    \item \textbf{Intermediates}: Internally produced and consumed materials. They are used to link tasks within the network. Intermediates cannot be purchased or sold externally.
    \item \textbf{Products}: Final deliverables for which external demand is specified. Products may generate revenue if delivered on time and incur penalties if demand is unmet.
    \item \textbf{Units}: Processing equipment on which tasks are scheduled. Each unit can process at most one task at a time and becomes available again after the task's processing duration.
    \item \textbf{Tasks}: Operations that transform a subset of input states into a subset of output states on a designated unit. Each task has a fixed processing time, specific unit assignment, defined stoichiometry, and utility consumption profile.
    \item \textbf{Utilities}: Operational resources (e.g., electricity, steam) with time-varying costs. Each task consumes a fixed amount of utilities per unit batch size, incurring operational costs accordingly.
    \item \textbf{Demand}: Defined only for product states, specifying \(d_{s,t}\) for state \(s\) at time \(t\). Demand fulfillment yields revenue, while shortages incur penalties.
\end{itemize}

\paragraph{Sets}
\begin{itemize}
    \item \( \mathcal{S} \): All material states (reactants, intermediates, products, equipments)
    \item \( \mathcal{S}^{\text{react}} \subset \mathcal{S} \): Reactants (raw materials)
    \item \( \mathcal{S}^{\text{int}} \subset \mathcal{S} \): Intermediates (internal materials)
    \item \( \mathcal{S}^{\text{prod}} \subset \mathcal{S} \): Products (deliverables)
    \item \( \mathcal{E} \): Equipment units
    \item \( \mathcal{S}^{\text{pend}} \subset \mathcal{S}^{\text{int}} \cup \mathcal{S}^{\text{prod}} 
    \cup \mathcal{E} \): States that are produced (i.e., replenished by tasks)
    \item \( \mathcal{I} \): Tasks
    \item \( \mathcal{K}_i \subset \mathcal{E} \): Set of units on which task \(i\) can be scheduled
    \item \( \mathcal{U} \): Utilities
    \item \( \mathcal{U}_i \subset \mathcal{U} \): Utilities consumed by task \(i\)
\end{itemize}

\paragraph{Parameters}
\begin{itemize}
    \item \(T\): Time horizon
    \item \(\tau_i\): Processing time of task \(i\)
    \item \(\tau_{\text{max}} = \max \limits_i \tau_i \): Maximum processing time
    \item \(V_{i,e}^{\min}, V_{i,e}^{\max}\): Batch size bounds for task \(i\) by using equipment \(e\).
    \item \(X_s^0, X_s^{\min}, X_s^{\max}\): Initial and bounded inventories for state \(s\)
    \item \(\nu_{i,s}\): Stoichiometric coefficient of state \(s\) in task \(i\) (negative if consumed, positive if produced)
    \item \(\text{cost}_s\): Unit cost of ordering reactant \(s \in \mathcal{S}^{\text{react}}\)
    \item \(\text{price}_p\): Unit price of product \(p \in \mathcal{S}^{\text{prod}}\)
    \item \(d_{p,t}\): Demand for product \(p\) at time \(t\)
    \item \(u_{u,t}\): Cost of utility \(u\) at time \(t\)
    \item \(\lambda_{\text{sanit}}\): Sanitization penalty coefficient (for equipment reuse)
    \item \(\mathbb{I}[\cdot]\): Indicator function
    \item \(\oplus\) : vector concatenation
\end{itemize}

\subsubsection{State Space}
At timestep \(t\), the agent observes:
\begin{itemize}
    \item Inventory vector \(X_t \in \mathbb{R}^{|\mathcal{S}|}\) : Inventory of all material states.
    \item Pending outputs \(p_t \in \mathbb{R}^{\tau_{\max} \times |\mathcal{S}^{\text{pend}}|}\) : Intermediates and products scheduled to be produced in future timesteps due to ongoing tasks.
    \item Future demand \(fd_t \in \mathbb{R}^{T \times |\mathcal{S}^{\text{prod}}|}\) : Demand for all product states from timestep \(t\) to \(T\), post-padded with \(0\)s to maintain consistent shapes.
\end{itemize}

\subsubsection{Action Space}
The action \(a_t \in [-1,1]^{|\mathcal{I} \times \mathcal{E}|}\) represents a normalized matrix over task-equipment pairs and is scaled to batch sizes via:
\[
a^{\text{scaled}}_{i,e,t} = \frac{a_{i,e,t} + 1}{2} \cdot (V_{i,e}^{\max} - V_{i,e}^{\min}) + V_{i,e}^{\min}
\]
For numerical stability, any \(|a_{i,e,t}| \leq 10^{-3} \) is set to \(0\).

\subsubsection{Transition Dynamics}
At each step:
\begin{enumerate}
    \item \textbf{Sanitize Action}: Prevent state violations and enforce unit availability by calculating the maximum available inventory for each input state, the maximum feasible batch size given current inventories, and clipping between the allowed batch size bounds accordingly.
    \begin{equation*}
        \begin{split}
        b_{i,s,t} \;&=\; \frac{\max(0, X_{s,t} - X_s^{\min})}{|\nu_{i,s}|} \\
        b_{i,t} \;&=\; \min \limits_{s \in \mathcal{S} : \nu_{i,s} < 0} b_{i,s,t} \\
        a_{i,e,t}^{\text{clip}} \;&=\;
        %\begin{cases}
            0 \;\; \text{if } b_{i,t} < V_{i,e}^{\min} \\
            %\min(a_{i,e,t}^{\text{scaled}}, V_{i,e}^{\max}), & \text{otherwise}
        %\end{cases} \\
        a_{i,e,t}^{\text{final}} \;&=\;
        \begin{cases}
            0, & \text{if } e \notin \mathcal{K}_i \text{ or } X_{e,t} = 0 \\
            a_{i,e,t}^{\text{clip}}, & \text{otherwise}
        \end{cases}
        \end{split}
    \end{equation*}

    \item \textbf{Inventory Update}: Immediately consumes the input states required by the activated tasks.
    \[
    X_{s,t+1} = X_{s,t} + \sum_{e} \sum_{i} \min(\nu_{i,s}, 0) \cdot a_{i,e,t}^{\text{final}}
    \]

    \item \textbf{Pending Outputs}: Add the output states of tasks to the pending output buffer and update inventories of materials delivered at the current timestep.
    \begin{equation}
        \begin{split}
            X_{s,t+1} \;&=\; X_{s,t} + p_{t,s,0} \\
            p_{t+1,s} \;&=\; p_{t,s,1 : \tau_{\max}} \oplus \sum_{e} \sum_{i} \max(\nu_{i,s}, 0) \cdot a_{i,e,t}^{\text{final}}
        \end{split}
    \end{equation}

    % \item \textbf{Equipment Update}:
    % \[
    % X_{e,t+1} = X_{e,t} - \sum_{i : e \in \mathcal{K}_i} \mathbb{I}[a_{i,e,t} > 0] + \sum_{i : e \in \mathcal{K}_i} \mathbb{I}[a_{i,e,t - \tau_i} > 0]
    % \]

    \item \textbf{Inventory Enforcement}: Enforces inventory bounds to prevent overflow or underflow. Violations of these bounds contribute to the constraint cost.
    \[
    X_{s,t+1} = \min(X_s^{\max}, \max(X_s^{\min}, X_{s,t+1}))
    \]
\end{enumerate}

\subsubsection{Cost Function}
\label{stn cost func}

The total cost at each timestep \(t\) is given by:
\[
C_t = C_t^{\text{lb}} + C_t^{\text{ub}} + C_t^{\text{eq}} + \lambda_{\text{sanit}} C_t^{\text{sanit}}
\]

\begin{itemize}
    \item \textbf{Bound Violation}: Penalizes material inventory going out of bounds. (Reactants going below their lower bounds are interpreted as triggering procurement costs.)
    \[
    C_t^{\text{lb}} = \sum_{s \in \mathcal{S}^{\text{int}} \cup \mathcal{S}^{\text{prod}}} \mathbb{I}[X_{s,t} < X_s^{\min}]
    \]
    \[
    C_t^{\text{ub}} = \sum_{s \in \mathcal{S}} \mathbb{I}[X_{s,t} > X_s^{\max}]
    \]

    \item \textbf{Equipment Feasibility Cost}: Penalizes task executions on unavailable units.
    \[
    C_t^{\text{eq}} = \sum_{i} \sum_{e \in \mathcal{K}_i} \mathbb{I}[a_{i,e,t}^{\text{final}} > 0] \cdot \mathbb{I}[X_{e,t} < 1]
    \]

    \item \textbf{Sanitization Cost}: Penalizes deviation between scaled and final actions due to sanitization (inventory or equipment infeasibility).
    \[
    C_t^{\text{sanit}} = \sum_{i,e} \left|a_{i,e,t}^{\text{final}} - a_{i,e,t}^{\text{scaled}}\right|
    \]
\end{itemize}

\subsubsection{Reward Function}
The total reward at each timestep \(t\) is:
\[
\Pi_t = \Pi_t^{\text{rev}} - \Pi_t^{\text{util}} - \Pi_t^{\text{unmet}} - \Pi_t^{\text{order}}
\]

\begin{itemize}
    \item \textbf{Revenue:} Earned by fulfilling product demand, constrained by available inventory.
    \[
    \Pi_t^{\text{rev}} = \sum_{p \in \mathcal{S}^{\text{prod}}} \min(X_{p,t} - X_p^{\min}, d_{p,t}) \cdot \text{price}_p
    \]

    \item \textbf{Unmet Demand Penalty:} Penalizes unmet product demand at \(1.5\times\) the product price.
    \[
    \Pi_t^{\text{unmet}} = 1.5 \cdot \sum_{p \in \mathcal{S}^{\text{prod}}} \max(d_{p,t} - (X_{p,t} - X_p^{\min}), 0) \cdot \text{price}_p
    \]

    \item \textbf{Reactant Ordering Cost:} Incurred when reactant inventory falls below zero (interpreted as external procurement).
    \[
    \Pi_t^{\text{order}} = \sum_{s \in \mathcal{S}^{\text{react}}} \max(-X_{s,t}, 0) \cdot \text{cost}_s
    \]

    \item \textbf{Utility Cost:} Accrued from task executions that consume utilities, priced per unit usage and per unit batch size.
    \[
    \Pi_t^{\text{util}} = \sum_{i,e} \sum_{j \in \mathcal{U}_i} a_{i,e,t}^{\text{final}} \cdot u_{j,t}
    \]

\end{itemize}
\subsubsection{Episode Termination}
The episode ends when \(t = T\). Truncation does not occur even under constraint violations.

\subsection{Unit Commitment (\texttt{UCEnv})}
\subsubsection{Overview}
\begin{figure}[]
    \centering
    \includegraphics[width=0.85\linewidth]{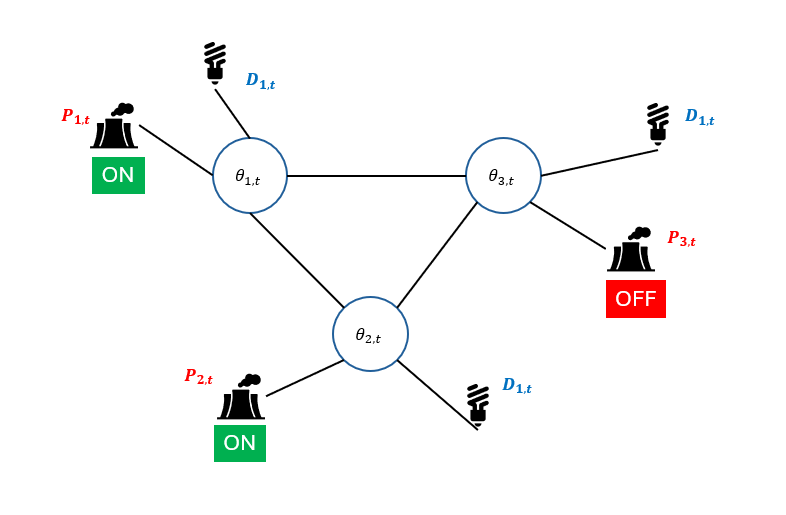}
    \caption{Schematic of UC in a single-bus system and 4-bus system at the time point $t$: Red lines indicate power inflow and outflow. Black lines denote transmission lines. At each time point, forecast demands, current power output, current voltage angle, current and previous on-off status (state) are used to infer the power output, voltage angle, on-off status at the next time point (action). }
    \label{fig:uc_schematic}
\end{figure}

The unit commitment problem is one of the most widely used optimization problems in power systems, aiming to minimize the operational costs of power generation over a specified planning horizon by determining optimal decisions for switching power units on or off and managing power dispatch \citep{knueven2020}. These decisions must comply with safety requirements and fulfill various operational goals. In practice, the unit commitment problem is solved in advance on a rolling basis, as future electricity demand is uncertain and must be forecasted continually once new information becomes available. The growth of renewable energy sources and fluctuations in the electricity markets further increase the uncertainties in demand forecasts. Due to the size of power systems in the real world, solving the resulting MILP or MIQCP problem on time can be prohibitively difficult. Therefore, the development of efficient solution methods that can respond to frequent forecast updates is of significant interest. In this study, we formulate the unit commitment problem as a CMDP and implement it as the environment \texttt{UCEnv}. To support safe sequential decision making in power scheduling and dispatch, practical constraints, such as minimum up and down time, ramping constraints, and reserve requirements, are incorporated into the environment. Given the complexity of unit commitment problem, the environment is provided in two versions. The \texttt{UCEnv-v0} version assumes a single-bus system, while \texttt{UCEnv-v1} requires an agent to take into account distributed power demands and power flows within a transmission network.  The schematic for this environment has been illustrated in \ref{fig:uc_schematic}.

\subsubsection{Problem Setup}

\paragraph{Sets}
\begin{itemize}
    \item \( \mathcal{G} \): the set of generators.
    \item \( \mathcal{N} \): the set of buses.
    \item \( \mathcal{K} \): the set of transmission lines.
    \item \(\mathcal{G}_n\): the set of generators that is connected to the $n$ bus.
    \item \(k(n)\): the from-bus for the $k$th transmission line.
    \item \(k(m)\): the to-bus for the $k$th transmission line.
    \item \(\delta^+(n)\): the set of transmission line for the to-bus $n$.
    \item \(\delta^-(n)\): the set of transmission line for the from-bus $n$.
\end{itemize}

\paragraph{Parameters}
\begin{itemize}
    \item \((a_i, b_i, c_i)\): quadratic, linear, and constant cost coefficients of power generation.
    \item \(C^v_i\): startup cost coefficients.
    \item \(C^w_i\): shutdown cost coefficients.
    \item \(UT_i\): minimum up time.
    \item \(DT_i\): minimum down time.
    \item \(RU_i\): ramp-up rate.
    \item \(RD_i\): ramp-down rate.
    \item \(SU_i\): start-up rate.
    \item \(SD_i\): shut-down rate.
    \item \(P_{\max i}\): maximum power output.
    \item \(P_{\min i}\): minimum power output.
    \item \(\Theta_{\max n}\): maximum voltage angle.
    \item \(\Theta_{\min n}\): minimum voltage angle.
    \item \(F_{\max k}\): maximum transmission capacity.
    \item \(F_{\min k}\): minimum transmission capacity.
    \item \( D_{n,t} \); time-varying power demand.
    \item \(C^{LS}\): load shedding cost coefficient for failure to meet load.
    \item \(C^{R}\): reserve shortfall cost coefficient for failure to meet the reserve requirement.
    \item  \(R\): system-wide reserve requirement
\end{itemize}

\subsubsection{State Space}
The state \( s_t \) includes the following components:

\begin{itemize}
    \item  \( u_{i,t}^{\text{seq}} = [u_{i,t}, u^{\text{old}}_{i,t}] \in \{0, 1\}^{\max(UT_i, DT_i)+ 1} \): a sequence of binary indicators about the on-off status of the generator $i \in \mathcal{G}$, where $u^{\text{old}}_{i,t}$ records the history in the most recent time periods before $t$, used to track compliance with the up/down time. The dimension of this variable is \( \sum_{i \in \mathcal{G}} ( \max(UT_i, DT_i)+ 1 ) \).
    \item \( p_{i, t} \in [P_{\min i}, P_{\max i}] \): power output of the generator $i \in \mathcal{G}$. The dimension of this variable is \( |\mathcal{G}| \).
    \item  \(\theta_{n, t} \in [\Theta_{\min n}, \Theta_{\max n}] \): voltage angle of the bus \( n\in \mathcal{N} \).  The dimension of this variable is \( |\mathcal{N}| \). In the simpler version, \texttt{UCEnv-v0}, the environment is assumed to be a single-bus system where the voltage angle is not considered. 
    \item \( D_{n, t+1:t+W} \in \mathbb{R}_{\geq 0}^W \): forecast demand of the bus \( n\in \mathcal{N} \) with a window length of $W$.  The dimension of this variable is \( |\mathcal{N}| \times W \). When the window  extends beyond the episode horizon, it uses the forecast demand from the following steps.
\end{itemize}

The full observation is flattened into a continuous vector space for reinforcement learning (RL) training.

\subsubsection{Action Space}
At each time step \( t \), the agent takes the following actions \( a_t \):

\begin{itemize}
    \item \( u_{i,t+1} \in \{0, 1\} \): on/off status of the generator $i \in \mathcal{G}$.  The dimension of this variable is \( |\mathcal{G}| \).
    \item \( p_{i,t+1} \in [P_{\min i}, P_{\max i}] \): power output of the generator $i \in \mathcal{G}$. The dimension of this variable is \( |\mathcal{G}| \).
    \item \(\theta_{n, t+1} \in [\Theta_{\min n}, \Theta_{\max n}] \): voltage angle of the bus \( n\in \mathcal{N} \), where $\theta_{1, t+1} = 0$ is always fixed as reference. The dimension of this variable is \( |\mathcal{N}| -1 \).
\end{itemize}

These decision variables are initially normalized to a continuous action space \( a_t \in [-1, 1]^n \) and then mapped to their actual values within their bounds.

\subsubsection{Transition Dynamics}

Given the action \( a_t \), the next state \( s_{t+1} \) is computed as follows.

\textbf{Intermeidate state, turn-on and turn-off status and sequence}:
First, we compute the intermediate states $v_{i, t+1}$, $w_{i, t+1}$ as well as the sequences $v_{i,t+1}^{\text{seq}}$ and $w_{i,t+1}^{\text{seq}}$. These intermediate states serve as a practical representation of whether a generator has recently been activated or deactivated, facilitating the subsequent computation of costs, violations, and rewards.
\[v_{i, t+1} = \max ( 0, u_{i,t+1} - u_{i,t} ) \]
\[w_{i, t+1} = - \min ( 0, u_{i, t+1} - u_{i, t} ) \]
\[ v_{i,t+1}^{\text{seq}} = \left[v_{i, t+1}, v^{\text{old}}_{i,t} \left[:-1 \right] \right] \] 
\[ w_{i,t+1}^{\text{seq}} = \left[w_{i, t+1}, w^{\text{old}}_{i,t} \left[:-1 \right] \right] \] 

\paragraph{Check and repair on-off status}
Next, we check the feasibility of the on-off status. The first inequality evaluates the minimum up-time constraint, ensuring the $i$th generator remains on for a specified number of time periods, $UT_i$, before shutdown. The second inequality evaluates the minimum down-time constraint, which requires the $i$th generator to have a minimum off duration, $DT_i$, before restart. These constraints prevent excessive wear-and-tear resulting from frequent cycling. The violation indicates the power generator that must be kept on and off, respectively. The values of $u_{i, t+1}$ of the must-on generators (must-off generators) are then corrected to $1$ ($0$).
\[
u_{i, t+1}^{\text{r}} = 
\begin{cases}
1, & \text{if }  \sum v_{i,t+1}^{\text{seq}} > u_{i, t+1} \\
0, & \text{elif }  \sum w_{i,t+1}^{\text{seq}} > 1 - u_{i, t+1} \\
u_{i, t+1}, & \text{otherwise }
\end{cases}
\]

\paragraph{Update sequence of on-off status}:
After checking and repairing the on-off status, their sequence is updated:
\[ \mathbf{u}_{i,t+1} = \left[u_{i, t+1}^{\text{r}}, u^{\text{old}}_{i,t} \left[:-1 \right] \right] \] 

\textbf{Check and repair power output}
We proceed to check the violations of the ramping constraints. The first inequality evaluates the ramp-up constraint, and the second inequality evaluates the ramp-down constraint. These constraints impose limits on the rate at which a generator increases or decreases its output between consecutive time steps, reflecting the physical capabilities of generators. The violation indicates the power generation that must be kept within a tighter bound based on the repaired on-off status.

\[
p_{i, t+1}^{\text{r}} = 
\begin{cases}
p_{i, t} + RU_i \cdot u_{i, t} + SU_i \cdot v_{i, t+1}^{\text{r}}, & \text{if }  p_{i, t+1} - p_{i, t} > RU_i \cdot u_{i, t} + SU_i \cdot v_{i, t+1}^{\text{r}}  \\
p_{i, t} - RD_i \cdot u_{i, t+1}^{\text{r}} - SD_i \cdot w_{i, t+1}^{\text{r}}, & \text{elif } p_{i, t} - p_{i, t+1} > RD_i \cdot u_{i, t+1}^{\text{r}} + SD_i \cdot w_{i, t+1}^{\text{r}} \\
p_{i, t+1}, & \text{otherwise }
\end{cases}
\]

\paragraph{Update power output}:
The power output is then updated by multiplying it with its on-off status.
\[p_{i, t+1} = u_{i, t+1}^{\text{r}} \cdot p_{i, t+1}^{\text{r}} \]

\paragraph{Update voltage angle}
\[\theta_{n, t+1} = \theta_{n, t+1}\]

\paragraph{Intermeidate state, load unfulfillment and power reserve}
We compute the following intermediate states to ease the computation of rewards. 
\[f_{k,t+1} = B_k (\theta_{k(n), t+1} - \theta_{k(m), t+1}) \]
\[f_{k,t+1}^{\text{r}} = \max \big( \min (F_{\max k}, f_{k,t+1}  ), F_{\min k} \big) \]
\[s_{n, t+1} = \max \big( D_{n, t+1} - \sum_{i \in \mathcal{G}_n} p_{i, t+1} - \sum_{k \in \delta^+(n)} f_{k,t+1}^\text{r} + \sum_{k \in \delta^-(n)} f_{k,t+1}^\text{r}, 0 \big) \]
\[r_{i, t+1} = \max \big( \min (P_{\max i} \cdot u_{i, t+1} - p_{i, t+1}, RU_{i} \cdot u_{i, t} + SU_{i} \cdot v_{i, t+1} ), 0 \big) \]
Although the power flow $f_{k,t+1}$ is clipped for subsequent computation, the voltage angles $\theta_{n,t+1}$ are not repaired accordingly because future states are independent of $\theta_{n, t+1}$ and they will not accumulate further violations. 

\paragraph{Demand forecast}:
\[D_{n, t+2:t+W+1} = \text{forecast}(D_{n, t+1:t+W})\]

\subsubsection{Cost Function}

\paragraph{Minimum up-time \& down-time violations}:
The penalty for correcting invalid on-off status decisions is the number of minimum up-time and down-time violations multiplied by a penalty factor. The violations represent that the agent attempted to turn off a power unit that must be kept on or turn on a power unit that must be kept off at the current time step. 
\[
C^{\text{UTDT}}_{t} = \sum_{i}
 P \cdot \mathbb{I}\big( \sum v_{i,t+1}^{\text{seq}}  > u_{i, t+1} \lor  \sum w_{i,t+1}^{\text{seq}}  > 1 - u_{i, t+1} \big)
\]

\paragraph{Ramp-up \& ramp-down violations}: The penalty for correcting invalid power output decisions is the magnitude of ramp-up and ramp-down violations multiplied by a penalty factor. The violations represent that the agent attempted to increase or decrease a power output too aggressively, which exceeds the generator's allowable ramping limits between consecutive time steps.
\[
C^{\text{Ramp}}_{t} =  P \cdot
\sum_{i} \max \big( p_{i, t+1} - p_{i, t} - RU_i \cdot u_{i, t} - SU_i \cdot v_{i, t+1}^{\text{r}}, p_{i, t} - p_{i, t+1} - RD_i \cdot u_{i, t+1}^{\text{r}} - SD_i \cdot w_{i, t+1}^{\text{r}}, 0 \big) 
\]

\paragraph{Transmission capacity violations}: The penalty for correcting invalid volatage angle decisions is the magnitude of transmission capacity violations multiplied by a penalty factor. The violations represent that the agent attempted to allocate an excessive or insufficient amount of power to other buses, exceeding or falling short of the transmission capacity. 

\[
C^{\text{Cap}}_{t} =  P \cdot
\sum_{k} \max \big( f_{k,t+1} - F_{\max k}, F_{\min k}- f_{k,t+1}, 0 \big) \\
\]

\paragraph{Total Cost}

\[
C_t = C^{\text{UTDT}}_t + C^{\text{Ramp}}_t + C^{\text{Cap}}_{t}
\]

\subsubsection{Reward Function}
The agent receives a reward equal to the negative of the production generation cost, startup cost, shutdown cost, load shedding cost and reserve shortfall cost. 

\textbf{Production generation reward}:
\[\Pi_t^{\text{pg}} = - \sum_{i}( a_i \cdot p_{i, t}^2 + b_i \cdot p_{i, t} + c_i)\]

\textbf{Startup reward}:
\[\Pi_t^{v} = - \sum_{i}( C^v_i \cdot v_{i,t})\]

\textbf{Shutdown reward}:
\[\Pi_t^{w} = - \sum_{i}( C^w_i \cdot w_{i,t})\]

\textbf{Load shedding reward}:
\[\Pi_t^{LS} = - C^{LS} \cdot \sum_{n} s_{n,t} \]
where $s_{n, t}$ represents the load unfulfillment at bus $n$.

\textbf{Reserve shortfall reward}:
\[\Pi_t^{R} = - C^{R} \cdot \max(R - \sum_{i} r_{i, t}, 0), \]
where $r_{i, t}$ represents the power that can be reserved at generator $i$.

\textbf{Total reward}:
\[
\Pi_t = \Pi_t^{\text{pg}} + \Pi_t^{v} + \Pi_t^{LS} + \Pi_t^{R}
\]

\subsubsection{Episode Dynamics}
The episode commences with the known sequence of the on-off status and power output from the preceding time step, and terminates after \( T \) time steps.

\subsection{Generation and Transmission Expansion Planning(\texttt{GTEPEnv})}
\subsubsection{Overview}

The generation and transmission expansion problem is a critical planning task in power systems, aiming to determine when and where to install new generators and transmission lines to ensure adequate power supply amidst growing and spatially distributed demand. This problem is inherently combinatorial and must satisfy constraints such as the maximum number of generators in a region and overall demand satisfaction \citep{li2022mixed}.

\begin{figure}[]
    \centering
    \includegraphics[width=0.5\linewidth]{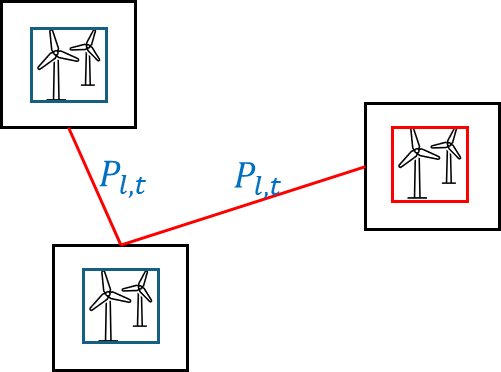}
    \caption{Schematic of GTEPEnv: This is a representation of a network of 3 regions where we install solar panels and transmission lines to deal with the demand for power. The solar panels with blue borders represent those already installed (state), and the one with the red border represents the one being installed in this time period (action)}
    \label{fig:GTEP_schematic}
\end{figure}

Because electricity demand varies over time and across regions, expansion decisions must be made periodically, turning the problem into a sequential decision-making process. Moreover, future demand is typically forecasted and therefore uncertain, necessitating a rolling planning framework in which decisions are continually updated as new information becomes available. This need is further heightened by the increasing integration of renewable energy sources and the growing volatility in energy consumption patterns, both of which introduce greater uncertainty into power system operations. As such, effective solutions must explicitly account for these uncertainties and remain flexible enough to adapt to frequent forecast revisions.
Keeping these considerations in mind, reinforcement learning (RL) offers a promising approach for addressing the generation and transmission expansion problem \citep{en17092167}. In particular, safe reinforcement learning techniques enable agents to learn actions that are more likely to be both feasible and near-optimal. To this end, we model the problem as a CMDP and implement it as the environment \texttt{GTEPEnv}, which supports decision-making regarding the installation of generators and transmission lines across different time periods and regions. We assume generators operate at full capacity and the cost for curtailment is negligible. A schematic of the environment is shown in Figure \ref{fig:GTEP_schematic}.
\subsubsection{Problem Setup}

We consider a multi-period generator and transmission expansion planning problem. The sets and known parameters used in the model are described below.

\paragraph{Sets}
\begin{itemize}
    \item \( \mathcal{T} = \{1, \ldots, T\} \): Set of discrete time periods.
    \item \( \mathcal{R} \): Set of regions.
    \item \( \mathcal{G} \): Set of generator types.
     \item \( \mathcal{L} \subseteq \mathcal{R} \times \mathcal{R} \): Set of candidate transmission lines, where each line is represented by a single directed pair \( (r_1, r_2) \) with \( r_1 \neq r_2 \). For any unordered pair \( \{r_1, r_2\} \), only one direction (e.g., \( (r_1, r_2) \)) is included in \( \mathcal{L} \). 
\end{itemize}

\paragraph{Parameters}
\begin{itemize}
    \item \( \text{Dem}_{r,t} \): Electricity demand in region \( r \in \mathcal{R} \) at time \( t \in \mathcal{T} \).
    \item \( \text{Cap}^{\text{gen}}_i \): Capacity of a single unit of generator type \( i \in \mathcal{G} \).
    \item \( C^{\text{inst,gen}}_i \): Installation cost of generator type \( i \).
    \item \( \text{Cap}^{\text{tl}}_l \): Transmission capacity of line \( l \in \mathcal{L} \).
    \item \( C^{\text{inst,tl}}_l \): Installation cost of transmission line \( l \).
    \item \( M_{i,r} \): Maximum number of generators of type \( i \) that can be installed in region \( r \).
    \item \( \lambda_0 \): Fixed penalty term (analogous to an \( \ell_0 \)-style penalty to encourage sparsity).
    \item \( \lambda_2 \): Quadratic penalty coefficient (analogous to an \( \ell_2 \)-style penalty to discourage overuse or smooth solutions).
    \item \(k\): Window length for demand forecast
     \item $\epsilon$: A small threshold used to ignore negligible power flows
\end{itemize}

\subsubsection{State Space}

The state at each time step \( t \in \mathcal{T} \), denoted \( s_t \), includes the following components:

\begin{itemize}
    \item \( n_{i,r,t} \in \mathbb{Z}_{\geq 0} \), bounded by \( [0, M_{i,r}] \): Number of generators of type \( i \in \mathcal{G} \) installed in region \( r \in \mathcal{R} \) at time \( t \in \mathcal{T} \).  
This component contributes a state dimension of \( |\mathcal{G}| \times |\mathcal{R}| \).
    
    \item \( nt_{l, t} \in \{0, 1\} \): Binary indicator for whether transmission line \( l \in \mathcal{L} \) is installed at time \( t \). This contributes a dimension of \( |\mathcal{L}| \).
    
    \item \( Dem_{r, t+1:t+k} \): Forecasted demand in region \( r \in \mathcal{R} \) from time \( t+1 \) to \( t+k \), where \( k \) is the forecasting window. This has a dimension of \( |\mathcal{R}| \times k \).  Entries are padded with zeros for time periods beyond the episode horizon.
    
    \item \( t \): Current time index, optionally included to provide temporal context.
\end{itemize}

All components are concatenated and flattened into a continuous vector for use in reinforcement learning.

\subsubsection{Action Space}

At each time step \( t \in \mathcal{T} \), the agent selects the following actions:

\begin{itemize}
    \item \( n^{\text{add}}_{i, r, t} \in [0, M_{i,r}] \): Number of generators of type \( i \in \mathcal{G} \) to install in region \( r \in \mathcal{R} \) at time \( t \). This defines a decision space of dimension \( |\mathcal{G}| \times |\mathcal{R}| \).
    
    \item \( P_{l, t} \in [-\text{Cap}^{\text{tl}}_l, \text{Cap}^{\text{tl}}_l] \): Power flow along transmission line \( l = (r_1, r_2) \in \mathcal{L} \) at time \( t \), where the direction of flow is from \( r_1 \) to \( r_2 \). This defines a space of dimension \( |\mathcal{L}| \).
\end{itemize}

The bounds of the action space are given by the physical constraints on installation limits and transmission capacities.

For reinforcement learning, actions are normalized to \( [-1, 1] \) and then scaled back to their original ranges using:
\[
a_{\text{actual}} = \frac{(a_{\text{normalized}} + 1)}{2} \cdot (a_{\text{high}} - a_{\text{low}}) + a_{\text{low}}
\]

\subsubsection{Transition Dynamics}

After scaling, we denote the set of actions by
\[
\left\{ n^{\text{add,action}}_{i,r,t},\quad 
P^{\text{action}}_{l,t} \right\}.
\]

We first further process the actions by rounding the number of generators added to the nearest integer and setting negligible power flows to 0. 
\[
n^{\text{add,prebound}}_{i,r,t} = \text{round}(n^{\text{add,action}}_{i,r,t})
\]

\[
P_{l,t} = 
\begin{cases}
0 & \text{if } |P^{\text{action}}_{l,t}| \leq \epsilon \\
P^{\text{action}}_{l,t} & \text{otherwise}
\end{cases}
\]

We then increment \( t \) to \( t + 1 \).
\paragraph{Checking action for generator bounds:}
Next, we proceed to check for violations of bounds in the state. The main focus is on checking for violations related to the number of generators in each region, as everything else has been accounted for in the scaling process. If the action results in a violation of the constraint, we adjust the action to ensure the state reaches the maximum possible configuration within the bounds.

\[
n^{\text{add}}_{i,r,t} = 
\begin{cases}
M_{i,r} - n_{i,r,t-1} & \text{if } n_{i,r,t-1} + n^{\text{add,prebound}}_{i,r,t} > M_{i,r} \\
n^{\text{add,prebound}}_{i,r,t} & \text{otherwise}
\end{cases}
\]

After sanitizing the actions and ensuring the state satisfies its bounds, the state is updated as follows:

\paragraph{Generator Updates:}
\[
n_{i,r,t} = n_{i,r,t-1} + n^{\text{add}}_{i,r,t}
\]

\paragraph{Transmission Line Installation:}
\[
nt_{l,t} = 
\begin{cases}
1 & \text{if } nt_{l,t-1} = 0 \text{ and } |P_{l,t}| > 0 \\
nt_{l,t-1} & \text{otherwise}
\end{cases}
\]

We then proceed to calculate the cost and reward. Finally, we shift the demand forecast by one step to \( \text{Dem}_{r,t+1:t+k} \).

\subsubsection{Cost Function}

The total cost at each time step consists of penalties for constraint violations, including those for the bounds of the state, and demand satisfaction violations. The components of the cost are defined as follows:

   \paragraph{Generator Bound Violations: }
     We apply a combination of \(L_0\) and \(L_2\) penalties for generating an action before sanitization that violates the maximum number of generators in a region:
    \[
    C^{\text{bound,gen}}_{i,r} = 
    \begin{cases}
    \lambda_0 + \lambda_2 \cdot \left( n^{\text{add,prebound}}_{i,r,t} + n_{i,r,t-1} - M_{i,r} \right)^2 & \text{if } n^{\text{add,prebound}}_{i,r,t} + n_{i,r,t-1} > M_{i,r} \\
    0 & \text{otherwise}
    \end{cases}
    \]

    \paragraph{Demand Violations:}
    The available power \( \text{Pow}^{\text{avail}}_{r,t} \) in region \( r \) at time \( t \) is calculated as:
    \[
    \text{Pow}^{\text{avail}}_{r,t} = \sum_i n_{i,r,t} \cdot  \text{Cap}^{\text{gen}}_i + \sum_{r'|(r,r') \in \mathcal{L}} P_{(r,r'),t} - \sum_{r'|(r',r) \in \mathcal{L}} P_{(r',r),t}
    \]
    \textbf{Note:} The power flow obtained from the agent is only in one direction. The power flow along the reverse direction can be interpreted simply the negative of the power flow in the original direction. Specifically, for any transmission line \( (r_1, r_2) \in \mathcal{L} \), the power flow from region \( r_1 \) to region \( r_2 \) is denoted by \( P_{(r_1, r_2), t} \), and the reverse flow from region \( r_2 \) to region \( r_1 \) is  \( P_{(r_2, r_1), t} = -P_{(r_1, r_2), t} \).

    We apply a combination of \(L_0\) and \(L_2\) penalties for unmet demand:

    \[C^{\text{demand}}_r = \begin{cases}
    \lambda_0 + \lambda_2 \cdot  \left( \text{Dem}_{r,t} - \text{Pow}^{\text{avail}}_{r,t} \right)^2  & \text{if } \text{Dem}_{r,t} > \text{Pow}^{\text{avail}}_{r,t} \\
    0 & \text{otherwise}
    \end{cases}\]

\paragraph{Total cost:}The total cost is the sum of the penalties for all violations, as follows:
\[
C_t = \sum_{i,r} C^{\text{bound,gen}}_{i,r} + \sum_{r} C^{\text{demand}}_r
\]
The total cost is used to guide the agent toward optimal decision-making.

\subsubsection{Reward Function}

The reward function consists of:
\paragraph{Contribution to reward from installation of generators}:
\[\pi^{gen}_t = - \sum_{i,r} n^{\text{add}}_{i,r,t} \cdot C^{\text{inst,gen}}_i\]

\paragraph{Contribution to reward from installation of transmission lines}:
\[\pi^{tl}_t = -  \sum_{l \in \mathcal{L}} C^{\text{inst,tl}}_l \cdot \mathbbm{1}[nt_{l,t-1} = 0 \wedge |P_{l,t}| > 0]\]
\paragraph{Total reward}:
\[
\pi_t = \pi^{gen}_t+\pi^{tl}_t
\]

\subsection{Multiperiod Blending Problem(\texttt{BlendingEnv})}
\subsubsection{Overview}

The multiperiod blending problem, a core challenge in chemical engineering, involves optimally blending multiple input streams to produce outputs with desired quality attributes over a series of time periods. Input streams, characterized by specific properties, are procured and stored in inventory vessels. These streams are then transferred to blenders, where they are mixed according to specified blending rules. The resulting output streams must meet property constraints (e.g., quality specifications) and are subsequently stored in output inventory vessels before being sold, subject to upper bounds on product quantities.

The goal is typically to maximize profit while satisfying constraints on inventory levels, property ranges, and operational rules—such as prohibiting simultaneous inflow and outflow in the same blender. These requirements lead to a highly nonlinear and constrained formulation, commonly modeled as a non-convex mixed-integer quadratically constrained program (MIQCP) \citep{Chen2020}.

\begin{figure}[htbp]
    \centering
    \includegraphics[width=0.85\linewidth]{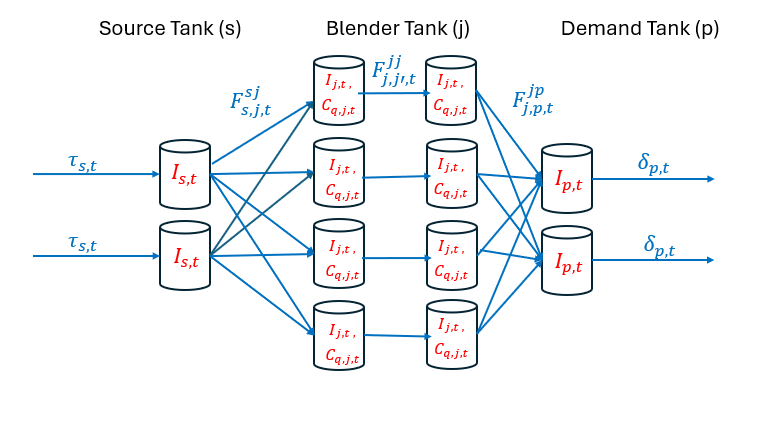}
    \caption{Schematic of BlendingEnv: This is a representation of a blending system with the blue arrows representing the flows between different components (action) and the red variables representing the different inventories and properties of the blender (state)}
    \label{fig:Blending_schematic}
\end{figure}

Due to the sequential decision-making structure of the problem and the uncertainty in demands or property variations over time, reinforcement learning (RL) emerges as a promising approach. In particular, safe RL techniques can guide agents to learn actions that are not only near-optimal but also likely to satisfy complex constraints. To that end, we model the problem as a Constrained Markov Decision Process (CMDP) and implement it as the environment \texttt{BlendingEnv}, which supports dynamic decisions on stream flows, purchasing, and selling over time.
A schematic of the environment is shown in Figure \ref{fig:Blending_schematic}.
\subsubsection{Problem Setup}

 The sets and known parameters used to describe the environment are shown below:

\paragraph{Sets}
\begin{itemize}
    \item \( \mathcal{T} = \{1, \ldots, T\} \): Set of discrete time periods.
    \item \( \mathcal{S} \): Set of source streams.
    \item \( \mathcal{J} \): Set of blenders.
    \item \( \mathcal{P} \): Set of demand streams.
    \item \( \mathcal{Q} \): Set of stream properties  (e.g., chemical or physical characteristics).
    \item \( \mathcal{F}^{s,j} \): Set of tuples representing possible directed flows from source streams to blenders.
\item \( \mathcal{F}^{j,j} \): Set of tuples representing possible directed flows between blenders. Flows are defined in such a way that only one way is possible.
\item \( \mathcal{F}^{j,p} \): Set of tuples representing possible directed flows from blenders to demand streams.
    
\end{itemize}

\paragraph{Parameters}
\begin{itemize}
    \item \( F^{\max} \): Upper bound for any flow between nodes.
    \item \( \sigma_{s,q} \): Value of property \( q \in \mathcal{Q} \) for source stream \( s \in \mathcal{S} \).
    \item \( [s^{\text{lb}}_s, s^{\text{ub}}_s] \): Lower and upper bounds on inventory of source \( s \in \mathcal{S} \).
    \item \( \tau^0_{s,t} \): Availability of source stream \( s \in \mathcal{S} \) at time \( t \in \mathcal{T} \).
    \item \( [b^{\text{lb}}_j, b^{\text{ub}}_j] \): Lower and upper inventory bounds for blender \( j \in \mathcal{J} \).
    \item \( [\sigma^{\text{lb}}_{p,q}, \sigma^{\text{ub}}_{p,q}] \): Lower and upper bounds on property \( q \in \mathcal{Q} \) for demand stream \( p \in \mathcal{P} \).
    \item \( [d^{\text{lb}}_p, d^{\text{ub}}_p] \): Lower and upper bounds on inventory of demand stream \( p \in \mathcal{P} \).
    \item \( \delta^0_{p,t} \): Maximum amount of demand stream \( p \in \mathcal{P} \) that can be fulfilled at time \( t \in \mathcal{T} \).
    \item \( \beta^d_p \): Unit selling price for demand stream \( p \in \mathcal{P} \).
    \item \( \beta^s_s \): Unit purchase cost of source stream \( s \in \mathcal{S} \).
    \item \( \beta \): Unit cost of intermediate flows (e.g., between nodes).
    \item \( \alpha \): Fixed cost of activating an intermediate flow.
    \item \(k\): Window length for source and demand forecast
   \item $strategy$: The strategy used in the environment to handle illegal actions caused by constraint violations. We support three options: $prop$, $disable$, and $none$. In the $prop$ strategy, we adjust a subset of actions when they violate the relevant constraints, with actions scaled or clipped depending on the specific constraint. In the $disable$ strategy,  a subset of actions is set directly to zero . The $none$ strategy applies no correction, allowing actions to remain unchanged regardless of constraint violations.
    \item \( \lambda_{0,B} \): a fixed cost for violating bound constraint (analogous to \( \ell_0 \) regularization).
    \item \( \lambda_{B} \): prefactor to the sum of \( \ell_0 \) and \( \ell_1 \)  violations for inventory bound constraint (analogous to \( \ell_0 \) regularization).
    \item \( \lambda_{0,M} \): a fixed cost for violating in-out rule constraint (analogous to \( \ell_0 \) regularization).
    \item \( \lambda_{0,Q} \): a fixed cost for violating property specifications (analogous to \( \ell_0 \) regularization).
   \item $\epsilon$: A small positive threshold used to disregard negligible violations of constraints.
    
\end{itemize}

\subsubsection{State Space}

The state \( s_t \) at time \( t \in \mathcal{T} \) includes the following components, with dimensionality expressed using the sets defined previously:

\begin{itemize}
    \item Source inventories: \( I^s_{s,t} \in [s^{\text{lb}}_s, s^{\text{ub}}_s] \), for all \( s \in \mathcal{S} \).  
    This has dimension \( |\mathcal{S}| \).

    \item Blender inventories: \( I^b_{j,t} \in [b^{\text{lb}}_j, b^{\text{ub}}_j] \), for all \( j \in \mathcal{J} \).  
    This has dimension \( |\mathcal{J}| \).

    \item Demand inventories: \( I^d_{p,t} \in [d^{\text{lb}}_p, d^{\text{ub}}_p] \), for all \( p \in \mathcal{P} \).  
    This has dimension \( |\mathcal{P}| \).

    \item Blender properties: \( C_{j,q,t} \in \mathbb{R} \), for all \( j \in \mathcal{J}, q \in \mathcal{Q} \).  
    This has dimension \( |\mathcal{J}| \times |\mathcal{Q}| \).

    \item  \( \tau^0_{s, t+1:t+k} \): Forecasted source availability for all \( s \in \mathcal{S}\) from time t+1 to t+k where \( k \) is the forecasting window.  
This component has a dimension of \( |\mathcal{S}| \times k \). Entries are padded with zeros for time periods beyond the episode horizon.

\item \( \delta^0_{p, t+1:t+k} \) Forecasted demand availability, for all \( p \in \mathcal{P}\) from time t+1 to t+k where \( k \) is the forecasting window.
This component has a dimension of \( |\mathcal{P}| \times k \). Entries are padded with zeros for time periods beyond the episode horizon.

    \item Time step: \( t \in \mathcal{T} \).  
    This is a scalar.
\end{itemize}

\subsubsection{Action Space}

At each time step \( t \in \mathcal{T} \), the agent selects the following actions:

\begin{itemize}
    \item Source purchases: \( \tau_{s,t} \in [0, \tau^0_{s,t}] \), for all \( s \in \mathcal{S} \).  
    This has dimension \( |\mathcal{S}| \). 

    \item Demand sales: \( \delta_{p,t} \in [0, \delta^0_{p,t}] \), for all \( p \in \mathcal{P} \).  
    This has dimension \( |\mathcal{P}| \). 

    \item Source-to-blender flows: \( F^{sj}_{s,j,t} \in [0, F^{\max}] \), for all \( (s,j) \in \mathcal{F}^{s,j} \).  
    This has dimension \( |\mathcal{F}^{s,j} |\).

    \item Blender-to-blender flows: \( F^{jj}_{j,j',t} \in [0, F^{\max}] \), for all \( (j,j') \in \mathcal{F}^{j,j}  \).  
    This has dimension \( |\mathcal{F}^{j,j} | \).

    \item Blender-to-demand flows: \( F^{jp}_{j,p,t} \in [0, F^{\max}] \), for all \( (j,p) \in \mathcal{F}^{j,p}  \).  
    This has dimension \( |\mathcal{F}^{j,p} | \).
\end{itemize}

Each action is initially normalized to the interval \( [-1, 1] \) and mapped to its actual value using affine transformation based on the lower and upper bounds of the corresponding action:
\[
a^{\text{actual}} = \frac{(a^{\text{normalized}} + 1)}{2} \cdot (a^{\text{high}} - a^{\text{low}}) + a^{\text{low}}
\]
Here, \( a^{\text{low}} \) and \( a^{\text{high}} \) are the lower and upper bounds for each action dimension as specified above.

\subsubsection{Transition Dynamics}
After scaling the actions, we increment \( t \) to \( t + 1 \).
Let us denote the set of actions at this point as
\[
\left\{\tau^{\text{prebound}}_{s,t},\quad 
\delta^{\text{prebound}}_{p,t},\quad 
F^{\text{sj,prebound}}_{s,j,t},\quad 
F^{\text{jj,preinout}}_{j,j',t},\quad 
F^{\text{jp,preinout}}_{j,p,t}\right\}.
\]
respectively.
\paragraph{Checking action for source inventory bounds:}
We next proceed to check if the actions provide a source inventory that violates the bounds. If there is a violation, we adjust the actions based on the strategy chosen.
\[I_{\text{new},s}^s = I^s_{s,t-1} - \sum_{(s,j) \in \mathcal{F}^{s,j}} F^{sj,prebound}_{s,j,t} + \tau^{prebound}_{s,t}\]
\[F^{sj}_{s,j,t|(s,j) \in \mathcal{F}^{s,j}} = \begin{cases}
     F^{sj,prebound}_{s,j,t}\left(\frac{I^s_{s,t-1} + \tau^{prebound}_{s,t} - s^{lb}_s}{\sum_{(s,j) \in \mathcal{F}^{s,j}} F^{sj,prebound}_{s,j,t}}\right) & \text{if  }  I_{\text{new},s}^s<s^{lb}_s-\epsilon, strategy = prop \\
     0 &  \text{if  }  I_{\text{new},s}^s<s^{lb}_s-\epsilon, strategy = disable \\
      F^{sj,prebound}_{s,j,t} & \text{otherwise}
\end{cases}\]

\[\tau_{s,t} = \begin{cases}
     \min(s^{ub}_s + \sum_{(s,j) \in \mathcal{F}^{s,j}} F^{sj,prebound}_{s,j,t} - I^s_{s,t-1}, \tau^0_{s,t}) & \text{if  }  I_{\text{new},s}^s>s^{ub}_s+\epsilon, strategy = prop \\
     0 &  \text{if  }  I_{\text{new},s}^s>s^{ub}_s+\epsilon, strategy = disable \\
     \tau^{prebound}_{s,t} & \text{otherwise}
\end{cases}\]
\paragraph{Source inventory updates}
We then update the source inventory as follows:
\[
I^s_{s,t} = \text{clip}(I^s_{s,t-1} - \sum_{(s,j) \in \mathcal{F}^{s,j}} F^{sj}_{s,j,t} + \tau_{s,t}, \; s^{lb}_s, \; s^{ub}_s)
\]
\paragraph{Checking action for in-out rule violations:}
We now check if the actions follow the in-out rule for blenders. That is, blenders should not have simultaneous inflow and outflow of streams. If there is such a case, we set all corresponding outflows to 0. The value of other flows are not changed.
\[\text{Inflow}_j = \sum_{(s,j) \in \mathcal{F}^{s,j}} F^{sj}_{s,j,t} + \sum_{(j',j) \in \mathcal{F}^{j,j}} F^{jj,preinout}_{j',j,t}\]
\[\text{Outflow}_j = \sum_{(j,j') \in \mathcal{F}^{j,j}}F^{jj,preinout}_{j,j',t}  +\sum_{(j,p) \in \mathcal{F}^{j,p}} F^{jp,preinout}_{j,p,t}\]
\[F^{jp,prebound}_{j,p,t|(j,p) \in \mathcal{F}^{j,p}} = \begin{cases}
    0 & \text{if }\text{Inflow}_j>\epsilon, \text{Outflow}_j>\epsilon, strategy \neq no \\
    F^{jp,preinout}_{j,p,t} & \text{otherwise}
\end{cases}\]
\[F^{jj,prebound}_{j,j',t|(j,j') \in \mathcal{F}^{j,j}} = \begin{cases}
    0 & \text{if }\text{Inflow}_j>\epsilon, \text{Outflow}_j>\epsilon, strategy \neq no \\
    F^{jj,preinout}_{j,j',t} & \text{otherwise}
\end{cases}\]

\paragraph{Checking action for blender inventory lower bound:}
We next proceed to check if the actions provide a blender inventory that violates the lower bounds. 

\begin{equation*}
\begin{split}
I_{\text{new},j}^b &= I^b_{j,t-1} 
+ \sum_{(s,j) \in \mathcal{F}^{s,j}} F^{sj}_{s,j,t} 
+ \sum_{(j',j) \in \mathcal{F}^{j,j}} F^{jj,prebound}_{j',j,t} \\
&\quad - \sum_{(j,j') \in \mathcal{F}^{j,j}} F^{jj,prebound}_{j,j',t}
- \sum_{(j,p) \in \mathcal{F}^{j,p}} F^{jp,prebound}_{j,p,t}
\end{split}
\end{equation*}
If there is a violation, we adjust the actions based on the strategy chosen. We keep the same actions otherwise.

\[F^{jp}_{j,p,t|(j,p) \in \mathcal{F}^{j,p}} = \begin{cases}
     F^{jp,prebound}_{j,p,t}F^{j,prop}_{j,t} & \text{if  }  I_{\text{new},j}^b<b^{lb}_j-\epsilon,\\ & strategy = prop \\
     0 &  \text{if  }  I_{\text{new},j}^b<b^{lb}_j-\epsilon, \\ & strategy = disable \\
      F^{jp,prebound}_{j,p,t} & \text{otherwise}
\end{cases}\]

\[F^{jj}_{j,j',t|(j,j') \in \mathcal{F}^{j,j}} = \begin{cases}
     F^{jj,prebound}_{j,j',t}F^{j,prop}_{j,t} & \text{if  }  I_{\text{new},j}^b<b^{lb}_j-\epsilon, \\ & strategy = prop \\
     0 &  \text{if  }  I_{\text{new},j}^b<b^{lb}_j-\epsilon, \\ & strategy = disable \\
      F^{jj,prebound}_{j,j',t} & \text{otherwise}
\end{cases}\]
\[F^{j,prop}_{j,t} = \frac{I^b_{j,t-1} + \sum_{(s,j) \in \mathcal{F}^{s,j}} F^{sj}_{s,j,t} + \sum_{(j',j) \in \mathcal{F}^{j,j}} F^{jj,prebound}_{j',j,t}-b^{lb}_j}{ \sum_{(j,j') \in \mathcal{F}^{j,j}}F^{jj,prebound}_{j,j',t} + \sum_{(j,p) \in \mathcal{F}^{j,p}} F^{jp,prebound}_{j,p,t}}\]
\paragraph{Blender inventory updates:}
After making the adjustments described above, we update the blender inventory.
\[
I^b_{j,t} = \text{clip}(I^b_{j,t-1} + \sum_{(s,j) \in \mathcal{F}^{s,j}} F^{sj}_{s,j,t} + \sum_{(j',j) \in \mathcal{F}^{j,j}} F^{jj}_{j',j,t} - \sum_{(j,j') \in \mathcal{F}^{j,j}}F^{jj}_{j,j',t} - \sum_{(j,p) \in \mathcal{F}^{j,p}} F^{jp}_{j,p,t}, \; b^{lb}_j, \; b^{ub}_j)
\]
\paragraph{Checking action for demand inventory lower bound:}
We now check if the actions provide a demand inventory that violates the lower bounds. 
\[I_{\text{new},p}^d = I^d_{p,t-1}+ \sum_{(j,p) \in \mathcal{F}^{j,p}} F^{jp}_{j,p,t} - \delta^{prebound}_{p,t}\]
If there is a violation, we adjust the actions based on the strategy chosen.
\[\delta_{p,t} = \begin{cases}
     I^d_{p,t-1}+ \sum_{(j,p) \in \mathcal{F}^{j,p}} F^{jp}_{j,p,t} - d^{lb}_p & \text{if  }  I_{\text{new},p}^d<d^{lb}_p-\epsilon, strategy = prop \\
     0 &  \text{if  }  I_{\text{new},p}^d<d^{lb}_p-\epsilon, strategy = disable \\
      \delta^{prebound}_{p,t} & \text{otherwise}
\end{cases}\]

\paragraph{Demand inventory updates:}
After making the adjustments described above, we update the demand inventory.
\[
I^d_{p,t} = \text{clip}(I^d_{p,t-1} + \sum_{(j,p) \in \mathcal{F}^{j,p}} F^{jp}_{j,p,t} - \delta_{p,t}, \; d^{lb}_p, \; d^{ub}_p)
\]
\paragraph{Blender property updates}
We then update the property of the materials in each blender as follows:

\[C_{j,q,t} = \begin{cases}
     = \frac{1}{I^b_{j,t}} \Big(C_{j,q,t-1} \cdot I^b_{j,t-1} + \sum_{(s,j) \in \mathcal{F}^{s,j}} F^{sj}_{s,j,t} \cdot \sigma_{s,q} & \\
+ \sum_{(j',j) \in \mathcal{F}^{j,j}}F^{jj}_{j',j,t} \cdot C_{j',q,t-1} &  \\
- \sum_{(j,j') \in \mathcal{F}^{j,j}} F^{jj}_{j,j',t} \cdot C_{j,q,t-1} & \\
- \sum_{(j,p) \in \mathcal{F}^{j,p}} F^{jp}_{j,p,t} \cdot C_{j,q,t-1} \Big) & \text{if } I^b_{j,t+1} > \epsilon \\
0 & \text{otherwise}
\end{cases}\]

We then proceed to calculate the cost and reward. Finally, we shift the future schedules forecast by one step to \( \tau^0_{s,t+1:t+k} \text{ and }\delta^0_{p,t+1,t+k}  \).

\subsubsection{Costs}

The total cost at each time step consists of penalties for constraint violations, including those for the violations of the bounds of the state and the in-out rule. The components of the cost are defined as follows:

\paragraph{Inventory Bound Violation Costs}:

Inventory levels at sources, blenders, and demands are required to remain within prescribed upper and lower bounds. Inventory levels computed with actions before the relevant upadates are used to assess constraint violations.

\textbf{Sources}:

\[
C^{source}_{s,t} = 
\begin{cases}
 \lambda_{B} \cdot (\lambda_{0,B}+I^s_{new,s} - s^{ub}_s), & \text{if } I^s_{new,s} > s^{ub}_s+\epsilon \\
\lambda_{B} \cdot (\lambda_{0,B}+s^{lb}_s-I^s_{new,s} ), & \text{if } I^s_{new,s} < s^{lb}_s-\epsilon \\
0, & \text{otherwise}
\end{cases}
\]

\textbf{Blenders}:

\[
C^{blender}_{j,t} = 
\begin{cases}
 \lambda_{B} \cdot (\lambda_{0,B}+I^b_{new,j} - b^{ub}_j), & \text{if } I^b_{new,j} > b^{ub}_j+\epsilon \\
\lambda_{B} \cdot (\lambda_{0,B}+b^{lb}_s-I^b_{new,j} ), & \text{if } I^b_{new,j} < b^{lb}_j-\epsilon \\
0, & \text{otherwise}
\end{cases}
\]

\textbf{Demands}:

\[
C^{demand}_{p,t} = 
\begin{cases}
\lambda_{B} \cdot (\lambda_{0,B}+I^d_{new,p} - d^{ub}_p), & \text{if } I^d_{new,p} > d^{ub}_p+\epsilon \\
\lambda_{B} \cdot (\lambda_{0,B}+d^{lb}_p-I^b_{new,j} ), & \text{if } I^d_{new,p} < d^{lb}_p-\epsilon \\
0, & \text{otherwise}
\end{cases}
\]

\paragraph{In-Out Rule Violation Costs}:

Blenders must not simultaneously receive and send flow within the same time step. If both incoming and outgoing flows are non-zero (beyond a tolerance \( \epsilon \)), a penalty is applied:

\[
C^{in-out}_{j,t} = 
\begin{cases}
\lambda_{0,M}, & \text{if } \text{Inflow}_j > \epsilon, \text{Outflow}_j > \epsilon \\
0, & \text{otherwise}
\end{cases}
\]

\paragraph{Property Violation Costs}:

Each demand \( p \) requires product properties \( q \) within bounds. If blender \( j \)'s content violates these bounds and is being delivered to \( p \), a penalty is applied:

\[
C^Q_{j,q,p,t} = 
\begin{cases}
\lambda_{0,Q}, & \text{if } (C_{j,q,t} < \sigma^{lb}_{p,q} - \epsilon \lor C_{j,q,t} > \sigma^{ub}_{p,q} + \epsilon) ,(j,p) \in \mathcal{F}^{j,p},F^{jp}_{j,p,t} > 0 \\
0, & \text{otherwise}
\end{cases}
\]

\paragraph{Total Cost}

The total penalty at time \( t \) is:

\[
C_t = \sum_{s \in S} C^{source}_{s,t} + \sum_{j \in J} C^{blender}_{j,t} + \sum_{p \in P} C^{demand}_{p,t} + \sum_{j \in J} C^{in-out}_{j,t} + \sum_{j \in J} \sum_{q \in Q} \sum_{p \in P} C^Q_{j,q,p,t}
\]

\subsubsection{Reward Function}
The reward function consists of:

\paragraph{Revenue from selling streams}:
\[
\pi^{sale}_t = \sum_{p \in P} \beta^d_p \cdot \delta_{p,t}
\]

\paragraph{Contribution to reward from buying streams}:
\[
\pi^{purchase}_t = -\sum_{s \in S} \beta^s_s \cdot \tau_{s,t}
\]

\paragraph{Contribution to reward from flows}:
\[
\pi^{flow}_t = -\alpha \cdot Q_{bin} - \beta \cdot Q_{float}
\]
where 
\[
Q_{bin} = \sum_{(s,j) \in \mathcal{F}^{s,j} }\mathbbm{1}[F^{sj}_{s,j,t} > 0] + \sum_{(j,p) \in \mathcal{F}^{j,p}}  \mathbbm{1}[F^{jp}_{j,p,t} > 0]
\]
\[
Q_{float} = \sum_{(s,j) \in \mathcal{F}^{s,j}} F^{sj}_{s,j,t} + \sum_{(j,p) \in \mathcal{F}^{j,p}}  F^{jp}_{j,p,t}
\]

\paragraph{Total Reward}:
\[
\pi_t = \pi^{sale}_t+ \pi^{purchase}_t + \pi^{flow}_t 
\]

\subsection{Multi-Echelon Inventory Management Environment (\texttt{InvMgmtEnv})}

\subsubsection{Overview}

The multi‐echelon inventory management problem involves coordinating replenishment orders across a five‐tier supply network—raw‐material suppliers, producers, distributors, retailers, and end‐markets—under time‐varying demand. At each period, the decision maker chooses continuous order quantities along each transportation route, with orders subject to fixed lead times before arrival. On‐hand and in‐transit inventories incur holding costs, while unmet demand is backlogged and penalized. The objective is to maximize cumulative net profit—total sales revenue minus procurement, operating, holding, and shortage penalty costs—over a finite planning horizon.

\begin{figure}[]
    \centering
    \includegraphics[width=0.85\linewidth]{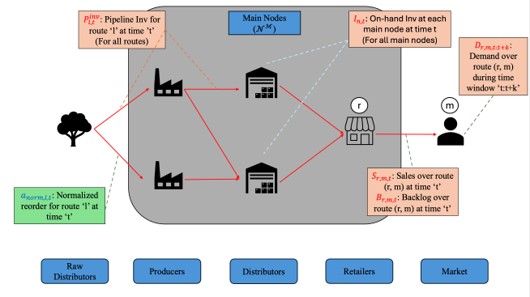}
    \caption{Schematic of \emph{InvMgmtEnv}. A multi‑echelon supply‑chain network with raw distributors, producers, distributors, and retailers serving a market. Directed red arrows depict transportation routes \(\ell\in\mathcal{L}\). Red callouts indicate observation components: \emph{on‑hand inventory} \(I_{n,t}\) for all main nodes \(n\in\mathcal{N}^{\mathcal{M}}\); \emph{pipeline inventory} \(P^{\mathrm{inv}}_{\ell,t}\) (time‑indexed along each route \(\ell\)); \emph{demand window} \(D_{r,m,t:t+k}\) on retail–market pairs \((r,m)\in\mathcal{RM}\); \emph{sales} \(S_{r,m,t}\); and \emph{backlog} \(B_{r,m,t}\). The blue callout marks the action—the \emph{normalized reorder} \(a_{\mathrm{norm},\ell,t}\) on each route \(\ell\). The shaded region highlights the main‑node set \(\mathcal{N}^{\mathcal{M}}\)}
    \label{fig:InvMgmt_schematic}
\end{figure}

We adopt a centralized, single‐product framework \citep{perez2021algorithmic} in which all replenishment decisions are made by a central planner, and customer demand at each retailer‐to‐market link is drawn from a known stationary distribution, such as a Gaussian with specified mean and variance. Transportation is lossless with deterministic lead times, and replenishment quantities are bounded by per‐route capacity, with any excess actions penalized. Episodes terminate after a fixed number of periods. A schematic of the environment is shown in figure \ref{fig:InvMgmt_schematic}.

\subsubsection{Problem Setup}

The sets and known parameters used to describe \texttt{InvMgmtEnv} are shown below.

\paragraph{Sets}
\begin{itemize}
  \item $\mathcal{T} = \{1,\dots,T\}$: set of discrete time periods.
  \item $\mathcal{M}$: set of market nodes.
  \item $\mathcal{R}$: set of retailer nodes.
  \item $\mathcal{D}$: set of distributor nodes.
  \item $\mathcal{P}$: set of producer nodes.
  \item $\mathcal{S}$: set of raw‐material supplier nodes.
  \item $\mathcal{N} = \mathcal{M}\cup\mathcal{R}\cup\mathcal{D}\cup\mathcal{P}\cup\mathcal{S}$: set of all nodes.
  \item $\mathcal{N^\mathcal{M}} = \mathcal{R}\cup\mathcal{D}\cup\mathcal{P}$: set of main nodes.

    \item $\mathcal{L}$: set of directed replenishment routes.
          Each route is written
          \[
            \ell
            \;=\;
            \bigl(\mathrm{orig}(\ell),\,\mathrm{dest}(\ell)\bigr),
          \]
          an ordered pair of nodes with
          \(\mathrm{orig}(\ell)\in\mathcal{N}\) the origin (shipping) node
          and \(\mathrm{dest}(\ell)\in\mathcal{N}\) the destination
          (receiving) node.
  \item $\mathcal{RM}$: set of retailer‐to‐market demand links.
\end{itemize}

\paragraph{Parameters}
\begin{itemize}
  \item $I^{\mathrm{init}}_{n}$, $n\in\mathcal{N}$: initial on‑hand inventory at node $n$.
  \item $\mathrm{Cap}^{\mathrm{route}}_{\ell}$, $\ell\in\mathcal{L}$: capacity limit of route $\ell$.
  \item $C^{\mathrm{hold,mat}}_{\ell}$, $\ell\in\mathcal{L}$: per‑unit holding cost for pipeline inventory on route $\ell$.
  \item $C_{\ell}$, $\ell\in\mathcal{L}$: procurement cost per unit ordered on route $\ell$.
  \item $LT_{\ell}$, $\ell\in\mathcal{L}$: fixed lead time (in periods) for route $\ell$.
  \item $C^{\mathrm{hold,inv}}_{n}$, $n\in\mathcal{N}$: per‑unit holding cost for on‑hand inventory at node $n$.
  \item $C^{\mathrm{oper}}_{p}$, $p\in\mathcal{P}$: operating cost per unit of production activity at producer $p$.
  \item $\eta_{p}$, $p\in\mathcal{P}$: production yield at producer $p$.
  \item $\mu_{r,m},\;\sigma_{r,m}$, $(r,m)\in\mathcal{RM}$: mean and standard deviation of demand on link $(r,m)$.
  \item $P_{r,m}$, $(r,m)\in\mathcal{RM}$: selling price per unit on link $(r,m)$.
  \item $C^{\mathrm{penalty}}_{r,m}$, $(r,m)\in\mathcal{RM}$: penalty cost per unit of unmet demand on link $(r,m)$.
  \item $k$: look‑ahead window length for demand forecast.
  \item $\varepsilon$: small threshold below which reorder quantities are treated as zero.
  \item $\phi_{\mathrm{action}}$: penalty factor for action‑bound violations.
  \item $\phi_{\mathrm{on\_hand}}$: penalty factor for on‑hand inventory violations.
  \item $\phi_{\mathrm{pipeline}}$: penalty factor for pipeline inventory violations.
  \item $\phi_{\mathrm{sales}}$: penalty factor for sales‑state violations.
  \item $\phi_{\mathrm{backlog}}$: penalty factor for backlog‑state violations.
\end{itemize}

\subsubsection{State Space}

The state $s_{t}$ at time $t\in\mathcal{T}$ is represented by the tuple
\[
s_{t} = \bigl(I_{n,t},\,P^{\mathrm{inv}}_{\ell,t},\,S_{r,m,t},\,B_{r,m,t},\,D_{r,m,t:t+k},\,t\bigr),
\]
where:
\begin{itemize}
  \item \emph{On‐hand inventory levels:} 
    \[
      I_{n,t}\in\mathbb{R}_{\ge0},\quad n\in \mathcal{N^\mathcal{M}},
    \]
    contributing $|\mathcal{N^\mathcal{M}}|$ dimensions.

  \item \emph{Pipeline inventory:} 
    \[
      P^{\mathrm{inv}}_{\ell,t}
      = \bigl(P_{\ell,1,t},\dots,P_{\ell,LT_{\ell},t}\bigr),\quad \ell\in\mathcal{L},
    \]
    contributing $\sum_{\ell\in\mathcal{L}}LT_{\ell}$ dimensions.

  \item \emph{Sales:} 
    \[
      S_{r,m,t}\in\mathbb{R}_{\ge0},\quad (r,m)\in\mathcal{RM},
    \]
    contributing $|\mathcal{RM}|$ dimensions.

  \item \emph{Backlog:} 
    \[
      B_{r,m,t}\in\mathbb{R}_{\ge0},\quad (r,m)\in\mathcal{RM},
    \]
    contributing $|\mathcal{RM}|$ dimensions.

    \item \emph{Demand window:} 
        \[
          D_{r,m,t:t+k}
          =\bigl(D_{r,m,t},\dots,D_{r,m,t+k}\bigr),\quad (r,m)\in\mathcal{RM},
        \]
        where, for offsets $h$ with $t+h > T$, the unavailable future demand
        $D_{r,m,t+h}$ is padded with $0$.  
        This contributes $|\mathcal{RM}|\times k$ dimensions.

  \item \emph{Time step:} (optional) the scalar $t$, contributing $1$ dimension.
\end{itemize}

All components are concatenated and flattened into a continuous observation vector of total dimension
\[
|\mathcal{N^\mathcal{M}}| \;+\;\sum_{\ell\in\mathcal{L}}LT_{\ell}\;+\;2\,|\mathcal{RM}|\;+\;|\mathcal{RM}|\times k\;+\;1.
\]

\subsubsection{Action Space}

At each decision epoch $t\in\mathcal{T}$ the agent outputs a
$|\mathcal{L}|$–tuple of normalized actions
\[
\mathbf{a}_{\mathrm{norm},t}
  \;=\;
  \bigl(a_{\mathrm{norm},\ell,t}\bigr)_{\ell\in\mathcal{L}}
  \;\in\;[-1,1]^{|\mathcal{L}|},
\]
whose components correspond one‑to‑one with the directed replenishment routes
$\ell\in\mathcal{L}$.

\paragraph{Scaling to Physical Reorder Quantities.}
Each normalized component is linearly mapped to a true reorder quantity,
respecting the capacity of its route:
\[
Q_{\ell,t}
  \;=\;
  \frac{a_{\mathrm{norm},\ell,t}+1}{2}\,
  \mathrm{Cap}^{\mathrm{route}}_{\ell},
  \qquad \ell\in\mathcal{L}.
\]
A small‑order cutoff $\varepsilon$ is then applied:
\[
Q_{\ell,t}
  \;=\;
  \begin{cases}
    0, & Q_{\ell,t}\le\varepsilon,\\[4pt]
    Q_{\ell,t}, & \text{otherwise}.
  \end{cases}
\]

Collecting all routes, the environment works with the physical reorder vector
\[
\mathbf{Q}_{t}
  \;=\;
  \bigl(Q_{\ell,t}\bigr)_{\ell\in\mathcal{L}}
  \;\in\;
  \bigl[0,\>\mathrm{Cap}^{\mathrm{route}}_{\ell}\bigr]^{|\mathcal{L}|}.
\]

\paragraph{Effective Action Space.}
After scaling and cutoff, the admissible actions lie in the $|\mathcal{L}|$‑dimensional box
\[
\mathcal{A}
  \;=\;
  \Bigl\{
    \mathbf{Q}_{t}
    \;\bigm|\;
    Q_{\ell,t}\in\bigl[0,\mathrm{Cap}^{\mathrm{route}}_{\ell}\bigr],
    \; \ell\in\mathcal{L}
  \Bigr\}.
\]

\subsubsection{Transition Dynamics}

At each time step $t\in\mathcal{T}$, after observing $s_{t}$ and selecting normalized actions $\mathbf{a}_{\mathrm{norm}}\in[-1,1]^{|\mathcal{L}|}$, the environment updates to $s_{t+1}$ as follows:

\paragraph{Action Processing, Clipping, and Penalty Recording.}  
Each component \(a_{\mathrm{norm},\ell}\) is first mapped to a preliminary reorder quantity
\[
Q^{\mathrm{pre}}_{\ell,t}
=\frac{a_{\mathrm{norm},\ell}+1}{2}\,\mathrm{Cap}^{\mathrm{route}}_{\ell}.
\]
A small‐order cutoff \(\varepsilon\) is applied:
\[
Q^{\mathrm{cut}}_{\ell,t}
=\begin{cases}
0, & Q^{\mathrm{pre}}_{\ell,t}\le\varepsilon,\\
Q^{\mathrm{pre}}_{\ell,t}, & \text{otherwise}.
\end{cases}
\]
Any remaining violation of the action bounds is then clipped and recorded:
\[
Q_{\ell,t}
=\begin{cases}
0, & Q^{\mathrm{cut}}_{\ell,t}<0,\\[3pt]
Q^{\mathrm{cut}}_{\ell,t}, & 0\le Q^{\mathrm{cut}}_{\ell,t}\le \mathrm{Cap}^{\mathrm{route}}_{\ell},\\[3pt]
\mathrm{Cap}^{\mathrm{route}}_{\ell}, & Q^{\mathrm{cut}}_{\ell,t}>\mathrm{Cap}^{\mathrm{route}}_{\ell},
\end{cases}
\]
and the action‐bound violation penalty is computed as
\[
C^{\mathrm{bound,action}}_{\ell,t}
=
\begin{cases}
\phi_{\mathrm{action}}\;\bigl|Q^{\mathrm{cut}}_{\ell,t}\bigr|, 
  & Q^{\mathrm{cut}}_{\ell,t}<0,\\[6pt]
\phi_{\mathrm{action}}\;\bigl|Q^{\mathrm{cut}}_{\ell,t}-\mathrm{Cap}^{\mathrm{route}}_{\ell}\bigr|, 
  & Q^{\mathrm{cut}}_{\ell,t}>\mathrm{Cap}^{\mathrm{route}}_{\ell},\\[6pt]
0, & \text{otherwise},
\end{cases}
\]
with \(\phi_{\mathrm{action}}\) the penalty factor for action violations.  Finally, increment the period: \(t\leftarrow t+1\).  

\paragraph{Pipeline Update.}  
After incrementing the period $t \leftarrow t + 1$ at the end of previous step, the in‐transit slots are shifted and the newest order is injected:
\[
P_{\ell,\tau,t}
= 
\begin{cases}
P_{\ell,\tau+1,\;t-1}, & \tau = 1,2,\dots,LT_{\ell}-1,\\[6pt]
Q_{\ell,\;t-1},       & \tau = LT_{\ell},\\[3pt]
0,                    & \text{otherwise.}
\end{cases}
\]
The total in‐transit inventory on route $\ell$ is
\[
P^{\mathrm{inv}}_{\ell,t}
=\sum_{\tau=1}^{LT_{\ell}}P_{\ell,\tau,t},
\]
and the arrivals at node $n$ are
\[
\mathrm{Arrivals}_{n,t}
=\sum_{\substack{\ell\in\mathcal{L}\\\mathrm{dest}(\ell)=n}}P_{\ell,1,t}.
\]

\paragraph{Arrival and Inventory Update.}  
For each node $n\in\mathcal{N}$,
\[
I_{n,t}
=I_{n,t-1}+\mathrm{Arrivals}_{n,t}.
\]

\paragraph{4. Demand Realization.}  
For each $(r,m)\in\mathcal{RM}$,
\[
D_{r,m,t}\sim\mathcal{N}(\mu_{r,m},\sigma_{r,m}).
\]

\paragraph{Sales and Backlog Update.}  
Sales are
\[
S_{r,m,t}
=\min\bigl\{D_{r,m,t}+B_{r,m,t-1},\,I_{r,t}\bigr\},
\]
then
\[
I_{r,t}\,\leftarrow\,I_{r,t}-S_{r,m,t},  
\quad
B_{r,m,t}
=D_{r,m,t}+B_{r,m,t-1}-S_{r,m,t}.
\]

\paragraph{Demand Forecast Shift.}  
\[
D_{r,m,t+1:t+k}
=\bigl(D_{r,m,t+1},\dots,D_{r,m,t+k}\bigr)
\quad\forall\,(r,m)\in\mathcal{RM}.
\]

\paragraph{Observation Clipping.}  
After constructing all next‐state components, any $x_{i,t+1}$ outside its bounds $[L_{i},U_{i}]$ is clipped to $L_{i}$ or $U_{i}$ as appropriate.  Each such clipping is recorded as an observation‐bound violation and will incur the corresponding penalty in the Cost Function.

\subsubsection{Cost Function}

The total cost at time \(t\) is the sum of the penalties incurred for any state‐observation violations:
\[
C_{t}
\;=\;
C^{\mathrm{on\_hand}}_{t}
\;+\;
C^{\mathrm{pipeline}}_{t}
\;+\;
C^{\mathrm{sales}}_{t}
\;+\;
C^{\mathrm{backlog}}_{t}
\]
Each category cost \(C^{c}_{t}\) (for \(c\in\{\mathrm{on\_hand},\mathrm{pipeline},\mathrm{sales},\mathrm{backlog}\}\)) is computed by applying the same piecewise linear penalty to each pre‐clipped component \(x_{i,t}\) in that category.  Let \(\mathcal{I}_{c}\) be the set of component indices for category \(c\), each with bounds \([L_{i},U_{i}]\) and penalty factor \(\phi_{c}\). For each category
    $c\in\{\mathrm{on\_hand},\mathrm{pipeline},\mathrm{sales},\mathrm{backlog}\}$,
    let $\phi_{c}$ denote its penalty factor (e.g.\ $\phi_{\mathrm{pipeline}}$
    for pipeline‐inventory violations).  Then
\[
C^{c}_{t}
=
\sum_{i\in\mathcal{I}_{c}}
\begin{cases}
\phi_{c}\,\bigl(L_{i}-x_{i,t}\bigr), 
  & x_{i,t}<L_{i},\\[6pt]
\phi_{c}\,\bigl(x_{i,t}-U_{i}\bigr), 
  & x_{i,t}>U_{i},\\[6pt]
0, & \text{otherwise},
\end{cases}
\]
so that any observation below its lower bound or above its upper bound contributes linearly to the total cost.  

\paragraph{Action‑bound penalty.}
If a preliminary order $Q^{\mathrm{cut}}_{\ell,t}$ violates the interval
$\bigl[0,\mathrm{Cap}^{\mathrm{route}}_{\ell}\bigr]$
(Step 1 of the transition dynamics),
it is clipped to the nearest bound and the deviation is recorded as an
\emph{action‑bound penalty} $C^{\mathrm{bound,action}}_{\ell,t}$,
weighted by the scalar factor $\phi_{\mathrm{action}}$.
The sum over all routes yields $C^{\mathrm{action}}_{t}$ is added to the total cost $C_{t}$ above.

\subsubsection{Reward Function}

We express the per‐step reward as the sum of five components, each denoted by \(\Pi\):

\[
\Pi_{t}
=
\Pi^{\mathrm{rev}}_{t}
\;-\;
\Pi^{\mathrm{proc}}_{t}
\;-\;
\Pi^{\mathrm{hold}}_{t}
\;-\;
\Pi^{\mathrm{oper}}_{t}
\;-\;
\Pi^{\mathrm{backlog}}_{t}.
\]

\begin{itemize}
  \item \(\displaystyle \Pi^{\mathrm{rev}}_{t}
    = \sum_{(r,m)\in \mathcal{RM}}
      S_{r,m,t}\;P_{r,m},
    \)
    revenue from sales.
    
  \item \(\displaystyle \Pi^{\mathrm{proc}}_{t}
    = \;\sum_{\ell\in \mathcal{L}}
      Q_{\ell,t}\;C_{\ell},
    \)
     procurement cost for orders.

  \item \(\displaystyle \Pi^{\mathrm{hold}}_{t}
    = \;\sum_{n\in \mathcal{N}}
        I_{n,t}\;C^{\mathrm{hold,inv}}_{n}
      \;+\;
      \sum_{\ell\in \mathcal{L}}
        P^{\mathrm{inv}}_{\ell,t}\;C^{\mathrm{hold,mat}}_{\ell},
    \)
     holding cost for on‐hand and pipeline inventories.

    \item \(\displaystyle 
            \Pi^{\mathrm{oper}}_{t}
            \;=\;
            \sum_{p\in\mathcal{P}}
              \frac{C^{\mathrm{oper}}_{p}}{\eta_{p}}
              \sum_{\substack{\ell\in\mathcal{L}\\\mathrm{orig}(\ell)=p}}
              Q_{\ell,t},
          \)
          operating cost at each producer \(p\), proportional to the total
          quantity \(Q_{\ell,t}\) dispatched along routes \(\ell\) that
          originate at \(p\); the factor \(1/\eta_{p}\) converts finished‑good
          output to required production input.
          
  \item \(\displaystyle \Pi^{\mathrm{backlog}}_{t}
    = \;\sum_{(r,m)\in \mathcal{RM}}
      B_{r,m,t}\;C^{\mathrm{penalty}}_{r,m},
    \)
     penalty for backlog.
\end{itemize}

\subsubsection{Episode Termination}

An episode ends when the final time step \(t = T\) is reached. It is \emph{not truncated}, even if the system encounters infeasible states.

\subsection{Grid-Integrated Energy Storage (\texttt{GridStorageEnv})}
\subsubsection{Overview}

The grid-integrated energy-storage environment (\texttt{GridStorageEnv}) models the hourly operation of a transmission-connected battery fleet co-located with conventional generators on a network subject to time-varying loads and deterministic line de-energisation schedules \citep{piansky2024long}. At each period, the agent simultaneously chooses (i) real-power outputs for every thermal generator, (ii) battery charge rates, (iii) battery discharge rates, (iv) deliberate load shedding, and (v) bus voltage angles at all non-reference buses (Bus 1 is fixed at 0 rad). Given these angles, line flows are computed through DC power-flow equations, and flows on deterministically de-energised lines are forced to zero.

Batteries follow a “bucket” state-of-charge (SOC) dynamic that applies charging/discharging inefficiency and an inter-period carry-over factor.

The objective is to minimise total cost over a finite horizon of \(T\) hours, comprising (a) generator fuel cost modelled as a polynomial in power output and (b) linear penalties on slack generation, unserved load, bus-angle limits, line-loading ratio, SOC and demand violations, as well as a nodal power-balance penalty that discourages infeasible angle choices. Physical limits on generators, transmission lines, buses, and batteries are enforced by clipping out-of-bounds actions or state variables and charging a proportional penalty.

\begin{figure}[ht]
    \centering
    \includegraphics[width=0.85\linewidth]{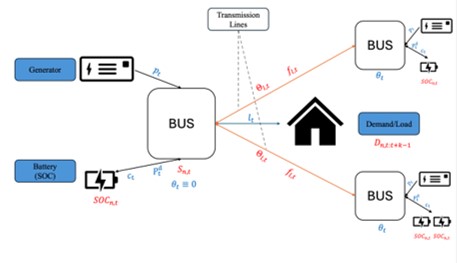}
    \caption{
        Schematic of \emph{GridStorageEnv}. A power grid with buses, generators, and battery storage, subject to time-varying demand and deterministic line de-energization. Agent actions (blue) at each time step include generator output $p_{g,t}$, battery charge $c_{n,t}$, discharge $p^d_{n,t}$, load shed $\ell_{n,t}$, and voltage angles $\theta_{n,t}$. Observed state (red) comprises battery SOC $\mathrm{SOC}_{n,t}$, line flows $f_{\ell,t}$, voltage-angle differences $\Theta_{\ell,t}$, slack $s_{n,t}$, and a $k$-period demand forecast $D_{n,t:t+k-1}$.
        \label{fig:GridStorage_Schematic}
    }
\end{figure}

We assume perfect foresight of hourly demand and line de-energisation schedules, a single central decision maker, lossless transmission on energised lines, and a fixed \(k\)-hour demand-forecast window embedded in the state. Episodes last exactly \(T\) steps and are never truncated. A schematic of the environment is shown in figure \ref{fig:GridStorage_Schematic}.

\subsubsection{Problem Setup}

The environment is defined by the following sets and parameters.

\paragraph{Sets}
\begin{itemize}
  \item $\mathcal{T}=\{1,\dots,T\}$: discrete time periods.
  \item $\mathcal{N}$: set of buses in the network, $|\mathcal{N}|=N$.
  \item $\mathcal{G}$: set of generators, $|\mathcal{G}|=G$.
  \item $\mathcal{L}\subseteq\mathcal{N}\times\mathcal{N}$: set of transmission lines.
  \item $\mathcal{D}_t\subseteq\mathcal{L}$: deterministic subset of lines de-energized at time $t$.
\end{itemize}

\paragraph{Parameters}
\begin{itemize}
  \item $\mathrm{BusGeneratorLink}$: a mapping $g\mapsto n$ for $g\in\mathcal{G}$, $n\in\mathcal{N}$, indicating which bus each generator sits on.
  \item $B_{ij}$, $(i,j)\in\mathcal{L}$: line susceptance.
  \item $\overline{f}_{\ell}$, $\ell\in\mathcal{L}$: maximum power‐flow on line $\ell$.
  \item $\underline{\theta}_{\ell},\,\overline{\theta}_{\ell}$, $\ell\in\mathcal{L}$: bounds on voltage‐angle difference.
  \item $d_{n,t}$, $(n,t)\in\mathcal{N}\times\{1,\dots,T\}$: demand at bus $n$ and time $t$.
  \item $p^{\min}_{g},\,p^{\max}_{g}$, $g\in\mathcal{G}$: generator output limits.
  \item $E^{\min}_{n},\,E^{\max}_{n}$, $n\in\mathcal{N}$: battery state‐of‐charge (SOC) bounds.
  \item $E_{n,0}$, $n\in\mathcal{N}$: initial SOC.
  \item $p^{c,\min}_{n},\,p^{c,\max}_{n}$, $n\in\mathcal{N}$: battery charge‐rate bounds.
  \item $p^{d,\min}_{n},\,p^{d,\max}_{n}$, $n\in\mathcal{N}$: battery discharge‐rate bounds.
  \item $\eta$: battery charge/discharge efficiency.
  \item $\gamma$: SOC carry‐over rate between periods.
  \item $\mathrm{PolynomialDegree}$: number of generator‑cost
        coefficients (the highest exponent is
        $\mathrm{PolynomialDegree}-1$).
  \item $C_{g,j}$, $g\in\mathcal{G}$, $j = 0,\dots,
        \mathrm{PolynomialDegree}-1$:   % <-- changed upper limit
        generator cost coefficients.
    \item $K_{\mathrm{slack}}$ \textnormal{(\texttt{Kslack})}:  
            per‑unit penalty for slack generation.
    
      \item $K_{\mathrm{ls}}$ \textnormal{(\texttt{Kls})}:  
            per‑unit penalty for load‑shedding.
    \item $\Theta^{\max}$: absolute bound on controllable bus voltage angle (radians).
   \item $\phi_{\mathrm{bal}}$: penalty factor for nodal power‑balance violations arising when injections do not match DC power flow.
   \item $\phi_{\theta,\text{act}}$: penalty factor for node‑angle
        \emph{action} clipping.

  \item $\phi_{\mathrm{power}}$: penalty factor for generator output bounds violations.
  \item $\phi_{\mathrm{charge}}$: penalty factor for battery charge‐rate bounds violations.
  \item $\phi_{\mathrm{discharge}}$: penalty factor for battery discharge‐rate bounds violations.
  \item $\phi_{\mathrm{slack}}$: penalty factor for slack generation bounds violations.
  \item $\phi_{\mathrm{shed}}$: penalty factor for load‐shedding bounds violations.
  \item $\phi_{\mathrm{soc}}$: penalty factor for state‐of‐charge observation bounds violations.
  \item $\phi_{\theta}$: penalty factor for voltage‑angle observation‑bound
      violations.
  \item $s^{\max}$: maximum slack‐generation allowed at each bus.
  \item $d^{\max}_{\text{global}}
          = \displaystyle\max_{n\in\mathcal{N},\,t\in\mathcal{T}} d_{n,t}$:
          system‑wide peak demand, used as the upper bound for all
          load‑shedding actions.
\end{itemize}

\subsubsection{State Space}

The state $s_{t}$ at time $t\in\mathcal{T}$ is the tuple
\[
  s_{t}
  =
  \bigl(
     \mathrm{SOC}_{n,t},\;
     \Theta_{\ell,t},\;
     f_{\ell,t},\;
     s_{n,t},\;
     D_{n,t:t+k-1},\;
     \tau_{t}
  \bigr),
\]

with components defined and sized as follows:

\begin{itemize}
  \item $\mathrm{SOC}_{n,t}=E_{n,t}/E^{\max}_{n}\in[0,1]$
        (normalized battery state of charge at bus $n$;
        dimension $N$).

  \item $\Theta_{\ell,t}=\frac{2(\theta_{i,t}-\theta_{j,t})-(\underline{\theta}_{\ell}+\overline{\theta}_{\ell})}{\overline{\theta}_{\ell}-\underline{\theta}_{\ell}}$
        (normalized voltage-angle difference for
        $\ell=(i,j)$; dimension $L$).

  \item $f_{\ell,t}\in[-\overline f_{\ell},\overline f_{\ell}]$
        (actual power flow on line $\ell$ in MW; dimension $L$).

  \item $s_{n,t}\in[0,s^{\max}]$
        (slack generation at bus $n$; dimension $N$).

  \item $D_{n,t:t+k-1}
        =(d_{n,t},\,d_{n,t+1},\dots,d_{n,t+k-1})$
        (length-$k$ demand-forecast window; dimension $N\times k$).

  \item $\tau_{t}=(t-1)/(T-1)\in[0,1]$
        (normalized time index; dimension $1$).
\end{itemize}

All components combine into a flattened observation vector of length
\[
  2N\;+\;2L\;+\;N\,k\;+\;1.
\]

\subsubsection{Action Space}

At the start of each period $t$ the agent chooses the vector
\[
  a_t
  =
  \bigl(
      p_t,\;
      c_t,\;
      p^{d}_t,\;
      \ell_t,\;
      \theta_t
  \bigr)^{\!\top},
  \qquad
  \theta_{1,t}\equiv 0.
\]

\vspace{3pt}
\noindent Its components obey the compact interval constraints
\[
\renewcommand{\arraystretch}{1.1}
\begin{array}{lll}
p_t \in\bigl[p^{\min},p^{\max}\bigr], &
  &(\text{generator outputs, dim. }G),\\[4pt]
c_t \in\bigl[p^{c,\min},p^{c,\max}\bigr], &
  &(\text{battery charge}, N),\\[4pt]
p^{d}_t \in\bigl[p^{d,\min},p^{d,\max}\bigr], &
  &(\text{battery discharge}, N),\\[4pt]
\ell_t \in[0,d^{\max}_{\text{global}}]^{N}, &
  &(\text{load shedding}, N),\\[4pt]
\theta_t \in[-\Theta^{\max},\Theta^{\max}]^{N-1}, &
  &(\text{bus angles}, N{-}1).
\end{array}
\]

\paragraph{Normalized interface.}
The policy operates in the cube \([-1,1]^{\,G+3N+(N-1)}\) and outputs
\(a_{\mathrm{norm},t}\).
An affine map rescales it to a preliminary action
\[
  a_{\mathrm{pre},t}
  =\tfrac12\bigl(a_{\mathrm{norm},t}+1\bigr)
   \odot(a_{\max}-a_{\min})
   +a_{\min},
\]
with block‑wise bounds
\[
  a_{\min}
  =\bigl(
        p^{\min},
        p^{c,\min},
        p^{d,\min},
        0_{N},
        -\Theta^{\max}1_{N-1}
    \bigr)^{\!\top}
\]
\[
  a_{\max}
  =\bigl(
        p^{\max},
        p^{c,\max},
        p^{d,\max},
        d^{\max}_{\text{global}}1_{N},
        \Theta^{\max}1_{N-1}
    \bigr)^{\!\top}
\]
Any element that exceeds its limits after mapping is clipped to the
nearest bound; the clipping distance is multiplied by the corresponding
penalty factor \(\phi_{\!\cdot}\) and accumulated into the action‑bound
penalty \(C^{\mathrm{action}}_{t}\).

\subsubsection{Transition Dynamics}

At each time step \(t\in\mathcal{T}\), the environment moves from state \(s_{t}\) to \(s_{t+1}\) after receiving a normalized action \(\mathbf{a}_{\mathrm{norm},t}\in[-1,1]^{G+3N+(N-1)}\). The update proceeds through seven ordered steps:

\paragraph{1. Action decoding, clipping, and penalty logging.}
Each normalized component is mapped back into its physical range:
\[
  a^{\mathrm{pre}}_{i,t}
  =\frac{a_{\mathrm{norm},i,t}+1}{2}\,
     (a^{\max}_{i}-a^{\min}_{i})
  +  a^{\min}_{i},
\]
and clipped to remain within bounds \([a^{\min}_{i},a^{\max}_{i}]\). Let the resulting action vector be
\[
  \mathbf{a}_{t}
  =\bigl\{p_{g,t}\bigr\}_{g\in\mathcal{G}}
   \;\Vert\;
   \bigl\{c_{n,t}\bigr\}_{n\in\mathcal{N}}
   \;\Vert\;
   \bigl\{p^{d}_{n,t}\bigr\}_{n\in\mathcal{N}}
   \;\Vert\;
   \bigl\{\ell_{n,t}\bigr\}_{n\in\mathcal{N}}
   \;\Vert\;
   \bigl\{\theta_{n,t}\bigr\}_{n\in\mathcal{N}\setminus\{1\}},
\]
with \(\theta_{1,t}\equiv 0\). Each clipping violation incurs a penalty multiplied by the corresponding \(\phi_{\cdot}\).

\paragraph{2. Battery state-of-charge update.}
The battery SOC at each bus \(n\) evolves as:
\[
  E_{n,t+1}
  = \gamma E_{n,t}
  + \eta c_{n,t}
  - \frac{1}{\eta} p^{d}_{n,t},
  \quad n\in\mathcal{N}.
\]

\paragraph{3. Load-shedding enforcement.}
Any load shedding exceeding the global maximum is clipped:
\[
\ell_{n,t} \leftarrow \min(\ell_{n,t}, d^{\max}_{\text{global}}),
\]
penalising excess with factor \(\phi_{\mathrm{shed}}\).

\paragraph{4. Power-flow calculation.}
Compute power flows from voltage angles, enforcing zero flow on de-energised lines:
\[
  f_{\ell,t}=\begin{cases}
  B_{ij}(\theta_{i,t}-\theta_{j,t}), & \ell\notin\mathcal{D}_t\\[6pt]
  0, & \ell\in\mathcal{D}_t
  \end{cases}, \quad \ell=(i,j).
\]

\paragraph{5. Slack generation calculation.}
Slack generation \(s_{n,t}\) is computed to enforce exact network balance:
\[
\begin{aligned}
  s_{n,t} &= \max\Bigl\{0,\;d_{n,t}-\ell_{n,t}
    -\sum_{g:\mathrm{BusGeneratorLink}[g]=n}p_{g,t}
    +c_{n,t}-p^{d}_{n,t}\\
    &\quad+\sum_{(i,n)\in\mathcal{L}}f_{(i,n),t}
    -\sum_{(n,j)\in\mathcal{L}}f_{(n,j),t}\Bigr\}\,.
\end{aligned}
\]

\paragraph{6. Net nodal-injection and power-balance penalty.}
Compute net nodal injection at each bus \(n\):
\[
  P_{n,t}
  = \sum_{g:\mathrm{BusGeneratorLink}[g]=n} p_{g,t}
    + s_{n,t}
    - d_{n,t}
    + \ell_{n,t}
    - c_{n,t}
    + p^{d}_{n,t}.
\]
The nodal power-balance residual is:
\[
  \Delta_{n,t}
  = P_{n,t} - \sum_{j\in\mathcal{N}} B_{nj}(\theta_{n,t}-\theta_{j,t}),
\]
and the network-balance penalty is:
\[
  C^{\mathrm{bal}}_{t}
  = \phi_{\mathrm{bal}} \sum_{n\in\mathcal{N}}\lvert\Delta_{n,t}\rvert.
\]

\paragraph{7. Demand-forecast and observation reconstruction.}
Update the forecast window at each bus \(n\):
\[
  D_{n,t:t+k-1}
  =(d_{n,t},d_{n,t+1},\dots,d_{n,\min\{t+k-1,T\}}),
\]
padded with zeros beyond horizon \(T\). Form the next state \(s_{t+1}\) from normalized SOC, voltage-angle differences, loading ratios, flows, slack generation \(s_{n,t}\), demand window, and normalized time \(\tau_{t+1}=t/(T-1)\). Components outside valid bounds are clipped, incurring penalties accordingly.

\subsubsection{Cost Function}

The total penalty cost at time \(t\) is
\[
  C_{t}
  \;=\;
   C^{\mathrm{soc}}_{t}
  +C^{\theta}_{t}
  +C^{\mathrm{flow\_ratio}}_{t}
  +C^{\mathrm{slack}}_{t}
  +C^{\mathrm{bal}}_{t}
  +C^{\mathrm{action}}_{t}
\]

\paragraph{Observation-bound terms.}
For each category \(c\in\{\text{soc},\theta,\text{flow\_ratio},\text{slack}\}\), let \(\mathcal I_{c}\) be the indices of the corresponding observation sub-vector and \([L_i,U_i]\) its valid range. Then
\[
  C^{c}_{t}
  \;=\;
  \sum_{i\in\mathcal I_{c}}
  \begin{cases}
     \phi_{c}\,(L_i-x_{i,t}), & x_{i,t}<L_i,\\[4pt]
     \phi_{c}\,(x_{i,t}-U_i), & x_{i,t}>U_i,\\[4pt]
     0, & \text{otherwise}.
  \end{cases}
\]

\paragraph{Network-balance term.}
\(C^{\mathrm{bal}}_{t}\) is defined in Transition-Step 5 as
\(
  \phi_{\mathrm{bal}}\sum_{n}|\Delta_{n,t}|
\),
penalising any mismatch between injections and DC power flow.

\paragraph{Action-bound term.}
\(C^{\mathrm{action}}_{t}\) aggregates the clipping penalties accrued in Transition-Step 1 for generator power, charge/discharge rates, load shedding, and bus angles (the latter weighted by \(\phi_{\theta,\text{act}}\)). It ensures every action component exceeding its hard limit is penalised proportionally to its violation.

\subsubsection{Reward Function}

We decompose the per‐step reward \(\Pi_{t}\) into three components, each denoted by \(\Pi\):

\[
\Pi_{t}
=
\Pi^{\mathrm{gen}}_{t}
\;+\;
\Pi^{\mathrm{slack}}_{t}
\;+\;
\Pi^{\mathrm{shed}}_{t}.
\]

\noindent where

\[
\Pi^{\mathrm{gen}}_{t}
= -\sum_{g\in\mathcal{G}}
  \sum_{j=0}^{\mathrm{PolynomialDegree}-1}
    C_{g,j}\,p_{g,t}^{\,j},
\]
negative generator fuel cost (polynomial in output \(p_{g,t}\)),

\[
\Pi^{\mathrm{slack}}_{t}
= -K_{\mathrm{slack}}
  \sum_{n\in\mathcal{N}}s_{n,t},
\]
negative cost of slack generation at each bus,

\[
\Pi^{\mathrm{shed}}_{t}
= -K_{\mathrm{ls}}
  \sum_{n\in\mathcal{N}} \ell_{n,t}.
\]

negative cost of load‐shedding.  

\subsubsection{Episode Termination}
An episode ends when the final time step $t = T$ is reached. It is \emph{not truncated}, even if the system encounters infeasible states.

\subsection{Integrated scheduling and maintenance (\texttt{SchedMaintEnv})}

\subsubsection{Overview}

Energy-intensive chemical processes leverage Demand Response (DR) to adjust electricity usage in response to price fluctuations, typically optimizing production on a rolling basis using forecasted demand and electricity prices. However, optimizing production scheduling alone can be detrimental, as it neglects the operational condition of essential equipment. Recent studies have addressed this by integrating condition-based maintenance into production optimization, notably for Air Separation Units (ASUs) \citep{Xenos2016Operational}, and natural gas plants \citep{Huang2020Multistage}.

\begin{figure}[H]
    \centering
    \includegraphics[width=0.85\linewidth]{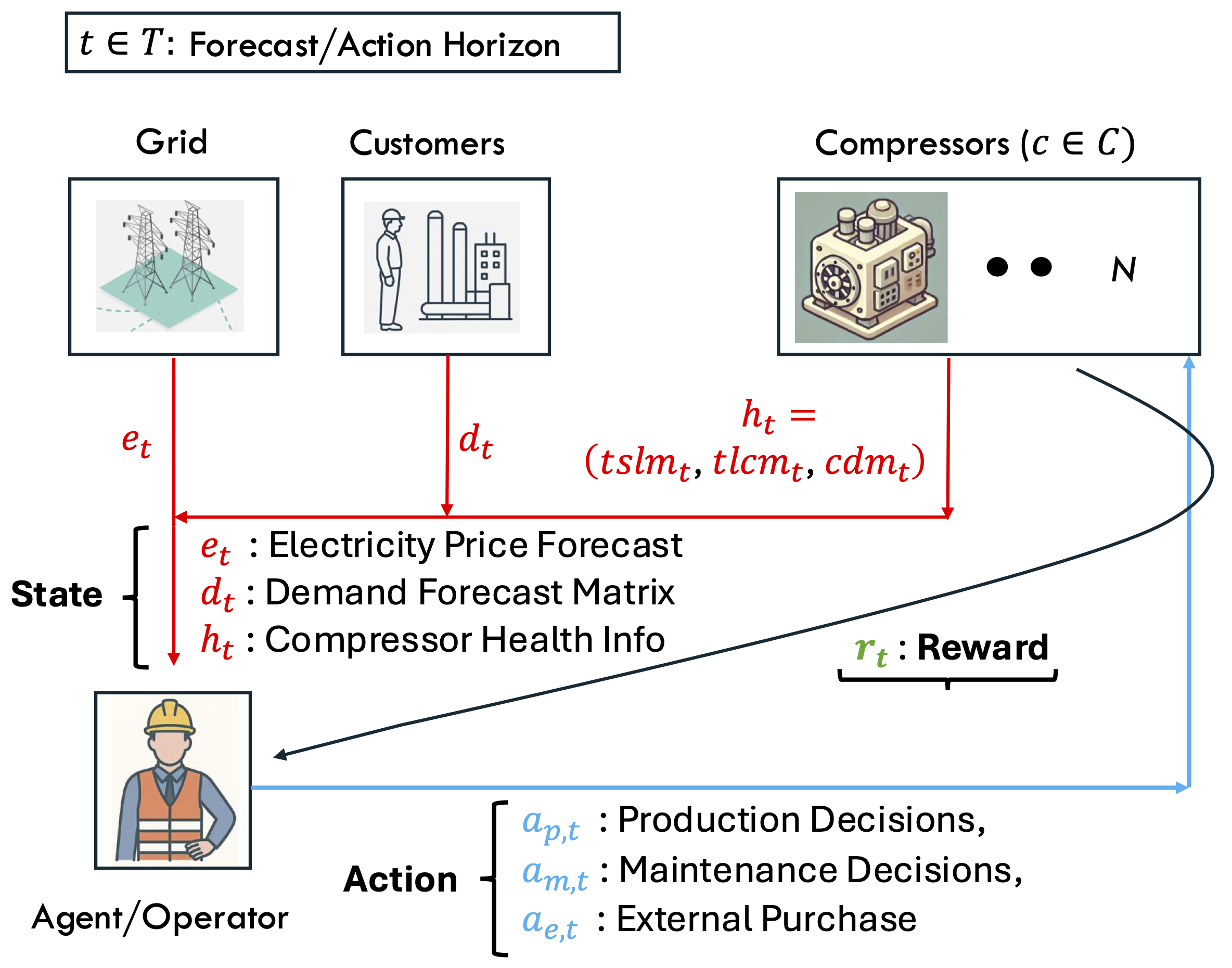}
        \caption{Schematic of \texttt{SchedMaintEnv}. At each time step \(t\), the agent observes the day-ahead electricity price forecast \(e_t\), the demand forecast \(d_t\), and the compressor health state \(h_t=(\mathrm{tslm}_t,\;\mathrm{tlcm}_t,\;\mathrm{cdm}_t)\). It then selects compressor production rates \(a_{p,t}\), maintenance scheduling flags \(a_{m,t}\), and external purchase fraction \(a_{e,t}\) to meet demand while managing cost and maintenance constraints, receiving reward \(r_t\) from the compressors.}
    \label{fig:ISM_schematic}
\end{figure}

In this study, we model an Air Separation Unit (ASU) comprising three compressors tasked with meeting aggregated gaseous nitrogen (GAN) and oxygen (GOX) demand over a 31‑day episode, leveraging a deterministic 30‑day rolling forecast of electricity prices and demands. Each day, the agent decides for each compressor whether to operate—producing at a chosen output level—or to undergo maintenance, incurring downtime and resetting its operational state. If total production falls short of demand, the deficit is met through external purchases at a fixed (though inflated) price. The objective is to minimize total operating expense—comprising electricity costs, downtime losses, and external purchase costs—by optimally trading off short‑term production gains against long‑term equipment health. The base environment, which is completely deterministic, is termed \texttt{SchedMaintEnv-v0}. To further resemble real‐world operations, we introduce uncertainty in compressor failure times. We refer to our stochastic variant of the base environment as \texttt{SchedMaintEnv-v1}. A schematic of the environment is shown in Figure \ref{fig:ISM_schematic}.

\subsubsection{Problem Setup}

The sets and known parameters used to describe the environment are
shown below:

\paragraph{Sets}
\begin{itemize}
    \item \( \mathcal{T} = \{0, \ldots, T\} \): Set of discrete time periods.
    \item \( \mathcal{C} \): Set of compressors.
\end{itemize}

\paragraph{Parameters}

\begin{itemize}
    \item \( n\): Number of compressors.
    \item \( S\): Forecast horizon.
    \item \( Cap_c\): Maximum daily production capacity of the compressor \( c\).
    \item \( \mathrm{SPEN}_c \): Specific energy of compressor \( c\) in KWh/t.
    \item \(\mathrm{MTTF}_c\): Mean time to failure represents the maximum number of consecutive operating days before maintenance is required for compressor \( c\).
    \item \(\mathrm{MTTR}_c\):  Mean time to repair represents fixed duration (in days) of any maintenance outage compressor \( c\).
    \item \(\mathrm{MNRD}_c\): Minimum no‐repair duration, i.e., the time that must elapse after maintenance before the next service can begin for compressor \( c\).
  \item \(tlcm_{c,0}\): Initial time left to complete maintenance for compressor \(c\) at start of the episode.
  \item \(tslm_{c,0}\): Initial time since last maintenance for compressor \(c\) at start of the episode.
  \item \(cdm_{c,0}\): Initial indicator of whether compressor \(c\) is eligible for maintenance at the start of the episode.
    \item \(\alpha_{ext}\): External purchase price of per unit of product.
    \item \(Q_{ext}\): Maximum possible purchase quantity for any given day.
    \item \(\texttt{D}\): Array of daily forecasted demand over the simulation horizon \(T + S \) in ton.
    \item \(\texttt{E}\): Array of daily forecasted electricity prices over the simulation horizon \(T + S \) in \$/KWh.
    \item \( \rho_{\text{MD}}, \rho_{\text{MF}}, \rho_{\text{EM}}, \rho_{\text{RP}}, \rho_{\text{D}}\): Various penalty parameters related to constraint violation.
\end{itemize}

\subsubsection{State Space}

The observation state at time \( t \in  \mathcal{T} \) is represented as a vector, 
\[
s(t) = \left( {d}_t, {e}_t, {tslm}_t, {tlcm}_t, {cdm}_t \right), \quad t \in [0,  T]
\]

The state vector $s(t)$ captures the essential operational and maintenance-related information for the Air Separation Unit (ASU) on day $t$. It includes forecasts of production demands and electricity prices over a fixed horizon (\textit{S} days), along with detailed maintenance indicators for each compressor. These components are defined as follows:

\begin{itemize}
  \item Demand Forecast ($d_t \in \mathbb{R}_+^{S \times 1})$: a vector of predicted demands from day $(t+1)$ to $(t+S)$, expressed in tons.
  
  \item Electricity Price Forecast ($e_t \in \mathbb{R}_+^{S \times 1}$): a vector of corresponding day-ahead electricity prices from day $t$ to $t+S-1$, measured in \(\$/kWh\)
  
  \item Time Since Last Maintenance ($\text{tslm}_t \in \mathbb{Z}_+^{n \times 1}$): the number of days since each compressor $c$ last underwent maintenance
  
  \item Time Left to Complete Maintenance ($\text{tlcm}_t \in \mathbb{Z}_+^{n \times 1}$): the remaining time (in days) required to complete maintenance for each compressor $c$; it is strictly positive only when maintenance is in progress.
  
  \item Can Do Maintenance Indicator ($\text{cdm}_t \in \{0, 1\}^n$): a binary vector where $\text{cdm}_{ct} = 1$ indicates that compressor $c$ is eligible for maintenance on day $t$
\end{itemize}

In the base \texttt{SchedMaintEnv}, the agent strives to learn the fixed failure time \(\mathrm{MTTF}_c\) of each compressor \(c\). In the stochastic variant, we assume each compressor can fail at \(\mathrm{MTTF}_c\), \(\mathrm{MTTF}_c - 1\), or \(\mathrm{MTTF}_c - 2\) with equal probability; hence, the agent should ideally learn a robust preventive maintenance policy. We implement the uncertainty by introducing the aforementioned stochasticity in failure times at the beginning of each episode.

\subsubsection{Action Space}

Given the received observation at the start of each day \(t \in \mathcal{T}\), the action space at time \(t\) consists of operational decisions related to maintenance scheduling, compressor utilization, and external product procurement. The physical description is as follows:

\begin{itemize}
  \item Compressor maintenance ($a_{\text{maintenance}}(t) \in \{0, 1\}^n$): a binary vector indicating whether each compressor is scheduled for maintenance at time $t$, with $a_{\text{maintenance},c}(t) = 1$ if compressor $c$ is under maintenance. We occasionally abbreviate this action as \( a_{\text{maint.}, c}(t) \).
  
  \item Compressor production rate ($a_{\text{production}}(t) \in [0, 1]^n$): a continuous vector representing the fraction of the maximum capacity $Cap_c$ utilized by each compressor $c$. We occasionally abbreviate this action as \( a_{\text{prod.}, c}(t) \).
  
    \item External purchase \(a_{\text{purchase}}(t)\in[0,1]\): the fraction of the maximum external product \(Q_{\mathrm{ext}}\) purchased to meet demand when internal production is insufficient.
\end{itemize}

The agent’s raw actions are clipped to remain within their specified bounds. In particular, each component of \(a_{\text{maintenance}}(t)\) is first generated as a scalar in \([0,1]\) and then rounded to \(\{0,1\}\), while all other actions are clipped directly to their respective intervals.

\subsubsection{Transition Dynamics}

Here we define how the environment state evolves in response to the agent’s actions at each discrete time step \(t \in \mathcal{T}\). Let \(s(t) = \bigl(d_t,\,e_t,\,\mathrm{tslm}_t,\,\mathrm{tlcm}_t,\,\mathrm{cdm}_t\bigr)\) be the observation vector at time \(t\), with \(s(0)\) denoting the initial observation. The transition to \(s(t+1)\) is governed by the following procedures:

\paragraph{Information State Update:}

This update incorporates changes in demand and electricity price signals. The updated states are retrieved from the simulated perfect-forecast arrays of demand (\texttt{D}) and electricity prices (\texttt{E}) for the next \(S\) days as follows:
\[
\textbf{d}_{t+1} \leftarrow \texttt{D}[t+1, t+S],
\quad e_{t+1} \leftarrow \texttt{E}[t+1, t+S] \quad \quad \forall t \in T
\]
It is worth noting that the simulated data is appropriately longer than the episode length to account for the state horizon; therefore, no padding is used at any point.

\paragraph{Compressor Physical Condition Transition:}

The following updates track the evolution of maintenance status and compressor readiness for each compressor \( c \in C \), based on operational decisions. The initial physical state at the start of the simulation is given by \( (tlcm_{c,0}, tslm_{c,0}, cdm_{c,0}) \), and future states are derived accordingly.

\[
\text{tslm}_{c,t+1} = 
\begin{cases}
0, & \text{if } a_{\text{maintenance},c}(t) = 1 \\
\text{tslm}_{ct} + 1, & \text{otherwise}
\end{cases}
\]

\[
\text{tlcm}_{c,t+1} = 
\begin{cases}
\text{MTTR}_c - 1, & \text{if } a_{\text{maintenance},c}(t) = 1 \land \text{cdm}_{ct} = 1 \\
\text{tlcm}_{ct} - 1, & \text{if } a_{\text{maintenance},c}(t) = 1 \land \text{cdm}_{ct} = 0 \\
\text{tlcm}_{ct}, & \text{otherwise}
\end{cases}
\]

\[
\text{cdm}_{c,t+1} = 
\begin{cases}
1, & \text{if } \text{tslm}_{c,t+1} \geq \text{MNRD}_c \\
0, & \text{otherwise}
\end{cases}
\]

To ensure feasibility and consistency with compressor state constraints, a sanitization step is applied to the raw agent actions before the state update, but after the associated violation costs are realized. For each compressor $c \in C$, the action is adjusted as follows:

\begin{itemize}
  \item If $\text{tslm}_{ct} \geq \text{MTTF}_c$ and $a_{\text{maintenance},c}(t) \neq 1$, then:
  \begin{align*}
  \Rightarrow \quad a_{\text{maintenance},c}(t) &\leftarrow 1 \quad 
            \text{and} \quad   a_{\text{production},c}(t) &\leftarrow 0
  \end{align*}
  
This rule enforces maintenance when it is overdue (i.e., \( \text{tslm}_{ct} \geq \text{MTTF}_c \)) but the agent has not scheduled it. Maintenance is forced, and production is halted to ensure feasibility and update the environment state accordingly.

  \item If $a_{\text{maintenance},c}(t) = 1$ and $a_{\text{production},c}(t) > 0$, then:
  \begin{align*}
  \Rightarrow \quad a_{\text{production},c}(t) &\leftarrow 0
  \end{align*}

This rule ensures that production is not allowed during maintenance.
  
  \item If $\text{cdm}_{ct} = 0$ and $\text{tlcm}_{ct} = 0$ and $a_{\text{maintenance},c}(t) = 1$, then:
  \begin{align*}
  \Rightarrow \quad a_{\text{maintenance},c}(t) &\leftarrow 0
  \end{align*}

To ensure maintenance is not performed after the required duration or prematurely before it is permitted.
  
  \item If $\text{tlcm}_{ct} > 0$ and $a_{\text{maintenance},c}(t) \neq 1$, then:
  \begin{align*}
  \Rightarrow \quad a_{\text{maintenance},c}(t) &\leftarrow 1\quad 
            \text{and} \quad  
               a_{\text{production},c}(t) &\leftarrow 0
  \end{align*}

To make sure maintenance remains active for the required maintenance duration.
  
\end{itemize}

\subsubsection{Cost Function}

The agent may incur various costs at each time step if system constraints are violated, encouraging it to learn an optimal policy. We reiterate that these costs are realized before the action is sanitized. The potential costs are as follows:\\

\textbf{Maintenance Duration Cost:} 

This cost is incurred if maintenance is interrupted before it is completed or prolonged, in which case \( \text{tlcm}_{ct} \) becomes negative. The cost is denoted by \( \text{C}^{\text{MI}}_{ct} \), where MI stands for maintenance interruption, defined as:

\[
\text{C}^{\text{MI}}_{ct} = 
\begin{cases} 
 & \text{if } \left( a_{\text{maint.}, c}(t) \neq 1 \text{ and } \text{tlcm}_{c t} > 0 \right) \\
\rho_{\text{MD}} \cdot \exp \left( | \text{tlcm}_{c t}| \right) & or \left( a_{\text{maint.}, c}(t) = 1 \text{ and } \text{tlcm}_{c t} = 0 \text{ and } \text{tslm}_{c t} = 0 \right) \\
 & or \left( a_{\text{maint.}, c}(t) = 1 \text{ and } \text{tlcm}_{c t} < 0 \right), \\
0 & \text{otherwise}.
\end{cases}
\]

\textbf{Maintenance Failure Cost:} 

This cost occurs if the \textit{Time Since Last Maintenance} (\( \text{tslm}_{ct} \)) exceeds the \textit{Mean Time to Failure} (\( \text{MTTF}_c \)) of the compressor. The cost is denoted by \( \text{C}^{\text{MF}}_{ct} \), where MF stands for maintenance failure, defined as:

\[
\text{C}^{\text{MF}}_{ct} = 
\begin{cases} 
\rho_{\text{MF}} \cdot (\text{tslm}_{c t} - \text{MTTF}_c) & \text{if } \left( a_{\text{maint.}, c}(t) = 0 \text{ and } \text{tslm}_{c t} > \text{MTTF}_c  \right), \\
\rho_{\text{MF}} & \text{if } \left( a_{\text{maint.}, c}(t) = 0 \text{ and } \text{tslm}_{c t} = \text{MTTF}_c  \right), \\
0 & \text{otherwise}.
\end{cases}
\]

\textbf{Early Maintenance Cost:} 

This cost is incurred if maintenance is performed on a compressor when it is not yet eligible for maintenance (i.e., when \( \text{cdm}_{c t} = 0 \), indicating that the compressor has recently undergone maintenance, and \( \text{tlcm}_{c t} = 0 \)). The cost is proportional to the \textit{Time Since Last Maintenance} (\( \text{tslm}_{c t} \)) of the compressor. The cost is denoted by \( \text{C}^\text{EM}_{ct}\), where EM stands for early maintenance, defined as:

\[
\text{C}^\text{EM}_{ct} = 
\begin{cases} 
-\rho_{\text{EM}} \cdot \text{tslm}_{c t} & \text{if } \left( a_{\text{maint.}, c}(t) = 1 \text{ and } \text{cdm}_{c t} = 0 \text{ and } \text{tlcm}_{c t} = 0 \right), \\
0 & \text{otherwise}.
\end{cases}
\]

\textbf{Ramp Cost:} 

This cost is incurred when a compressor is ramped up while under maintenance. 

\[
\text{C}^{ \text{Ramp}}_{ct} = 
\begin{cases} 
\rho_{\text{RP}} \cdot a_{\text{prod.}, c}(t) \cdot \text{Cap}_c & \text{if } \left( a_{\text{maint.}, c}(t) = 1 \text{ and } a_{\text{prod.}, c}(t) \neq 0  \right), \\
0 & \text{otherwise}.
\end{cases}
\]

\textbf{Demand Cost:}

This penalty is incurred if the total supply from production and external purchases does not meet the demand on the current day. The total supply is the sum of the production and external purchase, and if this is less than or greater than the demand, a penalty is imposed proportional to the absolute difference between demand and supply.

\[
\text{C}^{\text{Demand}}_{t} =   \rho_{\text{D}} \cdot | d_t - \sum_{c} (a_{\text{production}, c} \cdot Cap_{c}) |
\]

\textbf{Total Cost:}

\[
C^{\text{total}}_t = \sum_{c\in \mathcal{C}} (C^{\text{MI}}_{ct}+ C^{\text{MF}}_{ct}+ C^{\text{EM}}_{ct} + C^{\text{Ramp}}_{ct}) + C^{\text{Demand}}_{t}
\]

\subsubsection{Reward Function}

The reward function represents the cost incurred by the agent for making decisions related to production and external purchases. At each time step \( t \), the reward is defined as:
\begin{align*}
\Pi^{action}_{t}
&= -\bigl(\text{production cost}_t + \text{external purchase cost}_t\bigr) 
\end{align*}
where,

\begin{alignat*}{2}
& \text{production cost}_t
 &&= \sum_{c \in \mathcal{C}}
    \big(\mathrm{SPEN}_c\,a_{\text{production},c}(t)\,\mathrm{Cap}_c\,\mathbb{E}[t]\big)\\
\qquad
& \text{external purchase cost}_t
 &&= a_{\text{purchase}}(t)\,Q_{\mathrm{ext}}\,\alpha_{\mathrm{ext}}
\end{alignat*}

The production cost at time \( t \) is calculated using the production rate, compressor capacity, specific energy consumption, and electricity price. The external purchase cost is incurred when demand exceeds production capacity, calculated by multiplying the purchase amount by the external price. 

\subsubsection{Episode Termination}
The episode terminates when \(t+1 = T\), indicating the start of the day immediately after the final one.
\subsection{Production Scheduling in Air Separation Unit (\texttt{ASUEnv})}

\subsubsection{Overview}

The ASUEnv simulates a liquid air separation unit (ASU) producing liquid nitrogen, oxygen, and argon in a Gym-compatible reinforcement learning framework. We consider the ASU to have hourly production capacity while demand occurs only at 24-hour intervals. To that end, at each hour the agent selects production rates based on inventory levels to minimize electricity and storage costs while ensuring daily demand satisfaction. Production actions are restricted to the convex hull of historically observed operating points, guaranteeing industrial feasibility. We follow the simplified dynamics of Zhang et al.\ \citep{Zhang2015Air}, which capture core ASU behavior without the complexity of detailed mode-switching or maintenance constraints.

\begin{figure}[H]
    \centering
    \includegraphics[width=0.85\linewidth]{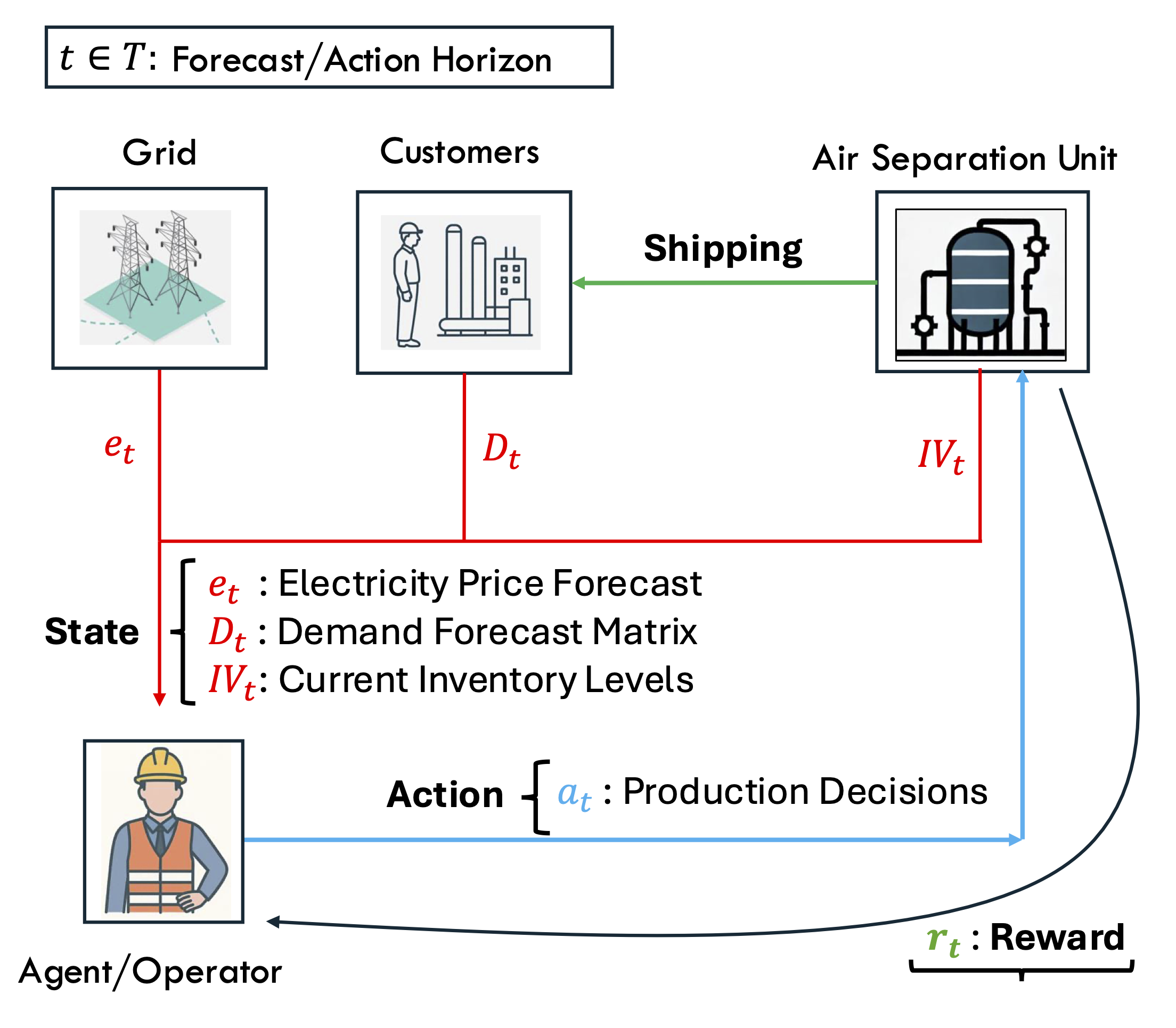}
        \caption{Schematic of \texttt{ASUEnv}. At each hour \(t\), the agent observes the electricity price forecast \(e_t\), demand forecast \(D_t\), and inventory levels \(IV_t\). It then selects production weights \(\lambda_t\) (convex‐combination coefficients of historical patterns) to meet demand and manage inventories, and receives reward \(r_t\) from the ASU.}
    \label{fig:ASU_schematic}
\end{figure}

We make several key assumptions to keep the environment tractable. Only two production modes—“Work” (active production) and “Off” (zero output)—are available each hour, and switching between them is instantaneous and cost-free. Electricity prices and product demands over a short rolling horizon are treated as perfectly known and error-free. Storage capacity for each product is finite and nonnegative, with penalties applied if inventory limits are exceeded or daily demand is unmet. Finally, we do not model any external purchasing option, assuming all demand falls within the ASU’s inherent production capability. The simulation runs for a total of seven days with a four-day lookahead, resulting in \(24 \times 7\) steps per episode. A schematic of the environment is shown in figure \ref{fig:ASU_schematic}.

\subsubsection{Problem Setup}

The sets and known parameters used to describe the environment are shown below:

\paragraph{Sets}
\begin{itemize}
    \item \(\mathcal{T} = \{0, \ldots, T\}\): Set of hours in the episode.
    \item \(\mathcal{D} = \{1, \ldots, D\}\): Set of days in the episode.
    \item \(\mathcal{J}\): Set of liquid products: liquid nitrogen (LIN), liquid oxygen (LOX) and liquid argon (LAR).
    \item \(\mathcal{X}\): Set of vertices of the convex hull derived from historical operational data.
\end{itemize}

\paragraph{Parameters}
\begin{itemize}
    \item \(T\): Episode length in hours.
    \item \(D\): Episode length in days.
    \item \(S\): Lookahead days used in the forecast.
    \item \(m\): Number of products in \(\mathcal{J}\).
    \item \([IV^{j,\mathrm{lb}},\,IV^{j,\mathrm{ub}}]\): Lower and upper bounds on the inventory level of each product \(j\).
    \item \(N\): Number of historical production data points.
    \item \(v_x\): Extreme points of the convex hull.
    \item \(\mathrm{HPQ}\): Historical hourly production quantities.
    \item \(k\): Number of extreme points of the convex hull.
    \item \(\bar{\texttt{D}}_{t}\): Matrix of hourly product demands used in the simulation, of dimension \( \mathbb{R}_{+}^{m\times 4(S+T+1) }\).
    \item \(\bar{\texttt{E}}_{t}\): Array of hourly electricity prices used in the simulation, of dimension \(\mathbb{R}_{+}^{24(S+T+1)}\).
    \item \(PQ_t\): Array of products produced based on the actions.
    \item \(DQ_d\): Array of dispatched product quantities at the end of each day in the simulation.
    \item \(\rho_{\mathrm{IV}}\): Penalty parameter for inventory overflow.
    \item \(\rho_{\mathrm{D}}\): Penalty parameter for unmet demand.
    \item \(C_{\text{fixed}}\): Fixed cost per hour to keep the plant operational.  
    \item \(C_{\text{unit}}\): Hourly unit production cost.  

\end{itemize}

\subsubsection{State Space}

At any hour \( t \), the observation state is represented by the vector

\[
s(t) = \left( e_t, D_t, IV_t \right), \quad \forall t \in \mathcal{T}
\]

The state vector \( s(t) \) captures all relevant information needed for the agent to make production planning decisions at hour \( t \) in the Air Separation Unit (ASU). It includes deterministic forecasts of electricity prices and product demands over a lookahead horizon of \( S \) future days; therefore, including the current day, the total forecasting horizon becomes \( S+1 \) days. It also includes the current inventory levels of liquid products. The components are defined as follows:

\begin{itemize}
  \item Electricity Price Forecast (\( e_t \in \mathbb{R}_+^{24(S+1) \times  1}  \)): a vector of day-ahead electricity prices for the next \( S+1 \) days, given at an hourly resolution (totaling \( 24(S+1) \) elements), measured in KWh.
  
    \item Demand Forecast (\( D_t \in \mathbb{R}_+^{ 24(S+1) \times  m} \)): a matrix of demands for each product \( j \) over the next \( S+1 \) days, where \( j \in \mathcal{J} \). Each row represents the hourly demand forecast for one product; however, the demand is non-zero only at 24-hour intervals.
    
  \item Inventory Levels (\( IV_t \in \mathbb{R}_+^{m\times  1} \)): the current inventory levels of all \( m \) liquid products at hour \( t \). These are bounded between predefined lower and upper capacity limits.
\end{itemize} 

\subsubsection{Action Space}

The action space at each time step \(t\) is defined as:
\[
a(t)=\lambda(t),
\]
where \(\lambda(t)\in[0,1]^k\) represents the weights of the extreme points in the convex hull of the possible production quantities. We assume historical hourly production quantities are given by
\[
\mathrm{HPQ}=\bigl\{Q_j^1,\,Q_j^2,\,\ldots,\,Q_j^N\mid j\in\mathcal{J}\bigr\},
\]
where \(N\) is the number of samples available. The feasible region \(\mathrm{FR}\) is then approximated as \(\texttt{convhull}(\mathrm{HPQ})\), with vertices \(v_{jx}\) for \(j\in\mathcal{J}\), \(x\in\mathcal{X}\), and \(|\mathcal{X}|=k\). The quantity of each product produced at time \(t\) is
\[
PQ_{jt}=\sum_{x\in\mathcal{X}}\lambda_x(t)\,v_{jx}.
\]
The received actions are clipped to their bounds \([0,1]\) and then normalized to enforce the convex‐sum property \(\sum_{x=1}^k\lambda_x(t)=1\) as described in the next section.

\subsubsection{Transition Dynamics}

Here we describe how the system state evolves in response to the agent's production decisions at each discrete hourly time step \( t \in T \). Let the observation vector at time \( t \) be \( s(t) = \left( e_t, D_t, IV_t \right) \), with \(s(0)\) denoting the initial observation. The transition to \( s(t+1) \) is governed by two primary updates: the shifting of forecast windows and the physical evolution of product inventories.

\paragraph{Forecast Update:}

The forecast arrays for electricity prices and product demand are deterministic and span a rolling horizon of \( S+1 \) days at hourly resolution. The environment updates the forecast component of the state differently depending on whether a new hour or a new day has begun.

\paragraph{Hourly Shifting:}
At each non-zero hour \( (t+1) \bmod 24 \neq 0 \), the environment updates the observation by shifting the forecast vectors leftward by one hour to discard outdated information. This is achieved via the \texttt{shift\_observation()} routine:
\[
{D}_{t+1} \leftarrow \texttt{ShiftLeft}({D}_{t}),
\quad 
{e}_{t+1} \leftarrow \texttt{ShiftLeft}({e}_{t})
\]
where \texttt{ShiftLeft} removes the earliest hour from the forecast vector and appends a placeholder value—such as the average of the remaining values (for electricity) or zeros (for demand)—to preserve the total horizon length \( 24(S+1) \). 

\paragraph{Daily Refresh:}  
At the start of each new day (i.e., when \( (t+1) \bmod 24 = 0 \)), the environment invokes \texttt{update\_demand\_and\_electricty\_state()} to refresh the entire forecast arrays for electricity prices and product demands. This function populates only the end-of-day demand values (i.e., the 24th hour of each day) while keeping the rest of the hourly entries zero, as demand is modeled to be daily:
 \[
D_{t+1} \;\leftarrow\; \bigl(\bar{\texttt{D}}[t+1],\,\bar{\texttt{D}}[t+2],\,\ldots,\,\bar{\texttt{D}}[\,t+24\,(S+1)\,]\bigr)
\]
\[
E_{t+1} \;\leftarrow\; \bigl(\bar{\texttt{E}}[t+1],\,\bar{\texttt{E}}[t+2],\,\ldots,\,\bar{\texttt{E}}[\,t+24\,(S+1)\,]\bigr)
\]
This rolling update mechanism enables the agent to make production decisions with awareness of upcoming price and demand trends while ensuring that outdated data does not persist in the observation state. The simulated data exceeds the episode length to cover the lookahead horizon, eliminating the need for padding during daily refresh.

\paragraph{Inventory State Transition:}  
Before updating the inventory, action sanitization rescales the raw weights \(\{\lambda_x(t)\}\) to enforce the convex‐sum constraint:
\[
\lambda_x(t)\;\leftarrow\;\frac{\lambda_x(t)}{\sum_{y=1}^k\lambda_y(t)} \quad \forall x \in \mathcal {X}, \quad  \text{s.t.} 
 \quad\sum_{x=1}^k\lambda_x(t)=1  
\]
The inventory vector is then updated based on the production action \(PQ_t\) at time \(t\), itself computed as a convex combination of feasible production profiles. If \(t\mod24=23\) (the last hour of the day), a portion of inventory is shipped to meet daily demand; otherwise, production is simply added:
\[
IV_{j, t+1} =
\begin{cases}
IV_{j,t} + PQ_{j,t} - DQ_{j,d}, & \text{if } t \bmod 24 = 23, d = \left\lfloor \frac{t+1}{24} \right\rfloor \\
IV_{j,t} + PQ_{j,t}        & \text{otherwise,}
\end{cases}
\]
where \( DQ_{j,d} = \min(IV_{j,t} + PQ_{j,t},\ D[j,23]) \) ensures that the shipment quantity does not exceed the sum of available inventory and production. 

Finally, state sanitization corrects any inventory overflow by checking whether the inventory penalty \( \text{C}^{\text{IV}}_{t} > 0 \). If so, the inventory vector is clipped element-wise to its upper bounds:
\[
IV_{t+1} \leftarrow \min\bigl(IV_{t+1},\,IV^{\text{ub}}\bigr),
\]
thereby preserving feasibility by preventing storage violations.

\subsubsection{Cost Function}

The agent may incur several types of costs at each time step \( t \) from production decisions or constraint violations. These costs guide the agent toward learning an efficient and feasible production policy. The individual cost components are as follows:\\

\textbf{Inventory Overflow Cost:}\\
This cost is incurred if the inventory of any product exceeds its maximum storage capacity:
\[
\text{C}^{\text{IV}}_{t}  = \rho_{\text{IV}} \cdot \sum_{j \in \mathcal{J}} \max(IV_{j,t} - IV_{j}^{\text{max}}, 0)
\]

\textbf{Demand Shortfall Cost:}\\
At the end of each day (i.e., every 24 hours), the environment evaluates whether the shipped quantity meets the daily demand. A cost is imposed for any shortfall:
\[
\text{C}^D_{t} = 
\begin{cases}
\rho_{\text{D}} \cdot \sum_{j \in J} \max\left(D[j,t]- \text{DQ}_{j,d}, 0\right) & \text{if } t \bmod 24 = 23, d = \left\lfloor \frac{t+1}{24} \right \rfloor \\
0 & \text{otherwise},
\end{cases}
\]
where \(D[j, 23] \) is the daily non-zero demand for product \( j \), and \( DQ_{j,d} \) is the quantity shipped. 
\paragraph{Total Cost}
\[  
\text{C}^{\text{total}}_{t} = \text{C}^{\text{IV}}_{t} + \text{C}^D_{t} 
\]
\subsubsection{Reward Function}

The reward at time \( t \) is defined as the negative of the production cost:
\[
\Pi^{production}_t = - \left( \text{production cost}_t \right)
\]
The production cost is formally given by:
\[
\text{production cost}_t  = 
\begin{cases}
 0& \text{if } \sum_{x} \lambda_x(t) = 0, \\
C_{\text{fixed}} + \left( \sum_{j \in J} PQ_{jt} \right) \cdot C_{\text{unit}} \cdot e[t] & \text{otherwise}
\end{cases}
\]
where \( PQ_{jt} \) is the production quantity of product \( j \), and \( e[t] \) is the electricity price at time \( t \). These costs arise from the fixed cost \( C_{\text{fixed}} \), incurred if any production occurs and a variable cost, proportional to the total production and electricity price at hour \( t \). 

\subsubsection{Episode Termination}

The episode terminates when \((t+1)=H\) or \((t+1) = 24D\), i.e., when the next starting hour is the first hour following the final day.
%\clearpage
\section{Results: Evaluation}\label{sec:appendix:realresults}
We have added the evaluation results in table format in Table \ref{table:evaluation} for readability.
\begin{sidewaystable}
\centering
\caption{Evaluation results for 10 episodes}
\label{table:evaluation}
\resizebox{\textheight}{!}{%
\begin{tabular}{l c cc cc cc cc cc cc cc cc}
\toprule
& \multirow{2}{*}{Optimal}
& \multicolumn{2}{c}{CPO}
& \multicolumn{2}{c}{DDPGLag}
& \multicolumn{2}{c}{OnCRPO}
& \multicolumn{2}{c}{P3O}
& \multicolumn{2}{c}{TRPOLag}
& \multicolumn{2}{c}{FOCOPS}
& \multicolumn{2}{c}{SACPID}
& \multicolumn{2}{c}{SACLag} \\
\cmidrule(lr){3-4}
\cmidrule(lr){5-6}
\cmidrule(lr){7-8}
\cmidrule(lr){9-10}
\cmidrule(lr){11-12}
\cmidrule(lr){13-14}
\cmidrule(lr){15-16}
\cmidrule(lr){17-18}
Environment 
& Reward 
& Reward & Cost 
& Reward & Cost 
& Reward & Cost
& Reward & Cost
& Reward & Cost
& Reward & Cost
& Reward & Cost
& Reward & Cost \\
\midrule
InvMgmtEnv 
& 11265.97 
& 7303.2 & 0 
& -4858.26 & 0
& 7598.66 & 0
& 1499.21 & 0
& 7198.26 & 0
& -6434.03 & 0
& 5555.74 & 0
& -14386.99 & 0 \\

SchedMaintEnv-v0 
& -1221.85 
& -1283.96 & 29.59
& -1700.98 & 11675.99
& -1277.08 & 25.56
& -1187.52 & 751.82
& -1272.61 & 19.85
& -1142.69 & 616.14
& -274.46 & 11461.29
& -274.46 & 11461.29 \\

SchedMaintEnv-v1 
& -1226.96 
& -1240.39 & 7.88
& -1700.98 & 11675.99
& -1244.07$\pm$1.49 & 103.21
& -1178.95$\pm$1.29 & 938.5$\pm$10.61
& -1217.24 & 21.35
& -1188.19$\pm$4.27 & 962.3$\pm$37.61
& -524.05 & 8600.12$\pm$31.62
& -274.46 & 11521.29$\pm$51.64 \\

UCEnv 
& -197258 
& -236553 & 5.56
& -337153 & 108
& -236989 & 8.55
& -556101 & 0
& -218334 & 9.19
& -220312 & 0
& -221922 & 41.68
& -298228 & 153.06 \\

GridStorageEnv 
& -54173 
& -95198.2 & 0.02
& -118704 & 120440
& -90589.6 & 0.02
& -111255 & 0.02
& -90265.5 & 0.02
& -126974 & 0.03
& -100407 & 2122.38
& -68690.5 & 0.03 \\

BlendingEnv 
& 1800 
& 0 & 190
& 0 & 32763.44
& 0.03 & 200
& 0.04 & 270.08
& 0.03 & 190
& 0.01 & 180.22
& 225.31 & 33101.34
& 523.75 & 33596.83 \\

RTNEnv 
& 363.78 
& -31.37 & 71
& -47.76 & 84
& -33.04 & 71
& -31.7 & 71
& -12.16 & 71
& -19.56 & 73
& -14.7 & 83
& -14.7 & 80 \\

STNEnv 
& 363.78 
& -36.76 & 71
& -10.38 & 84
& -37.01 & 71
& -30.83 & 73
& -11.34 & 71
& -28.79 & 73
& -14.7 & 81
& -14.7 & 81 \\

GTEPEnv
& -267.7
& -313 & 0
& -19 & 689140
& -298 & 0
& -346 & 79.28
& -352 & 0
& -367 & 0
& -403 & 10725.05
& -475 & 600
 \\

UCEnv-v1
& -264104
& -319469 & 0
& -575239 & 216
& -313897 & 0
& -1035242 & 0
& -553465 & 4.21
& -589754 & 0
& -390289 & 153.06
& -414884 & 195.9
 \\

ASUEnv
& -6597.5
& -7295.45 & 25563.34
& -7743.56 & 13130.41
& -7330.32 & 17706.08
& -7427.55 & 12624.76
& -7337.64 & 16018.8
& -7416.93 & 11747.68
& -7870.69 & 16713.63
& -8329.49 & 28864.44
 \\
% (your rows here unchanged)

\bottomrule
\end{tabular}}

\end{sidewaystable}
\clearpage
\section{Other Results}\label{sec:appendix:moreresults}

\subsection{Additonal Environments and Variants}

\begin{figure}[H]
    \centering

    \begin{subfigure}[t]{0.48\textwidth}
        \centering
        \includegraphics[width=0.49\textwidth]{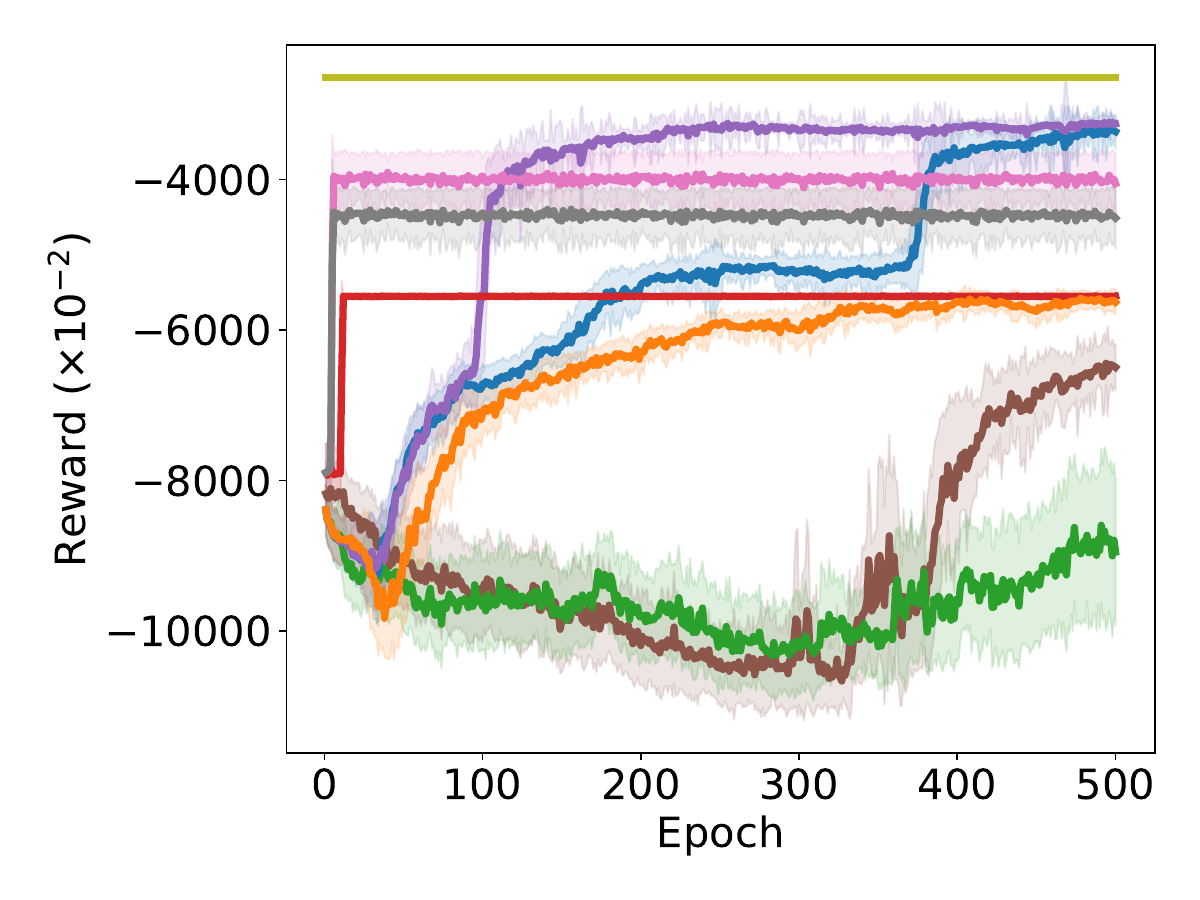}
        \includegraphics[width=0.49\textwidth]{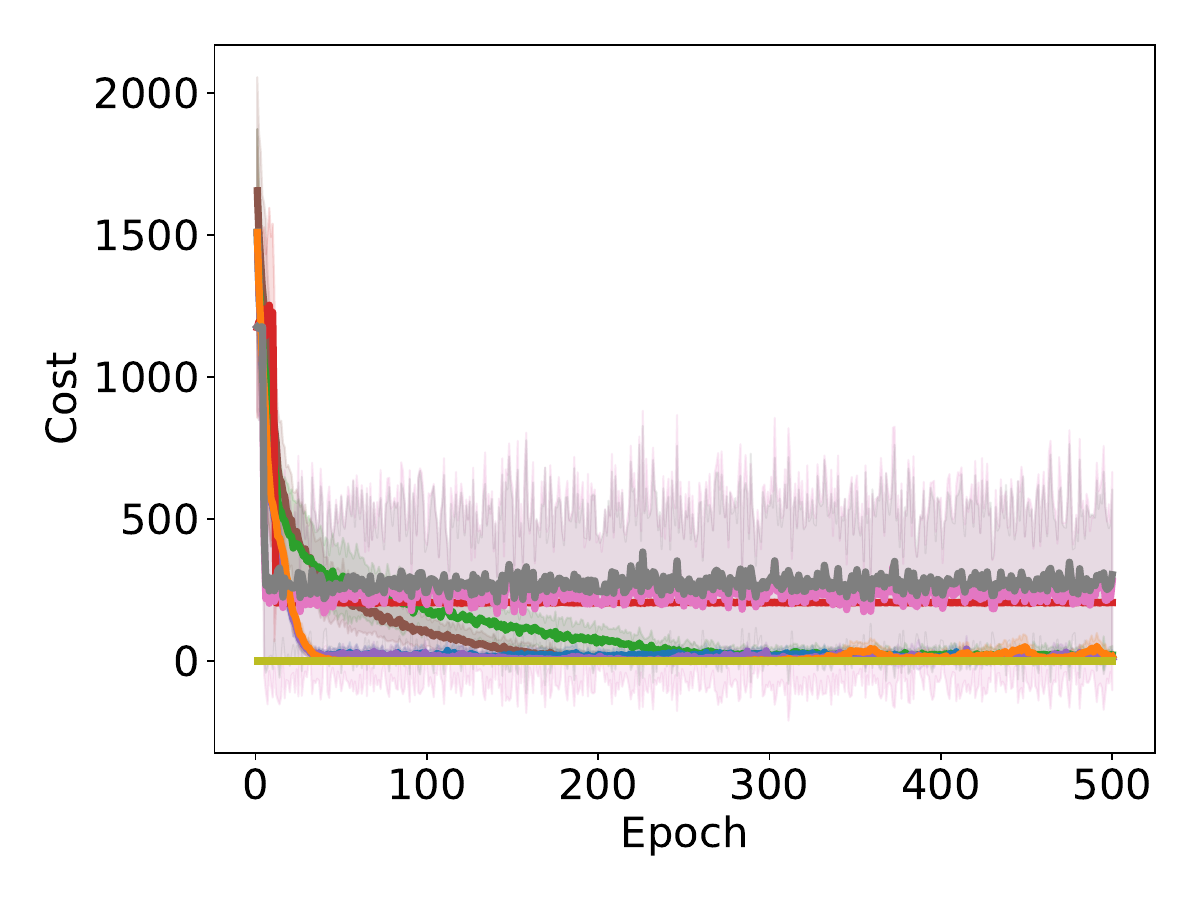}
        \caption{UCEnv-v1}
        \label{fig:UCEnv-v1}
    \end{subfigure}
    %ROw 1
    % Panel 1
    \begin{subfigure}[t]{0.48\textwidth}
        \centering
        \includegraphics[width=0.49\textwidth]{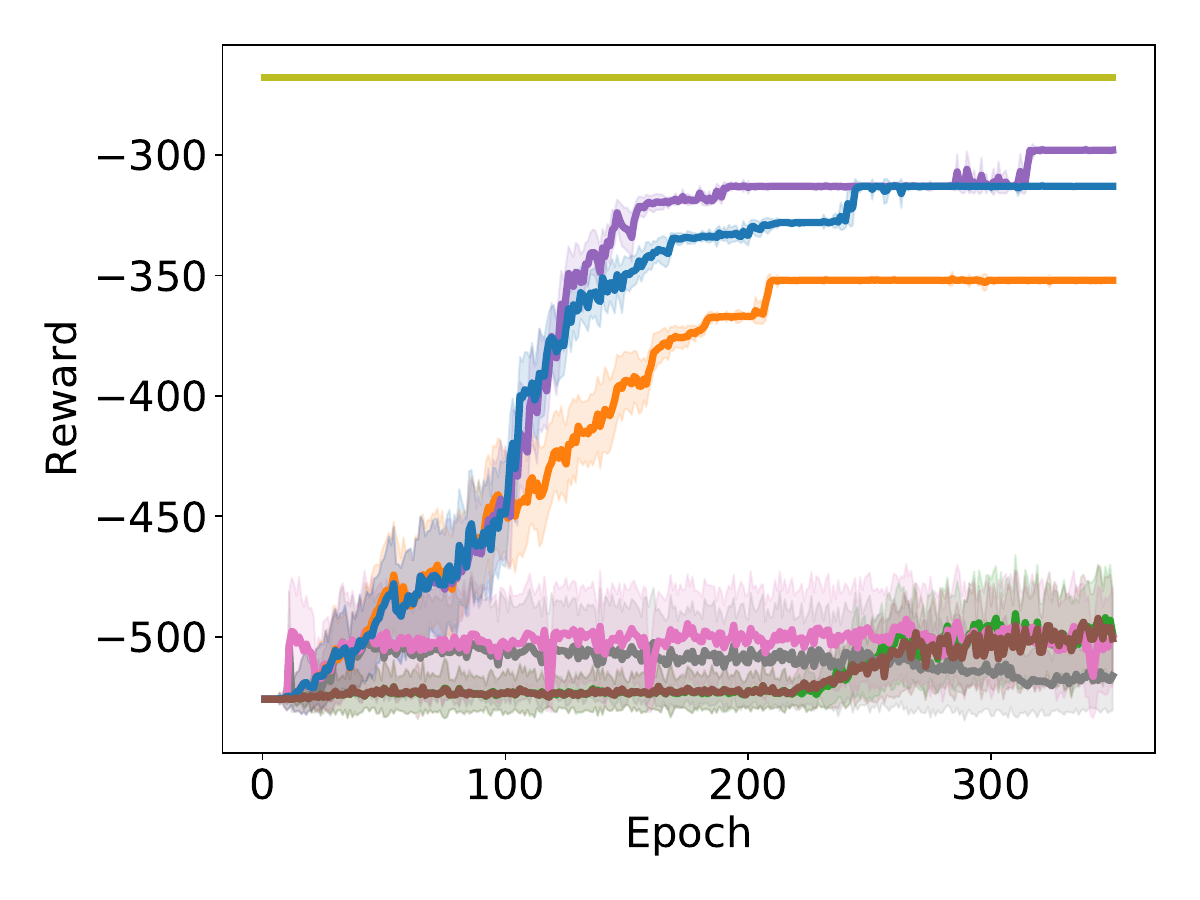}
        \includegraphics[width=0.49\textwidth]{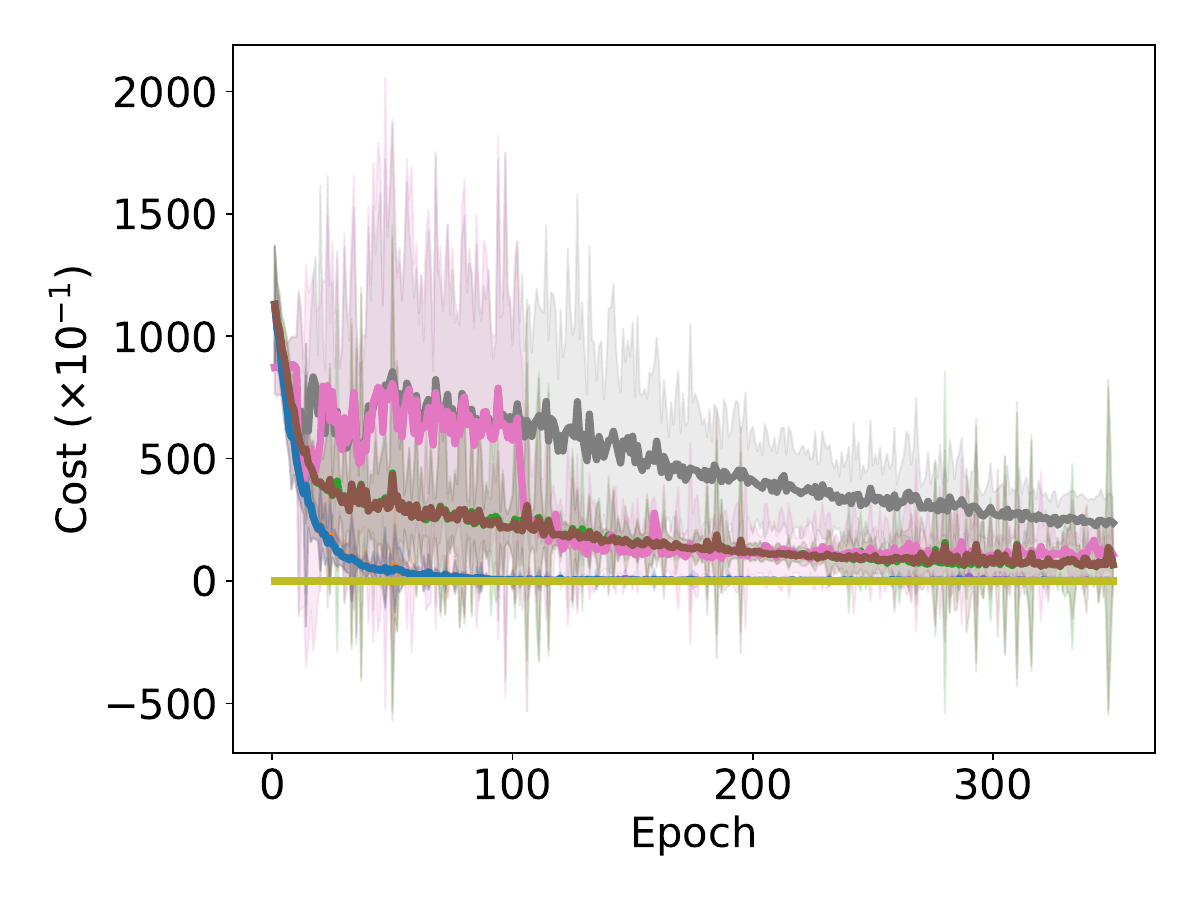}
        \caption{GTEPEnv}
        \label{fig:GTEPEnv}
    \end{subfigure}
    \hfill
    % Panel 2

    \vspace{1em}

%Row 3
    \begin{center}
        \begin{subfigure}[t]{0.48\textwidth}
            \centering
            \includegraphics[width=0.49\textwidth]{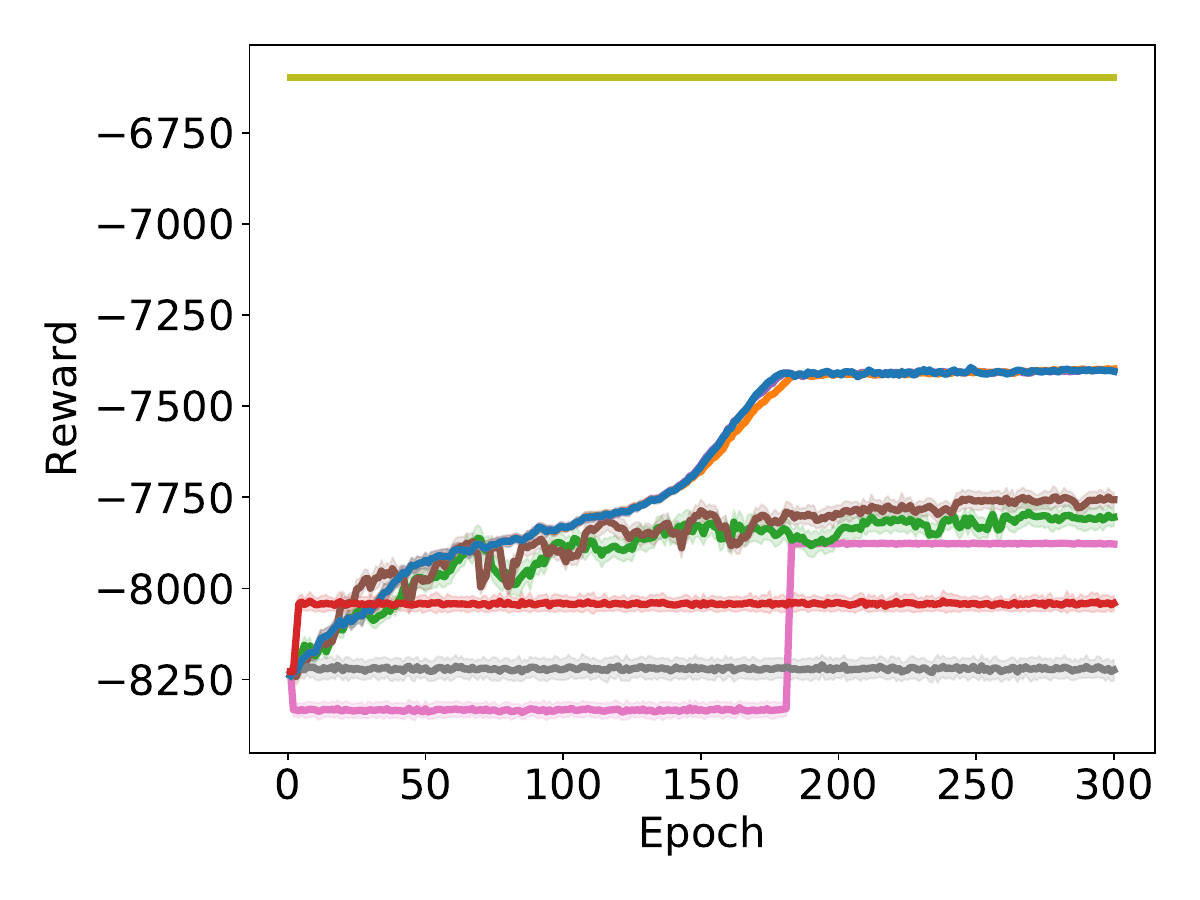}
            \includegraphics[width=0.49\textwidth]{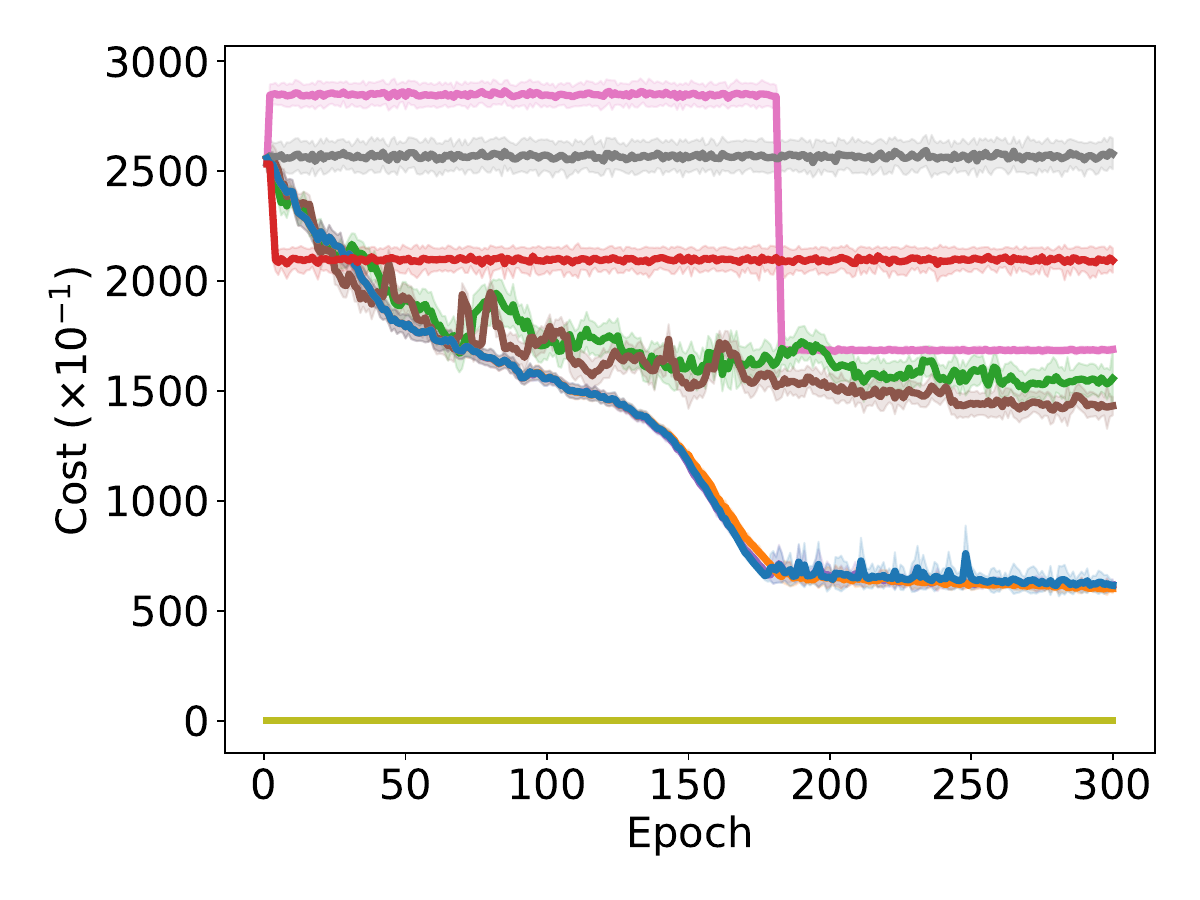}
            \caption{ASUEnv}
            \label{fig:Asu}
        \end{subfigure}
    \end{center}
    \hfill
    \vspace{0.5em}
    \centering
    \includegraphics[width=1\textwidth]{Images/shared_legend.pdf}
    \caption{Training curves of average reward and cost per episode across three additional environments and BlendingEnv variants with different strategies}
    \label{fig:all_panels_1}
\end{figure}

We consider results for three additional environments ASUEnv, UNEnv-v1, and GTEP. Figure~\ref{fig:all_panels_1} presents the average reward and cost per training epoch, with shaded regions indicating one standard deviation around the mean. 

Table~\ref{table:evaluation} includes the evaluation results, averaged over 10 episodes, for the additional environments. The optimal reward is calculated by solving an optimization counterpart of the environment at each step.

\subsubsection{Description of Environments}

\begin{itemize}

    \item \textbf{GTEPEnv:} This case study involves a five-region power system with two generators and possible transmission lines between all region pairs over a 10-period planning horizon. We train with 100 episodes per epoch. P3O shows a significant gap between training and evaluation in both reward and cost. Notably, DDPGLag learns a policy that achieves high rewards at the expense of significantly increased costs. To preserve clarity in visual comparisons, we omit the training curve for DDPGLag in Figure~\ref{fig:GTEPEnv}. We see a significant gap for training and evaluation costs in the P3O algorithm and for rewards in the P3O, FOCOPS, SACPID, and SACLag algorithms.

    \item \textbf{UCEnv-v1:} This unit commitment version models a multi-bus system with explicit power flows, enabling realistic evaluation of policies where generation and demand locations are critical. This example considers a multi-bus unit commitment power system with five generators distributed across four buses, connected by five transmission lines. It operates over a 24-hour horizon with hourly updates to demand forecasts, generator states, and network power flows. A significant gap is observed for training and evaluation costs in the SACPID and SACLag algorithms.

    \item \textbf{ASUEnv:} This example considers production scheduling in an air separation unit over a one-week horizon, where the agent takes hourly steps to meet daily product demands. We consider a planning horizon of 5 days, with each episode consisting of 168 steps, reflecting the operation of the ASU over a week. We train using 60 episodes per epoch. A significant gap is observed between the training and testing costs for CPO and TRPOLag.We see a significant gap for training and evaluation costs in the CPO, DDPGLag,OnCRPO, and TRPOLag algorithms.
    
\end{itemize}

\begin{figure}[t]
    \centering

    % -------- Row 1 --------
    \begin{subfigure}[t]{0.49\textwidth}
        \centering
        \includegraphics[width=\linewidth]{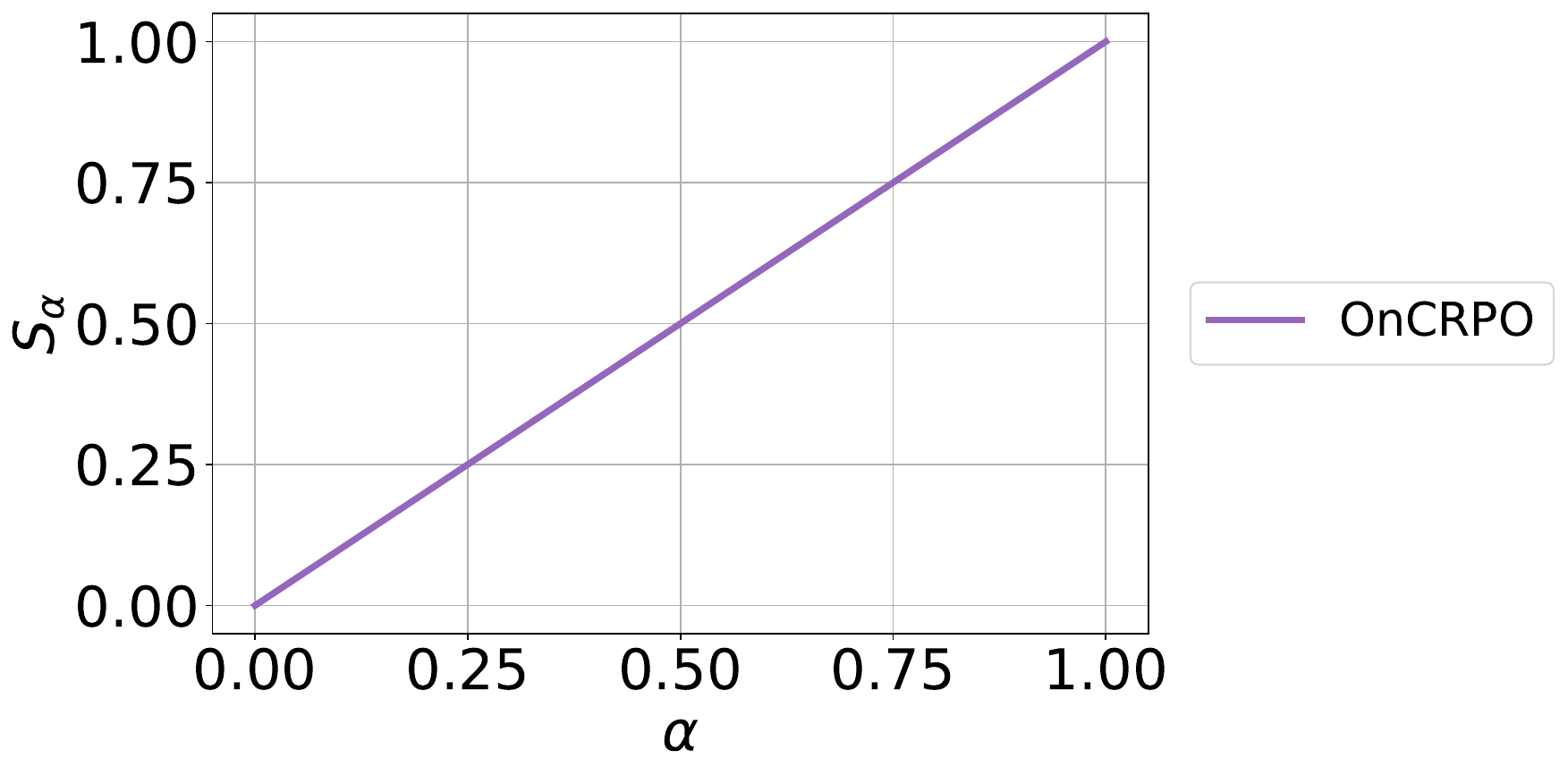}
        \caption{UCEnv-v1}
    \end{subfigure}
    \hfill
    \begin{subfigure}[t]{0.49\textwidth}
        \centering
        \includegraphics[width=\linewidth]{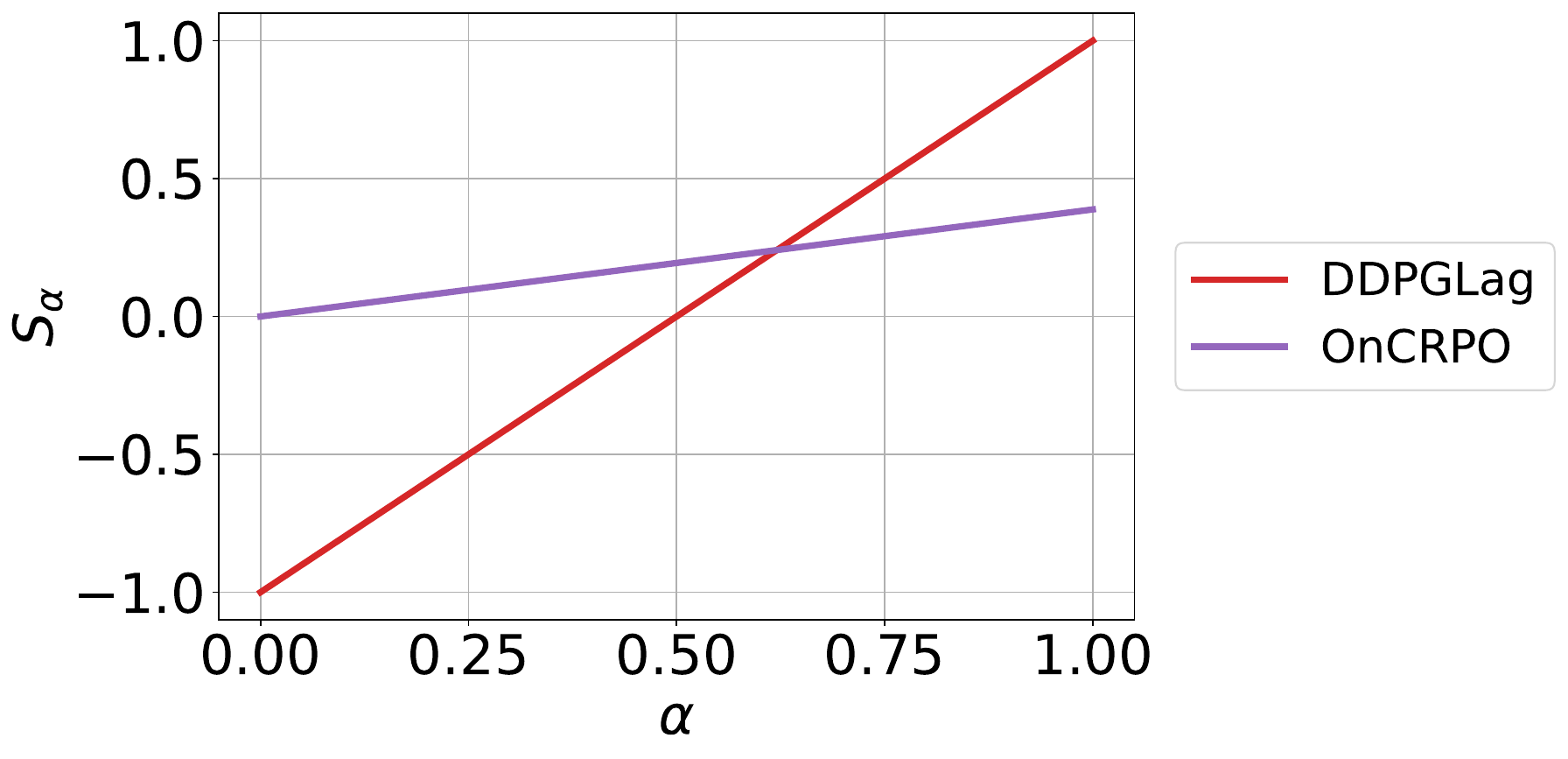}
        \caption{GTEPEnv}
    \end{subfigure}

    \vspace{1em}

    % -------- Row 2 --------
    \begin{subfigure}[t]{0.49\textwidth}
        \centering
        \includegraphics[width=\linewidth]{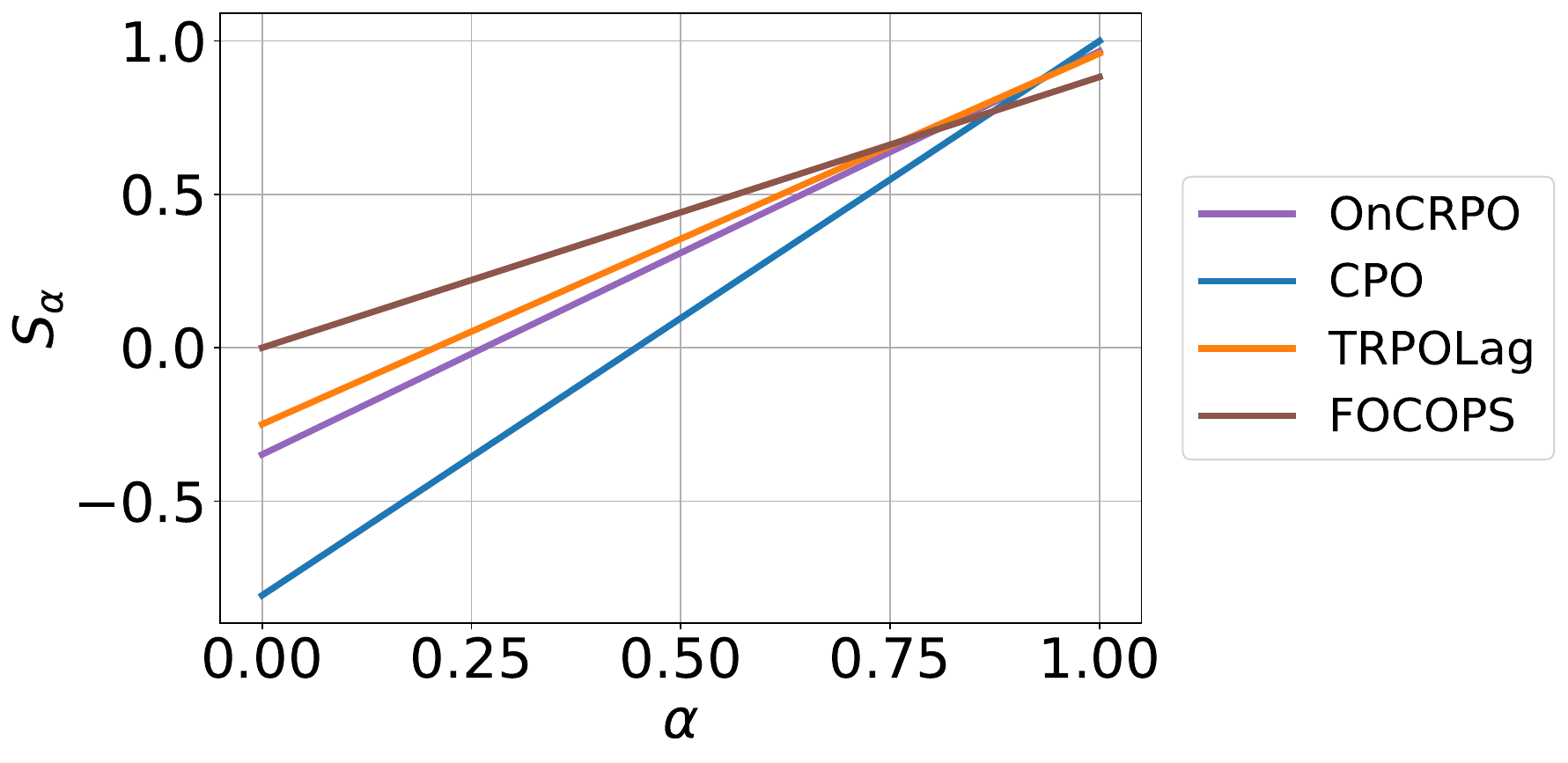}
        \caption{ASUEnv}
    \end{subfigure}

    \caption{Scalarized Pareto score across environments.}
    \label{fig:all_panels_alpha_extra}
\end{figure}

We also find out the Pareto efficient algorithms and the best performing algorithms as described in Section ~\ref{section:eval_criteria} and plot $S_\alpha$ in Figure ~\ref{fig:all_panels_alpha_extra}, as well as summarize those results in Table \ref{table:best_algo_extra}.

\begin{table}[htbp]
\centering
    \caption{Best performing algorithm}
    \label{table:best_algo_extra}
\begin{tabular}{lll}
Environment & Pareto efficient algorithms & Best performing algorithm \\
\hline 
UCEnv-v1 & OnCRPO & OnCRPO \\
GTEPEnv & DDPGLag, OnCRPO & OnCRPO \\
ASUEnv & CPO, FOCOPS, OnCRPO, TRPOLag & FOCOPS \\
\end{tabular}
\end{table}
\subsubsection{Discussion}

For the additional environments in Table~\ref{table:best_algo_extra}, we again observe environment-dependent multi-objective behavior. In \textbf{GTEPEnv}, the Pareto frontier consists of \textbf{DDPGLag} and \textbf{OnCRPO}, reflecting a trade-off between reward and constraint violation; among these, \textbf{OnCRPO} achieves the strongest scalarized performance and is therefore selected as best-performing. In \textbf{UCEnv-v1}, \textbf{OnCRPO} is the only Pareto-efficient method, indicating that it strictly dominates all other algorithms in the reward–violation space and provides the most favorable trade-off. In contrast, \textbf{ASUEnv} admits multiple Pareto-efficient algorithms—\textbf{CPO}, \textbf{FOCOPS}, \textbf{OnCRPO}, and \textbf{TRPOLag}—suggesting several competitive reward–safety trade-offs; among these, \textbf{FOCOPS} achieves the strongest overall scalarized performance. Overall, these results further emphasize that no single algorithm universally dominates across environments, and that performance must be interpreted through the lens of multi-objective trade-offs.

Furthermore, while GTEPEnv and UCEnv-v1 are trained to reasonable optimality, the evaluation costs for ASUEnv remain high for all the algorithms, indicating the issues faced by current algorithms to get good feasible solutions.  This further underscores the persistent challenges inherent in such constrained settings and highlights the urgent need for more refined and robust approaches to safe reinforcement learning in these domains.

% \begin{itemize}
% \item TO DO::::
% \item Explain the experiment (number of different stuff)
% \item Different versions of the reward
% \item Plot + insight

% \end{itemize}

% Note:  We use the help of ChatGPT to formulate the strategy for choosing the best and worst performing algorithms.

\subsection{Constraint-wise violation analysis across Safe-RL algorithms
}

In the following subsection we interpret the plots that report the \emph{mean episode-level constraint breaches per epoch} observed during training with eight Safe RL algorithms.
To help the reader connect each trend to its control philosophy, we first give a concise description of every algorithm and its reward–cost balancing mechanism.

\begin{itemize}
\item \textbf{Constrained Policy Optimization (CPO):} At each update, CPO solves a small trust-region \emph{quadratic program} (QP) that maximizes expected return while ensuring the new policy remains within a cumulative-cost constraint set. This provides theoretical guarantees of monotonic improvement under standard approximations in reward without violating the safety constraints.

\item \textbf{Trust-Region Policy Optimization with a Lagrange Multiplier (TRPOLag):} TRPOLag enhances standard \emph{Trust-Region Policy Optimization} (TRPO) by introducing an on-policy Lagrange multiplier, updated after each batch. This multiplier penalizes excessive costs within the Kullback–Leibler trust region, balancing reward optimization and safety.

\item \textbf{On-Policy Constrained Reinforcement Policy Optimization (OnCRPO):} OnCRPO alternates between maximizing reward using a standard \emph{Proximal Policy Optimization} (PPO) objective when constraints are satisfied, and minimizing costs through a dedicated surrogate objective when constraints are breached, thus explicitly balancing reward and safety.

\item \textbf{Penalty-based Proximal Policy Optimization (P3O):} P3O integrates an \emph{adaptive exterior penalty} into PPO's clipped-surrogate objective, dynamically adjusting penalty strength based on constraint violations. The penalty increases when cumulative cost exceeds its budget and decreases otherwise, gradually guiding the policy towards feasibility while prioritizing reward.

\item \textbf{Deep Deterministic Policy Gradient with a Lagrange Multiplier (DDPGLag):} DDPGLag employs an off-policy deterministic actor–critic architecture, augmented with stability enhancements from \emph{Twin Delayed Deep Deterministic Policy Gradients} (TD3). It concurrently learns a Lagrange multiplier, policy, and critic from replay-buffer data, effectively balancing reward and constraint satisfaction through deterministic policy gradients.

\item \textbf{Soft Actor–Critic with Lagrangian Penalty (SACLag):} An off‐policy actor–critic that augments the Soft Actor–Critic (SAC) objective with a learned Lagrange multiplier, adapting the penalty on expected cost online.  
  \item \textbf{Soft Actor–Critic with PID Control (SACPID):} Extends SAC with a proportional–integral–derivative controller on cumulative cost, automatically tuning penalty strength via PID updates to balance reward and safety.  
  \item \textbf{First‐Order Constrained Policy Optimization (FOCOPS):} A trust‐region method that linearizes both reward and cost objectives and solves a first‐order approximation via a closed‐form update, yielding a policy that enforces cost constraints with minimal computational overhead.  
\end{itemize}

\subsubsection{RTNEnv}

\begin{figure}[H]
    \centering
    \begin{subfigure}[b]{0.45\textwidth}
        \centering
        \includegraphics[width=\textwidth]{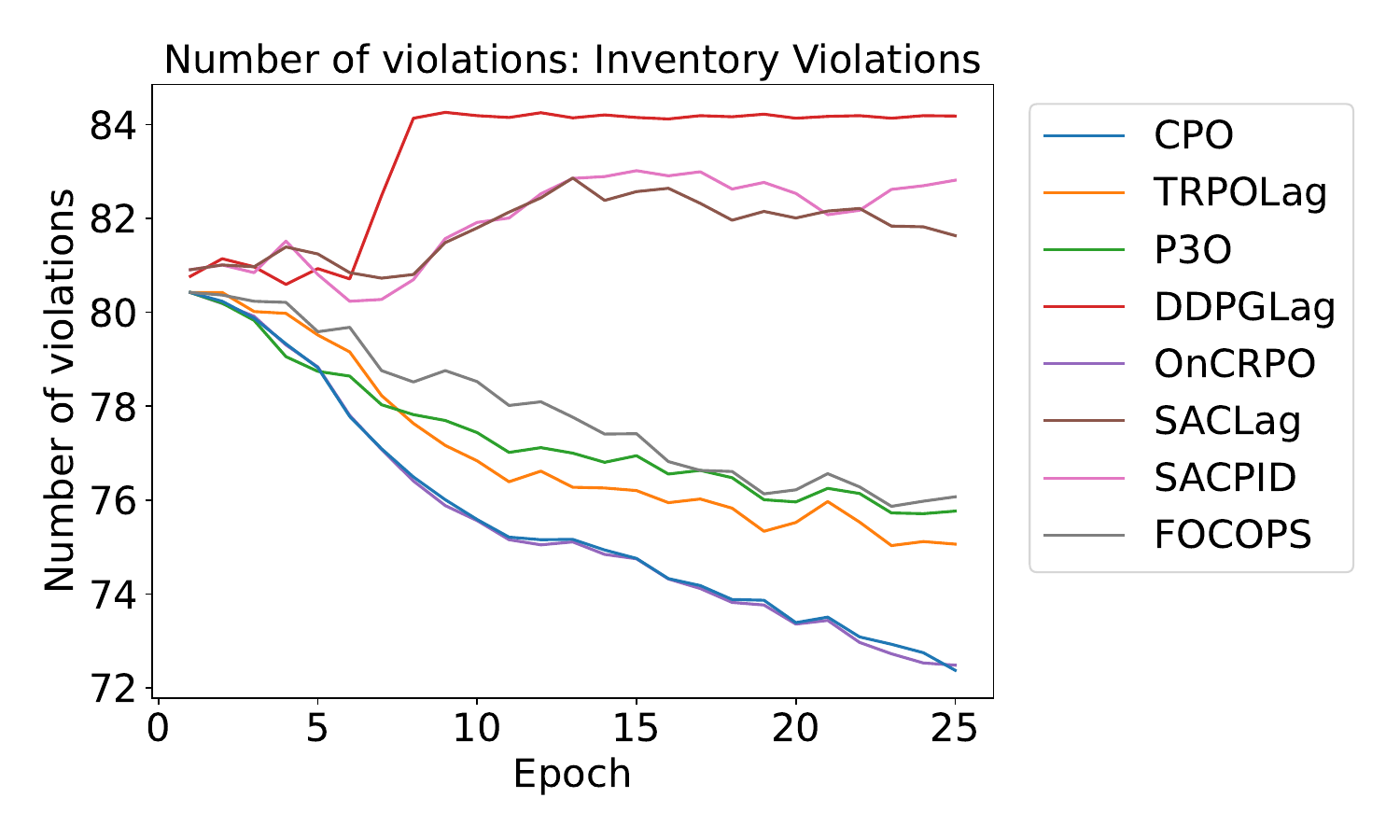}
    \end{subfigure}
    \hfill
    \begin{subfigure}[b]{0.45\textwidth}
        \centering
        \includegraphics[width=\textwidth]{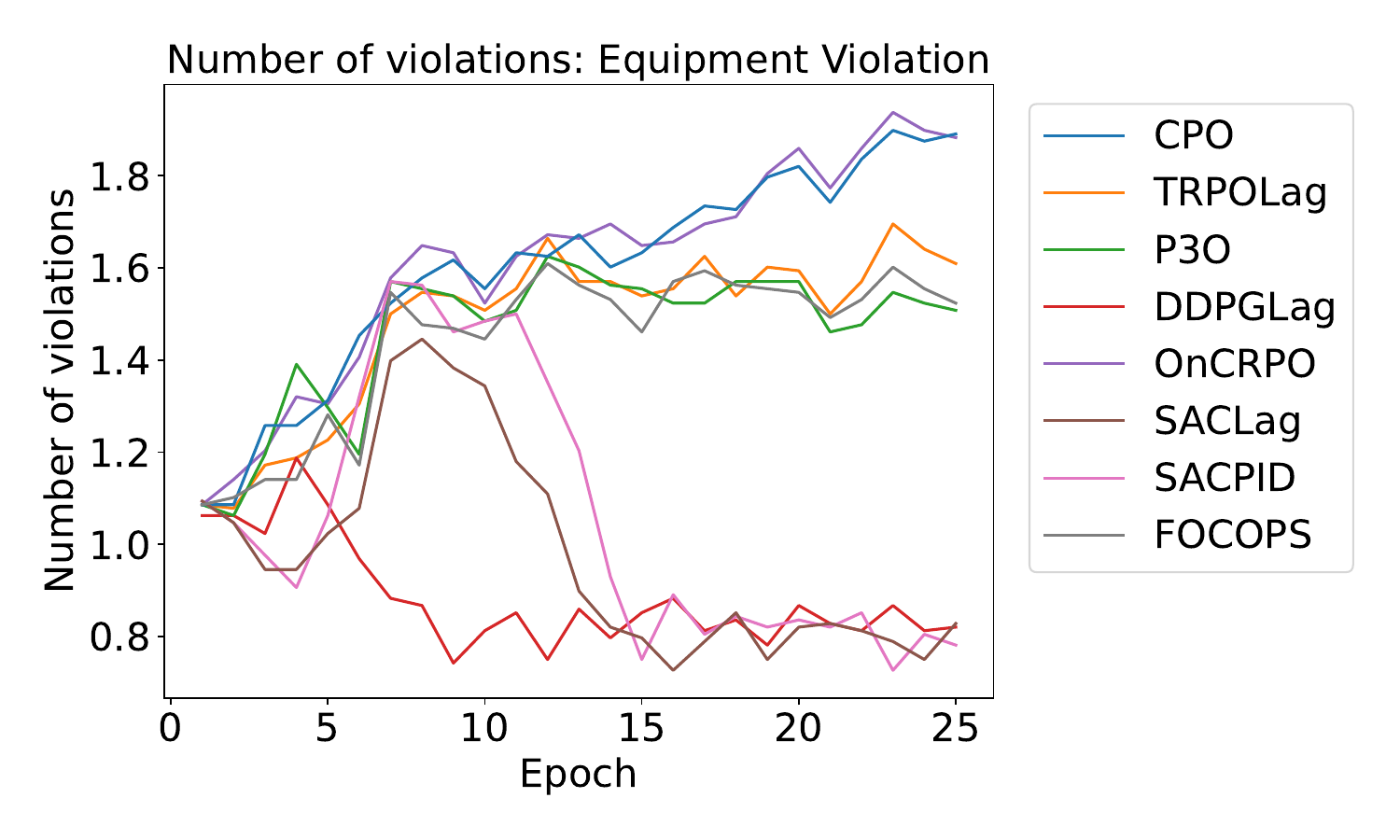}
    \end{subfigure}
    \caption{Average number of episode violations for different epochs for RTNEnv}
    \label{rtnenv_1}
\end{figure}

Figure~\ref{rtnenv_1} illustrates mean episode-level constraint violations per epoch for the RTNEnv. For \emph{inventory-level} constraints, projection-based algorithms—\textbf{CPO}, \textbf{OnCRPO}, and \textbf{TRPOLag}—consistently reduce violations, converging swiftly to minimal violation levels. \textbf{P3O} achieves comparable compliance more gradually, while \textbf{DDPGLag} consistently exhibits the highest residual violations. Both \textbf{SACLag} and \textbf{SACPID} demonstrate strong reductions in inventory violations, reaching compliance at rates comparable to projection-based methods, whereas \textbf{FOCOPS} shows slower improvement and stabilizes with moderately higher residual violations. \emph{Equipment-usage} violations are inherently less frequent owing to the simpler nature of the constraints compared to inventory management; all methods maintain near-baseline levels throughout training, with CPO, OnCRPO, TRPOLag, and the SAC-based methods (SACLag, SACPID) achieving compliance earliest, followed slightly later by P3O, and with DDPGLag and FOCOPS maintaining modestly higher yet infrequent violation rates.

\subsubsection{STNEnv}

\begin{figure}[H]
    \centering
    \begin{subfigure}[b]{0.45\textwidth}
        \centering
        \includegraphics[width=\textwidth]{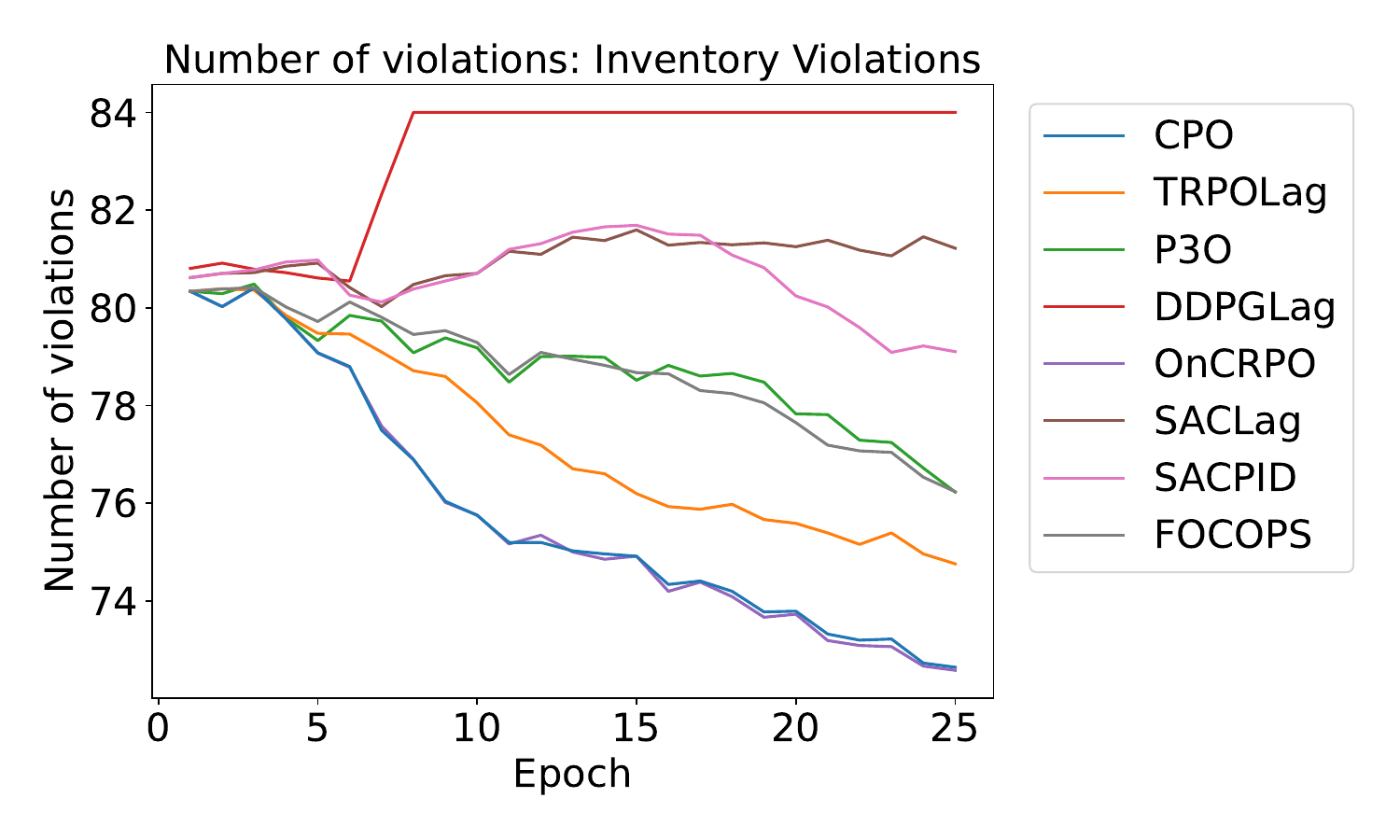}
    \end{subfigure}
    \hfill
    \begin{subfigure}[b]{0.45\textwidth}
        \centering
        \includegraphics[width=\textwidth]{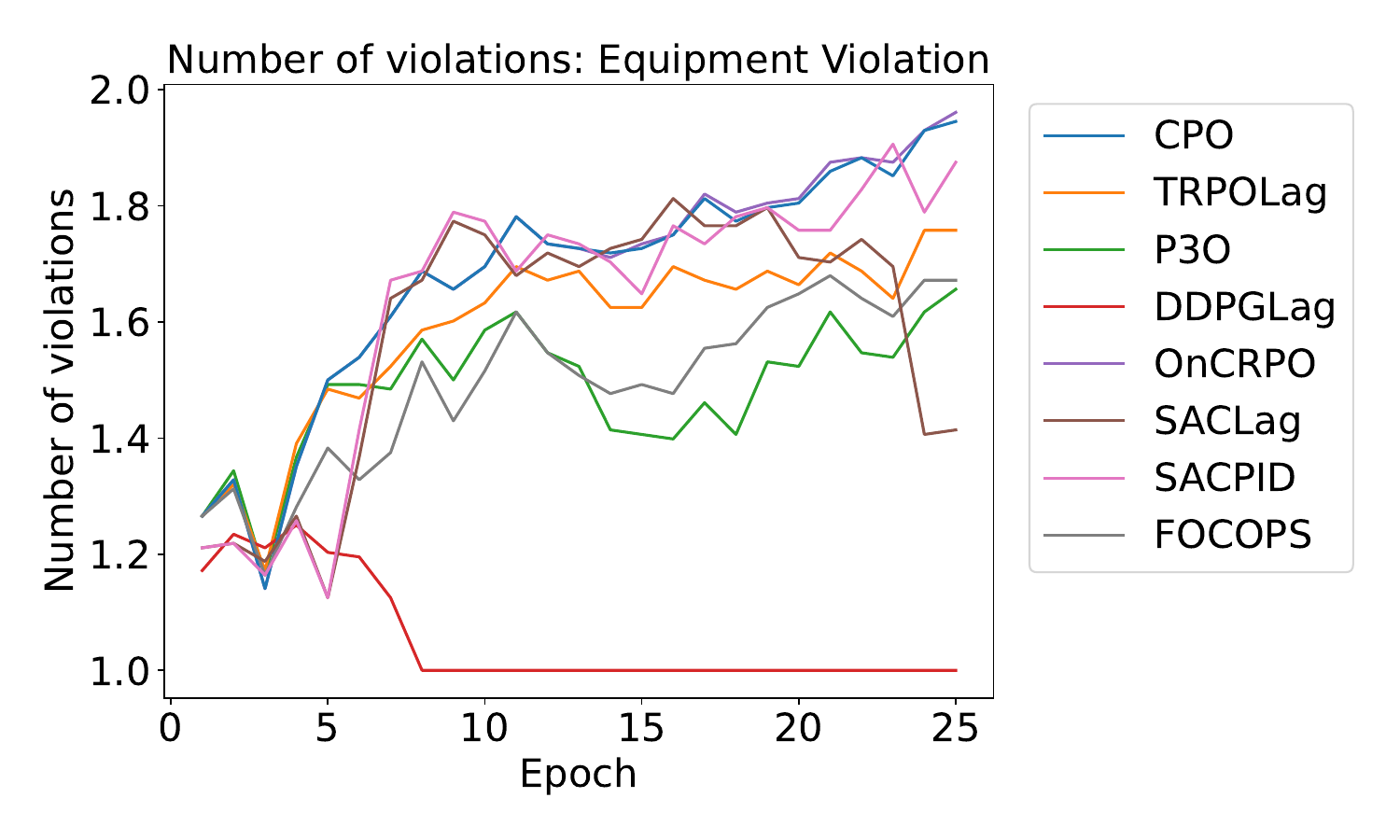}
    \end{subfigure}
    \caption{Average number of episode violations for different epochs for STNEnv}
    \label{stnenv_1}
\end{figure}

Figure~\ref{stnenv_1} presents mean episode-level constraint breaches per epoch for the STNEnv. Regarding \emph{inventory-level} safety, the projection-driven algorithms—\textbf{CPO}, \textbf{OnCRPO}, and \textbf{TRPOLag}—steadily minimize violations, stabilizing at the lowest counts observed. \textbf{P3O} demonstrates a slower convergence, while \textbf{DDPGLag} retains a significantly higher residual violation count. Both \textbf{SACLag} and \textbf{SACPID} show strong reduction in inventory violations, with convergence patterns comparable to projection-based methods, whereas \textbf{FOCOPS} achieves improvement but stabilizes at slightly higher violation levels. \emph{Equipment-usage} breaches are consistently low for all algorithms due to reasons similar to the RTN environment, with immediate practical compliance from CPO, OnCRPO, TRPOLag, and the SAC-based methods (SACLag, SACPID), followed shortly thereafter by P3O, while DDPGLag and FOCOPS maintain modestly higher yet infrequent violation rates.

\newpage
\subsubsection{UCEnv}
\begin{figure}[H]
    \centering
    \begin{subfigure}[b]{0.45\textwidth}
        \centering
        \includegraphics[width=\textwidth]{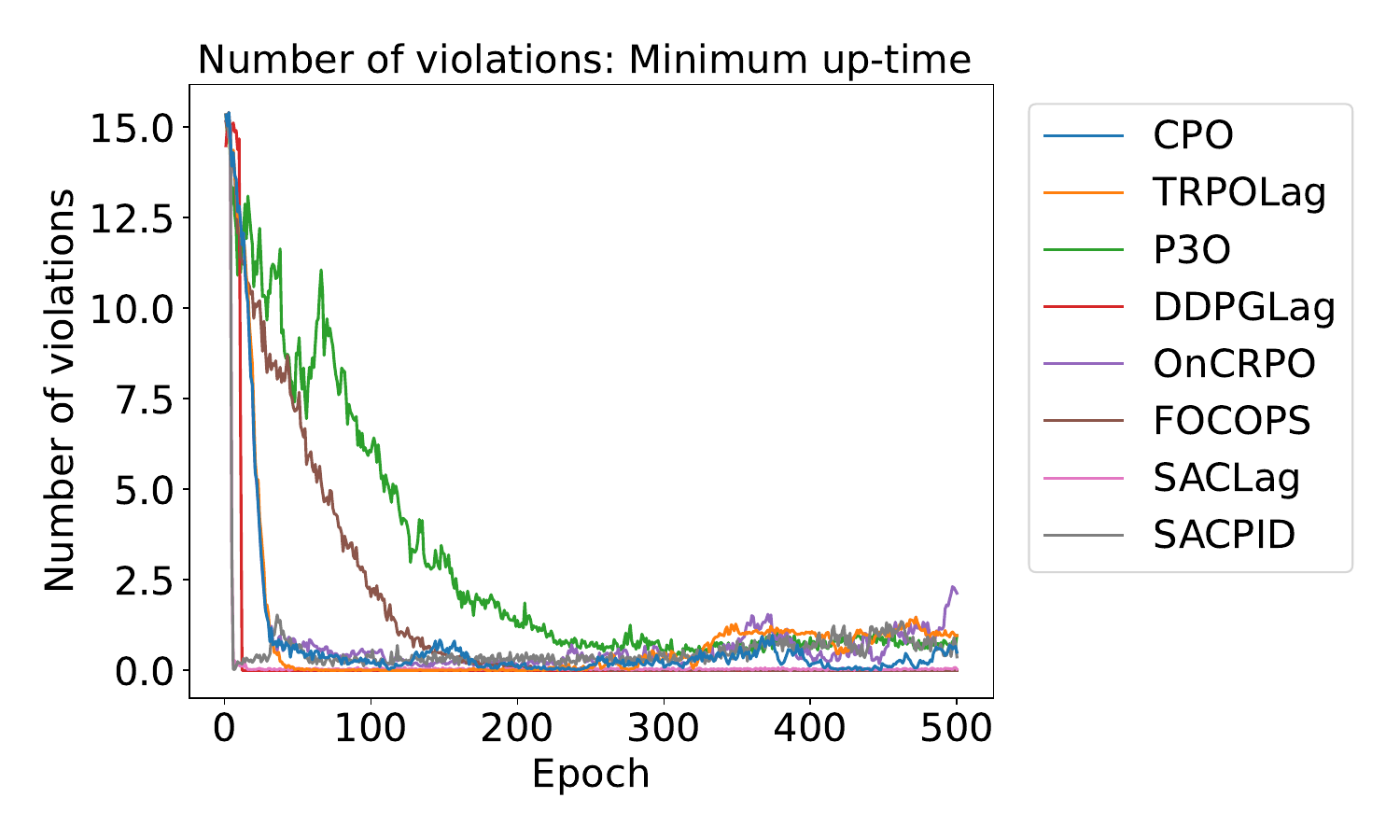}
    \end{subfigure}
    \hfill
    \begin{subfigure}[b]{0.45\textwidth}
        \centering
        \includegraphics[width=\textwidth]{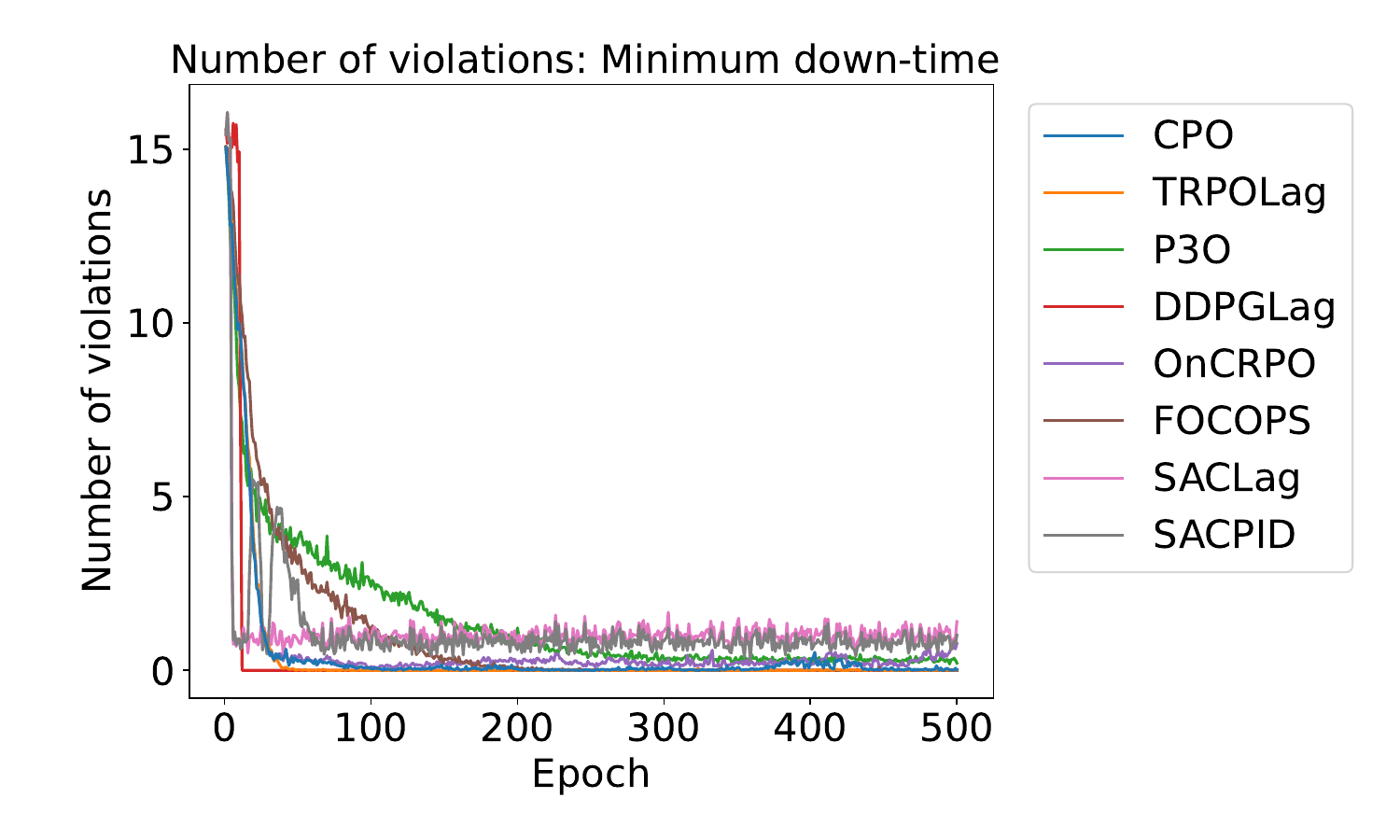}
    \end{subfigure}
        \begin{subfigure}[b]{0.45\textwidth}
        \centering
        \includegraphics[width=\textwidth]{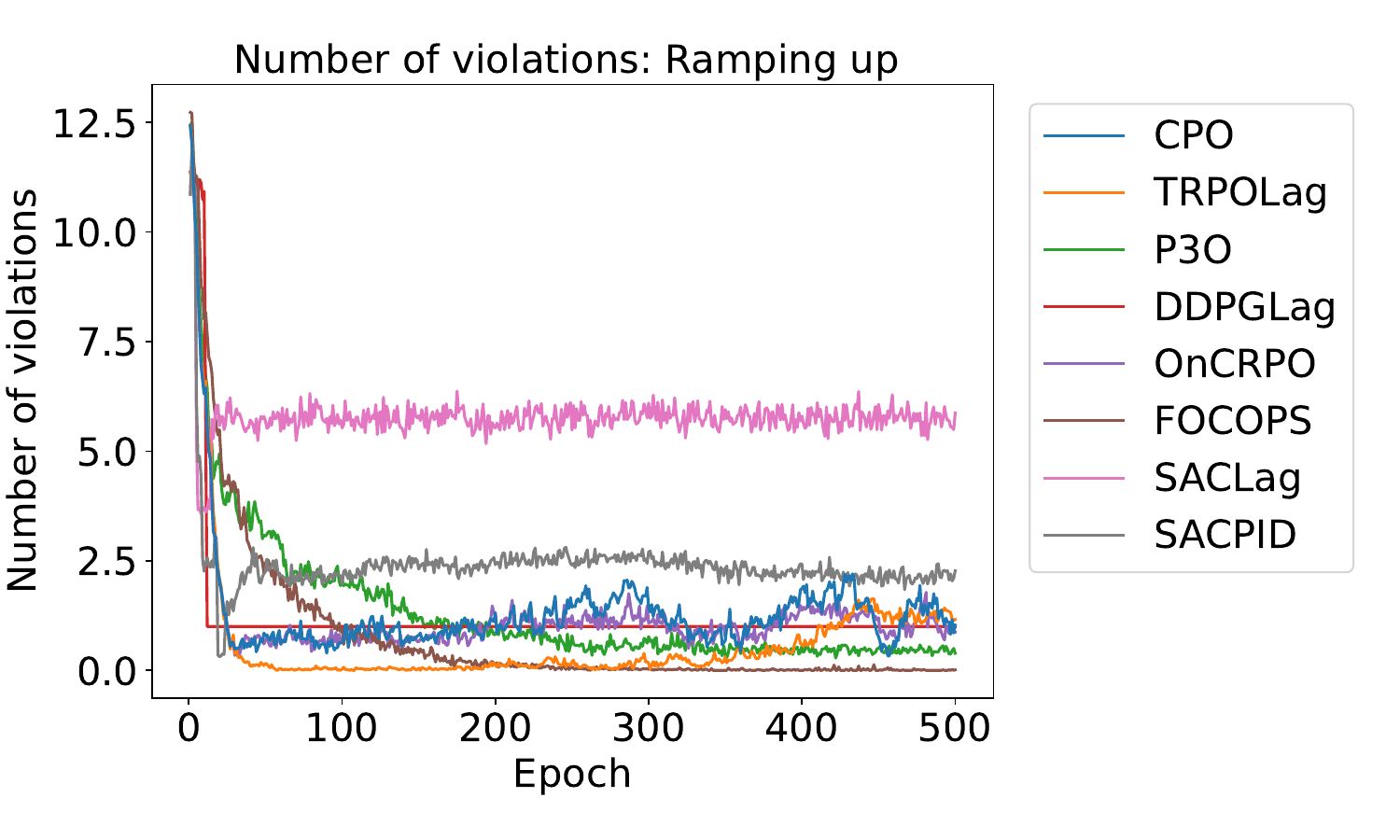}
    \end{subfigure}
    \begin{subfigure}[b]{0.45\textwidth}
        \centering
        \includegraphics[width=\textwidth]{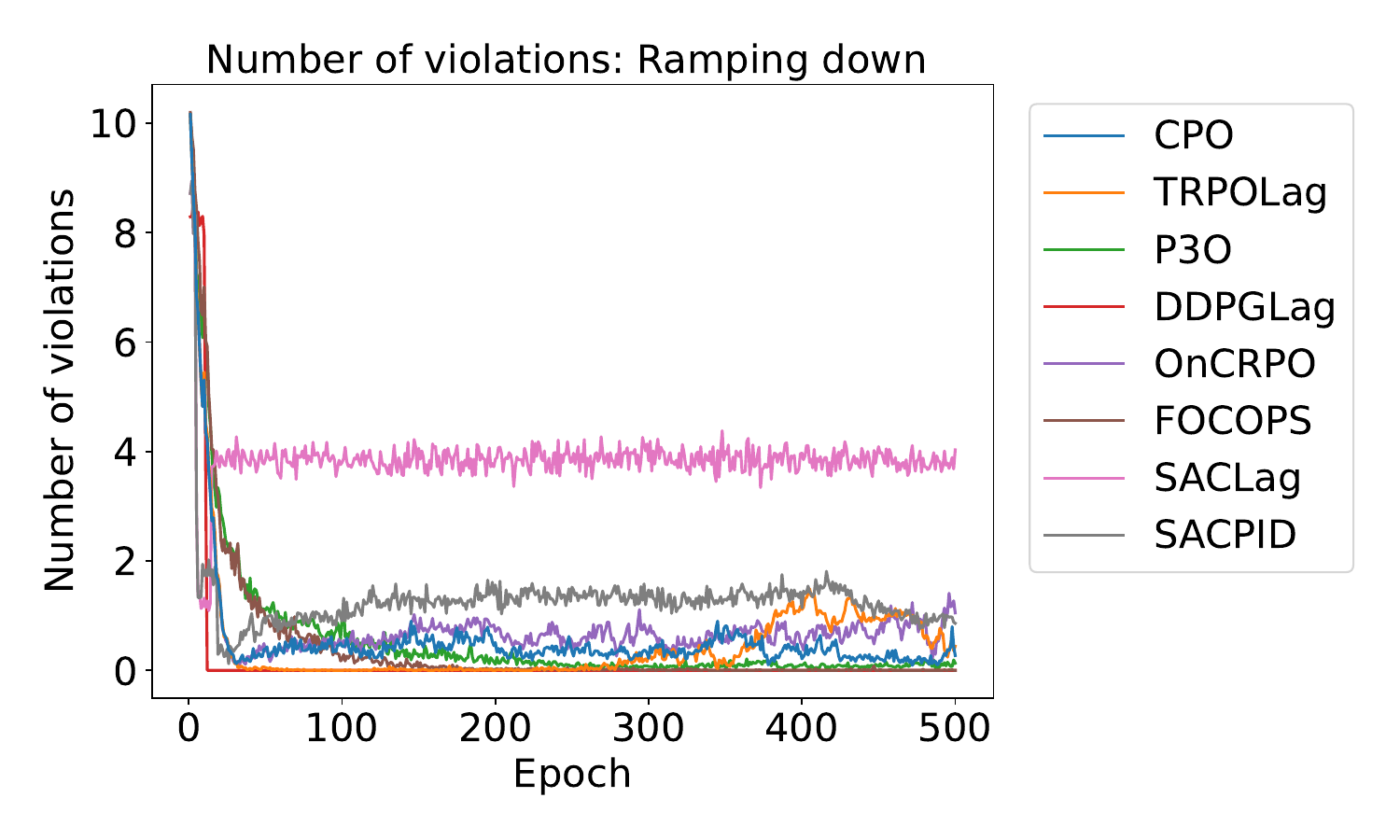}
    \end{subfigure}
    \caption{Average number of episode violations for different epochs for UCEnv-v0}
    \label{ucenv_1}
\end{figure}

\paragraph{Single-bus system without network constraints (UCEnv-v0):}
Figure~\ref{ucenv_1} reports mean episode-level constraint violations per epoch for the single-bus UCEnv problem, covering \emph{minimum up-time}, \emph{minimum down-time}, and \emph{ramping-rate} constraints. Projection-based algorithms—\textbf{CPO}, \textbf{OnCRPO}, \textbf{TRPOLag}, and \textbf{FOCOPS}—rapidly reduce violations (in about 100 epochs), converging to the lowest residual counts. \textbf{P3O} achieves compliance more gradually as its adaptive penalty strengthens, while \textbf{DDPGLag} is the quickest to reduce violations (in less than 10 epochs) early on due to aggressive Lagrangian penalty updates and off-policy learning but plateaus with ramp-up residual violations due to higher variance and instability. In contrast, SAC-based methods, \textbf{SACLag} and \textbf{SACPID}, exhibit considerably higher residual violation for the ramp-up/ramp-down constraints than other algorithms.

\begin{figure}[H]
    \centering
    \begin{subfigure}[b]{0.45\textwidth}
        \centering
        \includegraphics[width=\textwidth]{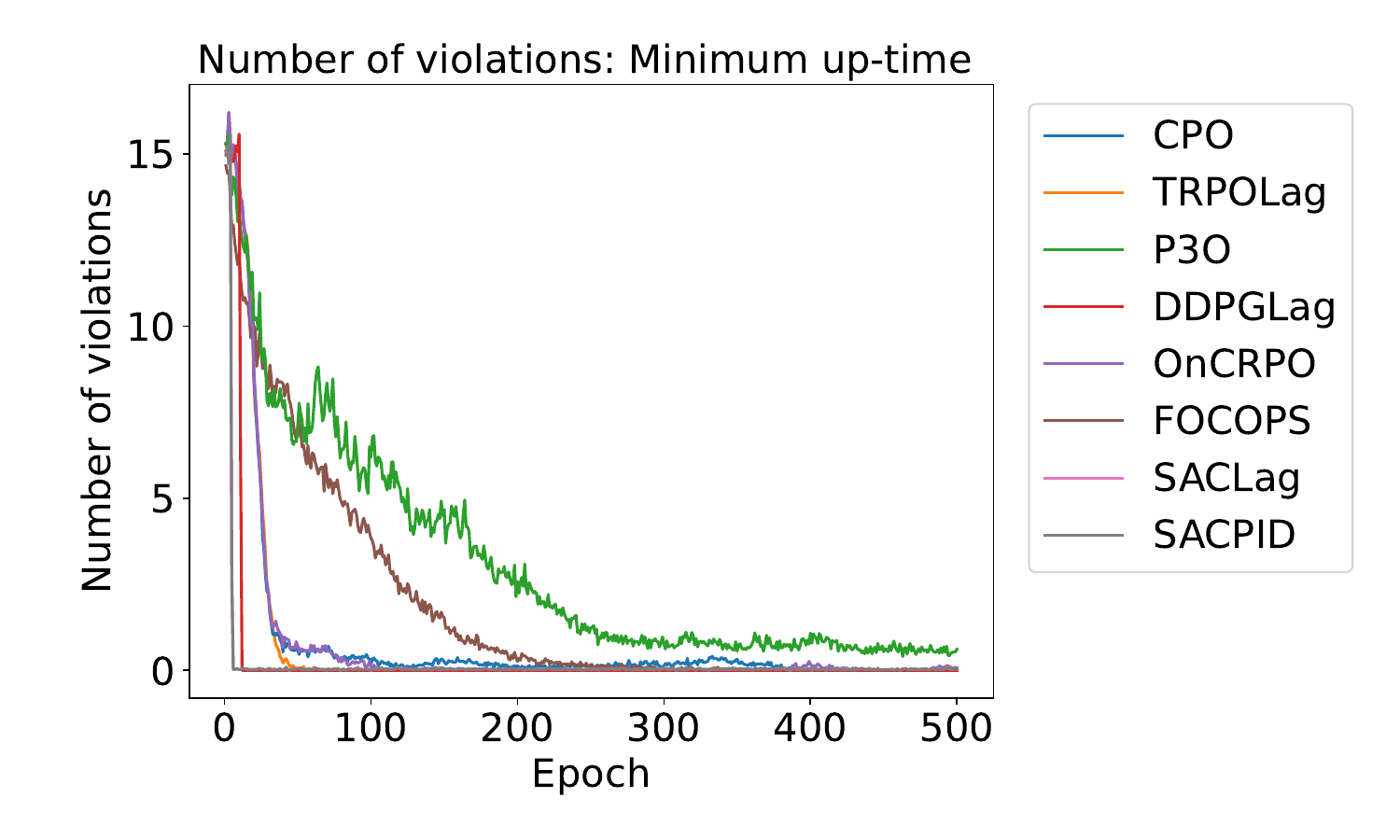}
    \end{subfigure}
    \hfill
    \begin{subfigure}[b]{0.45\textwidth}
        \centering
        \includegraphics[width=\textwidth]{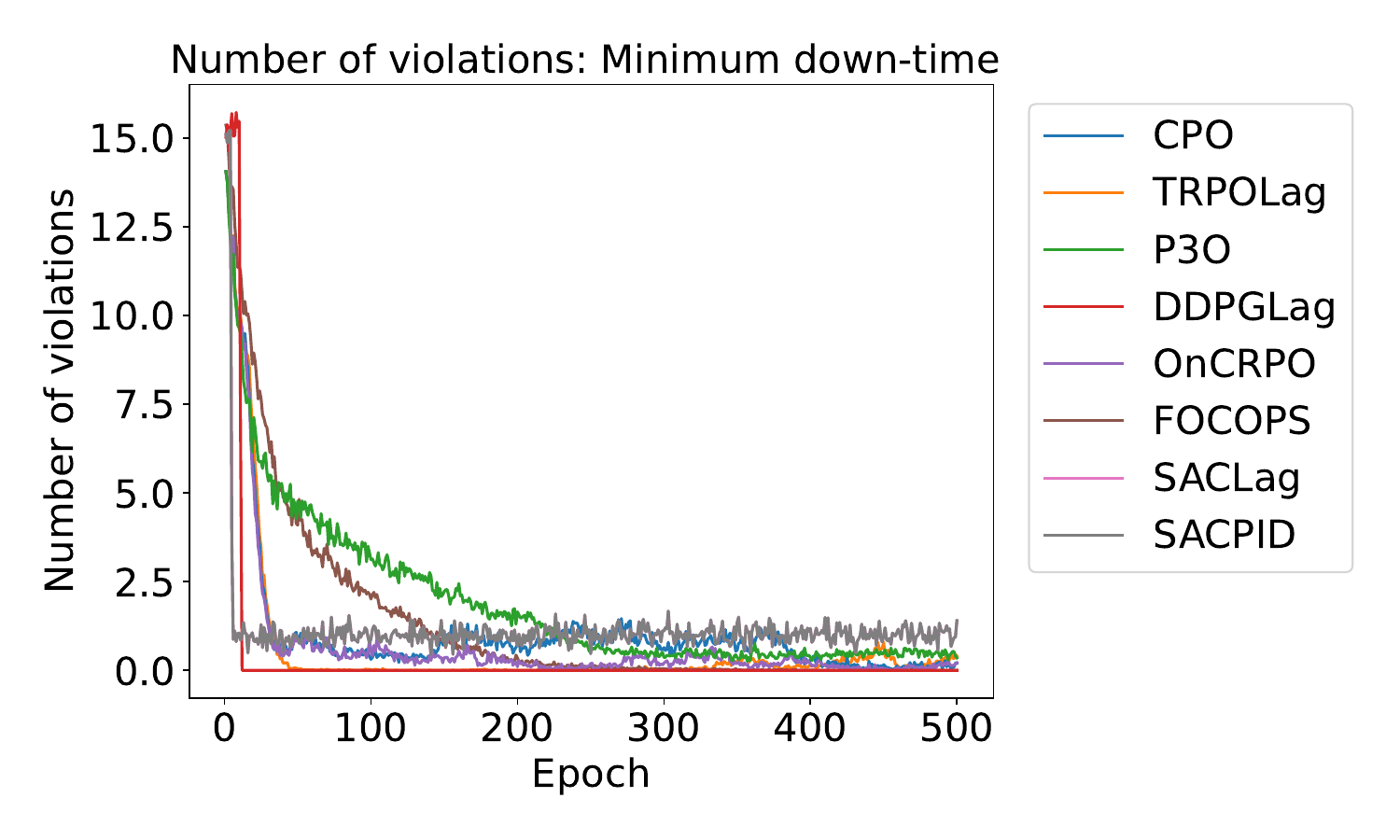}
    \end{subfigure}
        \begin{subfigure}[b]{0.45\textwidth}
        \centering
        \includegraphics[width=\textwidth]{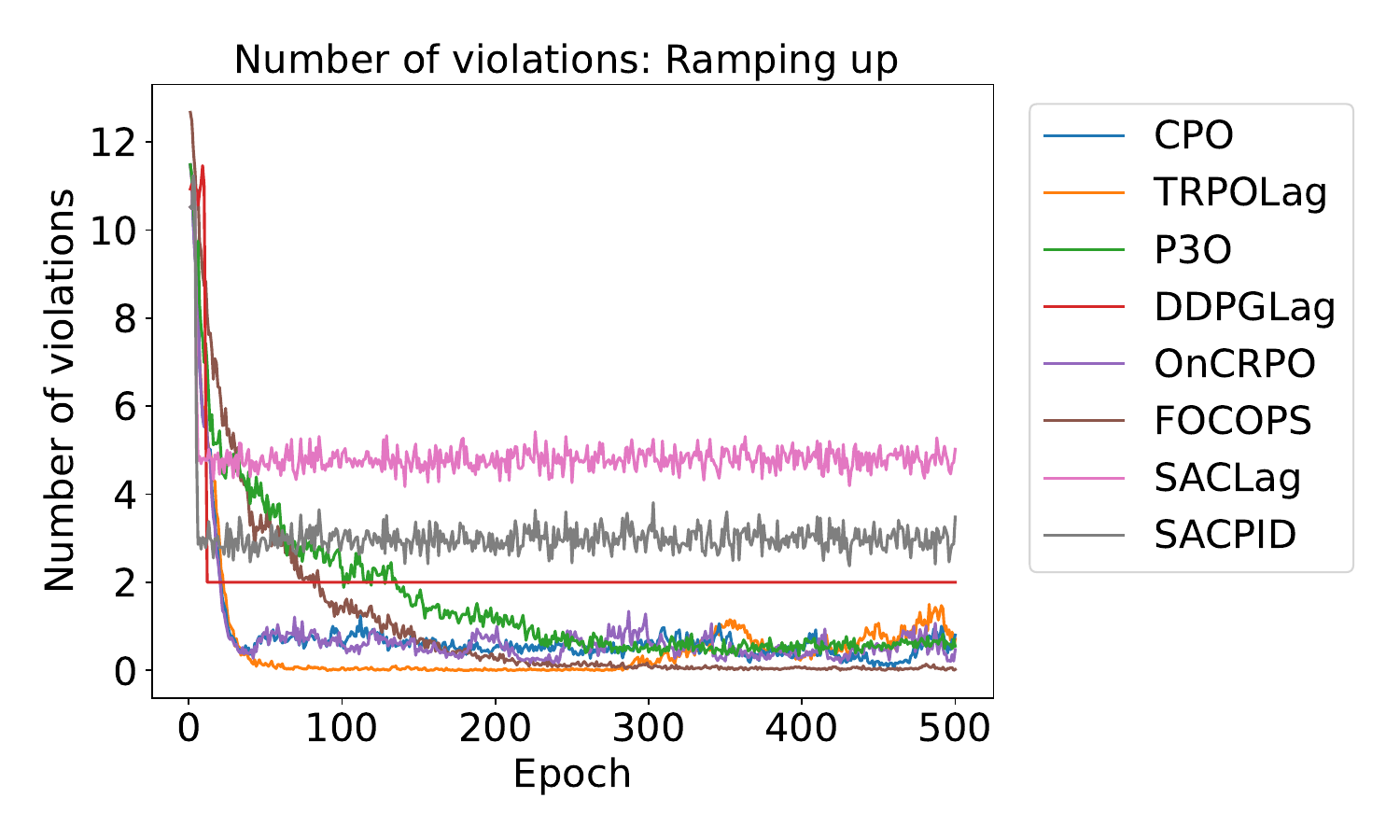}
    \end{subfigure}
    \begin{subfigure}[b]{0.45\textwidth}
        \centering
        \includegraphics[width=\textwidth]{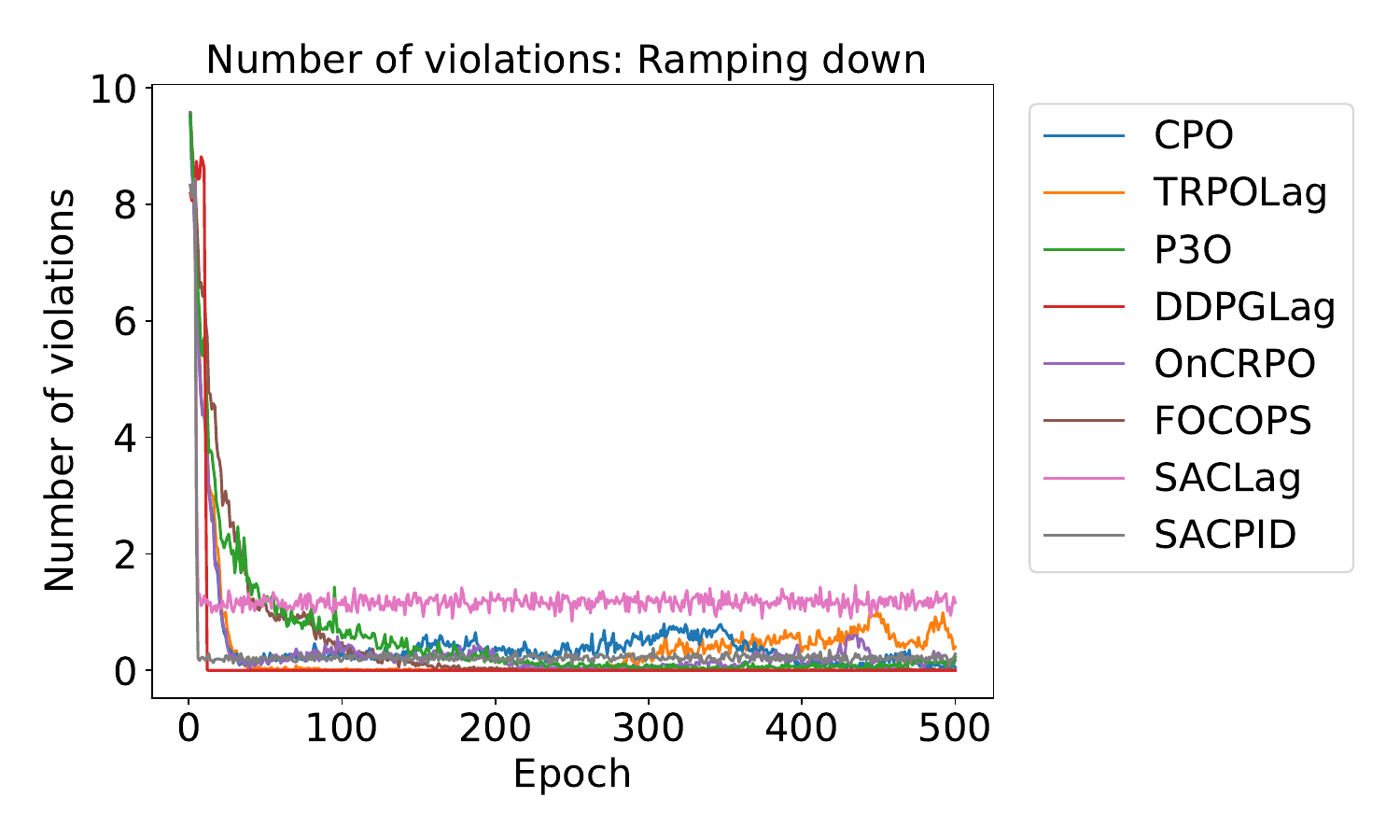}
    \end{subfigure}
    \caption{Average number of episode violations for different epochs for UCEnv-v1}
    \label{ucenv_2}
\end{figure}

\paragraph{Multiple-bus system with network constraints (UCEnv-v1):}
Figure~\ref{ucenv_2} illustrates mean episode-level breaches per epoch for the multi-bus UCEnv problem, incorporating \emph{minimum up-time}, \emph{minimum down-time}, \emph{ramping-rate}, and network feasibility constraints. The qualitative performance of the algorithms remains the same as described for UCEnv-v0.

\subsubsection{GTEPEnv}

\begin{figure}[H]
    \centering
    \begin{subfigure}[b]{0.45\textwidth}
        \centering
        \includegraphics[width=\textwidth]{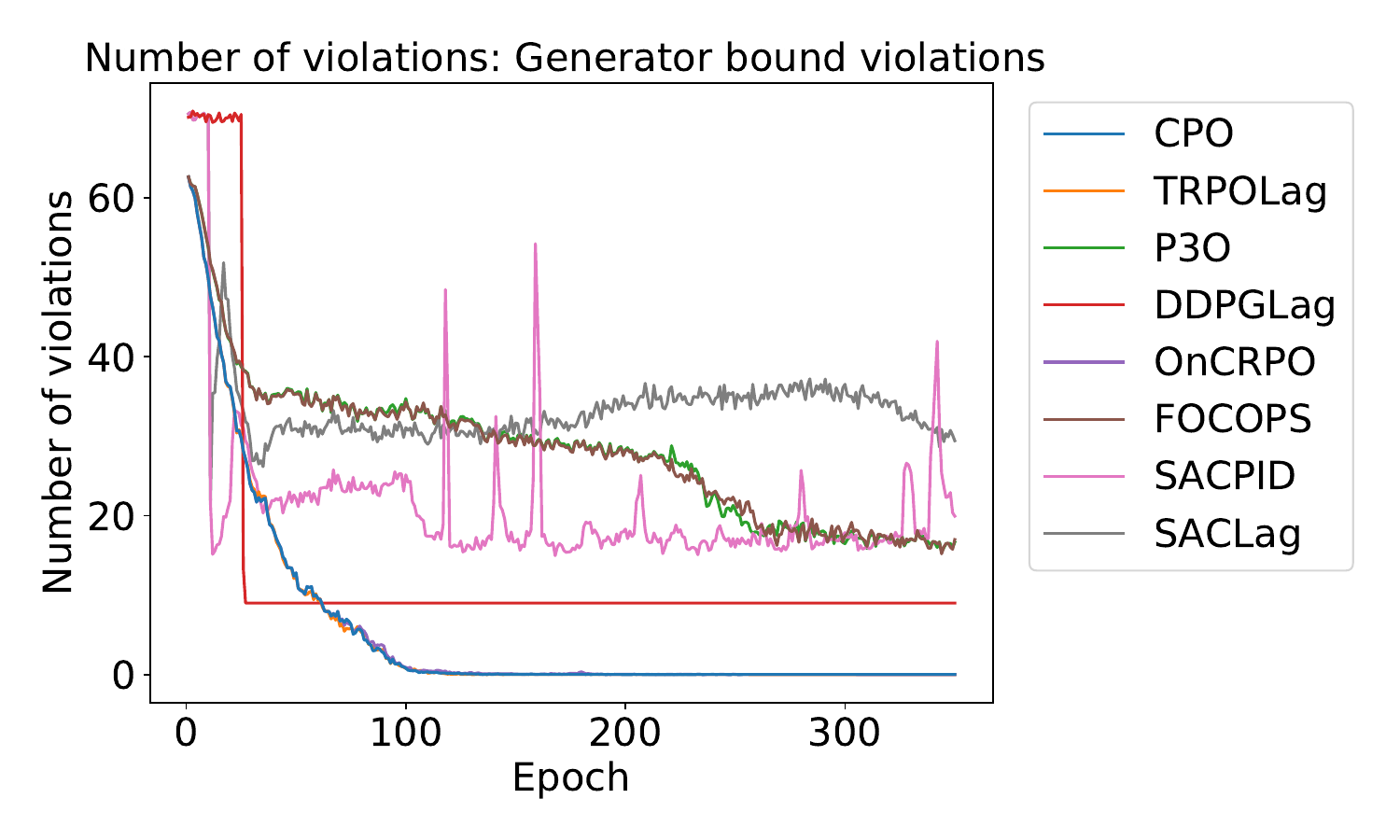}
    \end{subfigure}
    \hfill
    \begin{subfigure}[b]{0.45\textwidth}
        \centering
        \includegraphics[width=\textwidth]{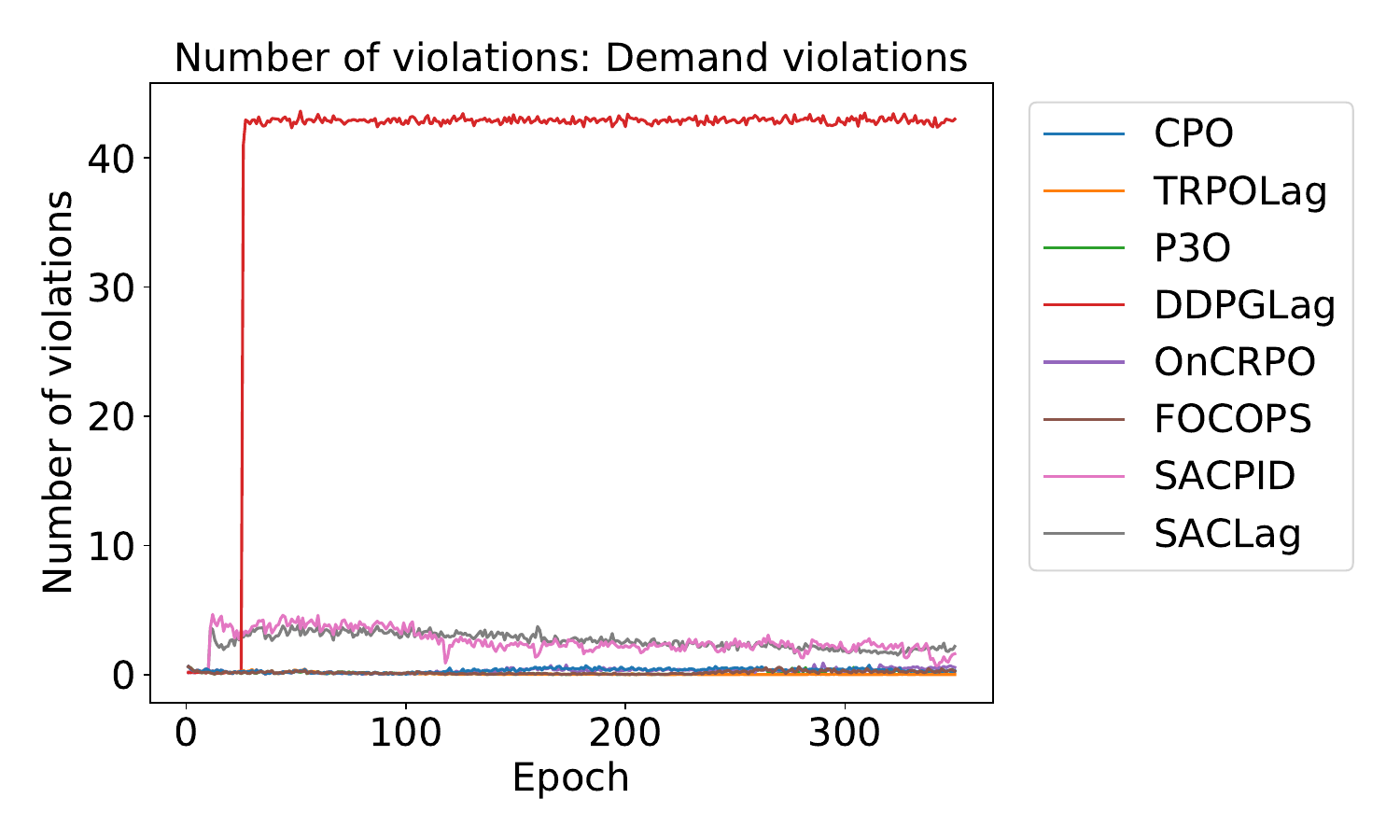}
    \end{subfigure}
    \caption{Average number of episode violations for different epochs for GTEPEnv }
    \label{gtep_1}
\end{figure}

Figure~\ref{gtep_1} illustrates mean episode-level constraint violations per epoch for the generation and transmission expansion planning task. The projection-based methods, \textbf{CPO}, \textbf{OnCRPO}, and \textbf{TRPOLag}, rapidly reduce generator-bound violations while keeping demand violations low, highlighting their effectiveness in per-update safety enforcement. \textbf{P3O} and \textbf{FOCOPS} achieve comparable compliance more gradually, converging to a non-zero level of bound violations. In contrast, \textbf{SACPID} and \textbf{SACLag} exhibit an initial reduction in generator-bound violations but persistently exceed the bounds thereafter, though they maintain relatively few demand violations. Finally, \textbf{DDPGLag} sustains non-zero bound violations and shows a persistent demand violation.

% \paragraph{GTEPEnv with transmission lines:}
% Figure~\ref{gtep_2} expands analysis to include active transmission-capacity constraints within a multi-bus network. Similar to the single-bus scenario, \textbf{CPO}, \textbf{OnCRPO}, and \textbf{TRPOLag} rapidly and effectively reduce generator-bound, demand-balance, and action-range violations. \textbf{P3O} follows this pattern more slowly, and \textbf{DDPGLag} maintains a modest yet persistent violation level, particularly notable in demand imbalance. For transmission constraints, projection and trust-region algorithms quickly achieve feasibility upon stabilizing generator dispatch, whereas \textbf{P3O} experiences brief overflow episodes before compliance, and \textbf{DDPGLag} demonstrates intermittent constraint breaches. Thus, network constraints reinforce the established safety hierarchy: per-update projection methods lead, adaptive penalties follow, and off-policy methods exhibit residual infractions.

% \begin{figure}[H]
%     \centering
%     \begin{subfigure}[b]{0.45\textwidth}
%         \centering
%         \includegraphics[width=\textwidth]{Images/Images_Asha/GTEP_gen_con_1.pdf}
%     \end{subfigure}
%     \hfill
%     \begin{subfigure}[b]{0.45\textwidth}
%         \centering
%         \includegraphics[width=\textwidth]{Images/Images_Asha/GTEP_gen_con_2.pdf}
%     \end{subfigure}
%     \caption{Average number of episode violations for different epochs for GTEPEnv with transmission lines}
%     \label{gtep_2}
% \end{figure}

 \subsubsection{BlendingEnv}
 
\begin{figure}[H]
    \centering
    \begin{subfigure}[b]{0.45\textwidth}
        \centering
        \includegraphics[width=\textwidth]{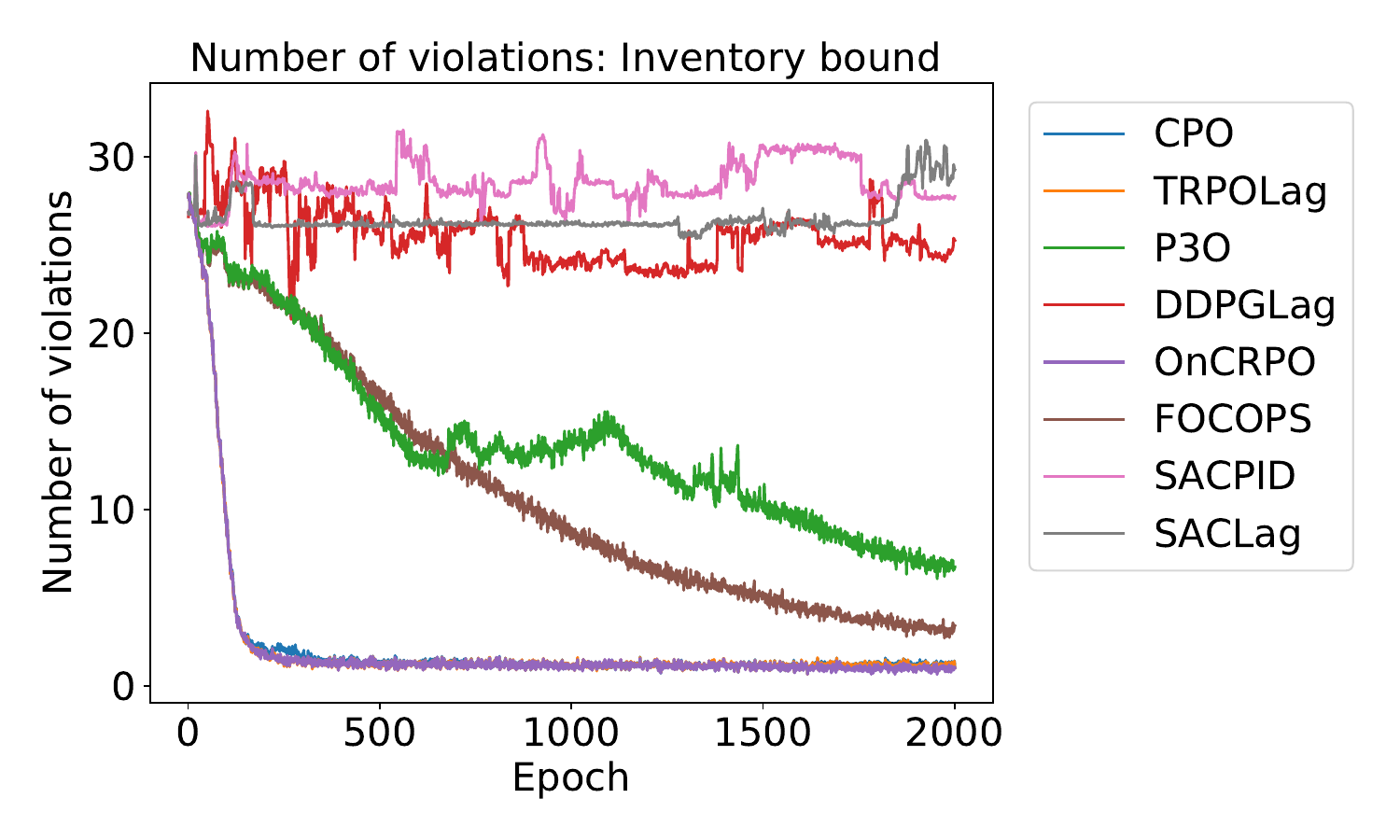}
    \end{subfigure}
    \hfill
    \begin{subfigure}[b]{0.45\textwidth}
        \centering
        \includegraphics[width=\textwidth]{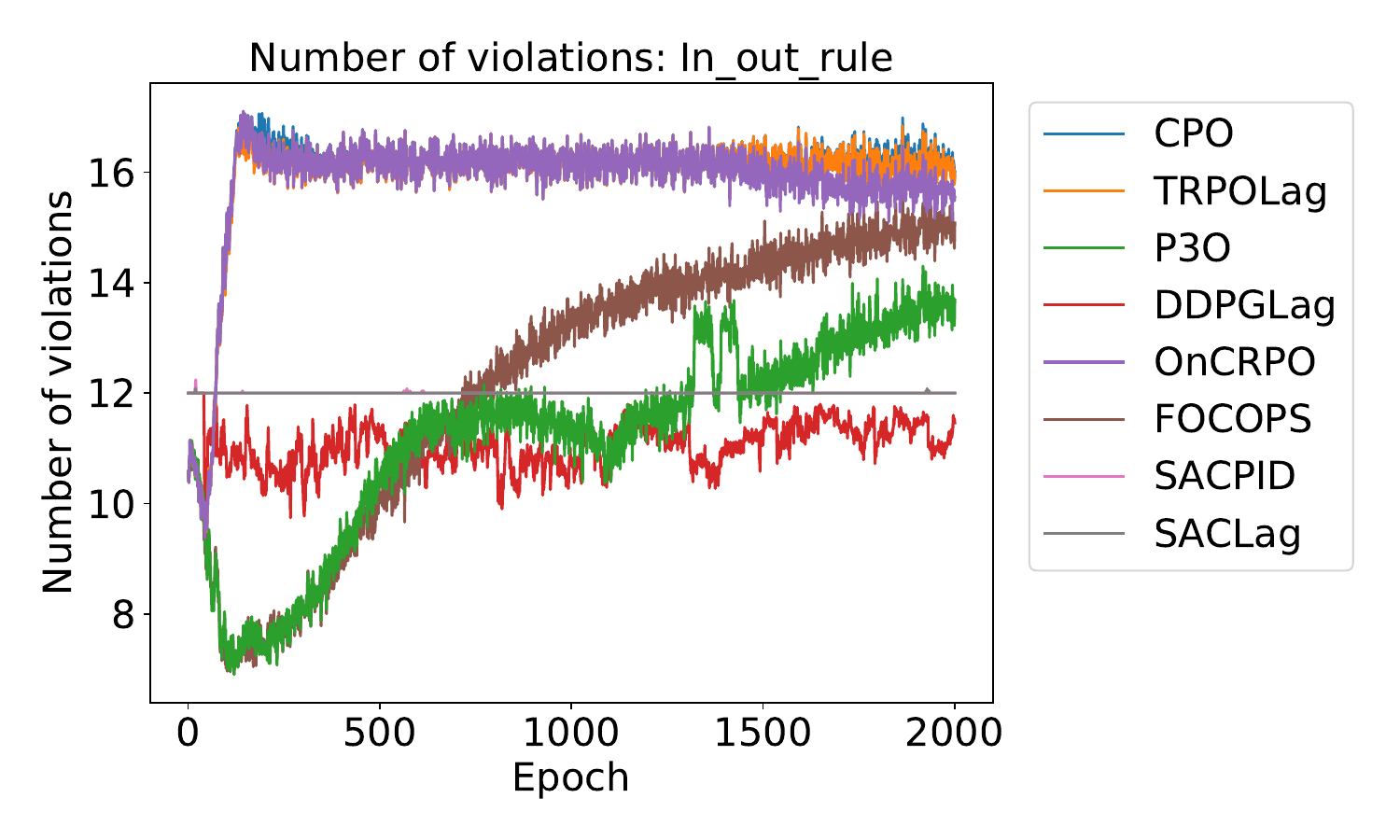}
    \end{subfigure}
   \hfill
    \begin{subfigure}[b]{0.45\textwidth}
        \centering
        \includegraphics[width=\textwidth]{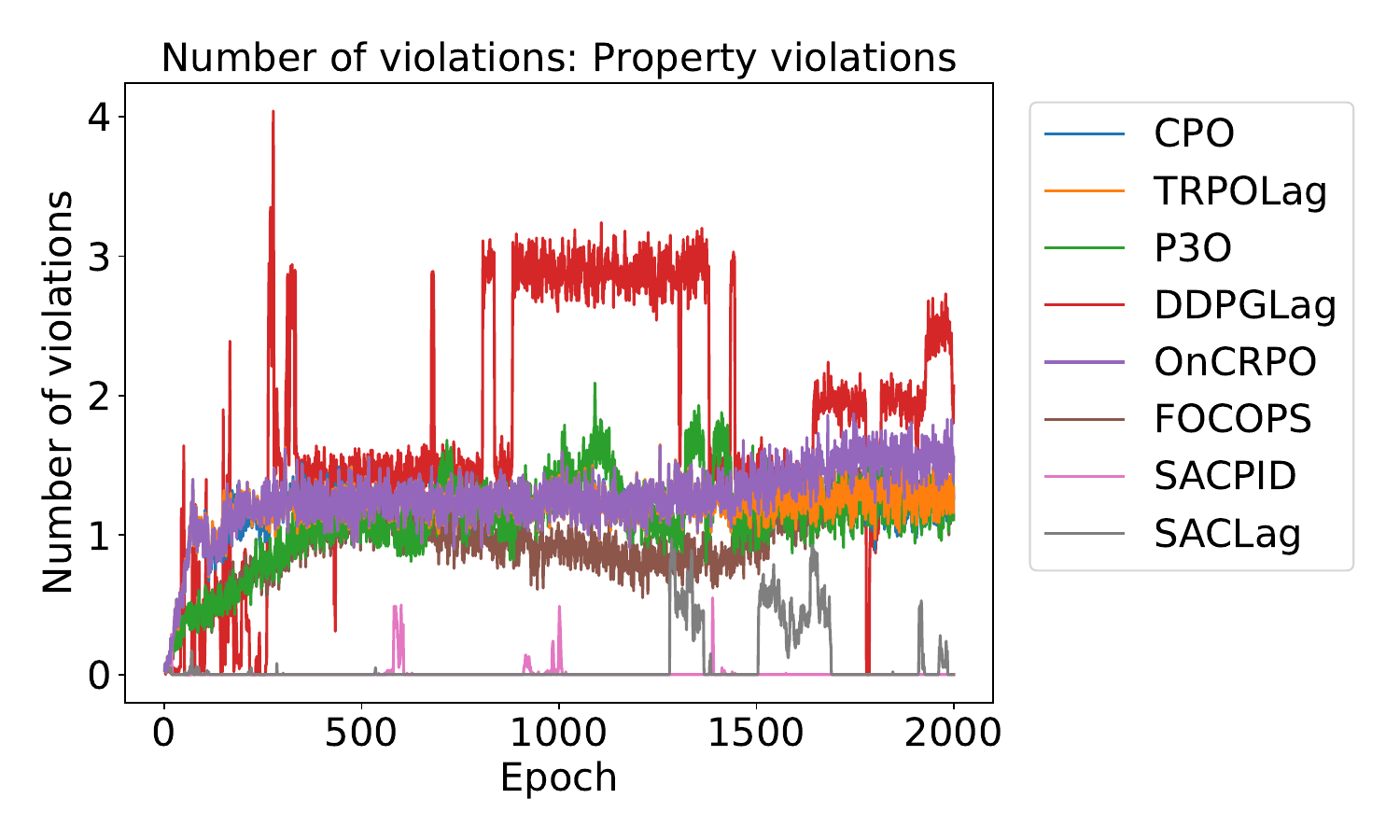}
    \end{subfigure}
    \caption{Average number of episode violations for different epochs for BlendingEnv with prop strategy}
    \label{blend:prop}
\end{figure}

Figure~\ref{blend:prop} shows the mean episode-level constraint violations per epoch during training for the Blending environment. For inventory bound violations, \textbf{CPO}, \textbf{TRPOLag}, and \textbf{OnCRPO} rapidly converge to zero violations. \textbf{P3O} and \textbf{FOCOPS} exhibits a gradual decline with minor oscillations before stabilizing at a small but non-zero level. \textbf{DDPGLag}, \textbf{SACPID} and \textbf{SACLag} consistently oscillates without clear convergence. In contrast, for the in-out rule violations, \textbf{CPO}, \textbf{TRPOLag}, and \textbf{OnCRPO} show significant increases, stabilizing at substantial violation levels. \textbf{P3O} and \textbf{FOCOPS} displays gradual increases with \textbf{P3O} showing pronounced occilations. \textbf{DDPGLag} maintains persistent oscillations around a steady level. \textbf{SACPID} and \textbf{SACLag} maintain a steady level with very occasional osccilations. Property violations slightly increase over time for \textbf{P3O}, \textbf{CPO}, \textbf{TRPOLag},  \textbf{OnCRPO} and \textbf{FOCOPS} with minor fluctuations, whereas \textbf{DDPGLag} demonstrates clear oscillations with an upward trend. In contrast, \textbf{SACPID} and \textbf{SACLag} osccilates around 0.

 \subsubsection{InvMgmtEnv}

\begin{figure}[H]
    \centering
    \begin{subfigure}[b]{0.45\textwidth}
        \centering
        \includegraphics[width=\textwidth]{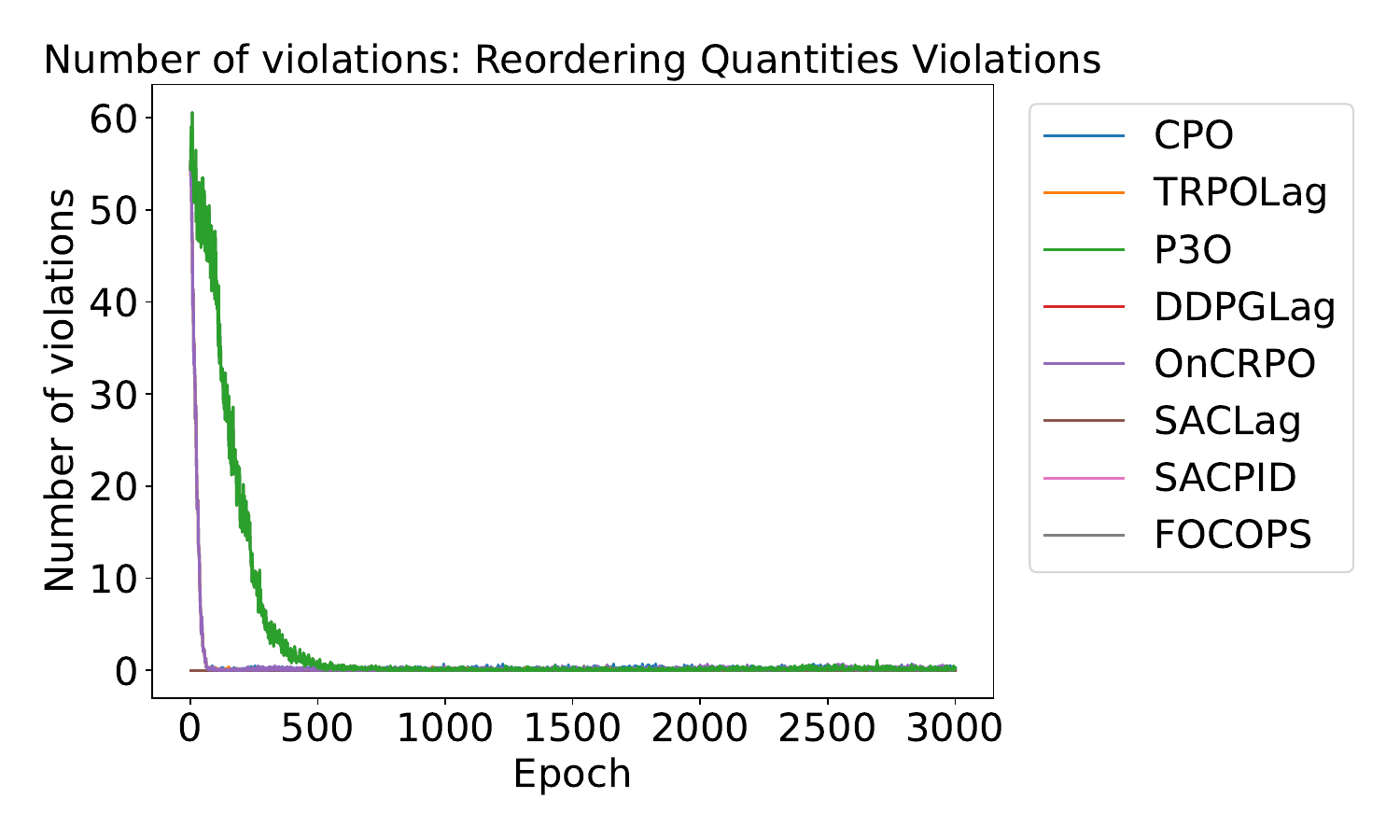}
    \end{subfigure}
    \caption{Average number of episode violations for different epochs for InvMgmtEnv}
    \label{invmgmtenv_1}
\end{figure}

Figure~\ref{invmgmtenv_1} plots episode-level reordering-quantity bound violations per epoch in InvMgmtEnv across several methods.

InvMgmtEnv includes bounds constraints on reordering quantities, on-hand inventory, pipeline inventory, backlog, and sales. The on-policy convex optimization method CPO and primal method OnCRPO, along with the on-policy primal-dual method TRPOLag, reduce violations rapidly and sustain low levels; the on-policy penalty function method P3O follows a similar path but needs a few extra epochs to recover from early spikes. The off-policy primal-dual methods DDPGLag and SACLag, the off-policy PID-based methods SACPID, and the on-policy convex optimization method FOCOPS also trend downward over training and exhibit zero to near-zero violations. All methods achieve near-zero violations for the on-hand inventory, pipeline inventory, sales, and backlog bounds constraints.
\subsubsection{GridStorageEnv}

\begin{figure}[H]
    \centering
    \begin{subfigure}[b]{0.45\textwidth}
        \centering
        \includegraphics[width=\textwidth]{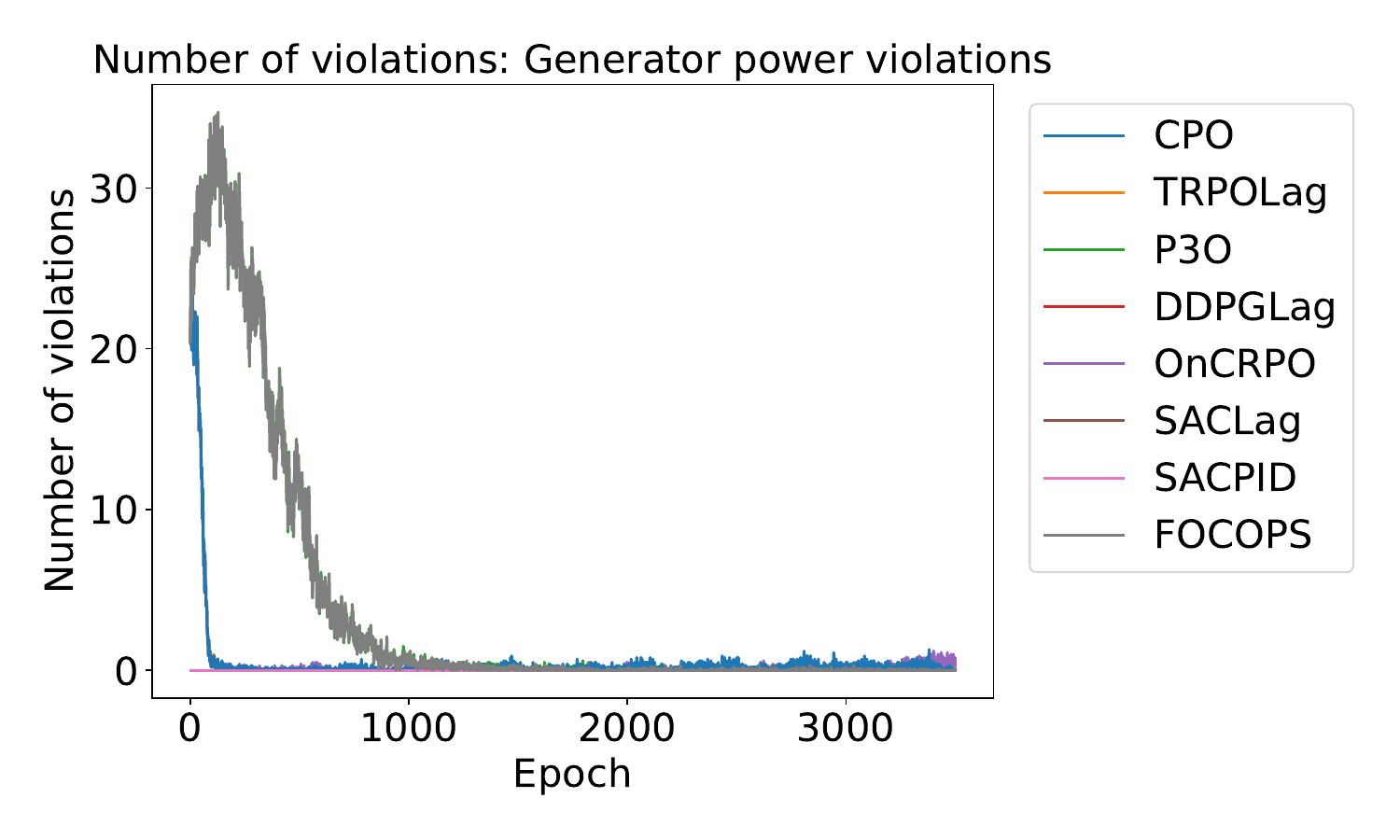}
    \end{subfigure}
    \hfill
    \begin{subfigure}[b]{0.45\textwidth}
        \centering
        \includegraphics[width=\textwidth]{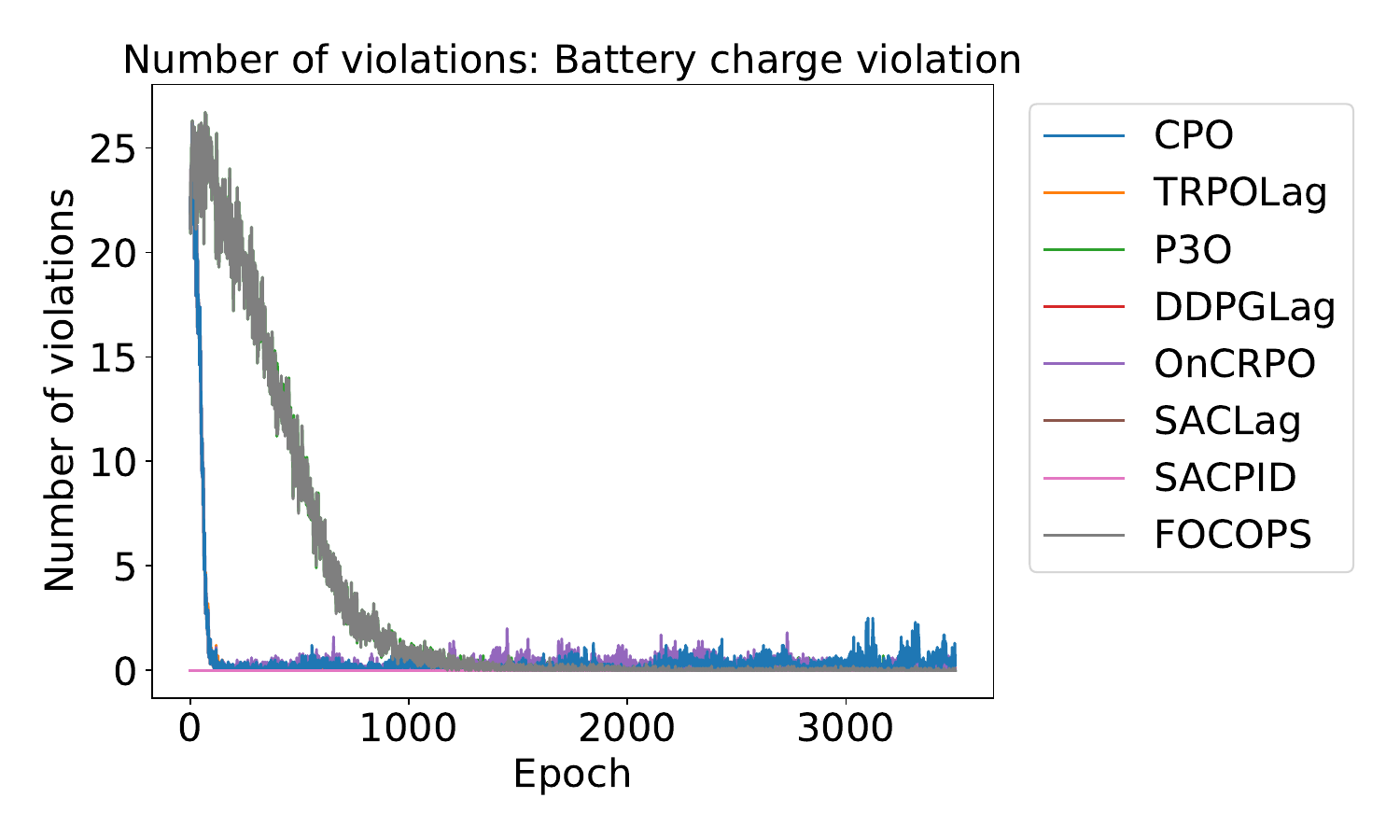}
    \end{subfigure}
        \begin{subfigure}[b]{0.45\textwidth}
        \centering
        \includegraphics[width=\textwidth]{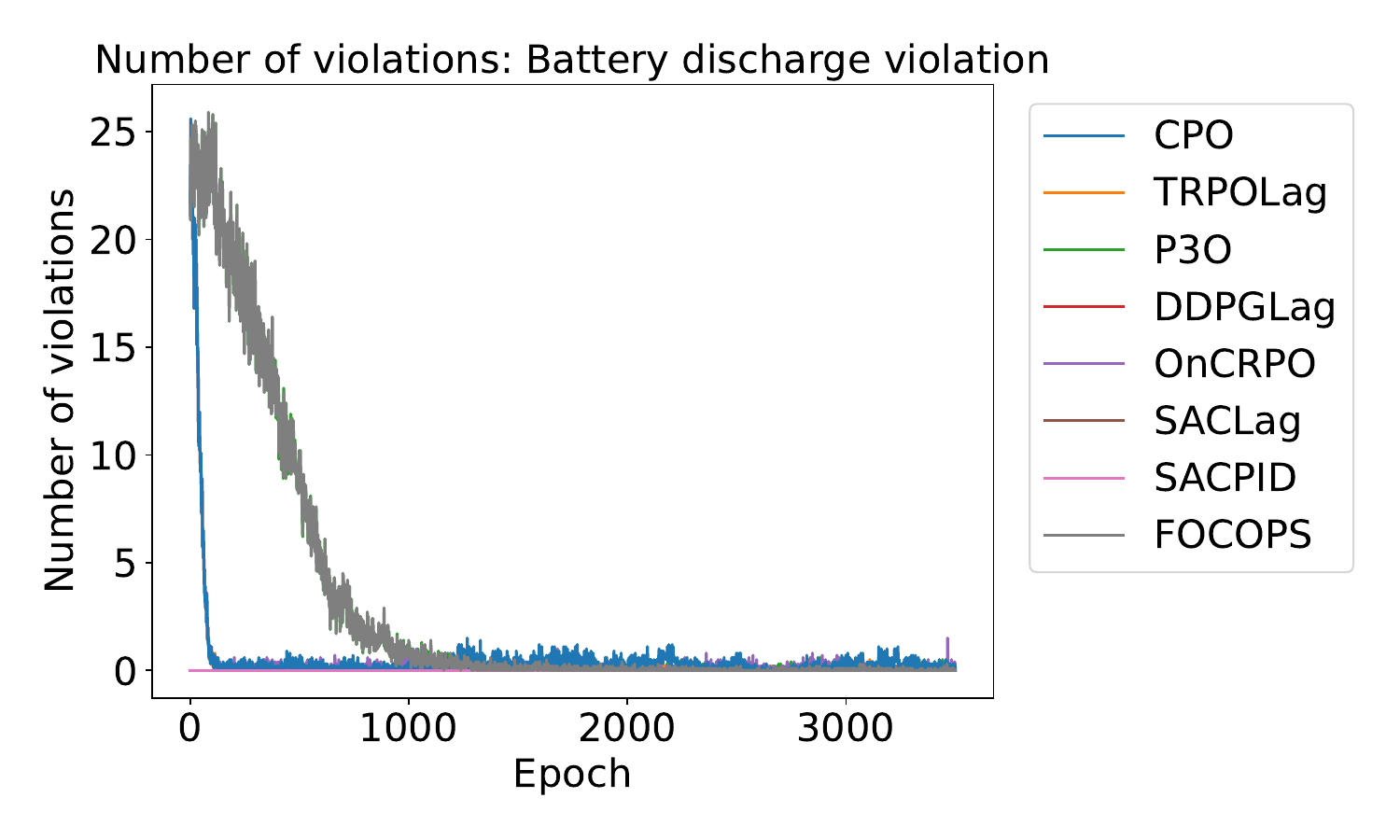}
    \end{subfigure}
    \begin{subfigure}[b]{0.45\textwidth}
        \centering
        \includegraphics[width=\textwidth]{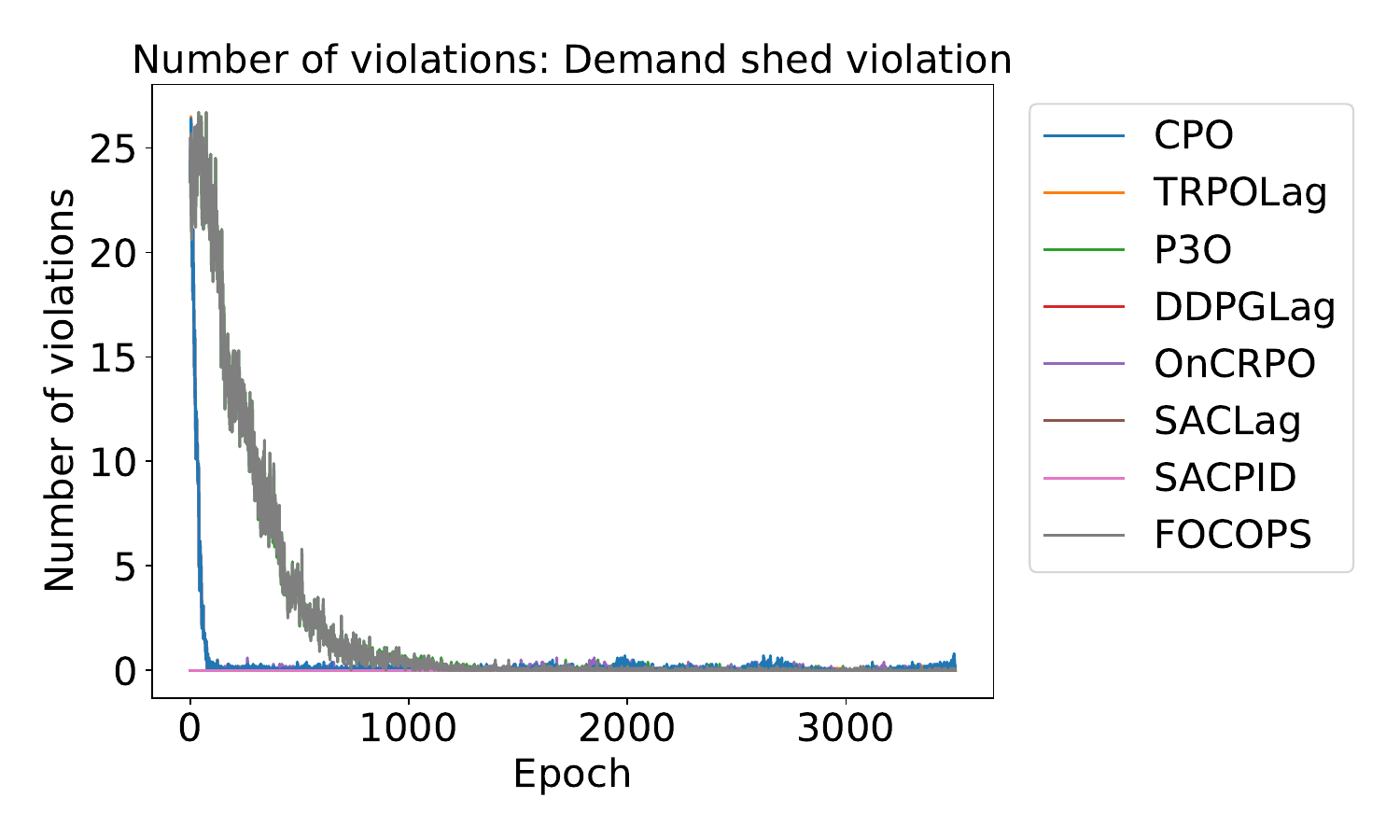}
    \end{subfigure}
    \begin{subfigure}[b]{0.45\textwidth}
        \centering
        \includegraphics[width=\textwidth]{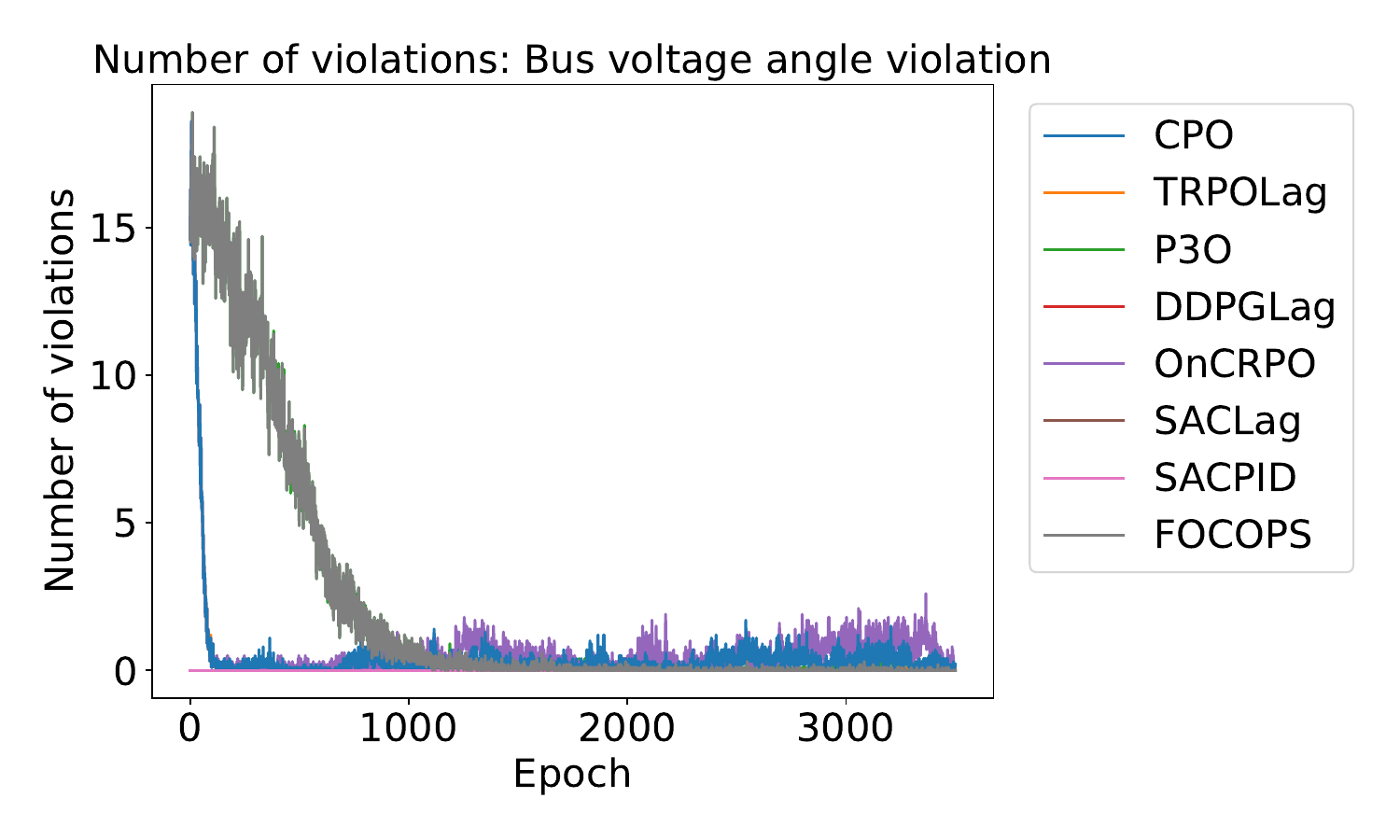}
    \end{subfigure}
    \caption{Average number of episode violations for different epochs for GridStorageEnv}
    \label{gridstorageenv_1}
\end{figure}

Figure~\ref{gridstorageenv_1} illustrates the mean episode-level violations per epoch for the GridStorageEnv, which includes bounds constraints on \emph{generator power limits}, \emph{battery charge rates}, \emph{battery discharge rates}, \emph{load shedding}, \emph{bus-angle bounds}, \emph{battery state of charge (SOC)}, and the \emph{slack-bus angle}. 

The on-policy convex optimization method CPO, primal method OnCRPO, and primal-dual method TRPOLag rapidly mitigate violations and quickly stabilize compliance; the on-policy penalty function method P3O converges more slowly, with temporary mid-training peaks before aligning with the leading methods. Among the other methods, apart from the on-policy convex optimization method FOCOPS, which in the initial stages shows an increasing trend in the number of violations, the off-policy primal-dual methods DDPGLag and SACLag, along with the off-policy PID-based method SACPID, remain consistently at (or near) zero across all violations. Violations for battery SOC and slack-bus voltage angles remain consistently at zero across all methods, indicating complete compliance from the outset.

\subsubsection{SchedMaintEnv}

\paragraph{Deterministic integrated scheduling and maintainence environemnt (SchedMaintEnv-v0):}

\begin{figure}[H]
    \centering
    \begin{subfigure}[b]{0.45\textwidth}
        \centering
        \includegraphics[width=\textwidth]{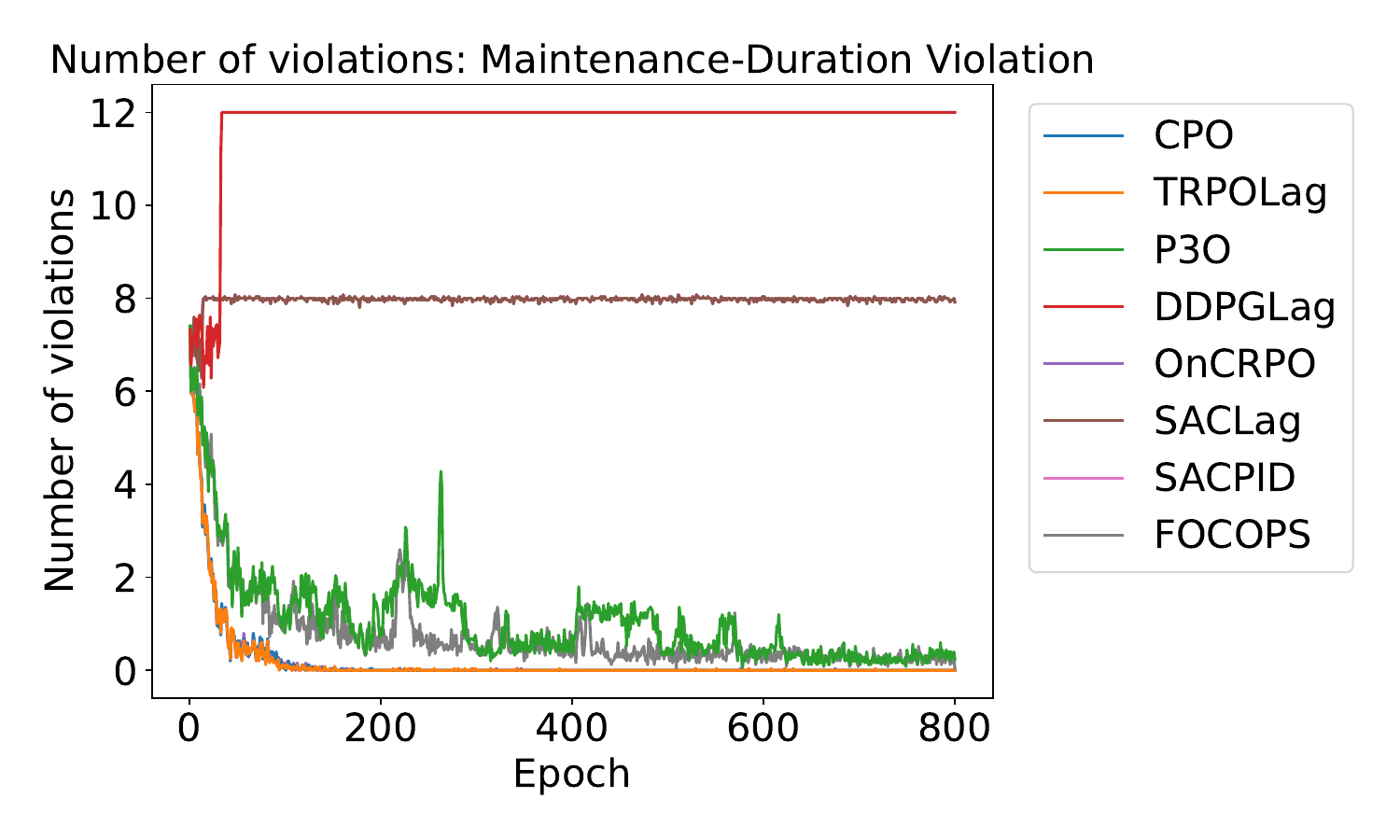}
    \end{subfigure}
    \hfill
    \begin{subfigure}[b]{0.45\textwidth}
        \centering
        \includegraphics[width=\textwidth]{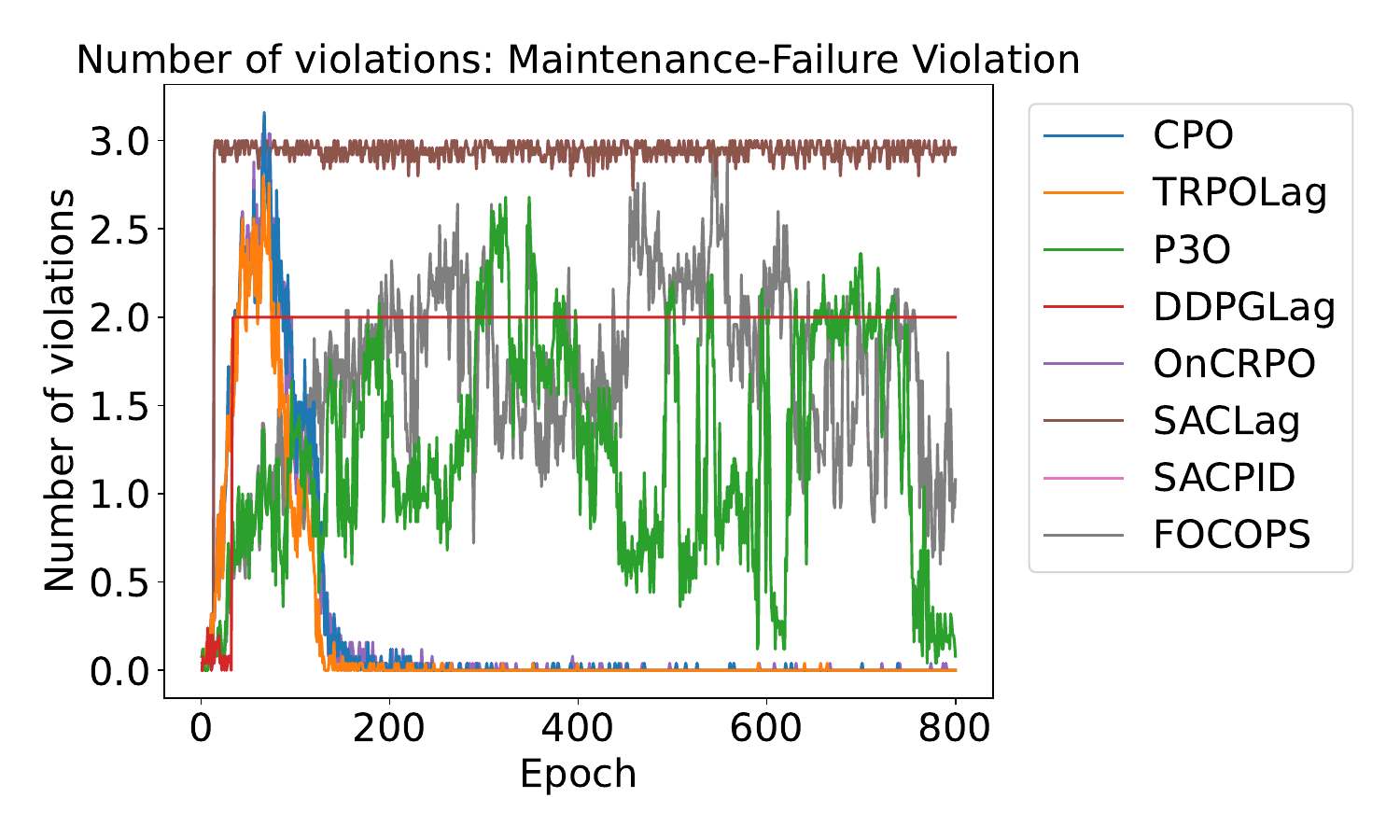}
    \end{subfigure}
        \begin{subfigure}[b]{0.45\textwidth}
        \centering
        \includegraphics[width=\textwidth]{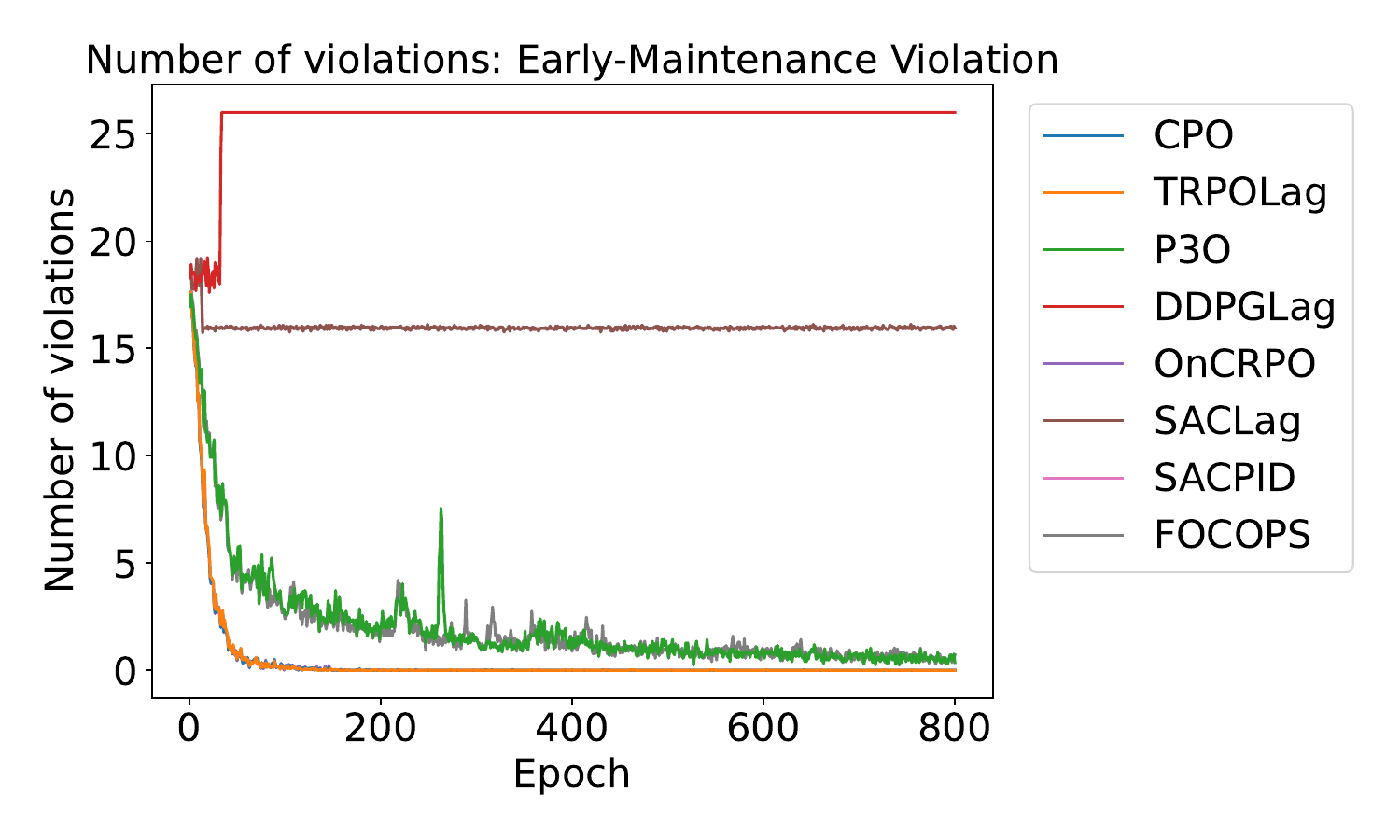}
    \end{subfigure}
    \begin{subfigure}[b]{0.45\textwidth}
        \centering
        \includegraphics[width=\textwidth]{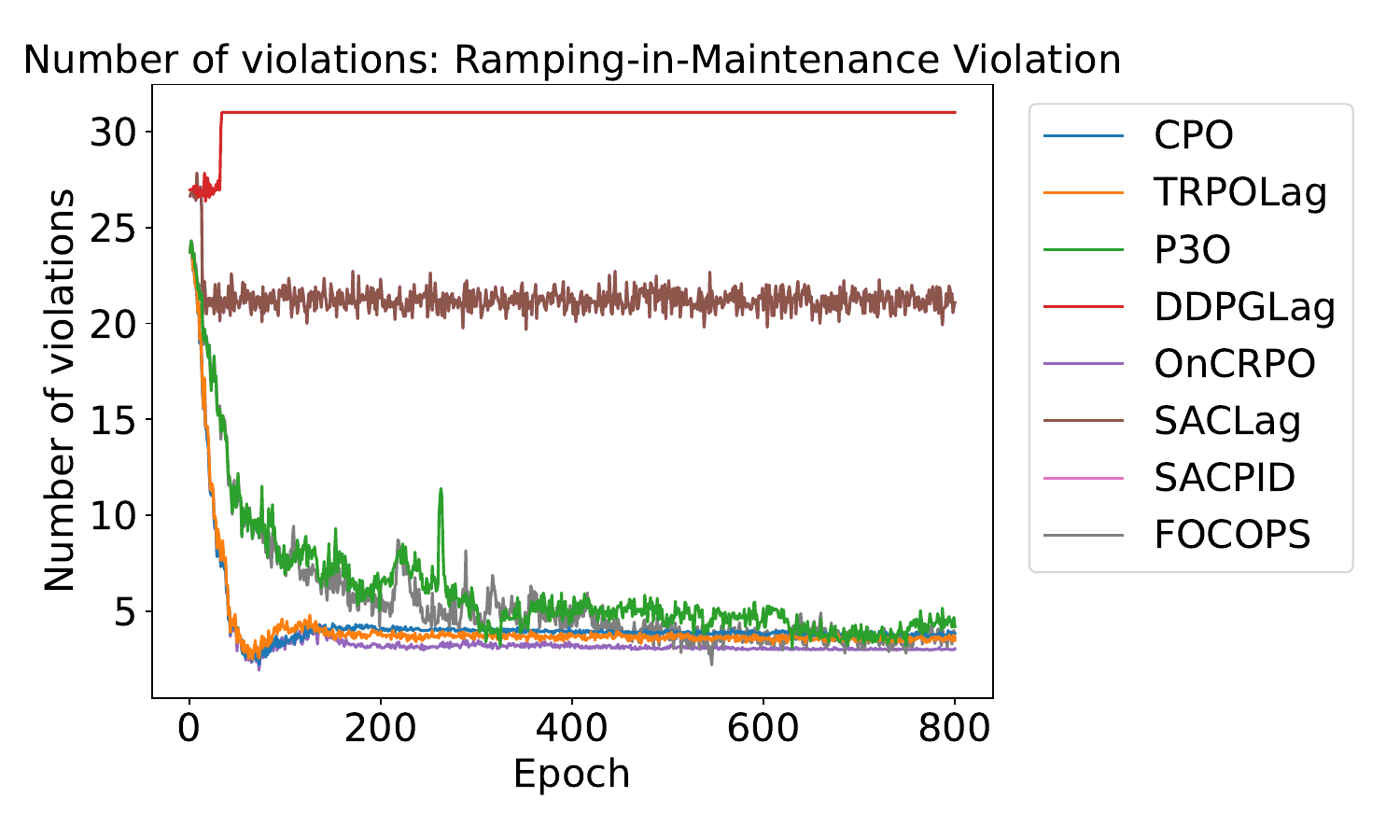}
    \end{subfigure}
    \begin{subfigure}[b]{0.45\textwidth}
        \centering
        \includegraphics[width=\textwidth]{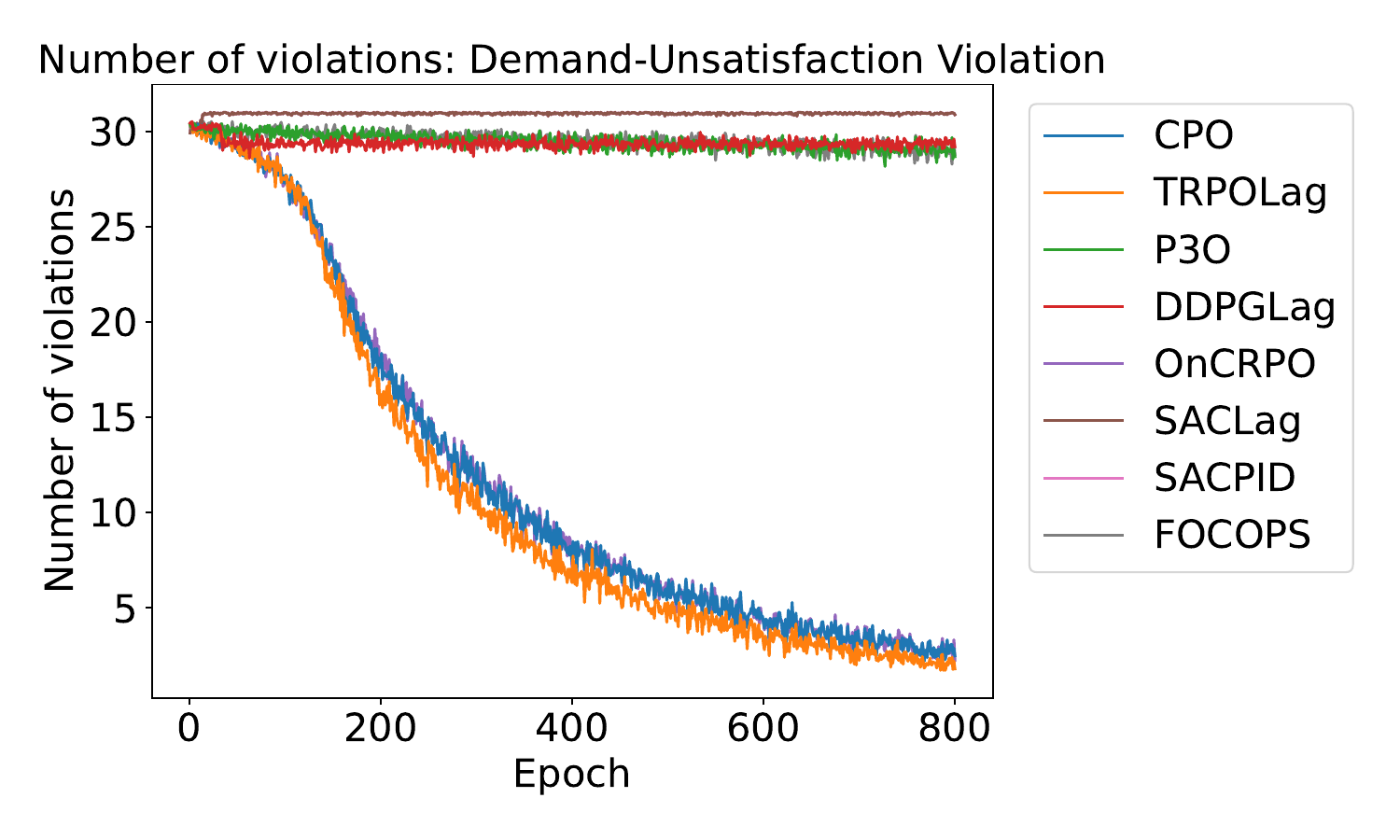}
    \end{subfigure}
    \caption{Average number of episode violations for different epochs for SchedMaintEnv-v0}
    \label{SchedMaintEnv_v0}
\end{figure}

Figure~\ref{SchedMaintEnv_v0} illustrates mean episode-level violations per epoch for the Integrated Scheduling \& Maintenance benchmark, covering constraints on \emph{maintenance-duration}, \emph{maintenance-failure}, \emph{early-maintenance}, \emph{ramping-in-maintenance}, and \emph{demand-unsatisfaction}. Projection-based methods—\textbf{CPO}, \textbf{OnCRPO}, and \textbf{TRPOLag}—swiftly reduce violations across all constraints, establishing stable and minimal breach levels. \textbf{P3O} follows a similar trend but exhibits transient spikes in violations for almost all constraints during mid-training, except for demand-unsatisfaction, where it consistently fails to learn full compliance. For the other constraints, it eventually achieves violation levels comparable to the projection-based algorithms. \textbf{DDPGLag} consistently displays higher residual violations, notably in duration and demand-unsatisfaction categories, highlighting variability from its replay-buffer updates. Maintenance-failure violations remain consistently low for all methods. Among the remaining methods, \textbf{FOCOPS} performs on par with \textbf{P3O}. \textbf{SACLag} is only slightly better than \textbf{DDPGLag}, but it persistently yields the highest rates of maintenance failures and unmet demand. \textbf{SACPID} has perfomance similar to \textbf{SACLag} (overlapping in this case).

\paragraph{Integrated scheduling and maintainence environemnt with stochasticity (SchedMaintEnv-v1):}

\pagebreak

Figure~\ref{SchedMaintEnv_v1} illustrates the mean episode-level breaches per epoch for the stochastic variant of the integrated scheduling and maintenance environment. The figure captures all maintenance-, production-, and demand-related constraints, as in Figure~\ref{SchedMaintEnv_v0}. The qualitative performance of the algorithms remains consistent with that observed for SchedMaintEnv-v0.

% --- First part of the figure ---
\begin{figure}[H]
    \centering
    \begin{subfigure}[b]{0.45\textwidth}
        \includegraphics[width=\textwidth]{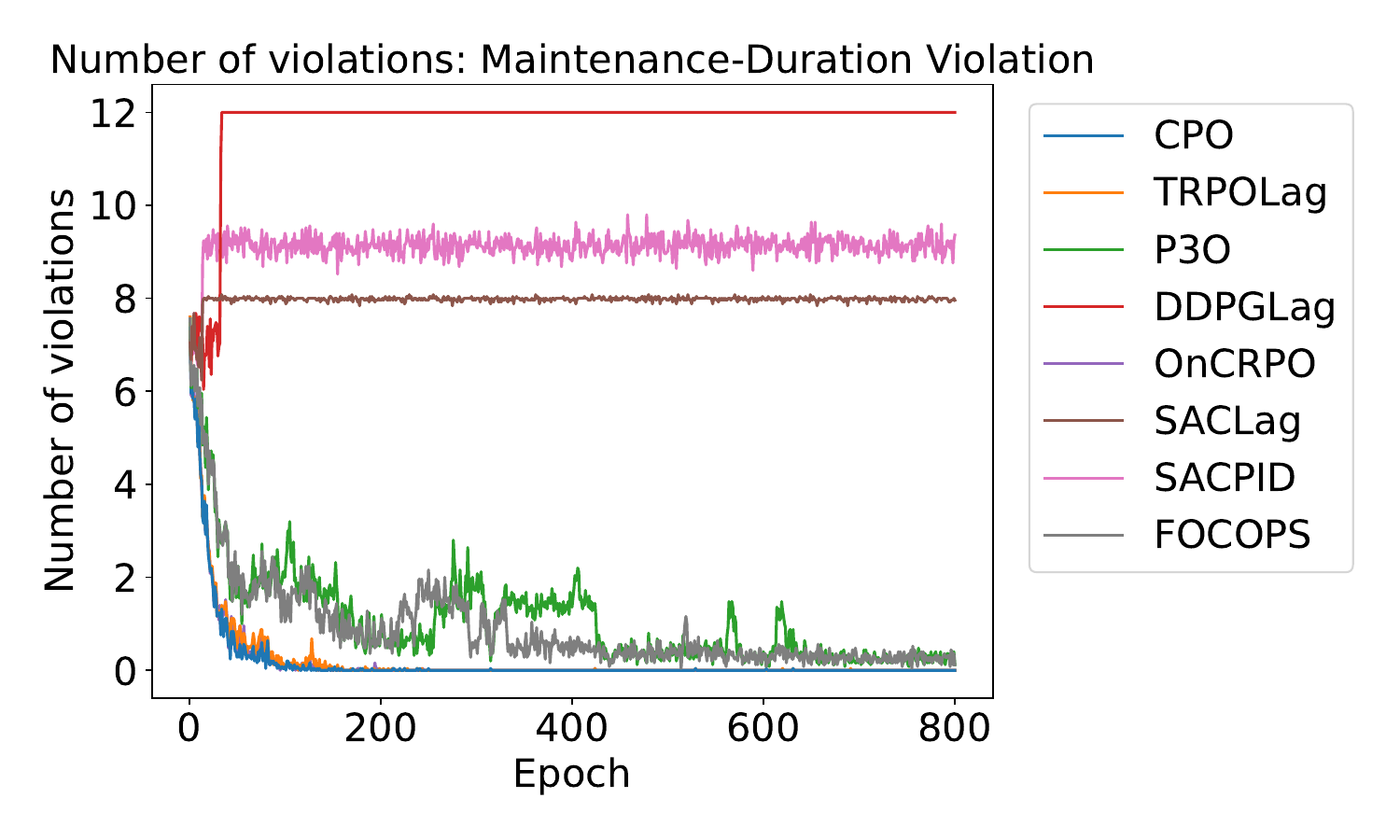}
    \end{subfigure}
    \hfill
    \begin{subfigure}[b]{0.45\textwidth}
        \includegraphics[width=\textwidth]{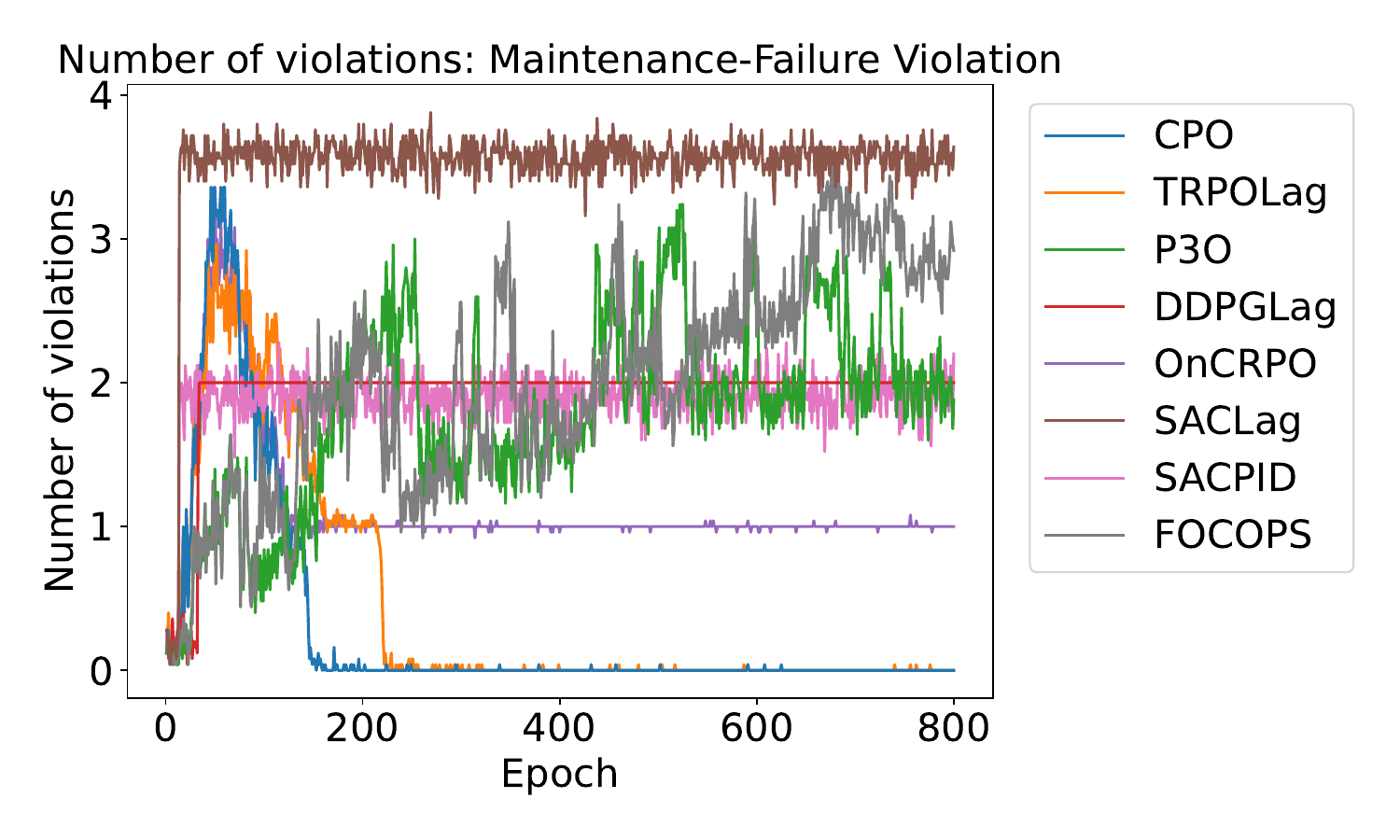}
    \end{subfigure}

    \vspace{1em}

    \begin{subfigure}[b]{0.45\textwidth}
        \includegraphics[width=\textwidth]{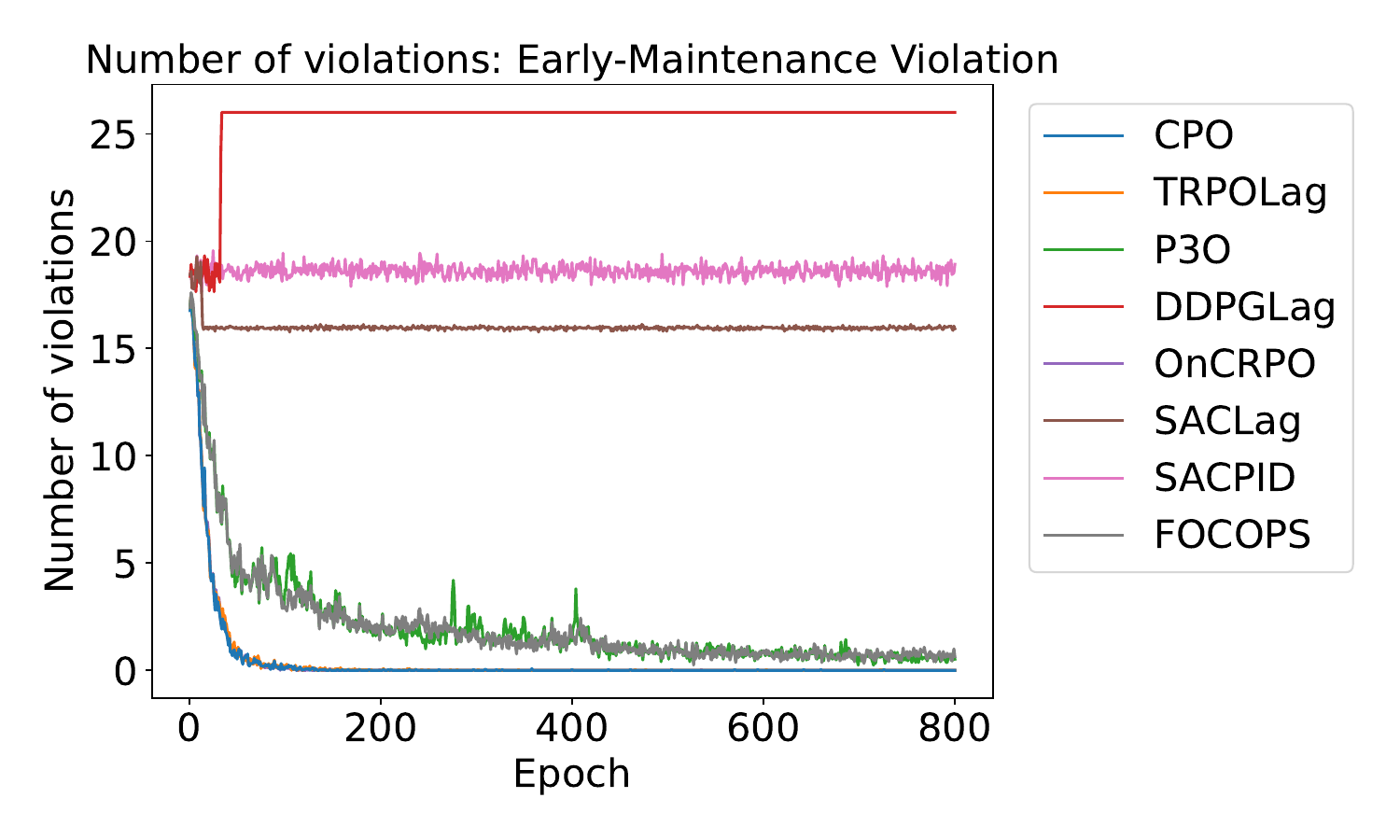}
    \end{subfigure}
    %\caption{Average number of episode violations for different epochs for \textbf{SchedMaintEnv-v1} (part 1).}
\end{figure}

% --- Second part of the figure (continued float) ---
\begin{figure}[H]\ContinuedFloat
    \centering
    \begin{subfigure}[b]{0.45\textwidth}
        \includegraphics[width=\textwidth]{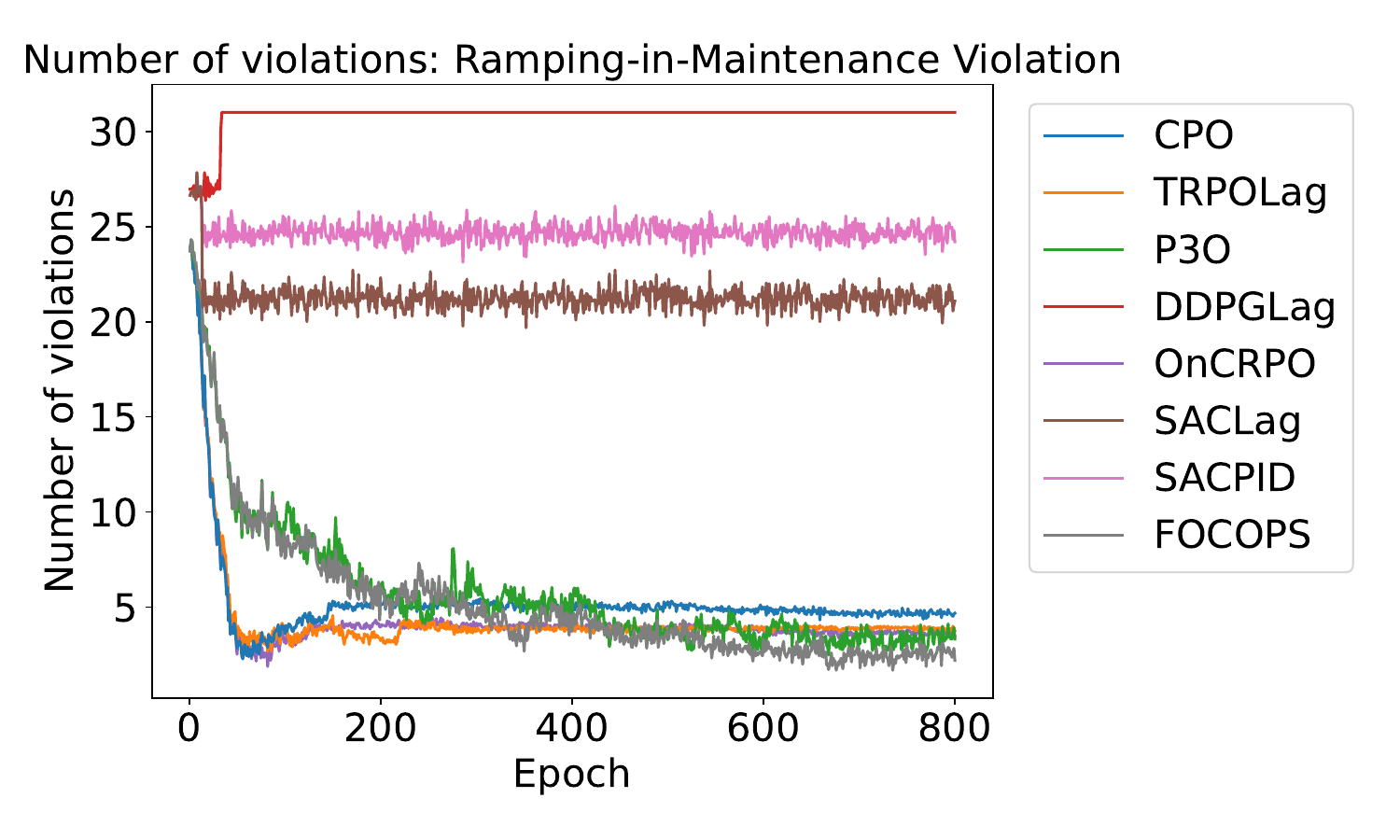}
    \end{subfigure}
    \hfill
    \begin{subfigure}[b]{0.45\textwidth}
        \includegraphics[width=\textwidth]{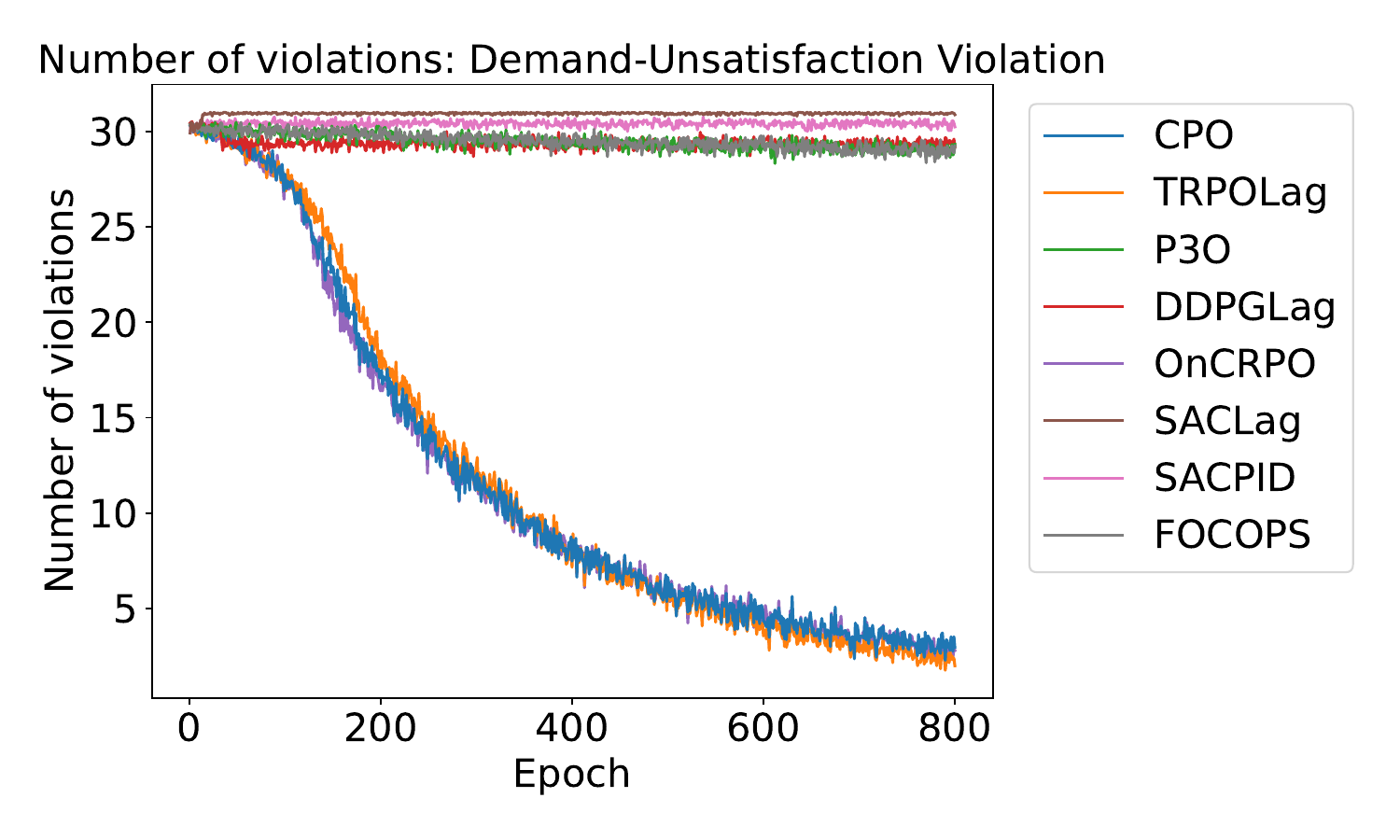}
    \end{subfigure}
    \caption{Average number of episode violations for different epochs for SchedMaintEnv-v1} 
    \label{SchedMaintEnv_v1}
\end{figure}

\subsubsection{ASUEnv}

% Figure~\ref{ASUEnv_1} shows the mean episode-level constraint violations per epoch in the Air Separation Unit environment, focusing on \emph{inventory limits} and \emph{demand satisfaction}. Projection-based methods—\textbf{CPO}, \textbf{OnCRPO}, and \textbf{TRPOLag}—quickly reduce inventory violations, reaching stable minimal levels. \textbf{P3O} exhibits a general downward trend in inventory violations but retains substantial residual violations by the end of training. \textbf{DDPGLag} maintains the highest violation counts, particularly for inventory constraints, consistent with the higher variance typical of off-policy algorithms. 

% Notably, demand violations emerge later in training—after approximately half the epochs—because the agent begins to encounter underproduction violations only once it has learned to avoid overproduction. Consequently, \textbf{P3O} and \textbf{DDPGLag}, which fail to sufficiently reduce inventory violations, rarely experience demand violations. In contrast, \textbf{CPO}, \textbf{TRPOLag}, and \textbf{OnCRPO} exhibit fewer than one demand violation episode on average per epoch. This underscores the importance of cost shaping to encourage the agent to explore all relevant cost signals early in training, enabling it to balance constraints effectively and converge to optimal policies.

\begin{figure}[H]
    \centering
    \begin{subfigure}[b]{0.45\textwidth}
        \centering
        \includegraphics[width=\textwidth]{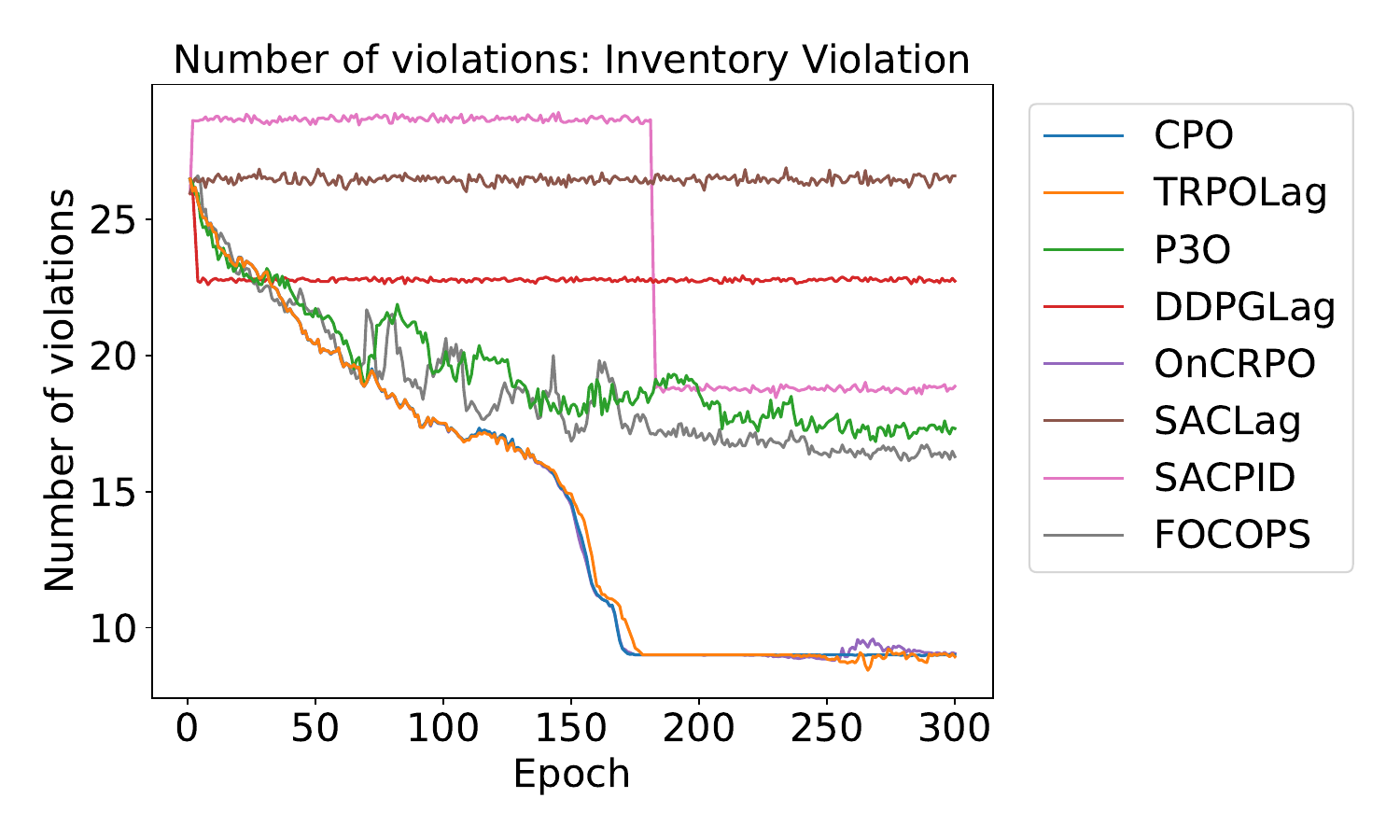}
    \end{subfigure}
    \hfill
    \begin{subfigure}[b]{0.45\textwidth}
        \centering
        \includegraphics[width=\textwidth]{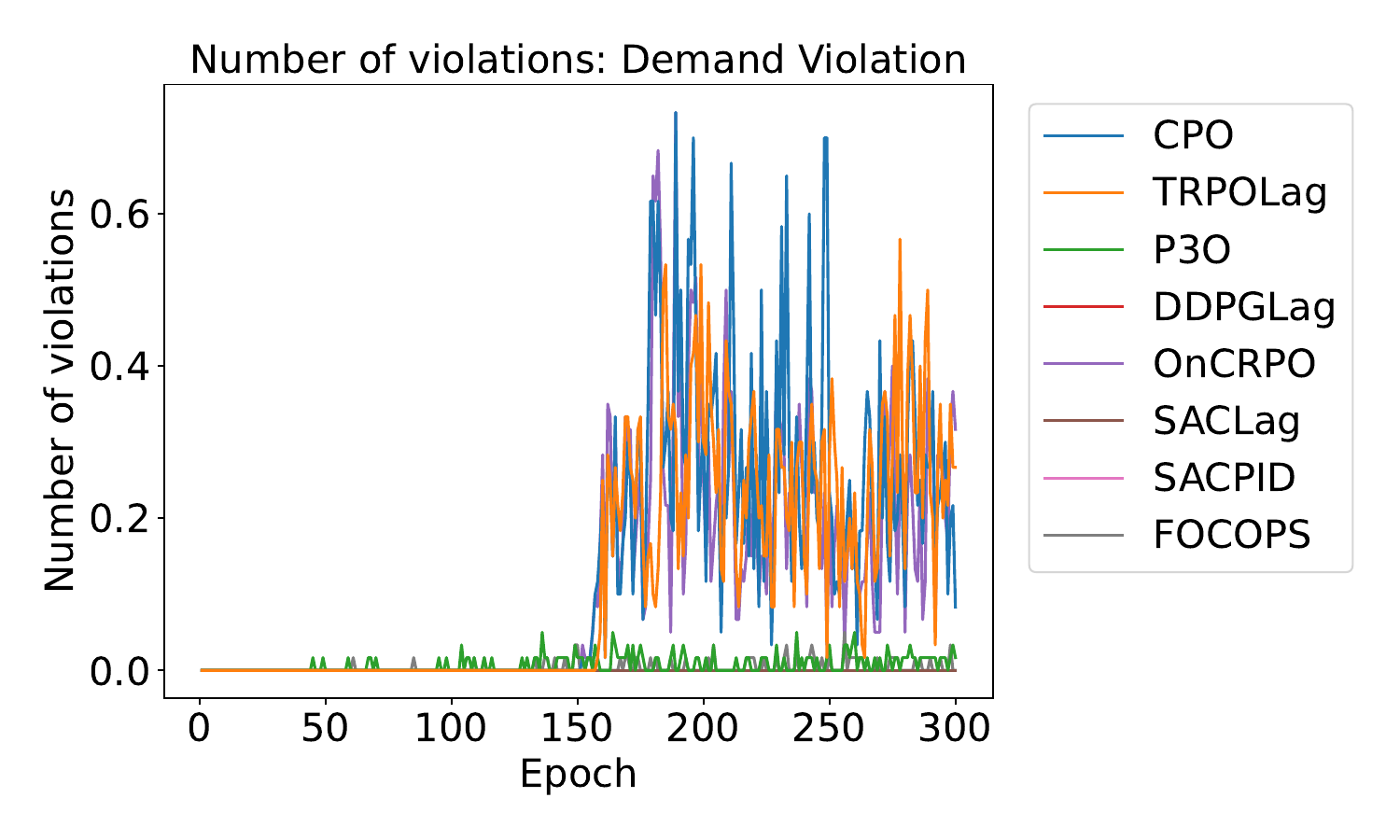}
    \end{subfigure}
    \caption{Average number of episode violations for different epochs for ASUEnv}
    \label{ASUEnv_1}
\end{figure}

Figure~\ref{ASUEnv_1} shows the mean episode-level constraint violations per epoch in the Air Separation Unit environment, focusing on \emph{inventory limits} and \emph{demand satisfaction}. Projection-based methods—\textbf{CPO}, \textbf{OnCRPO}, and \textbf{TRPOLag}—rapidly suppress inventory violations, converging to stable minimal levels (approximately $9$ episodes per epoch by around $170$ epochs). \textbf{P3O} and \textbf{FOCOPS} exhibit monotone declines but plateau with substantial residual violations (roughly $16$–$17$) by the end of training. Among off-policy methods, \textbf{DDPGLag} remains elevated (near $23$) throughout, \textbf{SACLag} is consistently the highest and flat (about $26$–$27$), and \textbf{SACPID} holds very high counts (around $28$) until a late drop near epoch $\sim 180$, after which it stabilizes at roughly $18$.

Demand violations emerge only in the latter half of training—after approximately $150$–$160$ epochs—because the agent begins to encounter underproduction violations once it has learned to avoid overproduction. Consequently, methods that tighten inventory most aggressively (\textbf{CPO}, \textbf{TRPOLag}, \textbf{OnCRPO}) exhibit sub-unit average demand-violation rates with intermittent spikes, whereas \textbf{P3O}, \textbf{FOCOPS}, \textbf{DDPGLag}, \textbf{SACLag}, and \textbf{SACPID} remain near zero—an artifact of maintaining slack inventories rather than superior constraint balancing. Overall, projection-based methods best reconcile the inventory–demand trade-off; off-policy methods—especially \textbf{SACLag} and \textbf{DDPGLag}—struggle to reduce inventory violations in \textsc{ASUEnv}, and \textbf{SACPID} improves late but does not match the leaders.

\subsection{Benchmarking Classical Rule-Based Policies in \texttt{InvMgmtEnv}}\label{sec:appendix:ssrq}

We evaluated two classical rule-based policies—\((s,S)\) and \((r,Q)\)—in \texttt{InvMgmtEnv}. Under an \((s,S)\) policy, an order is placed whenever the inventory position falls to or below a reorder point \(s\), and the order raises the position to a target level \(S\). Under an \((r,Q)\) policy, a fixed lot \(Q\) is ordered whenever the position falls to or below a threshold \(r\). We impose lower bounds \(s \ge s_{\min}\) and \(r \ge r_{\min}\) on these triggers; with \(s_{\min}=20\) the \((s,S)\) controller achieves a reward of \(7{,}610.42\), while with \(r_{\min}=20\) the \((r,Q)\) controller attains \(10{,}948.9\). With the optimal reward being \(11265.97493\); \((r,Q)\) lies within \(\approx 2.8\%\) of the optimum, whereas \((s,S)\) is \(\approx 32\%\) below. Relative to the learned SafeRL policies reported for this environment (e.g., ONCRPO \(\approx 7{,}599\), CPO \(\approx 7{,}303\), TRPOLag \(\approx 7{,}198\)), the \((s,S)\) baseline is comparable to the best-performing RL result, and the \((r,Q)\) baseline exceeds all RL methods by roughly \(40\%\!-\!50\%\). In our 10-episode evaluation for \texttt{InvMgmtEnv} (Table~\ref{table:evaluation}), \textsc{SACLag} achieves a reward of \(-14{,}386.99\) with cost \(0\), \textsc{SACPID} \(5{,}555.74\) with cost \(0\), and \textsc{FOCOPS} \(-6{,}434.03\) with cost \(0\); hence \((r,Q)\) improves upon \textsc{SACPID} by \(\approx 97\%\) and \((s,S)\) exceeds \textsc{SACPID} by \(\approx 37\%\), while both rule-based policies obtain higher rewards than \textsc{SACLag} and \textsc{FOCOPS}.

\subsection{Computational Time for Training Different Environments with Various Algorithms}

Table~\ref{tab:runtime} lists the wall-clock training time \textbf{(hours)} required by each algorithm on every environment.  
All experiments were run on an AWS \texttt{g4dn.xlarge} instance—4 vCPUs, 16 GB RAM, and a single NVIDIA T4 GPU.
\paragraph{Runtime trends across algorithms}
The projection–trust-region trio—\textbf{TRPO-Lag}, \textbf{On-CRPO}, and \textbf{CPO}—shows virtually identical runtimes, reflecting their shared on-policy update pattern with lightweight trust-region sub-problems.  
\textbf{P3O} matches this group closely, incurring only a modest overhead for its adaptive penalty update.  \textbf{FOCOPS} is also in this regime, with runtimes comparable to the projection methods thanks to its focused update structure that avoids heavy critic–actor coupling.   
By contrast, the off-policy \textbf{DDPG-Lag} is consistently the slowest: replay-buffer sampling, twin-critic evaluation, and deterministic actor updates roughly double the wall-clock time relative to the on-policy methods.  
\textbf{SAC-Lag} and \textbf{SAC-PID} exhibit similar off-policy behaviour, with entropy-regularised objectives and adaptive dual updates that inflate runtime beyond DDPG-Lag in some environments, though SAC-PID generally yields more stable trajectories at the cost of additional computation.  
Overall, the results indicate that enforcing safety through on-policy projections or dual updates delivers both strong constraint compliance and favourable computational efficiency, whereas off-policy dual learning trades longer runtimes for greater sample reuse.

\begin{sidewaystable}[t]
\centering
\scriptsize
\setlength{\tabcolsep}{5pt}
\caption{Wall-clock training time (hours) across environments and algorithms}
\label{tab:runtime}

\resizebox{\textheight}{!}{%
\begin{tabular}{lcccccccc}
\toprule
Environment 
& P3O 
& DDPGLag 
& TRPOLag 
& OnCRPO 
& CPO 
& FOCOPS 
& SACLag 
& SACPID \\
\midrule
RTNEnv             & 0.13 & 0.26 & 0.12 & 0.12 & 0.12 & 0.21 & 0.35 & 0.34 \\
STNEnv             & 0.13 & 0.26 & 0.13 & 0.13 & 0.13 & 0.17 & 0.35 & 0.36 \\
UCEnv-v0           & 4.66 & 9.18 & 4.43 & 4.43 & 4.46 & 3.71 & 7.71 & 7.52 \\
UCEnv-v1           & 4.58 & 3.71 & 4.44 & 4.44 & 4.43 & 3.24 & 5.72 & 5.91 \\
GTEPEnv            & 0.33 & 1.00 & 0.31 & 0.31 & 0.32 & 0.32 & 1.11 & 0.96 \\
BlendingEnv        & 1.20 & 3.62 & 1.16 & 1.18 & 1.23 & 1.11 & 3.49 & 3.51 \\
InvMgmtEnv         & 0.93 & 2.60 & 0.82 & 0.83 & 0.88 & 1.25 & 3.07 & 2.67 \\
GridStorageEnv     & 0.94 & 2.70 & 0.86 & 0.84 & 0.95 & 1.03 & 2.60 & 2.50 \\
SchedMaintEnv-v0   & 0.57 & 1.70 & 0.56 & 0.56 & 0.58 & 1.45 & 4.02 & 3.60 \\
SchedMaintEnv-v1   & 1.49 & 3.35 & 1.45 & 1.51 & 1.42 & 1.19 & 2.94 & 2.89 \\
ASUEnv             & 2.38 & 8.80 & 2.42 & 2.39 & 2.42 & 13.57 & 5.72 & 9.91 \\
\bottomrule
\end{tabular}%
}
\end{sidewaystable}

\clearpage
\section{Comparison with other Open-source Repositories}\label{sec:appendix:envcomp}
\paragraph{Discussion.}  
Table~\ref{tab:env_comparison_1} systematically evaluates four representative open‐source reinforcement learning environments that incorporate constraint management. We classify each framework by its API, enforcement mechanism (e.g.\ truncation, reward/cost penalties, or formal CMDP wrappers), application areas, compatibility with SafeRL algorithm libraries, the dimensionality of observation and action spaces, and support for nonconvex and mixed decision variables.  

\begin{table}[H]
\centering
\caption{Comparison of Gym environments with constraint handling. ``Mixed Space'' refers to presence of both discrete and continuous variables.}
\label{tab:env_comparison_1}

\footnotesize
\setlength{\tabcolsep}{1pt}     % ← tighter horizontal padding between columns (default ≈6pt)
\renewcommand{\arraystretch}{1.5} % ← more vertical space between rows (default 1.0)

% ---------- (a) Environment class, constraints, and domains ----------
\begin{subtable}[t]{\textwidth}
\centering
\caption{Environment class, constraints, and application domains}
\begin{tabular}{
  >{\raggedright\arraybackslash}p{0.23\linewidth}
  >{\raggedright\arraybackslash}p{0.20\linewidth}
  >{\raggedright\arraybackslash}p{0.24\linewidth}
  >{\raggedright\arraybackslash}p{0.30\linewidth}
}
\toprule
Work & Env.\ Class & Constraint Handling & Application Domain \\
\midrule
OR Gym     & Gynasium & Truncation, Reward & Classical OR problems \\
\addlinespace[2pt]
SustainGym & Gynasium & Reward Penalties   & Sustainable Energy Systems \\
\addlinespace[2pt]
% PC Gym     & Gynasium & Truncation         & Chemical Process Control \\
% \addlinespace[8pt]
SafeOR Gym &
  \shortstack[l]{Gynasium\\+ CMDP wrapper} &
  \shortstack[l]{Truncation,\\Reward Penalty,\\Cost} &
  \shortstack[l]{Supply Chain,\\Chemical Production,\\Network scheduling,\\Power Systems} \\
\bottomrule
\end{tabular}
\end{subtable}

\vspace{0.6em}

% ---------- (b) Compatibility, structure, and sizes ----------
\begin{subtable}[t]{\textwidth}
\centering
\caption{Compatibility, structural properties, and observation/action sizes}
\begin{tabular}{
  >{\raggedright\arraybackslash}p{0.10\linewidth}
  >{\centering\arraybackslash}p{0.20\linewidth}
  >{\centering\arraybackslash}p{0.20\linewidth}
  >{\centering\arraybackslash}p{0.20\linewidth}
  >{\centering\arraybackslash}p{0.30\linewidth}
}
\toprule
Work & \shortstack{Compatible with\\ SafeRL Algorithms} & \shortstack{Nonconvex\\constraints} & \shortstack{Mixed state\\action space} & \shortstack{Obs / Action\\(mean, max)} \\
\midrule
OR Gym     & \xmark     & \xmark     & \checkmark & (242, 2501) / (57, 200) \\
\addlinespace[2pt]
SustainGym & \xmark     & \xmark     & \checkmark & (79, 150) / (33, 72)    \\
\addlinespace[2pt]
% PC Gym     & \xmark     & \xmark     & \xmark     & (6.25, 2) / (16, 5)     \\
% \addlinespace[2pt]
SafeOR Gym & \checkmark & \checkmark & \checkmark & (86, 4280) / (32, 272)  \\
\bottomrule
\end{tabular}
\end{subtable}

\end{table}

The key differences of our work compared with 
\begin{enumerate}
    \item \textbf{Constraint handling.} OR-Gym and SustainGym mainly rely on simple truncation to enforce basic constraints, such as bounds on state variables. More complex constraints are handled indirectly by assigning negative rewards for violations. SafeOR-Gym uses a more principled approach:
    \begin{itemize}
        \item For non-safety-critical constraints (e.g., delayed product delivery in a supply chain problem), penalties are applied to the reward signal.
        \item For safety-critical or hard physical constraints (e.g., preventing negative inventory levels, which are physically infeasible), SafeOR-Gym introduces explicit costs to guide Safe RL algorithms to rigorously respect these constraints.
    \end{itemize}
    In contrast, the purely reward-based handling in other frameworks can lead to violations of safety-critical constraints.
    
    \item \textbf{Environment and algorithm compatibility.} Both OR-Gym and SustainGym are implemented in Gymnasium and are primarily compatible with standard RL algorithms (e.g., PPO, DDPG from Stable-Baselines), treating all problems as unconstrained Markov Decision Processes (MDPs). SafeOR-Gym extends Gymnasium with a Constrained MDP (CMDP) wrapper, enabling compatibility with Safe RL algorithms such as those provided in OmniSafe, which explicitly handle constraints.
    
    \item \textbf{Type of constraints.} SafeOR-Gym includes environments with nonlinear, nonconvex constraints and more complex logical relationships between variables. SustainGym focuses on linear and convex constraints, while OR-Gym environments are limited to relatively simple linear constraints.
    
    \item \textbf{Problem difficulty.} While the problem dimensions (observation and action space sizes) are of similar magnitude across the frameworks, SafeOR-Gym instances are significantly harder to solve due to their more intricate and realistic constraint structures.
    
    \item \textbf{Application domains.} OR-Gym implements classical OR problems such as knapsack and traveling salesman, where well-known greedy heuristics can often provide near-optimal solutions without violating constraints. SafeOR-Gym focuses on more complex, realistic problems where simple heuristics typically violate feasibility constraints. SustainGym targets sustainable energy systems, whereas SafeOR-Gym spans a wider set of domains, including supply chains, chemical production, scheduling, and power systems.
\end{enumerate}

\end{document}